\documentclass[]{grounding-template}
\input{compile_compat.tex}
\setcitestyle{sort}
\usepackage[shortlabels,inline]{enumitem} %
\usepackage{amsmath}
\usepackage{amssymb}

\usepackage{etoc}
\usepackage{wrapfig}

\usepackage{needspace}
\newsavebox{\ArxivNaturalTable}
\newcount\ArxivPreviousLines
\newcount\ArxivWrapSteps
\makeatletter
\newcommand{\FinishArxivWrap}{\par
  \ArxivWrapSteps=0
  \loop\ifnum\c@WF@wrappedlines>1
    \ArxivPreviousLines=\c@WF@wrappedlines
    \noindent\strut\par
    \advance\ArxivWrapSteps by 1
    \ifnum\c@WF@wrappedlines<\ArxivPreviousLines\else\WFclear\fi
    \ifnum\ArxivWrapSteps>80\WFclear\fi
  \repeat\WFclear}
\makeatother

\makeatletter
\newcommand{\CompactContents}{%
  \small\setlength{\parskip}{0pt}%
  \renewcommand*\l@section[2]{\@dottedtocline{1}{0em}{1.6em}{\bfseries ##1}{##2}}%
  \renewcommand*\l@subsection[2]{\@dottedtocline{2}{1.6em}{2.5em}{##1}{##2}}%
  \renewcommand*\l@subsubsection[2]{\@dottedtocline{3}{4.1em}{3.3em}{##1}{##2}}%
}
\makeatother

\makeatletter
\patchcmd{\WF@putfigmaybe}
  {\ifdim\dimen@>\@tempdimb\global\WF@floatfalse\pagebreak\fi}
  {\ifWF@float\else\pagebreak\fi}
  {}{\PackageError{arxiv-layout}{Could not install wrap fit guard}{}}
\apptocmd{\WF@startwrapping}{\ifinner\else\suppressfloats\fi}{}{}
\apptocmd{\@floatplacement}{\ifvoid\WF@box\else\global\@topnum\z@\global\@botnum\z@\fi}{}{}
\newcommand{\FinishPlacedArxivWrap}{\ifvoid\WF@box\FinishArxivWrap\fi}
\makeatother

\newsavebox{\ArxivInlineBox}
\newenvironment{ArxivInlineBlock}{%
  \par\FinishArxivWrap
  \begin{lrbox}{\ArxivInlineBox}\begin{minipage}{\textwidth}%
  \setlength{\parskip}{4pt}%
}{%
  \par\FinishArxivWrap\end{minipage}\end{lrbox}%
  \suppressfloats\Needspace{\dimexpr\ht\ArxivInlineBox+\dp\ArxivInlineBox+\baselineskip\relax}\suppressfloats%
  \noindent\usebox{\ArxivInlineBox}\par
}

\newcommand{\FrontContents}{%
  \etocdepthtag.toc{main}%
  \phantomsection\addcontentsline{toc}{section}{Abstract}%
  {\setlength{\parindent}{0pt}%
    \section*{Abstract}\abstractlist\par}%
  {\CompactContents
    \etocsettagdepth{main}{subsubsection}%
    \etocsettagdepth{appendix}{none}%
    \etocsettocstyle{\section*{Contents}}{}%
    \tableofcontents}%
  \clearpage
  {\CompactContents
    \etocsettagdepth{main}{none}%
    \etocsettagdepth{appendix}{subsubsection}%
    \etocsettocstyle{\section*{Appendix Contents}}{}%
    \tableofcontents}%
  \clearpage
}

\newcommand{\question}[2]{
    \begin{tcolorbox}[
        enhanced,
        frame hidden,
        colback=blue!5,
        borderline west={2pt}{0pt}{blue!70!black},
        sharp corners,
        boxsep=0pt,
        left=8pt,
        right=4pt,
        top=8pt,
        bottom=8pt,
        before skip=12pt plus 2pt minus 2pt,
        after skip=8pt plus 2pt minus 2pt,
    ]
        \phantomsection\label{question:#1}%
        \noindent\textcolor{blue!70!black}{\textbf{\textit{Question #1:}}} #2
    \end{tcolorbox}
}

\newcommand{\finding}[2]{
    \begin{tcolorbox}[
        colback=blue!5,
        colframe=blue!70!black,
        arc=5pt,
        boxsep=5pt,
        left=2pt,
        right=2pt,
        top=2pt,
        bottom=2pt,
        boxrule=0.8pt,
        drop shadow=gray!50!white,
        enhanced jigsaw,
        before skip=12pt plus 2pt minus 2pt, %
        after skip=12pt plus 2pt minus 2pt,  %
    ]
        \phantomsection\label{finding:#1}%
        \noindent\textbf{\textit{Finding #1:}} #2
    \end{tcolorbox}
}

\newcommand{\takeaway}[2]{
    \begin{tcolorbox}[
        colback=blue!5,
        colframe=blue!70!black,
        arc=5pt,
        boxsep=5pt,
        left=2pt,
        right=2pt,
        top=2pt,
        bottom=2pt,
        boxrule=0.8pt,
        drop shadow=gray!50!white,
        enhanced jigsaw,
        before skip=12pt plus 2pt minus 2pt, %
        after skip=12pt plus 2pt minus 2pt,  %
    ]
        \phantomsection\label{takeaway:#1}%
        \noindent\textbf{\textit{Takeaway #1:}} #2
    \end{tcolorbox}
}

\title{GroundingPI: A Grounding Foundation Model towards Physical Intelligence with Visual Primitives}

\author[1,2,*,\ddagger]{Qize~Yu}
\author[1,*]{Lianrui~Fan}
\author[1,3,7,*]{Boyu~Chen}
\author[2,*]{Jiaqi~Liang}
\author[1,*]{Xini~Ding}
\author[2]{Yue~Chen}
\author[1,2]{Zetian~Song}
\author[2,6]{Yuran~Wang}
\author[1]{Yi~Zou}
\author[3]{Kaixuan~Wang}
\author[3]{Tianxing~Chen}
\author[8]{Wenxuan~Song}
\author[2]{Bohan~Zhou}
\author[2]{Mingleyang~Li}
\let\titleauthorsbeforecompactrow\authorlist
\let\authorlist\empty
\author[7]{Siqiao~Huang}
\author[2]{Yuqi~Ye}
\author[1]{Caigao~Jiang}
\author[1]{Wei~Wei}
\author[4]{Ruihai~Wu}
\author[1]{Hang~Zhang}
\author[1]{Yixiao~Ge}
\author[1]{Shuchang~Zhou}
\let\titlecompactrowauthors\authorlist
\let\authorlist\empty
\author[5]{Shilong~Liu}
\author[1]{Xianming~Liu}
\author[3,\dagger]{Ping~Luo}
\author[1,\dagger]{Shiyu~Huang}
\let\titleauthorsaftercompactrow\authorlist
\newsavebox{\titlecompactrowbox}
\newcommand{\titlecompactauthorrow}{%
  \begingroup
    \sbox{\titlecompactrowbox}{\titlecompactrowauthors,}%
    \ifdim\wd\titlecompactrowbox>\linewidth
      \resizebox{\linewidth}{\height}{\usebox{\titlecompactrowbox}}%
    \else
      \usebox{\titlecompactrowbox}%
    \fi
  \endgroup
}
\renewcommand{\authorlist}{%
  \titleauthorsbeforecompactrow,\linebreak[4]%
  \titlecompactauthorrow\linebreak[4]%
  \titleauthorsaftercompactrow
}

\affiliation[1]{XPeng Inc.}
\affiliation[2]{Peking University}
\affiliation[3]{The University of Hong Kong}
\affiliation[4]{University of California, Berkeley}
\affiliation[5]{Princeton University}
\affiliation[6]{National University of Singapore}
\affiliation[7]{Tsinghua University}
\affiliation[8]{HKUST (GZ)}
\contribution[*]{Equal contribution}
\contribution[\ddagger]{Project lead}
\contribution[\dagger]{Corresponding authors}

\project{\href{https://groundingpi.github.io/}{\sffamily \fontsize{8.8pt}{11pt}\selectfont \texttt{https://groundingpi.github.io/}}}
\code{\href{https://github.com/groundingpi/GroundingPI}{\sffamily \fontsize{8.8pt}{11pt}\selectfont \texttt{https://github.com/groundingpi/GroundingPI}}}
\model{\href{https://huggingface.co/GroundingPI/GroundingPI}{\sffamily \fontsize{8.8pt}{11pt}\selectfont \texttt{https://huggingface.co/GroundingPI/GroundingPI}}}

\titleteaser{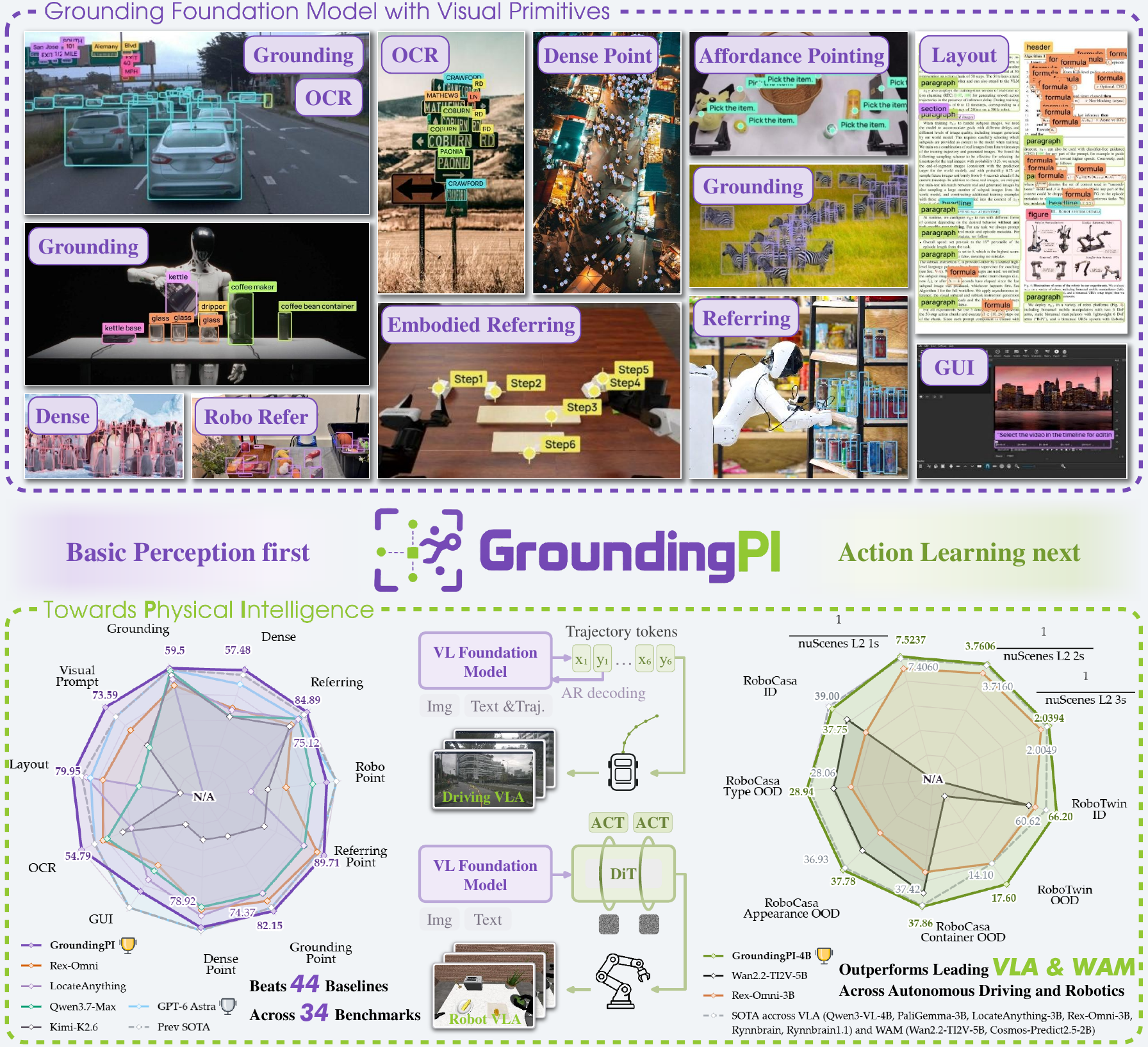}

\newlength{\titleauthorstoteaserskip}
\newlength{\titleteasertocaptionskip}
\newlength{\titlecaptiontoruleskip}
\abstract{Precise grounding matters. It specifies which object is the target and where that object is, even in clutter and for tiny objects, and it has to be fast enough for closed-loop control. Yet vision-language-action (VLA) and world-action models (WAMs) take perception from general-purpose vision-language and video-generation backbones, which still fail in these settings. We introduce GroundingPI, a 4B grounding foundation model that generates points and boxes as quantized coordinates in a shared vocabulary.  Training combines multimodal and spatial pretraining, supervised fine-tuning, and reinforcement learning with GRPO, using supervision from public datasets and dedicated data engines. Against 44 baselines across 34 grounding benchmarks spanning 11 perceptual capabilities, GroundingPI establishes a new state of the art, averaging 73.68\%, above the larger GPT-6 Astra (71.54\%). As a downstream visual backbone, GroundingPI improves performance on robotic manipulation and autonomous driving. Across RoboTwin 2.0 and RoboCasa-GR1, it outperforms all evaluated backbones in all four out-of-distribution settings, with relative gains of up to 24.8\% over the strongest baseline. On RoboCasa-GR1, GroundingPI trained with 50\% of the demonstrations outperforms those baselines trained with 75\%.
Its mean open-loop L2 error on nuScenes is 0.296 m.
We systematically analyze GroundingPI's pretraining in scale and data composition. Grounding quality and downstream manipulation improve with pretraining scale, with generalization tracking perceptual gains most closely. Analyzing the data recipe across these 11 perceptual capabilities shows dense grounding's substantial benefits for both, and OCR's potential as a catalyst for perceptual learning. These results support grounding as a perceptual foundation, and dedicated perceptual pretraining as a promising direction for foundation models of physical intelligence.

}

\begin{document}
\makeatletter
\renewcommand{\mymaketitle}{%
  \tcbset{enhanced,frame hidden}%
  \tcbset{left=0.5cm}%
  \tcbset{right=0.5cm}%
  \tcbset{top=0.26cm}%
  \tcbset{bottom=0.26cm}%
  \tcbset{arc=10pt}%
  \tcbset{colback=templatebackground}%
  \tcbset{before skip=0pt}%
  \tcbset{grow to left by=1.5pt}%
  \tcbset{grow to right by=1.5pt}%
  \begin{tcolorbox}
    \hypersetup{linkcolor=templatetitle,citecolor=templatetitle,urlcolor=templatetitle}%
    \setlength{\parindent}{0cm}%
    \setlength{\parskip}{0.5cm}%
    {\setlength{\parskip}{0cm}%
      \raggedright
      \nohyphens
      {\setstretch{1.618}\titlelist\par}%
      \vskip 0.12cm
      \authorlist\par
      \vskip 0.12cm
      \affiliationlist\par
      \contributionlist\par
    }%
    \ifdefempty{\teaserfile}{}{%
      {\setlength{\parskip}{0pt}%
        {\centering
          \includegraphics[width=\linewidth,page=1]{figures/teaser.pdf}\par
          {\captionsetup{type=figure,skip=5pt}%
            \caption{Overview of \textbf{GroundingPI}, featuring strong visual perception capabilities. Across 34 perception benchmarks, \textbf{GroundingPI} outperforms 44 baselines and leading VLA/WAM models in driving and robotics, suggesting that stronger perception can be a better foundation for physical intelligence than generality alone.}\label{fig:teaser}}%
        }%
        \templatedashedrule
      }%
    }%
    \ifdefempty{\metadatalist}{}{%
      \vskip 0.04cm
      \noindent
      \begin{minipage}[c]{\linewidth}
        \setlength{\parskip}{0cm}%
        \raggedright
        \metadatalist
      \end{minipage}\par
      \vskip 0.04cm
    }%
  \end{tcolorbox}%
  \tcbset{reset}%
  \FloatBarrier
}

\makeatother
\maketitle
\clearpage
\FrontContents

\newtcolorbox{introquestions}{
    enhanced,
    frame hidden,
    colback=blue!5,
    borderline west={2pt}{0pt}{blue!70!black},
    sharp corners,
    boxsep=0pt,
    left=8pt,
    right=8pt,
    top=6pt,
    bottom=4pt,
    before skip=7pt plus 2pt minus 1pt,
    after skip=10pt plus 2pt minus 2pt,
}
\newcommand{\introquestion}[2]{%
    \noindent\textcolor{blue!70!black}{\textbf{\textit{Question #1:}}} #2\par\vskip3pt\relax%
}

\section{Introduction}
\label{sec}

Visual grounding connects language to objects, locations, and interaction-relevant structures, making it a core perceptual capability of vision-language models (VLMs). Despite recent progress in grounding UI elements~\citep{lin2024showui,liu2025scalecua}, regions~\citep{lai2024lisa,ren2024pixellm}, and task-relevant entities~\citep{zhang2024llavagrounding,rexomni,yu2025perceptionr1}, achieving broad and precise grounding for physical intelligence~\citep{black2024pi0,openwam,openvla} remains challenging.

When tidying a cluttered desk, one finds the mug and its handle before deciding how to grasp and move it, establishing \emph{where to act} before determining \emph{how to act}. Vision-language-action (VLA) models~\citep{openvla,black2024pi0,nvidia2025gr00tn1} and world-action models (WAMs) ~\citep{cosmospolicy,dreamzero,openwam} increasingly build on general-purpose vision-language and video-generation backbones. However, these backbones are not primarily optimized for the precise grounding required by physical interaction and leave perception as a bottleneck for downstream action learning.

Our evaluations reveal substantial weaknesses in general-purpose VLMs and considerable room for improvement in dedicated models such as Rex-Omni and LocateAnything~\citep{rexomni,locateanything}. These limitations are particularly evident in demanding settings such as dense scenes and tiny objects, where Qwen3.5 models struggle and even GPT-6 Astra leaves room for improvement. Such perceptual gaps can leave scarce action data responsible for both perceptual and action learning. Compute and latency constraints further limit reliance on larger backbones for action execution ~\citep{nvidia2025gr00tn1,fastinslow}. These gaps motivate building a more capable grounding foundation model.

Recent policy recipes already incorporate perceptual learning: the progression from $\pi_0$~\citep{black2024pi0} to $\pi_{0.5}$~\citep{intelligence2025pi05visionlanguageactionmodelopenworld} adds bounding-box prediction within a broader knowledge-insulation recipe~\citep{black2024pi0,intelligence2025pi05visionlanguageactionmodelopenworld,driess2025knowledgeinsulatingvisionlanguageactionmodels}, while other approaches introduce auxiliary modules and learning objectives to enhance perception~\citep{affordancevla,reconvla,sgvla,chen2026pa3ff}. We argue that a more natural and scalable approach is to develop strong grounding as a native capability of the foundation model. We pursue a grounding foundation model built on visual primitives, aiming to advance broad, precise perception and investigate its value for physical intelligence. We ask:

\begin{introquestions}
    \introquestion{1}{\textbf{How far can we push the perceptual capabilities of grounding foundation models?}}
    \introquestion{2}{\textbf{What does a stronger grounding foundation contribute to physical intelligence, and which grounding capabilities matter most?}}
    \introquestion{3}{\textbf{What might these findings suggest about the design of future embodied foundation models?}}
\end{introquestions}

To address these questions, we introduce \textbf{GroundingPI}, a 4B-parameter grounding foundation model for a broad range of perception tasks, built on visual primitives including points and bounding boxes. A shared vocabulary with quantized coordinates unifies diverse perception tasks as language-conditioned structured generation, covering grounding, referring, pointing, OCR, GUI and layout grounding, and visual prompting (\Cref{fig:teaser}). Training combines multimodal and spatial pretraining, supervised fine-tuning, and reinforcement learning with GRPO, using supervision from public datasets and dedicated data engines.

Evaluated against 44 baselines across 34 grounding benchmarks spanning 11 perceptual capabilities, GroundingPI achieves 73.68\% on average, establishing a new state of the art, averaging 73.68\%, above the larger GPT-6 Astra (71.54\%). We compare with GPT-6 Astra specifically to explore the potential and limits of improving basic perceptual performance. We further evaluate transfer to autonomous driving on nuScenes and robotic manipulation on RoboTwin 2.0 and RoboCasa-GR1. Within each manipulation benchmark, we fix the Action DiT, action data, and training budget. GroundingPI demonstrates strong transfer to physical intelligence tasks, consistently outperforming all evaluated mainstream backbones, in all four OOD settings, with relative improvements of up to 24.8\% over the strongest baseline. Action-data efficiency also improves in robotic manipulation: on RoboCasa-GR1, GroundingPI trained with 50\% of demonstrations outperforms all compared baselines trained with 75\%.
These results support the value of a stronger perceptual foundation for downstream action learning.

Our analyses offer insights into perceptual pretraining and the design of embodied foundation models. Scaling and data ablations show that both pretraining scale and data composition matter for grounding and downstream transfer. Basic grounding provides an important foundation, dense grounding substantially benefits autonomous driving and robotic manipulation, and OCR shows potential as a catalyst for perceptual learning. Our design insights focus on perceptual foundations for System-1-style execution \citep{nvidia2025gr00tn1,fastinslow}. We discuss possible directions for designing such embodied foundation models to complement the high-level reasoning and planning of frontier models such as Astra. Visual primitives may further enable physical prompting, specifying targets, locations, and structures alongside language.

In summary, our major contributions are as follows:
\begin{itemize}[leftmargin=1.5em]

\item \textbf{A strong grounding foundation model.} We introduce GroundingPI, a 4B model built on visual primitives, with a staged training recipe and state-of-the-art grounding performance.

\item \textbf{Transfer toward physical intelligence.} Autonomous driving and robotic manipulation evaluations demonstrate the value of this perceptual foundation, including strong ID and OOD performance and improved action-data efficiency.

\item \textbf{Insights into perceptual pretraining and future embodied paradigms.} We analyze how pretraining scale and data composition shape grounding and transfer, and discuss implications for System-1 foundation-model design and its complementary role in future embodied systems.

\end{itemize}

\section{Related Work}
\label{sec:related_work}

\subsection{Visual Grounding Models}

Visual grounding has evolved from closed-set object detection \citep{yolo,DETRR50} to language-conditioned localization beyond fixed vocabularies through grounded image-text pretraining \citep{glip,groundingdino}. Generative grounding models adopt autoregressive next-token prediction to produce labels and quantized coordinates as structured sequences, providing a shared interface across diverse perception tasks \citep{chen2022pix2seqlanguagemodelingframework,xiao2024florence,rexomni}, while LocateAnything explores precise grounding with parallel box decoding \citep{rexomni,locateanything}. Spatial affordance prediction and spatial referring extend these capabilities toward robotic interaction \citep{RoboPoint13B,RoboRefer2B,wu2026sugarscalablehumanvideodrivengeneralizable}. These advances motivate studying broad and precise grounding as a transferable perceptual foundation for physical intelligence.

\subsection{Foundation Models for Physical Intelligence}

Physical-intelligence models increasingly inherit complementary priors from large-scale pretraining. VLA models build on VLMs to transfer broad visual-semantic knowledge and language grounding~\citep{openvla,black2024pi0}, while WAMs adapt video and world models to leverage spatiotemporal and visual-dynamics priors~\citep{cosmospolicy,dreamzero,openwam}. Embodied foundation models further combine semantic, dynamic, and embodied experience toward more general physical intelligence~\citep{intelligence2025pi05visionlanguageactionmodelopenworld,nvidia2025gr00tn1,dang2026rynnbrain,nvidia2026cosmos3}. However, these broad capabilities do not necessarily provide the precise, task-conditioned perception required for physical interaction, leaving a capability mismatch between pretraining and action learning. Together, this capability mismatch motivates closer study of how precise perception can support action learning.

\subsection{Perceptual Learning for Action Transfer}

Recent policy recipes strengthen perception through bounding-box prediction, task-relevant region reconstruction, affordance representations, and spatial auxiliary supervision~\citep{intelligence2025pi05visionlanguageactionmodelopenworld,reconvla,affordancevla,sgvla,you2026affordancewamaffordanceawarejointworldaction,shen2026ld4wamlearninglatentdynamics}. These approaches demonstrate the value of perception, but usually introduce it as a task-specific proxy, auxiliary objective, or intermediate representation within action learning. Such mechanisms cover selected perceptual signals, often depend on additional annotations or policy-specific components, and are difficult to scale across the diverse scenes and perceptual demands of physical intelligence. Together, these limitations motivate a perception-native foundation model: one that learns broad and precise perception before action adaptation, so downstream policies can build on perception rather than recover it from task-specific proxies.

\section{GroundingPI: A Grounding Foundation Model}
\label{sec:groundingpi}

\subsection{Model Architecture and Grounding Formulation}
\label{sec:architecture_formulation}

\begin{figure}[!tbp]
  \centering
  \includegraphics[width=1.0\linewidth]{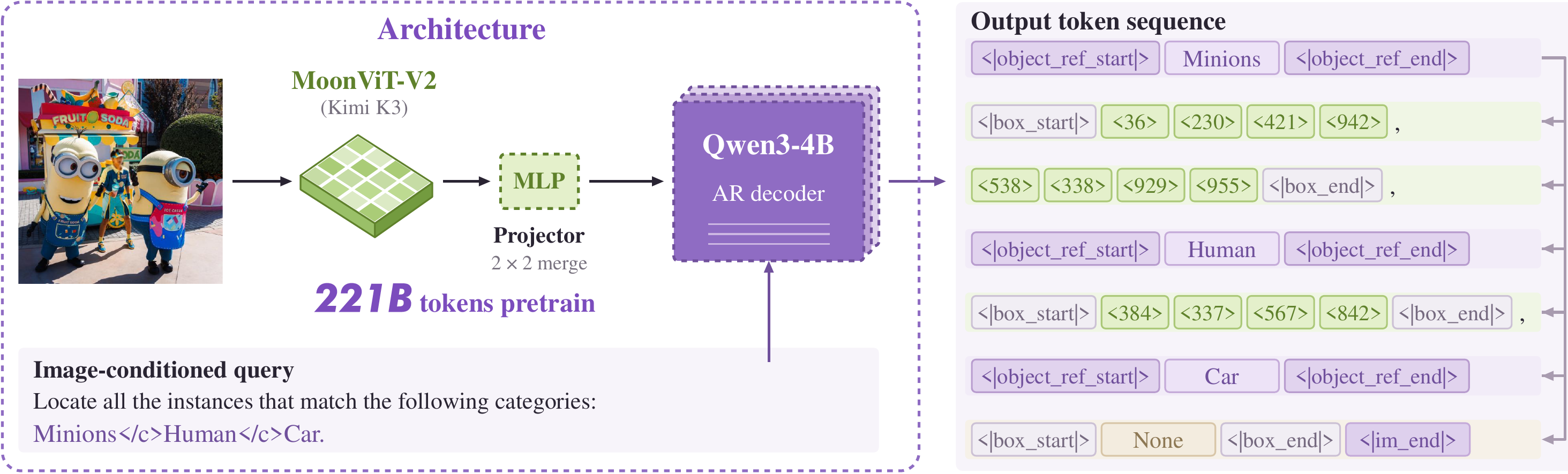}
  \caption{\textbf{GroundingPI architecture.} The grounding example includes two minions and one human and an absent car category (\texttt{None}). Architecture specifications and parameter counts are provided in \Cref{app:groundingpi_architecture}.}
  \label{fig:groundingpi-architecture}
\end{figure}

GroundingPI is an autoregressive grounding foundation model for language-conditioned perception. As shown in \Cref{fig:groundingpi-architecture}, it couples a MoonViT-V2 (Kimi K3) visual encoder~\citep{KimiK3} with a Qwen3-4B language decoder~\citep{Qwen34BLLM}. A learnable projector aggregates adjacent $2\times2$ visual features and maps them to the language embedding space through a two-layer MLP with GELU and output normalization.

Given an image $I$ and a query $P$, the encoder $E_\psi$ and projector $C_\phi$ produce visual embeddings $V=C_\phi(E_\psi(I))$. The decoder generates a variable-length response $Y$ autoregressively, with $p_\theta(Y\mid I,P)=\prod_{i=1}^{|Y|}p_\theta(y_i\mid V,P,y_{<i})$.
Semantic labels, protocol markers, and 1,000 quantized coordinate tokens (\texttt{<0>}--\texttt{<999>}) share the output vocabulary; responses are variable-length; an absent queried category yields \texttt{None}. Further input/output conventions are detailed in \Cref{app:groundingpi_protocol}.

\subsection{GroundingPI Data}
\label{sec:groundingpi_data}

\label{sec:public_datasets}
\label{sec:data_engines}

Training data combines public datasets with annotations produced by our data engine.

\begin{figure}[!tbp]
  \centering
  \includegraphics[width=1.0\linewidth]{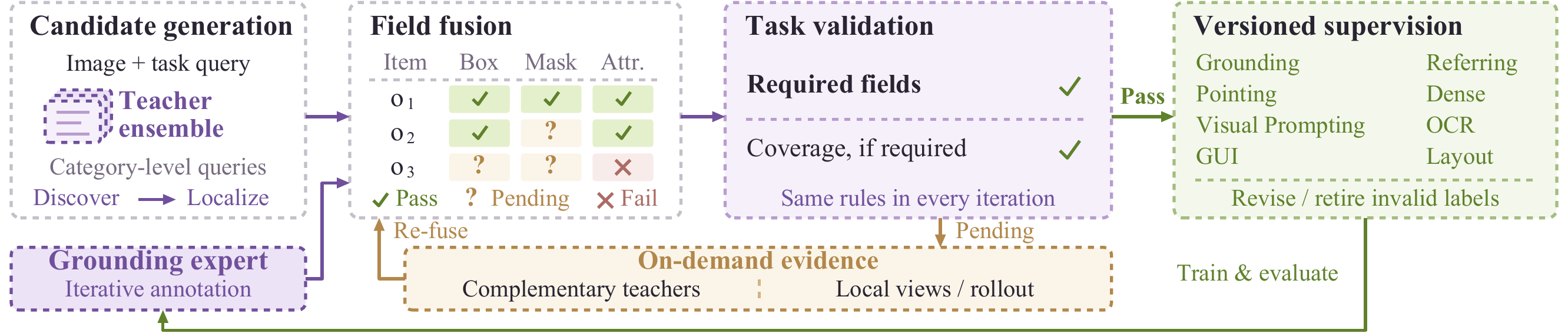}
  \caption{\textbf{Data engine.} Multi-teacher fusion and task-specific validation guide expert iteration, with targeted observations resolving uncertain labels.}
  \label{fig:data-engine}
  \label{fig:public-datasets}
\end{figure}

Our data engine fuses multi-teacher annotations at the field level and applies task-specific validation (\Cref{fig:data-engine}). Accepted labels train a unified grounding expert for iterative annotation, while complementary teachers and local observations resolve uncertain cases to expand supervision and refine existing labels. Field validation, task dependencies, and label revision are detailed in \Cref{app:groundingpi_data_engine}.

\subsection{Training Design}
\label{sec:training_design}

\subsubsection{Base VLM Training}
\label{sec:base_vlm_training}

Base training first aligns the projector using image-caption supervision while freezing both backbones. It then updates all modules in two stages: joint multimodal pretraining on text and image--text supervision, followed by general visual/video understanding through question answering and captioning. All stages use causal next-token prediction. Loss normalization and optimization details are provided in \Cref{app:groundingpi_vlm_training}.

\subsubsection{Supervised Fine-Tuning}
\label{sec:supervised_finetuning}

SFT aligns coordinates with visual locations and learns the shared output protocol. Under teacher forcing, we minimize $\mathcal L_{\mathrm{SFT}}=-|\mathcal S|^{-1}\sum_{t\in\mathcal S}\log p_\theta(y_t\mid I,P,y_{<t})$, where $\mathcal S$ contains assistant-response labels, protocol markers, and coordinates, excluding prompt and visual positions. Supervised adaptation first updates all modules, then freezes the vision encoder and projector for language-side refinement. Module schedules and hyperparameters are given in \Cref{app:groundingpi_vlm_training}.

\subsubsection{Reinforcement Post-Training}
\label{sec:reinforcement_posttraining}

\begin{figure}[!tbp]
  \centering
  \includegraphics[width=1.0\linewidth]{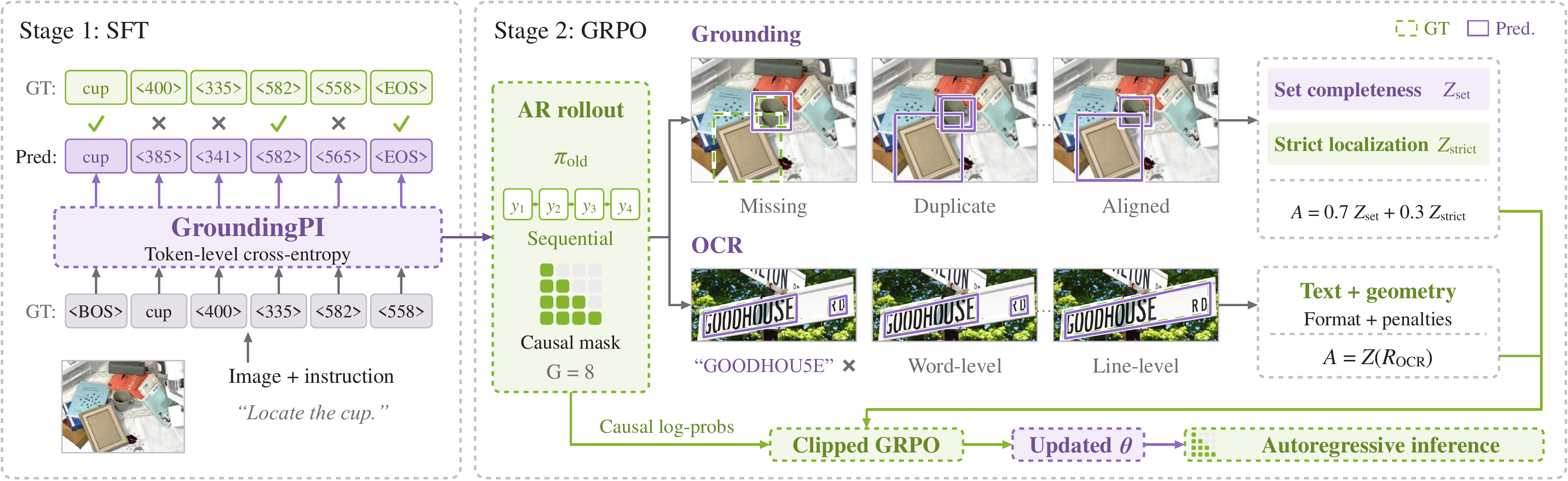}
  \caption{Overview of GroundingPI's two-stage training pipeline. The first stage uses supervised fine-tuning (SFT) to learn structured grounding outputs with token-level cross-entropy. The second stage applies group relative policy optimization (GRPO) with task-specific rewards that balance grounding completeness and localization, and assess OCR text--geometry consistency and output reliability. Illustrative rollouts highlight common prediction errors and alternative OCR granularities.}
  \label{fig:groundingpi-sft-rl}
\end{figure}

To directly optimize localization, target coverage, and text--geometry consistency, we apply group relative policy optimization (GRPO)~\citep{GRPO,ping2026longactharnessingintrinsicactivation} with $G=8$ autoregressive responses per image--query pair (\Cref{fig:groundingpi-sft-rl}).
For grounding, $R_{\mathrm{set}}$ measures coverage through reference-wise maximum-IoU matching with class validation. The complementary $R_{\mathrm{strict}}$ combines multi-threshold F1, localization, format, count, and ordering scores, penalizing duplicate and oversized boxes. The independently standardized rewards yield $\widetilde A_i=0.7Z(R_{\mathrm{set},i})+0.3Z(R_{\mathrm{strict},i})$. For OCR, Hungarian one-to-one matching evaluates text and geometry across complete word/line views or complementary references. Format gating and output penalties produce one composite reward, giving $\widetilde A_i=Z(R_{\mathrm{OCR},i})$. Here $Z$ standardizes each active reward within a prompt; the combined advantages are then standardized across the response batch. We optimize the clipped GRPO objective with a frozen SFT reference, updating only language parameters. Reward definitions and optimization details are provided in \Cref{app:groundingpi_grounding_reward,app:groundingpi_ocr_reward,app:groundingpi_vlm_rl}.

\section{GroundingPI Model towards Physical Intelligence}
\label{sec:physical_intelligence}

GroundingPI provides a strong perceptual interface for action learning. We study whether its precise, language-conditioned grounding representations translate into stronger action-learning capabilities in two downstream physical-intelligence paradigms: trajectory prediction for autonomous driving and continuous control for robot manipulation. Each domain retains its native action learner, while GroundingPI serves as the shared perceptual foundation.

\subsection{Autonomous Driving}
\label{sec:autonomous_driving}

We formulate autonomous driving on nuScenes~\citep{nuscenes} as language-conditioned trajectory prediction (\Cref{fig:teaser}). Given the current front-camera image and historical ego states, the token-based interface represents the future trajectory as six ordered waypoints at 0.5-second intervals over a three-second horizon. Each waypoint specifies a cumulative position in the current ego frame, with forward and left axes measured in meters. Fixed metric ranges map these coordinates to GroundingPI's existing \texttt{<0>}--\texttt{<999>} vocabulary. The language head generates the trajectory autoregressively, and response-only cross-entropy provides the adaptation objective. Decoding recovers metric waypoints while preserving their temporal order. Coordinate conventions and the open-loop L2 metric are detailed in \Cref{app:driving_setup}.

\subsection{Robot Manipulation}
\label{sec:robot_manipulation}

\paragraph{$\pi$-style robotics manipulation learning.}
For robot manipulation, we follow the $\pi$-style flow-matching policy~\citep{black2024pi0} and adopt the layer-wise connection used in StarVLA~\citep{community2026starvla} (\Cref{fig:teaser}). The pretrained foundation encodes the visual observation and language instruction once; its intermediate features are projected and resampled, then injected into the corresponding cross-attention layers of an Action DiT. Conditioned on these features and the robot state, the action expert denoises a noisy action chunk into continuous controls.

\paragraph{Controlled comparison across foundation models.}
To compare perceptual foundations under the same action-learning setup, we follow the two adaptation pathways implemented in StarVLA: Wan-$\pi$ for video-generation backbones and the StarVLA $\pi$-style pathway for vision-language foundations. This yields four controlled model families: \textbf{general-purpose vision-language models}, \textbf{video-generation backbones}, \textbf{grounding models}, and \textbf{embodied foundation models}. Within each benchmark, we keep the Action DiT, robot data, action representation, optimization budget, and inference procedure fixed, varying only the pretrained foundation. We compare their action acquisition and generalization, with full implementation details in \Cref{app:manipulation_setup}.

\section{Experiments}
\label{sec:experiments}

\ifdefined\BenchmarkTableFont\else
\newcommand{\BenchmarkTableFont}{\normalfont}
\newcommand{\BenchHead}[1]{\begin{tabular}{@{}c@{}}#1\end{tabular}}
\colorlet{benchmarktype}{black!8}
\definecolor{benchmarkpurple}{HTML}{F5F1FA}
\definecolor{benchmarkgreen}{HTML}{F0F3E8}
\fi

\ifdefined\GPIBoxHelpersLoaded\else
\def\GPIBoxHelpersLoaded{1}
\ifdefined\tcolorbox
  \providecommand{\finding}[2]{%
    \begin{tcolorbox}[colback=blue!4,colframe=blue!55!black,
      boxrule=0.6pt,arc=2pt,left=5pt,right=5pt,top=3pt,bottom=3pt,
      before skip=5pt,after skip=5pt]\small
      \textbf{\textit{Finding #1:}} #2
    \end{tcolorbox}}
  \providecommand{\takeaway}[2]{%
    \begin{tcolorbox}[colback=blue!4,colframe=blue!55!black,
      boxrule=0.6pt,arc=2pt,left=5pt,right=5pt,top=3pt,bottom=3pt,
      before skip=5pt,after skip=5pt]\small
      \textbf{\textit{Takeaway #1:}} #2
    \end{tcolorbox}}
\else
  \providecommand{\finding}[2]{\begin{gpiquestions}\textbf{\textit{Finding #1:}} #2\end{gpiquestions}}
  \providecommand{\takeaway}[2]{\begin{gpiquestions}\textbf{\textit{Takeaway #1:}} #2\end{gpiquestions}}
\fi
\fi

In this section, we ask:
\begin{introquestions}
    \introquestion{1}{\textbf{Q1: Perception.} How far can a compact grounding foundation advance broad and precise perception?}
    \introquestion{2}{\textbf{Q2: Transfer.} Does stronger perception support stronger physical intelligence?}
    \introquestion{3}{\textbf{Q3: Design.} Which perceptual capabilities and base-model choices help explain this transfer, and what might they suggest for embodied foundations?}
\end{introquestions}

\subsection{Grounding Benchmark Results}
\label{sec:grounding_results}

\paragraph{Implementation Details and Evaluation Setup.}
We evaluate GroundingPI against 44 baselines on 34 benchmarks spanning 11 perceptual capabilities, covering specialized detectors, general-purpose VLMs, grounding specialists, and embodied foundations. All numerical results and conclusions involving our models are based on the mean of ten runs, using five random seeds with two runs per seed. Owing to the high computational cost, other baselines that we evaluate locally for the main leaderboard are averaged over three runs, using three random seeds with one run per seed. GPT-6 Astra is evaluated with thinking effort set to \texttt{High}. We use F1mIoU for box grounding and OCR, F1@Point for object pointing, and task-specific accuracy for spatial and robo and GUI grounding.

\FinishArxivWrap
\begin{wrapfigure}{R}{0.65\textwidth}
\raggedleft
  \includegraphics[width=\linewidth]{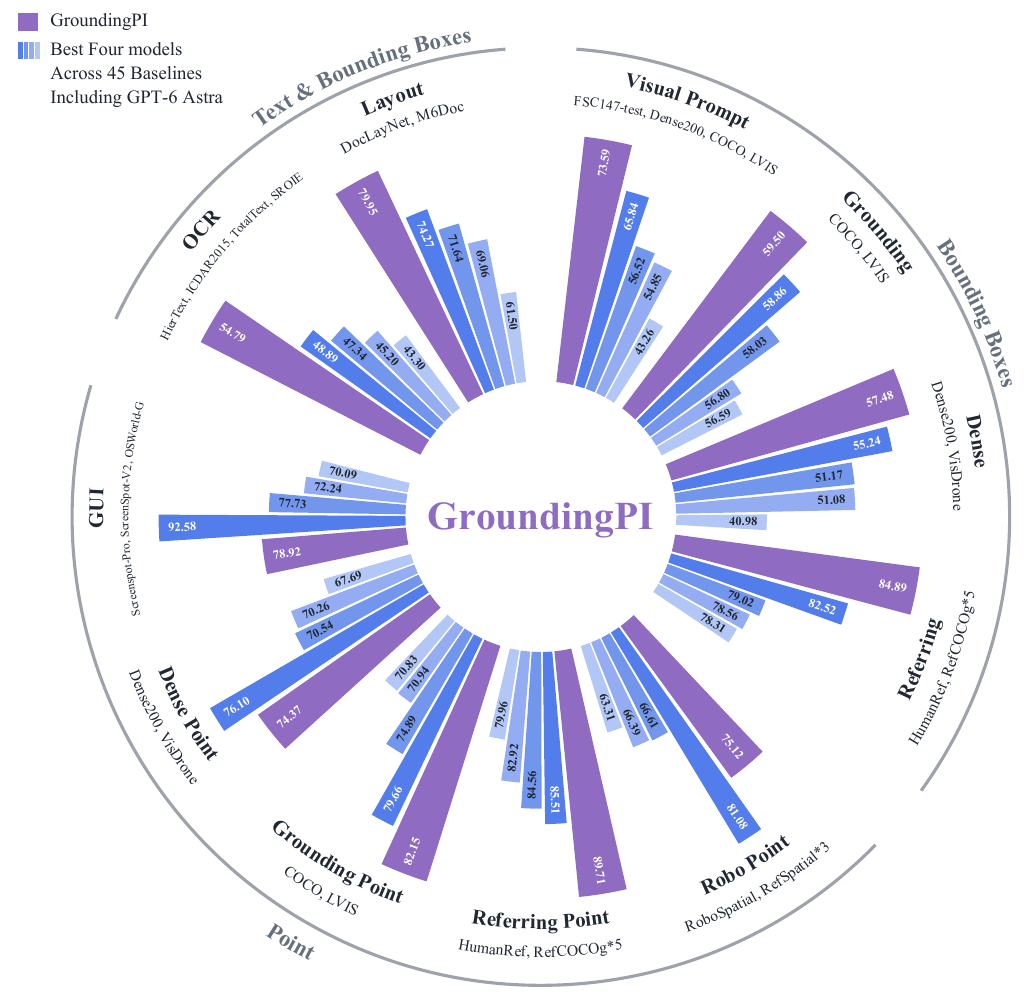}
  \caption{\textbf{Grounding across perceptual capabilities.} GroundingPI and leading baselines across box, point, text, and exemplar-conditioned tasks.
  }
  \label{fig:groundingpi-benchmark-results}
\end{wrapfigure}

\paragraph{Benchmark suite.}
Our evaluation covers common and long-tailed detection on COCO~\citep{COCO} and LVIS~\citep{LVIS}, and dense and tiny-object detection on Dense200~\citep{rexomni} and VisDrone~\citep{VisDrone}. Referring grounding uses RefCOCO and RefCOCO+~\citep{RefCOCO} and RefCOCOg~\citep{RefCOCOg,RefCOCOgUMD}, together with HumanRef~\citep{HumanRef}. Spatial and GUI grounding are evaluated on RoboSpatial~\citep{RoboSpatial}, RefSpatial~\citep{RoboRefer2B}, ScreenSpot-V2~\citep{ScreenSpotV2}, ScreenSpot-Pro~\citep{ScreenSpotPro}, and OSWorld-G~\citep{JEDI3B}. OCR uses HierText~\citep{HierText}, ICDAR2015~\citep{ICDAR2015}, TotalText~\citep{TotalText}, and SROIE~\citep{SROIE}; document layout grounding uses DocLayNet~\citep{DocLayNet} and M6Doc~\citep{M6Doc}. We additionally evaluate exemplar-based visual prompting on FSC147~\citep{FSC147} and detection benchmarks described above.

\paragraph{Main Results.}

GroundingPI achieves 73.68\% Avg, outperforming similarly sized models and remaining competitive with GPT-6 Astra (71.54\%). \Cref{fig:groundingpi-benchmark-results} summarizes this breadth. The selected comparisons below distinguish gains in coverage and localization from the remaining task-specific limitations.

\paragraph{Reporting conventions.}
\label{sec:benchmark_reporting}
We report percentage scores for selected baselines; bold marks column bests, including ties (lower is better only for parse error). Model-name stars denote externally reported rows; entry-level stars denote source/support exceptions. \texttt{--}, \texttt{N/A}, and \texttt{UNK} indicate unreported values, unsupported evaluations, and unspecified zero-shot status, respectively. Daggers flag prompt/parser uncertainty, including DeepSeek GUI; affected scores are descriptive rather than definitive capability estimates. Under our evaluation protocols, GroundingDINO lacks GUI/OCR/layout interfaces, and Kimi-K3 lacks compatible pointing/OCR/layout outputs. SenseNova-Vision's GUI evaluation is incompatible. Detailed exceptions appear in \Cref{app:grounding_results}.

\paragraph{Detection and referring grounding.}
In \Cref{tab:main-detection}, GroundingPI reaches 74.53 on Dense200, versus Astra's 65.04 and Qwen3-VL-4B's 14.02, directly addressing the dense-scene perceptual gap, although VisDrone remains challenging. RefCOCO avg is the unweighted mean of RefCOCO, RefCOCOg, and RefCOCOplus.

\begin{table}[htbp]
  \centering
  \BenchmarkTableFont
  \caption{\textbf{Detection and referring grounding (F1mIoU).} External rows are from \citep{rexomni}. Full results appear in Appendices~\ref{app:common_longtailed_detection}--\ref{app:referring_detection}.}
  \label{tab:main-detection}
  \begingroup
  \fontsize{8}{9.7}\selectfont
  \setlength{\tabcolsep}{3pt}
  \renewcommand{\arraystretch}{1.15}
  \resizebox{\linewidth}{!}{%
  \begin{NiceTabular}{@{}>{\raggedright\arraybackslash}p{177pt}cccccccc@{}}
  \CodeBefore
    \rowcolor{benchmarktype}{3}
    \rowcolor{benchmarktype}{6}
    \rowcolor{benchmarktype}{8}
    \rowcolor{benchmarkpurple}{15}
    \rowcolor{benchmarktype}{16}
    \rowcolor{benchmarktype}{19}
  \Body
  \toprule
  \textbf{Model} & \multicolumn{1}{c}{\textbf{Common}} & \multicolumn{1}{c}{\textbf{Long-tailed}} & \multicolumn{2}{c}{\textbf{Dense \& Tiny}} & \multicolumn{4}{c}{\textbf{Referring grounding}} \\
  \cmidrule(lr){2-2}\cmidrule(lr){3-3}\cmidrule(lr){4-5}\cmidrule(lr){6-9}
   & \BenchHead{COCO} & \BenchHead{LVIS} & \BenchHead{Dense200} & \BenchHead{VisDrone} & \BenchHead{HumanRef} & \BenchHead{RefCOCOg\\val} & \BenchHead{RefCOCOg\\test} & \BenchHead{RefCOCO\\avg} \\
  \midrule
  \multicolumn{9}{@{}l}{\hspace{0.4em}\strut\textbf{Closed-set Specialized Detectors}} \\
  DINO-R50\textsuperscript{*}\hspace{0.35em}\citep{DINOR50} & 55.60 & -- & -- & -- & -- & -- & -- & -- \\
  DETR-R50\textsuperscript{*}\hspace{0.35em}\citep{DETRR50} & 48.30 & -- & -- & -- & -- & -- & -- & -- \\
  \addlinespace[2pt]
  \multicolumn{9}{@{}l}{\hspace{0.4em}\strut\textbf{Open-set Specialized Detectors}} \\
  GroundingDINO\hspace{0.35em}\citep{groundingdino} & 60.56 & 52.61 & 24.92 & 34.47 & 46.13 & 49.77 & 50.43 & 45.15 \\
  \addlinespace[2pt]
  \multicolumn{9}{@{}l}{\hspace{0.4em}\strut\textbf{Vision-Language Models (<10B)}} \\
  Qwen3-VL-4B\hspace{0.35em}\citep{qwen3vl} & 46.53 & 49.86 & 14.02 & 31.14 & 70.06 & 75.27 & 75.88 & 75.24 \\
  Qwen3.5-9B\hspace{0.35em}\citep{qwen35} & 51.99 & 48.87 & 30.02 & 32.84 & 73.51 & 76.20 & 76.28 & 76.08 \\
  Rex-Omni\hspace{0.35em}\citep{rexomni} & 56.28 & 46.74 & 53.29 & 27.19 & 79.87 & 73.90 & 74.76 & 69.29 \\
  LocateAnything Hybrid\hspace{0.35em}\citep{locateanything} & 59.12 & 49.56 & 50.07 & 28.57 & 78.52 & 76.43 & 77.67 & 78.06 \\
  SenseNova-Vision\hspace{0.35em}\citep{SenseNovaVision7BMoT} & 57.49 & \textbf{56.12} & 68.13 & \textbf{42.35} & 79.34 & 78.69 & 79.48 & 77.55 \\
  RynnBrain1.1\hspace{0.35em}\citep{RynnBrain112B} & 35.15 & 26.01 & 0.12 & 8.29 & 52.07 & 67.66 & 68.24 & 63.63 \\
  GroundingPI & \textbf{62.98} & 56.02 & \textbf{74.53} & 40.44 & \textbf{88.56} & \textbf{85.62} & \textbf{84.36} & \textbf{83.92} \\
  \addlinespace[2pt]
  \multicolumn{9}{@{}l}{\hspace{0.4em}\strut\textbf{Vision-Language Models (10B--1T)}} \\
  SEED1.5-VL\textsuperscript{*}\hspace{0.35em}\citep{SEED15VL} & 51.40 & 46.70 & 53.20 & 27.40 & 81.60 & 71.90 & 73.20 & -- \\
  Qwen3.8-27B\hspace{0.35em}\citep{qwen38} & 60.84 & 49.99 & 34.30 & 33.33 & 77.26 & 76.03 & 77.37 & 77.89 \\
  \addlinespace[2pt]
  \multicolumn{9}{@{}l}{\hspace{0.4em}\strut\textbf{Vision-Language Models (>1T)}} \\
  Qwen3.7-Max\hspace{0.35em}\citep{qwen37} & 62.79 & 53.28 & 31.20 & 41.59 & 66.88 & 80.54 & 81.71 & 80.20 \\
  Kimi-K2.6 & 61.16 & 51.19 & 42.67 & 29.95 & 73.76 & 70.33 & 71.87 & 69.66 \\
  Kimi-K3\hspace{0.35em}\citep{KimiK3} & 60.89 & 47.73 & 51.64 & 30.31 & 79.89 & 73.39 & 73.99 & 72.86 \\
  GPT-6 Astra & 62.75 & 54.97 & 65.04 & 37.12 & 83.01 & 74.98 & 78.91 & 77.81 \\
  \addlinespace[2pt]
  \bottomrule
  
  \end{NiceTabular}%
  }
  \endgroup
\end{table}

\paragraph{Robot, spatial, and GUI grounding.}
\Cref{tab:main-spatial-gui} shows strong spatial transfer. RefSpatial averages its Location and Placement splits. GroundingPI also improves over the selected compact baselines on all benchmarks, while Astra retains a substantial advantage, exposing a remaining limit.
\begin{table}[htbp]
  \centering
  \BenchmarkTableFont
  \caption{\textbf{Robot, spatial, and GUI grounding (accuracy).} External RefSpatial and JEDI/UI-R1 scores are from \citep{rexomni}, respectively; GUI-Owl scores are from \citep{locateanything}. Full results appear in \Cref{app:robot_spatial_pointing,app:gui_grounding}.}
  \label{tab:main-spatial-gui}
  \begingroup
  \fontsize{8}{9.7}\selectfont
  \setlength{\tabcolsep}{3pt}
  \renewcommand{\arraystretch}{1.15}
  \resizebox{\linewidth}{!}{%
  \begin{NiceTabular}{@{}>{\raggedright\arraybackslash}p{177pt}cccccc@{}}
  \CodeBefore
    \rowcolor{benchmarktype}{3}
    \rowcolor{benchmarktype}{5}
    \rowcolor{benchmarkpurple}{15}
    \rowcolor{benchmarktype}{16}
    \rowcolor{benchmarktype}{22}
  \Body
  \toprule
  \textbf{Model} & \multicolumn{3}{c}{\textbf{Robot and spatial pointing}} & \multicolumn{3}{c}{\textbf{GUI grounding}} \\
  \cmidrule(lr){2-4}\cmidrule(lr){5-7}
   & \BenchHead{RefSpatial\\(avg)} & \BenchHead{RefSpatial\\Unseen} & \BenchHead{RoboSpatial\\Context} & \BenchHead{ScreenSpot-Pro} & \BenchHead{ScreenSpot-V2} & \BenchHead{OSWorld-G} \\
  \midrule
  \multicolumn{7}{@{}l}{\hspace{0.4em}\strut\textbf{Open-set Specialized Detectors}} \\
  GroundingDINO\hspace{0.35em}\citep{groundingdino} & 14.25 & 4.33 & 4.92 & N/A\textsuperscript{*} & N/A\textsuperscript{*} & N/A\textsuperscript{*} \\
  \addlinespace[2pt]
  \multicolumn{7}{@{}l}{\hspace{0.4em}\strut\textbf{Vision-Language Models (<10B)}} \\
  JEDI\textsuperscript{*}\hspace{0.35em}\citep{JEDI3B} & -- & -- & -- & 36.10 & 88.60 & -- \\
  UI-R1\textsuperscript{*}\hspace{0.35em}\citep{UIR13B} & -- & -- & -- & 17.80 & 85.40 & -- \\
  Qwen3-VL-4B\hspace{0.35em}\citep{qwen3vl} & 49.00 & 27.27 & 64.75 & 56.74 & 92.30 & 56.91 \\
  Qwen3.5-9B\hspace{0.35em}\citep{qwen35} & 55.92 & 37.01 & 60.66 & 53.13 & 90.57 & 60.99 \\
  Rex-Omni\hspace{0.35em}\citep{rexomni} & 51.75 & 37.01 & 59.02 & 36.75 & 88.29 & 46.10 \\
  LocateAnything Hybrid\hspace{0.35em}\citep{locateanything} & 36.17 & 20.78 & 14.75 & 57.05 & 89.94 & 60.46 \\
  SenseNova-Vision\hspace{0.35em}\citep{SenseNovaVision7BMoT} & 20.38 & 8.54 & 0.82 & N/A\textsuperscript{\textdagger} & N/A\textsuperscript{\textdagger} & N/A\textsuperscript{\textdagger} \\
  RynnBrain1.1\hspace{0.35em}\citep{RynnBrain112B} & 50.60 & 36.90 & 54.10 & 34.66 & 70.44 & 33.33 \\
  RoboRefer\textsuperscript{*}\hspace{0.35em}\citep{RoboRefer2B} & 50.00 & 39.00 & -- & -- & -- & -- \\
  GroundingPI & 75.50 & 75.32 & \textbf{73.77} & 65.78 & 96.15 & 74.82 \\
  \addlinespace[2pt]
  \multicolumn{7}{@{}l}{\hspace{0.4em}\strut\textbf{Vision-Language Models (10B--1T)}} \\
  RoboPoint\textsuperscript{*}\hspace{0.35em}\citep{RoboPoint13B} & 16.10 & 8.40 & -- & -- & -- & -- \\
  GUI-Owl\textsuperscript{*}\hspace{0.35em}\citep{GUIOwl32B} & -- & -- & -- & 58.00 & -- & -- \\
  Gemini-2.5-Pro\textsuperscript{*}\hspace{0.35em}\citep{Gemini2.5Pro} & 35.60 & 27.10 & -- & -- & -- & -- \\
  Molmo-72B\textsuperscript{*}\hspace{0.35em}\citep{Molmo} & 30.25 & 21.20 & -- & -- & -- & -- \\
  Qwen3.8-27B\hspace{0.35em}\citep{qwen38} & 60.00 & 46.75 & 63.93 & 58.76 & 94.50 & 63.48 \\
  \addlinespace[2pt]
  \multicolumn{7}{@{}l}{\hspace{0.4em}\strut\textbf{Vision-Language Models (>1T)}} \\
  Qwen3.7-Max\hspace{0.35em}\citep{qwen37} & 68.75 & 57.14 & 69.67 & 55.06 & 81.13 & 49.29 \\
  Kimi-K2.6 & 41.33 & 41.56 & 30.33 & 6.07\textsuperscript{\textdagger} & 52.36\textsuperscript{\textdagger} & 10.11\textsuperscript{\textdagger} \\
  Kimi-K3\hspace{0.35em}\citep{KimiK3} & 58.92 & 54.98 & 54.92 & 25.36\textsuperscript{\textdagger} & 82.70\textsuperscript{\textdagger} & 68.26\textsuperscript{\textdagger} \\
  GPT-6 Astra & \textbf{85.93} & \textbf{81.93} & 65.69 & \textbf{93.17} & \textbf{97.88} & \textbf{86.70} \\
  \addlinespace[2pt]
  \bottomrule
  
  \end{NiceTabular}%
  }
  \endgroup
\end{table}

\paragraph{OCR and document layout.}
\Cref{tab:main-ocr-layout} extends the same structured interface to text and document regions: GroundingPI reaches 72.47 on SROIE and 74.82 on M6Doc, compared with Astra's 53.57 and 60.59. The gains are not uniform: Astra remain stronger on TotalText, and SenseNova-Vision slightly leads on DocLayNet.
\begin{table}[htbp]
  \centering
  \BenchmarkTableFont
  \caption{\textbf{OCR and document layout (F1mIoU).} External rows are from \citep{rexomni}. SenseNova-Vision uses published HierText/ICDAR2015 scores~\citep{SenseNovaVision7BMoT} and locally evaluated TotalText/SROIE scores. Full results appear in \Cref{app:ocr,app:layout_grounding}.}
  \label{tab:main-ocr-layout}
  \begingroup
  \fontsize{8}{9.7}\selectfont
  \setlength{\tabcolsep}{3pt}
  \renewcommand{\arraystretch}{1.15}
  \resizebox{\linewidth}{!}{%
  \begin{NiceTabular}{@{}>{\raggedright\arraybackslash}p{177pt}cccccc@{}}
  \CodeBefore
    \rowcolor{benchmarktype}{3}
    \rowcolor{benchmarktype}{6}
    \rowcolor{benchmarkpurple}{13}
    \rowcolor{benchmarktype}{14}
    \rowcolor{benchmarktype}{17}
  \Body
  \toprule
  \textbf{Model} & \multicolumn{4}{c}{\textbf{OCR}} & \multicolumn{2}{c}{\textbf{Layout grounding}} \\
  \cmidrule(lr){2-5}\cmidrule(lr){6-7}
   & \BenchHead{HierText} & \BenchHead{ICDAR2015} & \BenchHead{TotalText} & \BenchHead{SROIE} & \BenchHead{DocLayNet} & \BenchHead{M6Doc} \\
  \midrule
  \multicolumn{7}{@{}l}{\hspace{0.4em}\strut\textbf{Closed-set Specialized Detectors}} \\
  DocLayout-YOLO\textsuperscript{*}\hspace{0.35em}\citep{DocLayoutYOLO} & -- & -- & -- & -- & 81.10 & -- \\
  PaddleOCRv5\textsuperscript{*}\hspace{0.35em}\citep{PaddleOCRv5} & 30.50 & 25.60 & 25.70 & 58.60 & -- & -- \\
  \addlinespace[2pt]
  \multicolumn{7}{@{}l}{\hspace{0.4em}\strut\textbf{Vision-Language Models (<10B)}} \\
  Qwen3-VL-4B\hspace{0.35em}\citep{qwen3vl} & 23.48 & 28.41 & 38.35 & 40.41 & 40.81 & 24.73 \\
  Qwen3.5-9B\hspace{0.35em}\citep{qwen35} & 29.63 & 29.90 & 37.26 & 26.74 & 34.65 & 17.35 \\
  Rex-Omni\hspace{0.35em}\citep{rexomni} & 34.46 & 45.65 & 52.35 & 48.35 & 68.06 & 54.95 \\
  LocateAnything Hybrid\hspace{0.35em}\citep{locateanything} & 26.65 & 27.48 & 45.49 & 30.05 & 77.34 & 65.94 \\
  SenseNova-Vision\hspace{0.35em}\citep{SenseNovaVision7BMoT} & 31.20\textsuperscript{*} & 49.50\textsuperscript{*} & 11.40 & 36.26 & \textbf{85.53} & 35.62 \\
  RynnBrain1.1\hspace{0.35em}\citep{RynnBrain112B} & 2.07 & 17.81 & 16.90 & 3.59 & 6.02 & 4.53 \\
  GroundingPI & 41.70 & \textbf{55.68} & 49.32 & \textbf{72.47} & 85.08 & \textbf{74.82} \\
  \addlinespace[2pt]
  \multicolumn{7}{@{}l}{\hspace{0.4em}\strut\textbf{Vision-Language Models (10B--1T)}} \\
  SEED1.5-VL\textsuperscript{*}\hspace{0.35em}\citep{SEED15VL} & 12.00 & 18.70 & 19.50 & 28.10 & 28.70 & 28.00 \\
  Qwen3.8-27B\hspace{0.35em}\citep{qwen38} & 32.95 & 39.72 & 41.57 & 37.24 & 42.21 & 19.92 \\
  \addlinespace[2pt]
  \multicolumn{7}{@{}l}{\hspace{0.4em}\strut\textbf{Vision-Language Models (>1T)}} \\
  Qwen3.7-Max\hspace{0.35em}\citep{qwen37} & \textbf{42.95} & 43.07 & 48.64 & 38.54 & 46.93 & 32.31 \\
  Kimi-K2.6 & 25.26\textsuperscript{\textdagger} & 32.25\textsuperscript{\textdagger} & 40.99\textsuperscript{\textdagger} & 46.83\textsuperscript{\textdagger} & 15.59\textsuperscript{\textdagger} & 13.50\textsuperscript{\textdagger} \\
  GPT-6 Astra & 39.58 & 48.87 & \textbf{53.55} & 53.57 & 77.54 & 60.59 \\
  \addlinespace[2pt]
  \bottomrule
  
  \end{NiceTabular}%
  }
  \endgroup
\end{table}

\paragraph{Object pointing.}
Following Rex-Omni~\citep{rexomni}, SAM-derived object masks~\citep{SAM} determine point correctness, with F1@Point balancing missed objects and false positives. GroundingPI leads six of the seven columns in \Cref{tab:main-object-pointing}; Astra leads Dense200 pointing, despite GroundingPI's stronger box result, showing that point selection and boundary precision remain distinct challenges.
\begin{table}[htbp]
  \centering
  \BenchmarkTableFont
  \caption{\textbf{Object pointing (F1@Point).} External rows are from \citep{rexomni}, where Molmo denotes Molmo-7B-D. Full results appear in \Cref{app:object_pointing}.}
  \label{tab:main-object-pointing}
  \begingroup
  \fontsize{8}{9.7}\selectfont
  \setlength{\tabcolsep}{3pt}
  \renewcommand{\arraystretch}{1.15}
  \resizebox{\linewidth}{!}{%
  \begin{NiceTabular}{@{}>{\raggedright\arraybackslash}p{177pt}ccccccc@{}}
  \CodeBefore
    \rowcolor{benchmarktype}{3}
    \rowcolor{benchmarktype}{5}
    \rowcolor{benchmarkpurple}{13}
    \rowcolor{benchmarktype}{14}
    \rowcolor{benchmarktype}{17}
  \Body
  \toprule
  \textbf{Model} & \multicolumn{3}{c}{\textbf{Referring object pointing}} & \multicolumn{2}{c}{\textbf{Common / long-tailed}} & \multicolumn{2}{c}{\textbf{Dense / tiny}} \\
  \cmidrule(lr){2-4}\cmidrule(lr){5-6}\cmidrule(lr){7-8}
   & \BenchHead{HumanRef} & \BenchHead{RefCOCOg val} & \BenchHead{RefCOCOg test} & \BenchHead{COCO} & \BenchHead{LVIS} & \BenchHead{Dense200} & \BenchHead{VisDrone} \\
  \midrule
  \multicolumn{8}{@{}l}{\hspace{0.4em}\strut\textbf{Open-set Specialized Detectors}} \\
  GroundingDINO\hspace{0.35em}\citep{groundingdino} & 46.85 & 49.34 & 49.97 & 70.41 & 55.07 & 32.93 & 39.45 \\
  \addlinespace[2pt]
  \multicolumn{8}{@{}l}{\hspace{0.4em}\strut\textbf{Vision-Language Models (<10B)}} \\
  Qwen3-VL-4B\hspace{0.35em}\citep{qwen3vl} & 66.89 & 76.43 & 77.64 & 65.33 & 55.08 & 21.72 & 23.50 \\
  Qwen3.5-9B\hspace{0.35em}\citep{qwen35} & 78.21 & 77.59 & 77.85 & 72.21 & 64.00 & 65.35 & 44.73 \\
  Rex-Omni\hspace{0.35em}\citep{rexomni} & 83.40 & 84.96 & 85.32 & 79.74 & 70.04 & 76.66 & 51.97 \\
  LocateAnything Hybrid\hspace{0.35em}\citep{locateanything} & 71.44 & 75.89 & 76.65 & 73.78 & 64.89 & 78.07 & 57.30 \\
  SenseNova-Vision\hspace{0.35em}\citep{SenseNovaVision7BMoT} & 74.04 & 74.63 & 75.42 & 72.96 & 62.66 & 78.71 & 61.81 \\
  RynnBrain1.1\hspace{0.35em}\citep{RynnBrain112B} & 62.53 & 74.42 & 74.17 & 25.70 & 17.56 & 3.95 & 13.18 \\
  Molmo-7B\textsuperscript{*}\hspace{0.35em}\citep{Molmo} & 70.00 & 83.70 & 83.60 & 77.30 & 40.30 & 33.10 & 29.20 \\
  GroundingPI & \textbf{88.79} & \textbf{90.26} & \textbf{90.06} & \textbf{84.79} & \textbf{79.51} & 81.27 & \textbf{67.47} \\
  \addlinespace[2pt]
  \multicolumn{8}{@{}l}{\hspace{0.4em}\strut\textbf{Vision-Language Models (10B--1T)}} \\
  SEED1.5-VL\textsuperscript{*}\hspace{0.35em}\citep{SEED15VL} & 83.10 & 83.60 & 84.20 & 78.20 & 70.70 & 72.10 & 56.70 \\
  Qwen3.8-27B\hspace{0.35em}\citep{qwen38} & 74.16 & 75.84 & 75.86 & 74.01 & 67.65 & 74.55 & 51.16 \\
  \addlinespace[2pt]
  \multicolumn{8}{@{}l}{\hspace{0.4em}\strut\textbf{Vision-Language Models (>1T)}} \\
  Qwen3.7-Max\hspace{0.35em}\citep{qwen37} & 81.30 & 71.21 & 72.40 & 72.13 & 66.94 & 66.28 & 59.04 \\
  Kimi-K2.6 & 55.97\textsuperscript{\textdagger} & 39.59\textsuperscript{\textdagger} & 39.60\textsuperscript{\textdagger} & 30.65\textsuperscript{\textdagger} & 25.51\textsuperscript{\textdagger} & 35.18\textsuperscript{\textdagger} & 13.63\textsuperscript{\textdagger} \\
  GPT-6 Astra & 83.83 & 87.80 & 84.90 & 82.17 & 77.14 & \textbf{86.57} & 65.62 \\
  \addlinespace[2pt]
  \bottomrule
  
  \end{NiceTabular}%
  }
  \endgroup
\end{table}

\paragraph{Additional visual-prompt capability.} Exemplar-based visual prompting on FSC147, Dense200, COCO, and LVIS is evaluated as an additional capability; the complete comparisons are provided in Appendix~\ref{app:visual_prompting}.

\subsection{Physical Intelligence Performance}
\label{sec:physical_intelligence_performance}

\begin{figure}[!tbp]
  \centering
  \includegraphics[width=\linewidth]{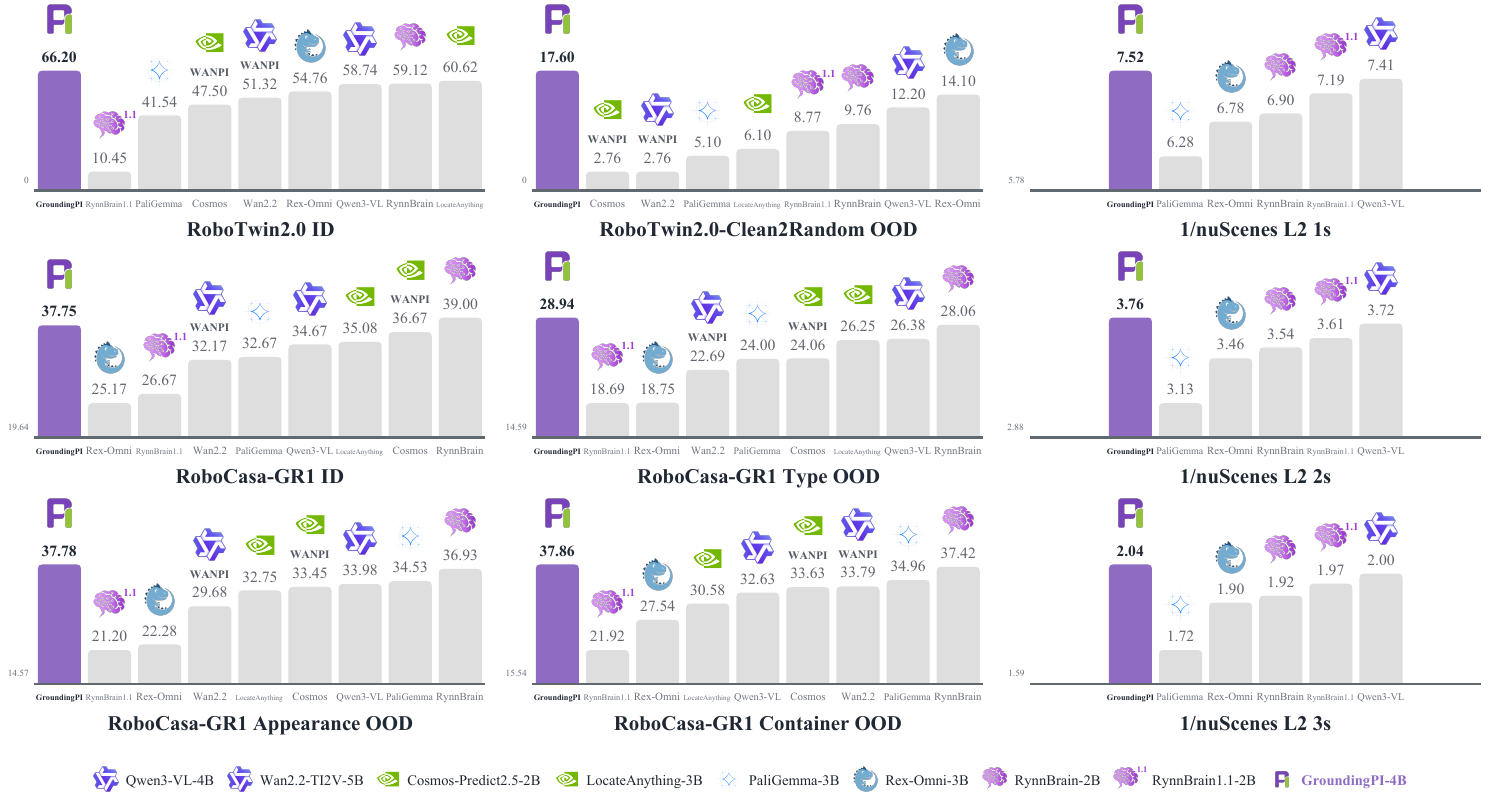}
  \caption{\textbf{Transfer to physical intelligence.} Manipulation success rates and reciprocal nuScenes open-loop L2 errors (higher is better in all panels). Manipulation comparisons fix the action expert, action data, and training budget within each benchmark.}
  \label{fig:physical-intelligence-benchmarks}
\end{figure}

We evaluate autonomous driving on nuScenes~\citep{nuscenes} and manipulation on RoboTwin 2.0~\citep{chen2025robotwin} and RoboCasa-GR1~\citep{nasiriany2024robocasa,nvidia2025gr00tn1}. \Cref{fig:physical-intelligence-benchmarks} shows lower driving error at every reported horizon and the highest success rate in five of six manipulation settings, including all four OOD settings. These results highlight the value of grounding perception for physical intelligence across distinct action learners and embodiments.

\subsubsection{Autonomous Driving Performance}
\label{sec:driving_performance}

GroundingPI attains an average open-loop L2 error of 0.296\,m, improving on Qwen3-VL-4B (0.301\,m) and RynnBrain (0.308\,m); the trajectory interface and metric are defined in \Cref{app:driving_setup}. This modest, consistent gain suggests that embodied specialization alone need not supply the perception most useful for driving. Reading road signs also motivates accurate and timely OCR in vision-based driving, although this evaluation does not isolate sign reading or measure its latency. For latency-sensitive execution, these results motivate compact perceptual foundations; they do not establish closed-loop performance or a deployment limit for larger models.

\subsubsection{Robot Manipulation Performance}
\label{sec:manipulation_performance}

\question{2.1}{Can a perceptual foundation support both action learning and generalization?}

\paragraph{Controlled Backbone Comparison.}
We evaluate the vision-language and video-generation foundations summarized in Figure~\ref{fig:physical-intelligence-benchmarks}, spanning the backbone families used in VLA and WAM systems. To compare their contribution to action learning under controlled downstream conditions, we couple every backbone to the same layer-wise Action DiT. Following the conditioning design of $\pi_0$~\citep{black2024pi0}, intermediate features from a single backbone forward pass are projected and resampled to a common interface, then injected into the corresponding cross-attention blocks of the action expert. Within each benchmark, all runs keep the Action DiT architecture, action-training data, optimization budget, action representation, and inference procedure fixed, and vary only the pretrained foundation. This controlled comparison allows us to assess how effectively each foundation supports downstream action learning and how well the resulting policies generalize out of distribution. \Cref{app:manipulation_setup} provides the implementation.

\paragraph{Evaluation Protocol.}
We report task success rate (SR) across two representative embodiments: fixed-base bimanual tabletop manipulation on RoboTwin2.0 and humanoid dexterous-hand manipulation on RoboCasa-GR1. In-distribution (ID) evaluation follows the full training and evaluation protocol of each benchmark. For out-of-distribution (OOD) evaluation, policies are trained on ID data and tested under held-out conditions.

\begin{itemize}[leftmargin=1.5em]
    \item \textbf{Fixed-base bimanual --- RoboTwin2.0.} This widely adopted benchmark spans 50 tabletop manipulation tasks~\citep{chen2025robotwin}. \textbf{RoboTwin2.0-Full} trains and evaluates on both Clean and Randomized data, whereas \textbf{RoboTwin2.0-Clean2Random} trains only on Clean demonstrations and evaluates on Randomized scenes. The released Randomized setting retains the same manipulation tasks but varies scene clutter, lighting, table and background textures, tabletop height, and language instructions. It therefore primarily evaluates scene robustness under environmental variation.
    \item \textbf{Humanoid dexterous-hand --- RoboCasa-GR1.} This benchmark evaluates fine-grained manipulation from head-camera observations without wrist cameras~\citep{nasiriany2024robocasa,nvidia2025gr00tn1}. Its full protocol measures ID performance, while OOD evaluation uses three held-out suites~\citep{chen2026unit}: \textbf{Unseen Appearance} applies novel textures to familiar object--container pairs; \textbf{Unseen Combinations} places seen objects in novel container pairings; and \textbf{Unseen Object Types} introduces novel object categories. These suites evaluate entity generalization across object appearances, object categories, and object--container combinations.
\end{itemize}

Together, the two protocols evaluate scene robustness and entity generalization, two complementary requirements of an embodied foundation model. Both require stable, task-conditioned perception before action learning, making them direct tests of whether a pretrained foundation provides reusable representations for manipulation.

\paragraph{Overall Performance.}
Under the matched downstream architecture and training protocol, Figure~\ref{fig:physical-intelligence-benchmarks} compares backbones with distinct pretraining objectives. We organize the compared backbones into four groups: \textbf{general-purpose VLMs} (Qwen3-VL-4B~\citep{qwen3vl} and PaliGemma-3B~\citep{Paligemma}), \textbf{video-generation backbones} (Wan2.2-TI2V-5B~\citep{WAN} and Cosmos-Predict2.5-2B~\citep{CosmosPredict25}), \textbf{grounding models} (LocateAnything-3B~\citep{locateanything} and Rex-Omni-3B~\citep{rexomni}), and \textbf{embodied foundation models} (RynnBrain~\citep{dang2026rynnbrain} and RynnBrain1.1~\citep{RynnBrain112B}). Across these heterogeneous pretraining sources, GroundingPI ranks first in five of the six evaluations, including RoboTwin2.0 Full and every OOD setting. It achieves 66.20\% on RoboTwin2.0 Full and remains competitive on RoboCasa-GR1 Full at 37.75\%, compared with RynnBrain's 39.00\%. These results support precise spatial perception as an effective foundation for action learning across two robot embodiments, with consistent performance under distribution shift.

\paragraph{Generalization under Complementary Shifts.}
RoboTwin2.0-Clean2Random evaluates scene robustness, while RoboCasa-GR1 evaluates entity generalization across appearances, object types, and object--container relations. GroundingPI ranks first in all four OOD settings. On RoboTwin2.0, it reaches 17.60\%, with Rex-Omni second at 14.10\%. On RoboCasa-GR1, RynnBrain ranks second across Type, Appearance, and Container OOD. Rex-Omni's strength in scene robustness is consistent with its emphasis on detection, referring, and coordinate prediction~\citep{rexomni}. RynnBrain's strength in entity generalization is consistent with its broader language-conditioned embodied and semantic understanding, including spatiotemporal localization and physically grounded reasoning~\citep{dang2026rynnbrain}. These observations highlight the complementary strengths of precise perceptual grounding and broader embodied semantic understanding.

\begin{figure}[!tbp]
  \centering
  \includegraphics[width=1.0\linewidth]{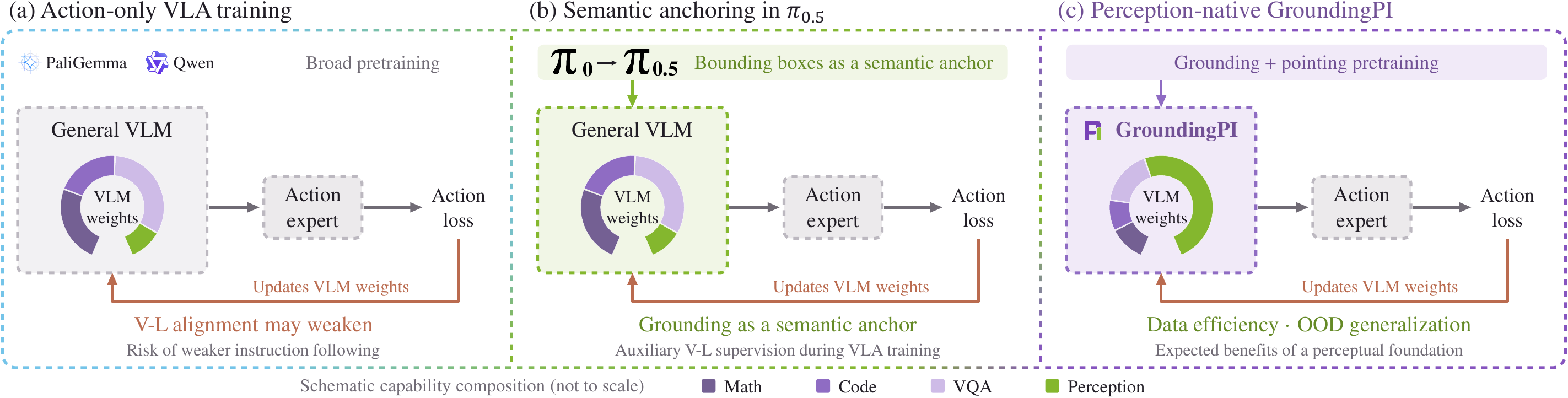}
  \caption{\textbf{Perception before action.} A conceptual progression toward a foundation with native grounding capabilities. Segment sizes are illustrative and do not represent measured parameter or data proportions.}
  \label{fig:vla-representation-concept}
\end{figure}

\paragraph{Perception Before Action.}
\Cref{fig:vla-representation-concept} illustrates a conceptual progression from action-only adaptation of general VLMs, through perceptual supervision in the $\pi$ family, to grounding as a native foundation capability~\citep{black2024pi0,intelligence2025pi05visionlanguageactionmodelopenworld,driess2025knowledgeinsulatingvisionlanguageactionmodels}. GroundingPI retains broad VQA-style, language-conditioned semantic knowledge while emphasizing precise spatial perception through grounding and pointing supervision.  Visual primitives provide an interface for physical prompting, specifying targets, locations, and structures alongside language. Together with precise spatial perception, this interface supports scene robustness and entity generalization. Our transfer results evaluate the pretrained foundation for physical intelligence rather than a separate prompting intervention. The data-composition ablation in \Cref{sec:grounding_to_action_transfer} further examines which forms of grounding supervision transfer to action.

\finding{2.1}{Grounding provides an effective perceptual interface that supplies \emph{where to act} for learning \emph{how to act}, improving action learning and generalization.}

\subsubsection{Data Efficiency}
\label{sec:data_efficiency}

\question{2.2}{Does a perception-native foundation enable more efficient action learning from limited robot demonstrations?}

\FinishArxivWrap
\begin{wrapfigure}{R}{0.48\textwidth}
\centering
    \includegraphics[width=\linewidth]{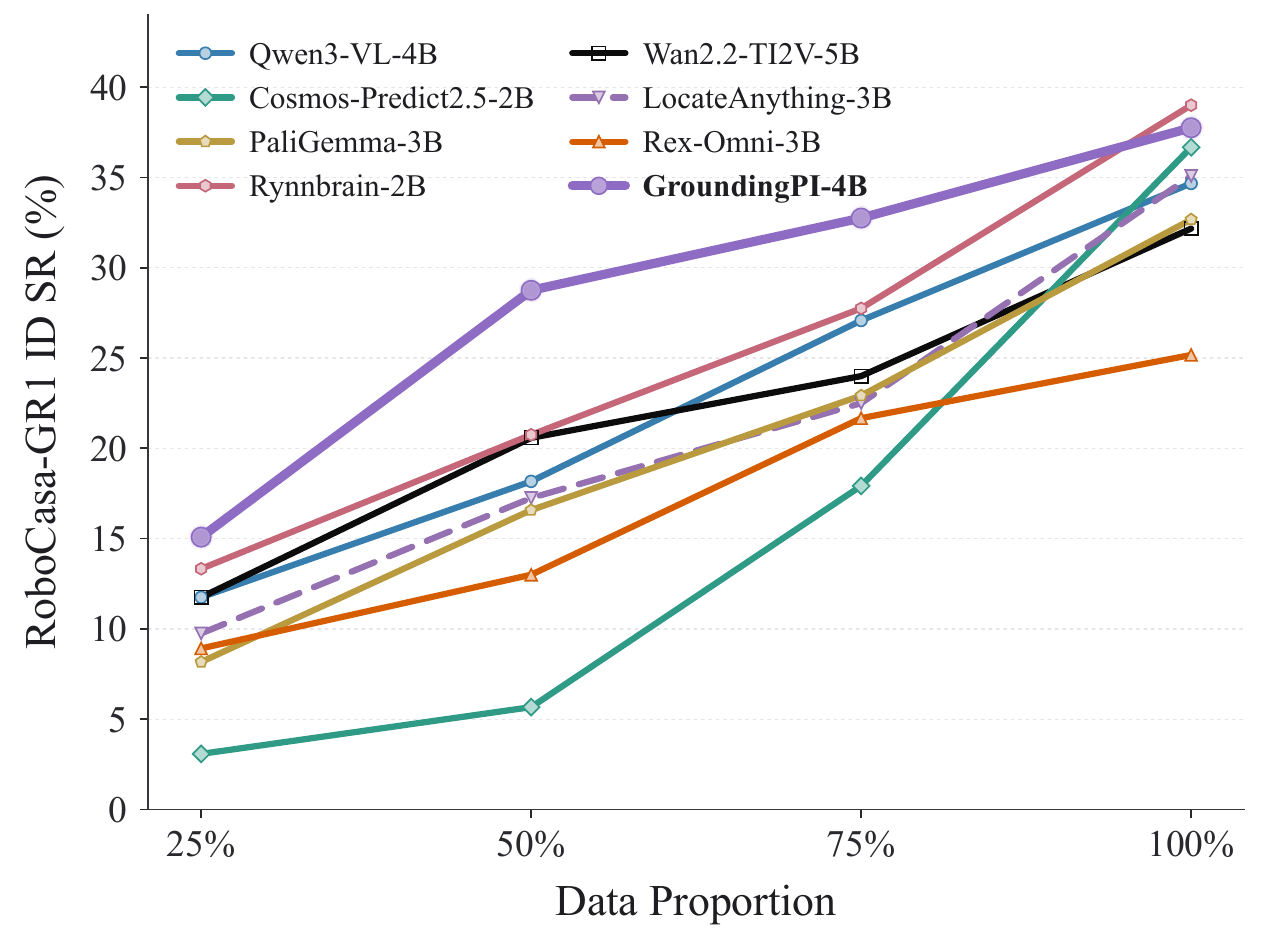}
    \caption{\textbf{Action-data efficiency on RoboCasa-GR1 ID.} Success rate at four demonstration fractions with the remaining action-learning setup fixed.}
    \label{fig:data-efficiency}
\end{wrapfigure}

Figure~\ref{fig:data-efficiency} evaluates data efficiency under the RoboCasa-GR1 Full/ID protocol using 25\%, 50\%, 75\%, and 100\% of the robot demonstrations. We vary only the amount of action data, keeping the architecture, action-learning approach, and other training choices unchanged. GroundingPI leads all three reduced-data settings. At 50\% data, its 28.75\% SR exceeds every baseline at 75\%, where RynnBrain achieves the highest score of 27.75\%.

These results support prioritizing precise spatial perception before action learning. The same physical-prompting interface allows scarce action demonstrations to focus on control rather than relearning basic perception. Together with the preceding results, this suggests that precise grounding supports effective and generalizable action learning as well as more data-efficient adaptation.

\FinishArxivWrap

\finding{2.2}{Precise, language-conditioned grounding provides an effective, generalizable, and data-efficient interface for action learning.}

\takeaway{1}{A foundation model with broad and precise grounding can provide a strong perceptual basis for physical intelligence.}

\subsection{Ablations and Analysis}
\label{sec:ablations}

\subsubsection{Effects of Grounding Training}
\label{sec:effects_grounding_training}

\question{3.1}{How does pretraining scale affect perceptual capability and its transfer to action learning?}

\FinishArxivWrap
\begin{wrapfigure}[17]{r}{0.48\textwidth}
\centering
    \includegraphics[width=\linewidth]{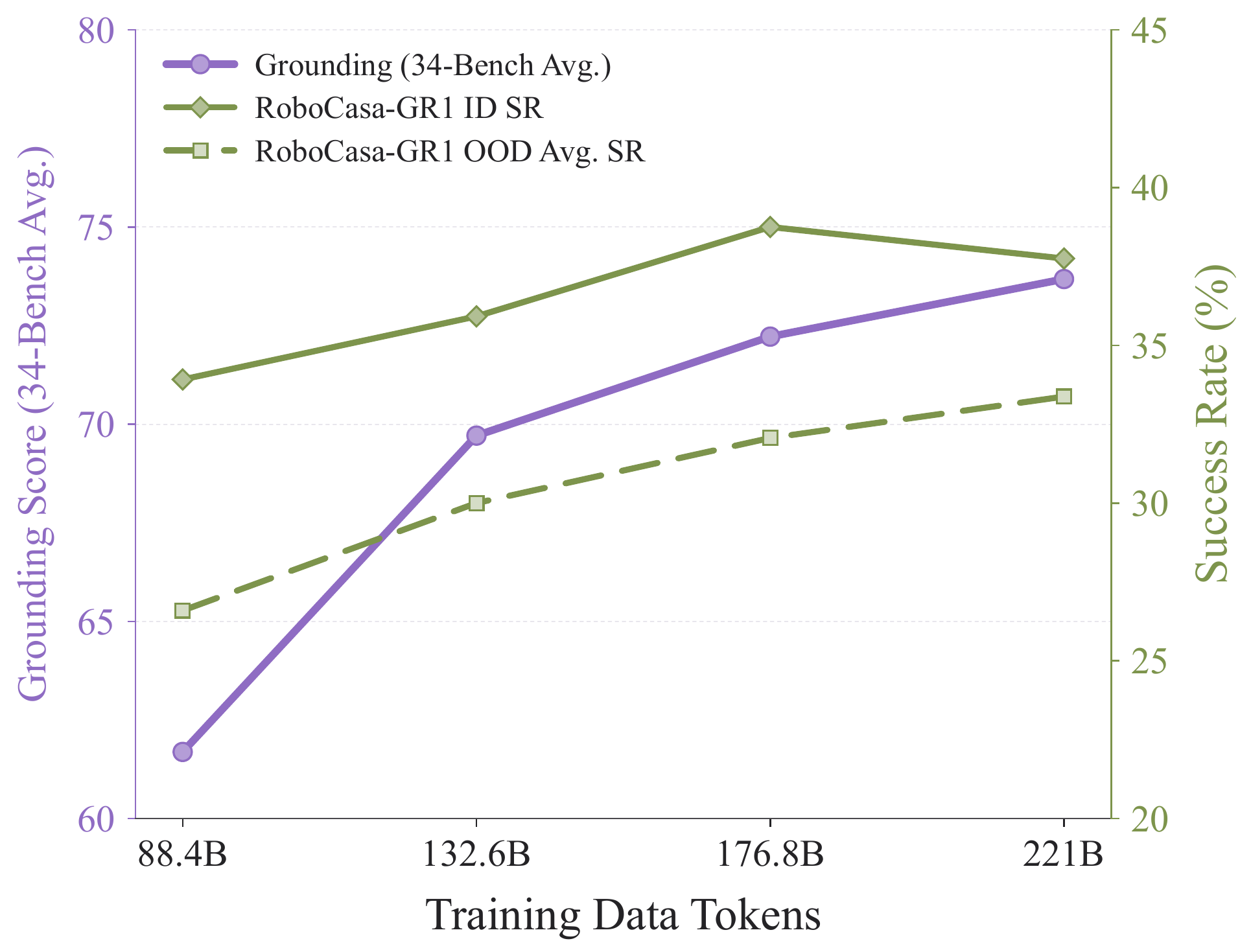}
    \caption{\textbf{Grounding and action scaling.} Grounding Avg and RoboCasa-GR1 ID/OOD success as grounding-training exposure increases under a fixed downstream recipe.}
    \label{fig:pretraining-scaling}
\end{wrapfigure}

Figure~\ref{fig:pretraining-scaling} tracks the aggregate grounding score together with manipulation success on the RoboCasa-GR1 ID and OOD splits as pretraining grows from 88.4B to 221B tokens. This scaling sweep examines how perception and downstream action learning evolve together.

\paragraph{Grounding and Action Learning Improve with Scale.}
As pretraining scale increases, the grounding score rises steadily and robotic manipulation performance improves on both splits. Scaling grounding pretraining therefore strengthens precise perception and transfers to more effective and generalizable action learning.

\paragraph{Grounding Gains Track Action-Learning Gains.}
The curves move together across the scaling trajectory, with generalization tracking grounding quality most closely. This correspondence links the two observations: better grounding provides a stronger perceptual foundation for downstream action learning.

\FinishArxivWrap

\finding{3.1}{Scaling grounding pretraining strengthens both grounding quality and downstream action learning, with generalization tracking the perceptual gains most closely.}

\subsubsection{What Transfers from Grounding to Action?}
\label{sec:grounding_to_action_transfer}

\question{3.2}{Which grounding data compositions best support effective and generalizable downstream action learning?}

\begin{figure}[!tbp]
  \centering
  \includegraphics[width=\linewidth]{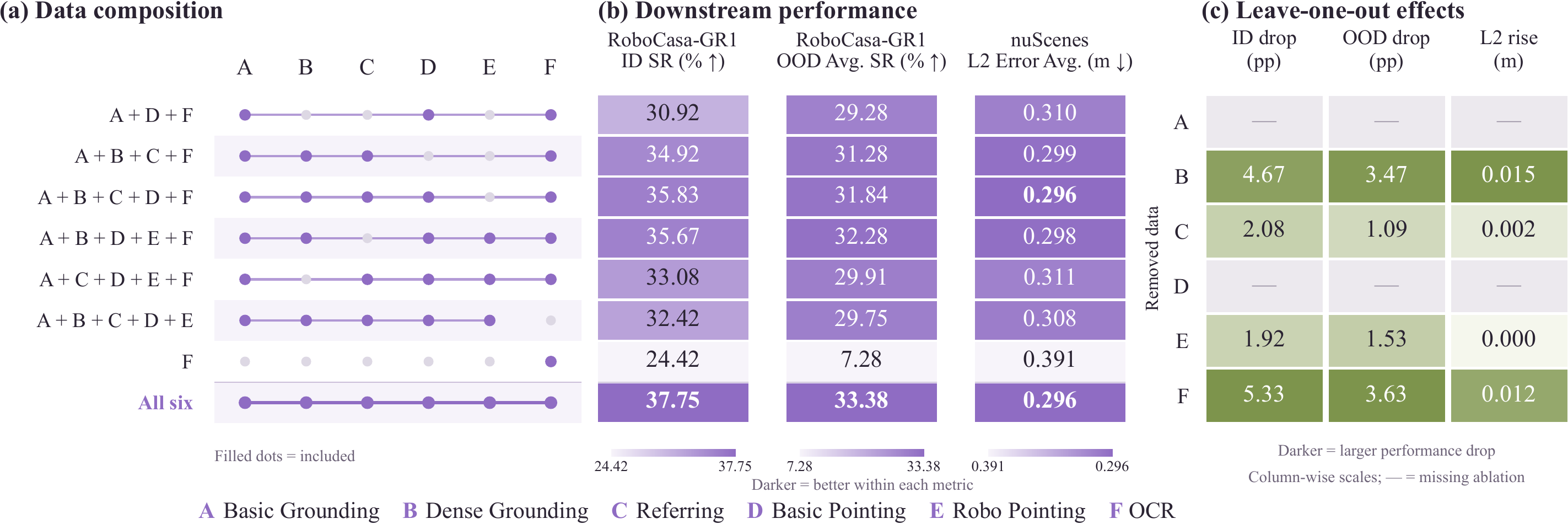}
  \caption{\textbf{Which perceptual supervision transfers?} Included task groups and downstream results. Leave-one-out differences compare with all six groups; -- denotes an unavailable ablation.}
  \label{fig:grounding-data-mix}
\end{figure}

The scaling study in Section~\ref{sec:effects_grounding_training} establishes that more grounding pretraining improves downstream action learning. We now ask what that data should contain. Figure~\ref{fig:grounding-data-mix} decomposes the training mixture into basic and dense grounding, referring, pointing, and auxiliary OCR, layout, and GUI supervision.

\paragraph{Dense Grounding Provides the Spatial Core.}
In the available leave-one-out comparisons, removing dense grounding degrades both manipulation splits and nuScenes performance more than removing referring or robo pointing. In driving and manipulation, relevant targets may be densely packed or small relative to the image, coupling dense and tiny-object perception through a shared need for fine-grained spatial discrimination. Dense multi-object supervision therefore provides the core precise spatial perception that transfers across downstream domains.

\paragraph{Auxiliary Perception Amplifies Grounding: OCR as a Potential Catalyst.}
Removing the OCR-containing group causes the largest manipulation losses on both splits (5.33/3.63 pp), yet this group alone transfers poorly. We hypothesize that OCR catalyzes perceptual learning: sharp character boundaries demand local precision, while transcription binds visual regions to semantics, providing a localized captioning proxy task for vision--language alignment. Such supervision could strengthen fine-grained ViT features and amplify the benefits of dense spatial supervision for grounding and downstream action learning. Additional OCR mixing experiments favor combining cold-start and batch-level supervision for manipulation transfer; the OCR-specific mechanism remains a hypothesis (\Cref{app:ocr_catalyst}).

\paragraph{Complementary Supervision Builds the Strongest Interface.}
No reduced mixture matches the full mixture simultaneously across manipulation performance and nuScenes localization. Dense grounding supplies the spatial core, auxiliary perception sharpens the visual representation, and referring and pointing connect that representation to language and action-relevant targets. Together with the scaling study, these results show that scale determines how much grounding capability is learned, while composition determines whether it forms an effective and generalizable interface for action learning.

\finding{3.2}{Dense grounding provides the strongest spatial core in the available leave-one-out comparisons, while auxiliary OCR, layout, and GUI supervision is most valuable in combination; the full mixture performs best across manipulation and driving.}

\takeaway{2}{Grounding transfer depends on both scale and composition: scale strengthens the perceptual foundation, while complementary supervision determines how effectively it supports downstream action.}

\subsubsection{Discussion}
\label{sec:discussion}

\begingroup
\setlength{\emergencystretch}{1em}
\paragraph{Effects of Base Model.}
\label{sec:effects_base_model}

The backbone comparisons suggest possible sensitivities beyond supervision. RynnBrain\allowbreak{}1.1 trails RynnBrain in all six manipulation settings despite improving driving. Its Qwen3.5 base combines Gated DeltaNet with gated attention, whereas RynnBrain uses Qwen3-VL~\citep{RynnBrain112B,dang2026rynnbrain,qwen35}. Recurrent compression and output gating may affect access to spatial features under a new action objective; \Cref{app:gated_transfer} derives conditional memory and gradient effects~\citep{GatedDeltaNet,GatedAttention}.
\par\endgroup

DeepStack enriches visual evidence through intermediate feature injection, but may also increase the demands of aligning several feature levels with an action readout~\citep{qwen3vl,DeepStack}. GroundingPI instead couples MoonViT-V2 to a full-attention Qwen3-4B decoder through a single visual interface. Since RynnBrain and Qwen3-VL both use DeepStack, their driving difference cannot isolate this mechanism (\Cref{app:deepstack_transfer}). The older Qwen2.5-VL foundations of Rex-Omni and LocateAnything also leave base-model quality as a possible factor. These comparisons motivate controlled studies of feature accessibility and alignment; they do not establish architectural causes of the observed rankings.

\takeaway{3}{Base-model selection should consider architectural effects on training dynamics and the transfer of perceptual capabilities to action.}

\paragraph{Spatial Supervision: A First-Principles Hypothesis.}
From first principles, a useful starting point is what spatial information the visual input can actually determine. An image records projected structure, while its absolute metric scale may remain ambiguous. Human perception illustrates this limitation: we may readily identify and localize a distant object yet struggle to judge whether it is 25\,m or 30\,m away. A raw $L_1$ depth loss nevertheless assigns a 5\,m error, which need not reflect the perceptual difficulty; a fixed metric tolerance also has different implications for close-range manipulation and distant scenes. Under perspective projection, small image-space errors can further translate into larger metric depth and 3D localization errors at longer ranges. These considerations motivate a hypothesis: \emph{grounding may help elicit and develop spatial intelligence in VLMs by forcing spatial understanding to become explicit through precise localization}. Predicting boxes and points ties language to specific visual regions, providing low-level supervision that may compel the model to preserve and use fine-grained spatial evidence. Relative depth, such as scene-normalized values in $[0,1]$, could offer complementary supervision of depth ordering and scene structure without requiring absolute scale. We therefore conjecture that prioritizing these visually grounded targets may better cultivate transferable perception, with metric calibration learned for downstream action. This is a hypothesis about perceptual pretraining, and the proposed advantage over absolute metric supervision remains untested in our study.

\paragraph{Design of Embodied Foundation Models.}

A useful distinction may be between deliberative planning (System~2) and fast perception--action execution (System~1), as explored in dual-system robotics~\citep{nvidia2025gr00tn1,RoboDual}. Planning can benefit from extensive knowledge, coding, and long-horizon reasoning; execution requires timely feedback, precise interaction perception, and reliable control. The analogy to coding agents is functional: sophisticated reasoning has limited practical value when generated programs repeatedly fail to run, just as strong planning can be constrained by unreliable physical execution.

General VLMs, including Qwen and the PaliGemma base of $\pi_0$~\citep{qwen3vl,black2024pi0}, offer valuable starting points, but may not best serve every execution role. GroundingPI explores a perception-native alternative: learning \emph{where to interact} before limited action data teach \emph{how to act}. Astra provides a frontier reference for grounding quality; comparison with it probes the perceptual frontier rather than its suitability for direct VLA adaptation. Such frontier models may instead serve higher-level planning roles. Our results motivate stronger perceptual foundations for System~1; a deployed dual-system controller and its latency remain to be evaluated (\Cref{app:sys1_design}).

\takeaway{4}{Planning and execution may benefit from different foundations: broad reasoning models for System~2, and perception-native foundations for reliable System~1 execution, as explored by GroundingPI.}

\subsubsection{Ablation Study}
\label{sec:ablation_study}

\FinishArxivWrap
\begin{wrapfigure}[16]{r}{0.48\textwidth}
\centering
    \includegraphics[width=\linewidth]{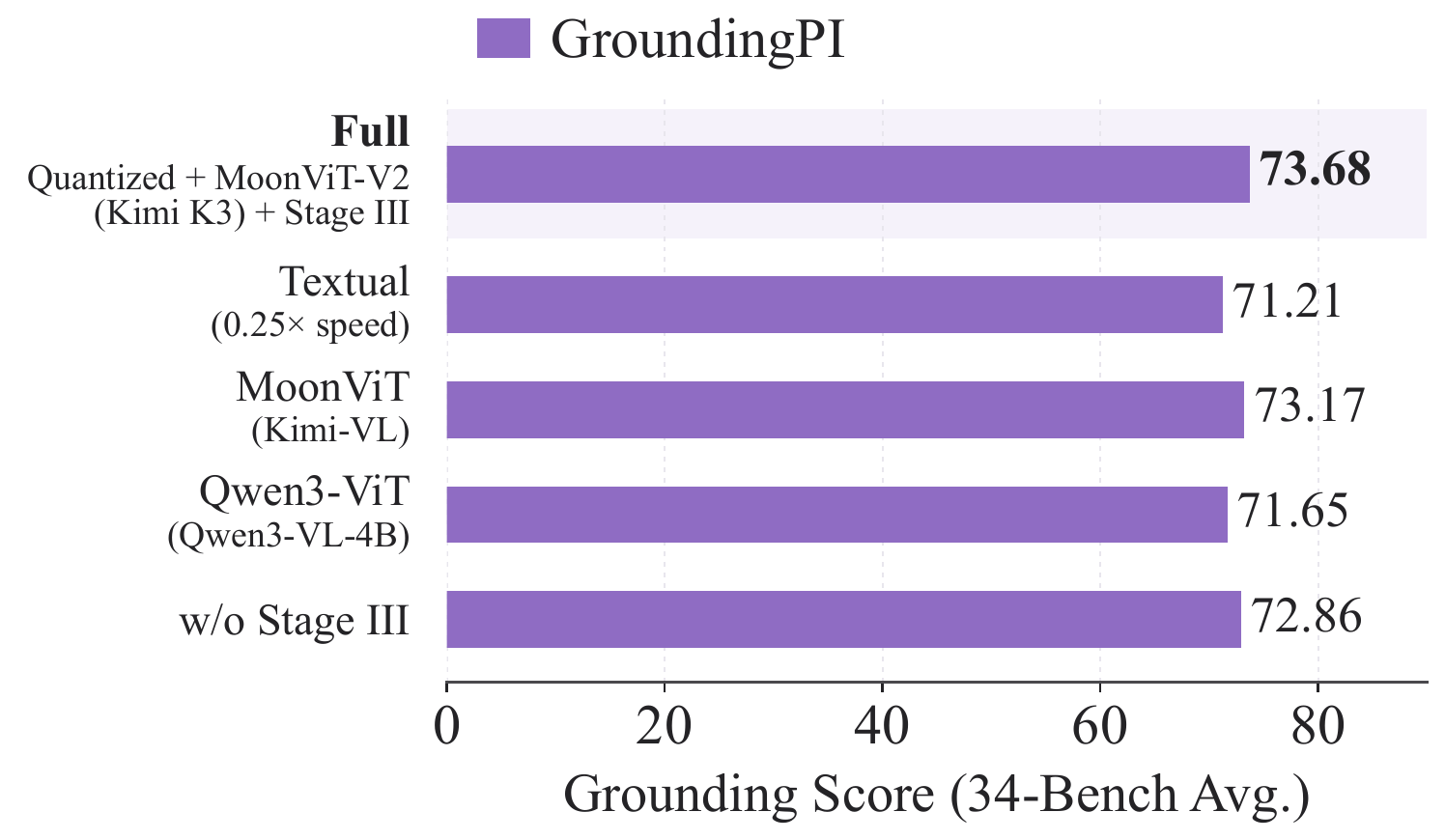}
    \caption{\textbf{Grounding-model ablations.} Effects of coordinate representation, visual encoder, and Stage III reinforcement post-training on grounding Avg.}
    \label{fig:grounding-model-ablations}
\end{wrapfigure}

\Cref{fig:grounding-model-ablations} supports the chosen grounding architecture and training recipe. Quantized coordinates improve Avg from 71.21 to 73.68, while textual coordinates run at 0.25$\times$ the relative speed. MoonViT-V2 reaches 73.68, versus 73.17 with MoonViT and 71.65 with Qwen3-ViT; potential implications for visual alignment are discussed in \Cref{app:deepstack_transfer}. Removing Stage III reduces Avg to 72.86. These comparisons support the complete design choices without isolating every architectural difference.

\begin{figure}[!tbp]
  \centering
  \includegraphics[width=\linewidth]{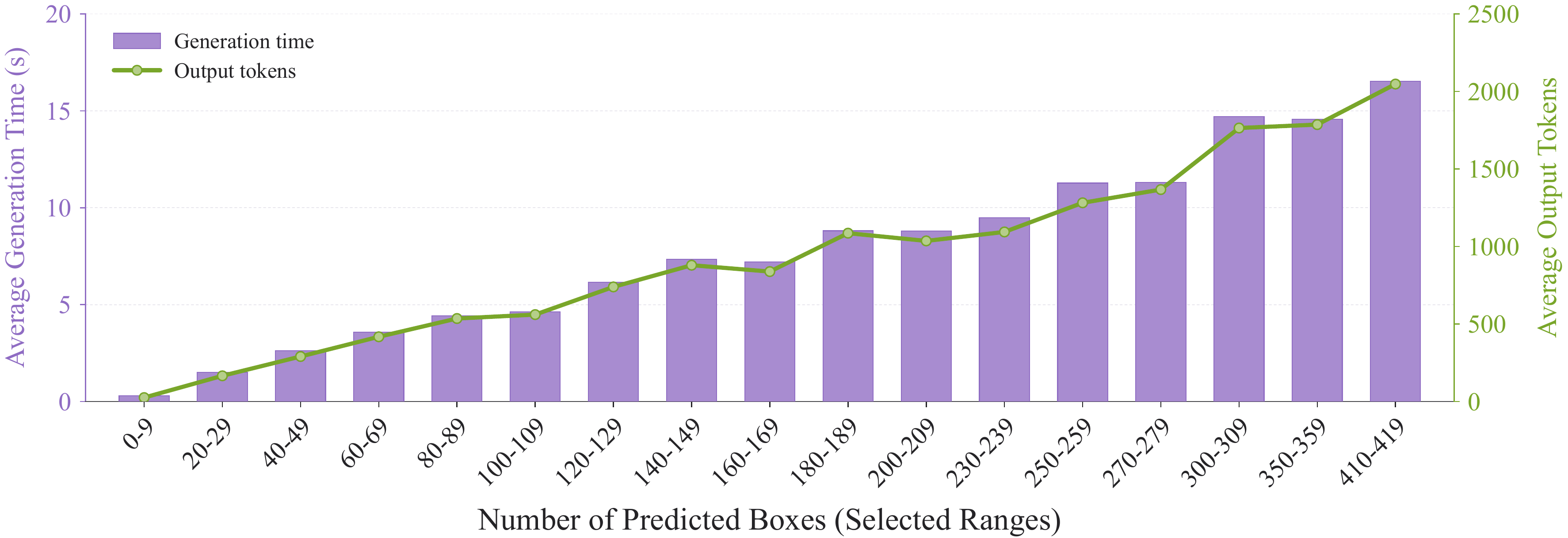}
  \caption{\textbf{Generation cost versus predicted object count.} GroundingPI generation time and output-token count across box-count ranges.}
  \label{fig:efficiency-report}
\end{figure}

\Cref{tab:grounding-token-efficiency} shows compact serialization: GroundingPI uses 7.6/5.1 tokens per box on COCO/Dense200, versus 148.8/74.5 for SEED1.5-VL. \Cref{fig:efficiency-report} separately relates GroundingPI's generation time and output length to predicted object count. Shared protocol overhead is amortized in dense outputs, while autoregressive generation cost remains (\Cref{app:efficiency_analysis}).

\FinishArxivWrap

\Needspace{5\baselineskip}
\section{Conclusion}
\label{sec:conclusion}

We introduced GroundingPI, a 4B-parameter grounding foundation model for broad and precise perception. Across 34 grounding benchmarks, it outperforms similarly sized models and remains competitive with GPT-6 Astra, while transferring effectively to driving and manipulation.

\begin{wraptable}[9]{r}{0.6\textwidth}
\centering
  \BenchmarkTableFont
  \caption{\textbf{Output token efficiency.} SEED1.5-VL values are from \citep{rexomni}. This cross-model comparison measures serialization rather than matched latency.}
  \label{tab:grounding-token-efficiency}
  \begingroup
  \small
  \setlength{\tabcolsep}{5pt}
  \renewcommand{\arraystretch}{1.15}
  \definecolor{efficiencypurple}{HTML}{F5F1FA}
  \definecolor{efficiencygreen}{HTML}{F0F3E8}
  \resizebox{\linewidth}{!}{%
    \begin{NiceTabular}{@{}lcccccc@{}}
      \CodeBefore
        \rowcolor{efficiencypurple}{4}
      \Body
      \toprule
      & \multicolumn{3}{c}{\textbf{COCO}} & \multicolumn{3}{c}{\textbf{Dense200}} \\
      \cmidrule(lr){2-4}\cmidrule(lr){5-7}
      Model & Boxes/img & Tokens/img & Tokens/box & Boxes/img & Tokens/img & Tokens/box \\
      \midrule
      SEED1.5-VL\hspace{0.35em}\citep{SEED15VL} & 4.2 & 631.0 & 148.8 & 73.1 & 5446.3 & 74.5 \\
      GroundingPI & \textbf{6.0} & \textbf{45.6} & \textbf{7.6} & \textbf{87.2} & \textbf{444.7} & \textbf{5.1} \\
      \bottomrule
    \end{NiceTabular}%
  }
  \endgroup
\end{wraptable}

Reliable physical interaction requires connecting instructions to precise spatial targets, a capability that general semantic understanding alone does not guarantee and scarce action supervision may not adequately develop. Dedicated grounding pretraining directly addresses this perceptual gap, supporting better manipulation generalization and action-data efficiency. Our analyses highlight basic grounding as an important foundation, dense grounding as valuable for physical interaction, and OCR as a potential catalyst for perceptual learning. These findings motivate distinct embodied foundations: broad reasoning for System~2 planning, and precise perception for reliable System~1 execution, the direction explored by GroundingPI.

\subsection*{AI Use Statement}
In this work, we used generative AI tools to assist with code development, to polish the writing, and to produce some of the figures. We also used generative AI models as part of the data annotation pipeline to generate pseudo-labels for model training. We did not use generative AI tools to develop the research ideas or methodology, to design or interpret the experiments, or to outline the paper. We have reviewed all AI-assisted work. We take responsibility for the final content of this work, including text, claims, data annotations, or artifacts produced with the aid of generative AI.

\subsection*{Ethics Statement}
This work does not involve human-subject studies. All grounding datasets and evaluation benchmarks used in this work were obtained from publicly available or appropriately licensed sources in accordance with their respective terms and licenses. Driving was evaluated offline on nuScenes, and robot manipulation in simulation; no real-world vehicles or robots were deployed in these experiments. GroundingPI is developed as a general-purpose visual grounding foundation model for research, with downstream experiments intended to study how precise spatial perception transfers to physical-intelligence tasks rather than to demonstrate real-world autonomous deployment. We encourage appropriate safety evaluation and human oversight before applying such models to real-world embodied systems. The authors declare no conflicts of interest.

\subsection*{Reproducibility Statement}

We will publicly release the code, model weights, and evaluation data for GroundingPI to support systematic and reproducible evaluation. The release will include training configurations and evaluation scripts with standardized task prompts, output parsers, and metric implementations. The input/output protocol, coordinate conventions, architecture, tokenizer, and visual processing are detailed in \Cref{app:groundingpi_protocol,app:groundingpi_architecture}; data construction and validation in \Cref{app:groundingpi_data_engine}; and base-model training, spatial supervised fine-tuning, and optimization hyperparameters in \Cref{app:groundingpi_vlm_training}. Downstream interfaces, the shared action architecture, matched training budgets, and driving and manipulation evaluation protocols are specified in \Cref{app:driving_setup,app:manipulation_setup}. Data-efficiency, scaling, training-mixture, and OCR-mixing studies are documented in \Cref{app:data_efficiency_analysis,app:scaling_analysis,app:composition_analysis,app:ocr_catalyst}. The benchmark suite, evaluation metrics, reporting conventions, and complete grounding results are provided in \Cref{app:grounding_results}, including the visual-prompting evaluations in \Cref{app:visual_prompting}.

\subsection*{Research Scope, Data Use, and Institutional Disclaimer}

This work originated from exploratory academic research undertaken by the project leader Qize Yu during their internship at Xpeng Inc. The project was conducted exclusively for scientific investigation and academic publication and does not involve commercial applications, product development, or commercial deployment. All data used in this project were used solely for academic research and maintained under strict segregation from the company’s commercial model development and deployment activities. No project data were used to train, fine-tune, evaluate, or otherwise support commercial models, products, or services. Internal legal review of the dataset materials was completed on September 21, 2026. This work neither uses nor discloses business data containing users’ private or personally identifiable information. The research-only scope described here does not modify or supersede the applicable terms and licenses of the source datasets.

The views, methods, findings, and conclusions presented in this paper are those of the authors and do not represent the official positions, technical direction, product roadmap, or commercial commitments of Xpeng Inc. The company’s support for this research should not be construed as endorsement of any commercial application. Neither the research findings nor their publication constitute a claim of readiness, safety, or suitability for commercial deployment.

\subsection*{Acknowledgments}

We thank Xpeng Inc. for providing computational and data resources in support of this academic research, and the data team for their assistance with data preparation and research support. We are particularly grateful to Professor Ping Luo for his guidance on the research ideas and manuscript writing. We also thank Xinghang Li, Qing Li, Baiqiao Yin, Xinyu Wei, Jade Yang, Hang Ye, Weiyang Jin, Ruowen Zhao, Mengfan Dong, Xirui Liang, Tianshu Wu, Daiheng Gao, Jiadi You, Linhao Zhou, Qiman Wu, Ziteng Cui, Haojun Zhang, Min Chen, Hao Li, Jiarui Lu, Zhengyu Jia, Zhihui Shang, Hanzhen Zhang and Zhuo Li for their valuable suggestions and constructive feedback.
\FinishArxivWrap
\FloatBarrier
\bibliography{paper}

\begin{thebibliography}{96}
\providecommand{\natexlab}[1]{#1}
\providecommand{\url}[1]{\texttt{#1}}
\expandafter\ifx\csname urlstyle\endcsname\relax
  \providecommand{\doi}[1]{doi: #1}\else
  \providecommand{\doi}{doi: \begingroup \urlstyle{rm}\Url}\fi

\bibitem[Ali et~al.(2025)Ali, Bai, Bala, Balaji, Blakeman, Cai, Cao, Cao, Cha, Chao, et~al.]{CosmosPredict25}
[1] Ali, A., Bai, J., Bala, M., Balaji, Y., Blakeman, A., Cai, T., Cao, J., Cao, T., Cha, E., Chao, Y.-W., et~al.
\newblock World simulation with video foundation models for physical ai.
\newblock \emph{arXiv preprint arXiv:2511.00062}, 2025.

\bibitem[Bai et~al.(2025{\natexlab{a}})Bai, Cai, Chen, Chen, Chen, Cheng, Deng, Ding, Gao, Ge, Ge, Guo, Huang, Huang, Huang, Hui, Jiang, Li, Li, Li, Li, Lin, Lin, Liu, Liu, Liu, Liu, Liu, Liu, Lu, Luo, Lv, Men, Meng, Ren, Ren, Song, Sun, Tang, Tu, Wan, Wang, Wang, Wang, Wang, Xie, Xu, Xu, Xu, Yang, Yang, Yang, Yang, Yu, Zhang, Zhang, Zhang, Zheng, Zhong, Zhou, Zhou, Zhou, Zhu, and Zhu]{qwen3vl}
[2] Bai, S., Cai, Y., Chen, R., Chen, K., Chen, X., Cheng, Z., Deng, L., Ding, W., Gao, C., Ge, C., Ge, W., Guo, Z., Huang, Q., Huang, J., Huang, F., Hui, B., Jiang, S., Li, Z., Li, M., Li, M., Li, K., Lin, Z., Lin, J., Liu, X., Liu, J., Liu, C., Liu, Y., Liu, D., Liu, S., Lu, D., Luo, R., Lv, C., Men, R., Meng, L., Ren, X., Ren, X., Song, S., Sun, Y., Tang, J., Tu, J., Wan, J., Wang, P., Wang, P., Wang, Q., Wang, Y., Xie, T., Xu, Y., Xu, H., Xu, J., Yang, Z., Yang, M., Yang, J., Yang, A., Yu, B., Zhang, F., Zhang, H., Zhang, X., Zheng, B., Zhong, H., Zhou, J., Zhou, F., Zhou, J., Zhu, Y., and Zhu, K.
\newblock Qwen3-vl technical report, 2025{\natexlab{a}}.
\newblock URL \url{https://arxiv.org/abs/2511.21631}.

\bibitem[Bai et~al.(2025{\natexlab{b}})Bai, Chen, Liu, Wang, Ge, Song, Dang, Wang, Wang, Tang, Zhong, Zhu, Yang, Li, Wan, Wang, Ding, Fu, Xu, Ye, Zhang, Xie, Cheng, Zhang, Yang, Xu, and Lin]{qwen25vl}
[3] Bai, S., Chen, K., Liu, X., Wang, J., Ge, W., Song, S., Dang, K., Wang, P., Wang, S., Tang, J., Zhong, H., Zhu, Y., Yang, M., Li, Z., Wan, J., Wang, P., Ding, W., Fu, Z., Xu, Y., Ye, J., Zhang, X., Xie, T., Cheng, Z., Zhang, H., Yang, Z., Xu, H., and Lin, J.
\newblock Qwen2.5-vl technical report, 2025{\natexlab{b}}.
\newblock URL \url{https://arxiv.org/abs/2502.13923}.

\bibitem[Beyer et~al.(2024)Beyer, Steiner, Pinto, Kolesnikov, Wang, Salz, Neumann, Alabdulmohsin, Tschannen, Bugliarello, et~al.]{Paligemma}
[4] Beyer, L., Steiner, A., Pinto, A.~S., Kolesnikov, A., Wang, X., Salz, D., Neumann, M., Alabdulmohsin, I., Tschannen, M., Bugliarello, E., et~al.
\newblock Paligemma: A versatile 3b vlm for transfer.
\newblock \emph{arXiv preprint arXiv:2407.07726}, 2024.

\bibitem[Black et~al.(2024)Black, Brown, Driess, Esmail, Equi, Finn, Fusai, Groom, Hausman, Ichter, et~al.]{black2024pi0}
[5] Black, K., Brown, N., Driess, D., Esmail, A., Equi, M., Finn, C., Fusai, N., Groom, L., Hausman, K., Ichter, B., et~al.
\newblock $\pi_0$: A vision-language-action flow model for general robot control.
\newblock \emph{arXiv preprint arXiv:2410.24164}, 2024.

\bibitem[Bu et~al.(2025)Bu, Li, Chen, Cai, Zeng, Cui, Yao, and Qiao]{RoboDual}
[6] Bu, Q., Li, H., Chen, L., Cai, J., Zeng, J., Cui, H., Yao, M., and Qiao, Y.
\newblock Towards synergistic, generalized, and efficient dual-system for robotic manipulation, 2025.
\newblock URL \url{https://arxiv.org/abs/2410.08001}.

\bibitem[Caesar et~al.(2020)Caesar, Bankiti, Lang, Vora, Liong, Xu, Krishnan, Pan, Baldan, and Beijbom]{nuscenes}
[7] Caesar, H., Bankiti, V., Lang, A.~H., Vora, S., Liong, V.~E., Xu, Q., Krishnan, A., Pan, Y., Baldan, G., and Beijbom, O.
\newblock nuscenes: A multimodal dataset for autonomous driving.
\newblock In \emph{2020 IEEE/CVF conference on computer vision and pattern recognition (CVPR)}, pp.\  11618--11628. IEEE, 2020.

\bibitem[Carion et~al.(2020)Carion, Massa, Synnaeve, Usunier, Kirillov, and Zagoruyko]{DETRR50}
[8] Carion, N., Massa, F., Synnaeve, G., Usunier, N., Kirillov, A., and Zagoruyko, S.
\newblock End-to-end object detection with transformers.
\newblock In \emph{European conference on computer vision}, pp.\  213--229. Springer, 2020.

\bibitem[Chen et~al.(2026{\natexlab{a}})Chen, Chen, Qiu, Bai, Ge, and Ge]{chen2026unit}
[9] Chen, B., Chen, Y., Qiu, L., Bai, J., Ge, Y., and Ge, Y.
\newblock {UniT}: Toward a unified physical language for human-to-humanoid policy learning and world modeling.
\newblock \emph{arXiv preprint arXiv:2604.19734}, 2026{\natexlab{a}}.
\newblock \doi{10.48550/arXiv.2604.19734}.
\newblock URL \url{https://arxiv.org/abs/2604.19734}.

\bibitem[Chen et~al.(2026{\natexlab{b}})Chen, Liu, Gu, Liu, Zhang, Li, He, Guo, Fu, Zhang, et~al.]{fastinslow}
[10] Chen, H., Liu, J., Gu, C., Liu, Z., Zhang, R., Li, X., He, X., Guo, Y., Fu, C.-W., Zhang, S., et~al.
\newblock Fast-in-slow: a dual-system vla model unifying fast manipulation within slow reasoning.
\newblock \emph{Advances in Neural Information Processing Systems}, 38:\penalty0 98049--98083, 2026{\natexlab{b}}.

\bibitem[Chen et~al.(2022)Chen, Saxena, Li, Fleet, and Hinton]{chen2022pix2seqlanguagemodelingframework}
[11] Chen, T., Saxena, S., Li, L., Fleet, D.~J., and Hinton, G.
\newblock Pix2seq: A language modeling framework for object detection, 2022.
\newblock URL \url{https://arxiv.org/abs/2109.10852}.

\bibitem[Chen et~al.(2025)Chen, Chen, Chen, Cai, Liu, Li, Liang, Lin, Ge, Gu, et~al.]{chen2025robotwin}
[12] Chen, T., Chen, Z., Chen, B., Cai, Z., Liu, Y., Li, Z., Liang, Q., Lin, X., Ge, Y., Gu, Z., et~al.
\newblock {RoboTwin 2.0}: A scalable data generator and benchmark with strong domain randomization for robust bimanual robotic manipulation.
\newblock \emph{arXiv preprint arXiv:2506.18088}, 2025.

\bibitem[Chen et~al.(2026{\natexlab{c}})Chen, Jiang, Zheng, Liang, Tie, Lu, Wu, and Dong]{chen2026pa3ff}
[13] Chen, Y., Jiang, M., Zheng, K., Liang, J., Tie, C., Lu, H., Wu, R., and Dong, H.
\newblock Pa3ff:learning part-aware dense 3d feature field for generalizable articulated object manipulation.
\newblock In \emph{International Conference on Learning Representations}, volume 2026, 2026{\natexlab{c}}.

\bibitem[Cheng et~al.(2023)Cheng, Zhang, Wu, Zhang, Zhu, Xie, Li, Ding, and Jin]{M6Doc}
[14] Cheng, H., Zhang, P., Wu, S., Zhang, J., Zhu, Q., Xie, Z., Li, J., Ding, K., and Jin, L.
\newblock M$^{6}$doc: A large-scale multi-format, multi-type, multi-layout, multi-language, multi-annotation category dataset for modern document layout analysis, 2023.
\newblock URL \url{https://arxiv.org/abs/2305.08719}.

\bibitem[Ch'Ng \& Chan(2017)Ch'Ng and Chan]{TotalText}
[15] Ch'Ng, C.~K. and Chan, C.~S.
\newblock Total-text: A comprehensive dataset for scene text detection and recognition.
\newblock In \emph{2017 14th IAPR international conference on document analysis and recognition (ICDAR)}, volume~1, pp.\  935--942. IEEE, 2017.

\bibitem[Comanici et~al.(2025)Comanici, Bieber, Schaekermann, Pasupat, Sachdeva, Dhillon, Blistein, Ram, Zhang, Rosen, et~al.]{Gemini2.5Pro}
[16] Comanici, G., Bieber, E., Schaekermann, M., Pasupat, I., Sachdeva, N., Dhillon, I., Blistein, M., Ram, O., Zhang, D., Rosen, E., et~al.
\newblock Gemini 2.5: Pushing the frontier with advanced reasoning, multimodality, long context, and next generation agentic capabilities.
\newblock \emph{arXiv preprint arXiv:2507.06261}, 2025.

\bibitem[Community(2026)]{community2026starvla}
[17] Community, S.
\newblock Starvla: A lego-like codebase for vision-language-action model developing.
\newblock \emph{arXiv preprint arXiv:2604.05014}, 2026.

\bibitem[Covert et~al.(2025)Covert, Sun, Zou, and Hashimoto]{LocalityAlignment}
[18] Covert, I., Sun, T., Zou, J.~Y., and Hashimoto, T.
\newblock Locality alignment improves vision-language models.
\newblock In \emph{International Conference on Learning Representations}, volume 2025, pp.\  83127--83165, 2025.

\bibitem[Cui et~al.(2025)Cui, Sun, Lin, Gao, Zhang, Liu, Wang, Zhang, Zhou, Liu, Zhang, Lv, Huang, Zhang, Zhang, Zhang, Liu, Yu, and Ma]{PaddleOCRv5}
[19] Cui, C., Sun, T., Lin, M., Gao, T., Zhang, Y., Liu, J., Wang, X., Zhang, Z., Zhou, C., Liu, H., Zhang, Y., Lv, W., Huang, K., Zhang, Y., Zhang, J., Zhang, J., Liu, Y., Yu, D., and Ma, Y.
\newblock Paddleocr 3.0 technical report, 2025.
\newblock URL \url{https://arxiv.org/abs/2507.05595}.

\bibitem[Dai et~al.(2021)Dai, Chen, Xiao, Chen, Liu, Yuan, and Zhang]{DyHeadR50}
[20] Dai, X., Chen, Y., Xiao, B., Chen, D., Liu, M., Yuan, L., and Zhang, L.
\newblock Dynamic head: Unifying object detection heads with attentions.
\newblock In \emph{2021 IEEE/CVF Conference on Computer Vision and Pattern Recognition (CVPR)}, pp.\  7369--7378. ieee, 2021.

\bibitem[Dang et~al.(2026)Dang, Guo, Hou, Leng, Li, Li, Liu, Mao, Wang, Yuan, Zhu, Lin, Bai, Jiang, Zhao, Zeng, Gao, Jiang, Cen, Huang, Wang, Zhang, Liu, Yang, Lu, and Zhao]{dang2026rynnbrain}
[21] Dang, R., Guo, J., Hou, B., Leng, S., Li, K., Li, X., Liu, J., Mao, Y., Wang, Z., Yuan, Y., Zhu, M., Lin, X., Bai, Y., Jiang, Q., Zhao, Y., Zeng, M., Gao, J., Jiang, Y., Cen, J., Huang, S., Wang, L., Zhang, W., Liu, C., Yang, J., Lu, S., and Zhao, D.
\newblock {RynnBrain}: Open embodied foundation models.
\newblock \emph{arXiv preprint arXiv:2602.14979}, 2026.
\newblock \doi{10.48550/arXiv.2602.14979}.
\newblock URL \url{https://arxiv.org/abs/2602.14979}.

\bibitem[Deitke et~al.(2025)Deitke, Clark, Lee, Tripathi, Yang, Park, Salehi, Muennighoff, Lo, Soldaini, et~al.]{Molmo}
[22] Deitke, M., Clark, C., Lee, S., Tripathi, R., Yang, Y., Park, J.~S., Salehi, M., Muennighoff, N., Lo, K., Soldaini, L., et~al.
\newblock Molmo and pixmo: Open weights and open data for state-of-the-art vision-language models.
\newblock In \emph{2025 IEEE/CVF Conference on Computer Vision and Pattern Recognition (CVPR)}, pp.\  91--104. IEEE, 2025.

\bibitem[Deng et~al.(2025)Deng, Zhu, Li, Gou, Li, Wang, Zhong, Yu, Nie, Song, Shi, and Fan]{BAGEL7BMoT}
[23] Deng, C., Zhu, D., Li, K., Gou, C., Li, F., Wang, Z., Zhong, S., Yu, W., Nie, X., Song, Z., Shi, G., and Fan, H.
\newblock Emerging properties in unified multimodal pretraining, 2025.
\newblock URL \url{https://arxiv.org/abs/2505.14683}.

\bibitem[Driess et~al.(2025)Driess, Springenberg, Ichter, Yu, Li-Bell, Pertsch, Ren, Walke, Vuong, Shi, and Levine]{driess2025knowledgeinsulatingvisionlanguageactionmodels}
[24] Driess, D., Springenberg, J.~T., Ichter, B., Yu, L., Li-Bell, A., Pertsch, K., Ren, A.~Z., Walke, H., Vuong, Q., Shi, L.~X., and Levine, S.
\newblock Knowledge insulating vision-language-action models: Train fast, run fast, generalize better, 2025.
\newblock URL \url{https://arxiv.org/abs/2505.23705}.

\bibitem[Guo et~al.(2025)Guo, Wu, Zhu, Leng, Shi, Chen, Fan, Wang, Jiang, Wang, et~al.]{SEED15VL}
[25] Guo, D., Wu, F., Zhu, F., Leng, F., Shi, G., Chen, H., Fan, H., Wang, J., Jiang, J., Wang, J., et~al.
\newblock Seed1. 5-vl technical report.
\newblock \emph{arXiv preprint arXiv:2505.07062}, 2025.

\bibitem[Gupta et~al.(2019)Gupta, Dollar, and Girshick]{LVIS}
[26] Gupta, A., Dollar, P., and Girshick, R.
\newblock Lvis: A dataset for large vocabulary instance segmentation.
\newblock In \emph{2019 IEEE/CVF Conference on Computer Vision and Pattern Recognition (CVPR)}, pp.\  5351--5359. IEEE, 2019.

\bibitem[Han et~al.(2026)Han, Li, Deng, Chen, Shi, Wang, Li, Wang, Xie, You, Quan, Cai, Diao, Liu, Yang, Lin, and Wang]{SenseNovaVision7BMoT}
[27] Han, X., Li, J., Deng, K., Chen, Z., Shi, X., Wang, S., Li, B., Wang, L., Xie, S., You, X., Quan, J., Cai, Z., Diao, H., Liu, Z., Yang, L., Lin, D., and Wang, Q.
\newblock Vision as unified multimodal generation, 2026.
\newblock URL \url{https://arxiv.org/abs/2607.06560}.

\bibitem[Huang et~al.(2019)Huang, Chen, He, Bai, Karatzas, Lu, and Jawahar]{SROIE}
[28] Huang, Z., Chen, K., He, J., Bai, X., Karatzas, D., Lu, S., and Jawahar, C.
\newblock Icdar2019 competition on scanned receipt ocr and information extraction.
\newblock In \emph{2019 International Conference on Document Analysis and Recognition (ICDAR)}, pp.\  1516--1520. IEEE, 2019.

\bibitem[Intelligence et~al.(2025)Intelligence, Black, Brown, Darpinian, Dhabalia, Driess, Esmail, Equi, Finn, Fusai, Galliker, Ghosh, Groom, Hausman, Ichter, Jakubczak, Jones, Ke, LeBlanc, Levine, Li-Bell, Mothukuri, Nair, Pertsch, Ren, Shi, Smith, Springenberg, Stachowicz, Tanner, Vuong, Walke, Walling, Wang, Yu, and Zhilinsky]{intelligence2025pi05visionlanguageactionmodelopenworld}
[29] Intelligence, P., Black, K., Brown, N., Darpinian, J., Dhabalia, K., Driess, D., Esmail, A., Equi, M., Finn, C., Fusai, N., Galliker, M.~Y., Ghosh, D., Groom, L., Hausman, K., Ichter, B., Jakubczak, S., Jones, T., Ke, L., LeBlanc, D., Levine, S., Li-Bell, A., Mothukuri, M., Nair, S., Pertsch, K., Ren, A.~Z., Shi, L.~X., Smith, L., Springenberg, J.~T., Stachowicz, K., Tanner, J., Vuong, Q., Walke, H., Walling, A., Wang, H., Yu, L., and Zhilinsky, U.
\newblock $\pi_{0.5}$: a vision-language-action model with open-world generalization, 2025.
\newblock URL \url{https://arxiv.org/abs/2504.16054}.

\bibitem[Jiang et~al.(2025)Jiang, Wu, Zeng, Ren, Xiong, Chen, Qin, and Zhang]{HumanRef}
[30] Jiang, Q., Wu, L., Zeng, Z., Ren, T., Xiong, Y., Chen, Y., Qin, L., and Zhang, L.
\newblock Referring to any person.
\newblock In \emph{2025 IEEE/CVF International Conference on Computer Vision (ICCV)}, pp.\  21667--21678. IEEE, 2025.

\bibitem[Jiang et~al.(2026)Jiang, Huo, Chen, Xiong, Zeng, Chen, Ren, Yu, and Zhang]{rexomni}
[31] Jiang, Q., Huo, J., Chen, X., Xiong, Y., Zeng, Z., Chen, Y., Ren, T., Yu, J., and Zhang, L.
\newblock Detect anything via next point prediction.
\newblock In \emph{Proceedings of the IEEE/CVF Conference on Computer Vision and Pattern Recognition}, pp.\  25472--25483, 2026.

\bibitem[Karatzas et~al.(2015)Karatzas, Gomez-Bigorda, Nicolaou, Ghosh, Bagdanov, Iwamura, Matas, Neumann, Chandrasekhar, Lu, Shafait, Uchida, and Valveny]{ICDAR2015}
[32] Karatzas, D., Gomez-Bigorda, L., Nicolaou, A., Ghosh, S., Bagdanov, A., Iwamura, M., Matas, J., Neumann, L., Chandrasekhar, V.~R., Lu, S., Shafait, F., Uchida, S., and Valveny, E.
\newblock Icdar 2015 competition on robust reading.
\newblock In \emph{2015 13th International Conference on Document Analysis and Recognition (ICDAR)}, pp.\  1156--1160, 2015.
\newblock \doi{10.1109/ICDAR.2015.7333942}.

\bibitem[Kim et~al.(2024)Kim, Pertsch, Karamcheti, Xiao, Balakrishna, Nair, Rafailov, Foster, Lam, Sanketi, et~al.]{openvla}
[33] Kim, M.~J., Pertsch, K., Karamcheti, S., Xiao, T., Balakrishna, A., Nair, S., Rafailov, R., Foster, E., Lam, G., Sanketi, P., et~al.
\newblock Openvla: An open-source vision-language-action model.
\newblock \emph{arXiv preprint arXiv:2406.09246}, 2024.

\bibitem[Kim et~al.(2026)Kim, Gao, Lin, Lin, Ge, Lam, Liang, Song, Liu, Finn, et~al.]{cosmospolicy}
[34] Kim, M.~J., Gao, Y., Lin, T.-Y., Lin, Y.-C., Ge, Y., Lam, G., Liang, P., Song, S., Liu, M.-Y., Finn, C., et~al.
\newblock Cosmos policy: Fine-tuning video models for visuomotor control and planning.
\newblock \emph{arXiv preprint arXiv:2601.16163}, 2026.

\bibitem[Kirillov et~al.(2023)Kirillov, Mintun, Ravi, Mao, Rolland, Gustafson, Xiao, Whitehead, Berg, Lo, et~al.]{SAM}
[35] Kirillov, A., Mintun, E., Ravi, N., Mao, H., Rolland, C., Gustafson, L., Xiao, T., Whitehead, S., Berg, A.~C., Lo, W.-Y., et~al.
\newblock Segment anything.
\newblock In \emph{2023 IEEE/CVF international conference on computer vision (ICCV)}, pp.\  3992--4003. IEEE, 2023.

\bibitem[Lai et~al.(2024)Lai, Tian, Chen, Li, Yuan, Liu, and Jia]{lai2024lisa}
[36] Lai, X., Tian, Z., Chen, Y., Li, Y., Yuan, Y., Liu, S., and Jia, J.
\newblock Lisa: Reasoning segmentation via large language model, 2024.
\newblock URL \url{https://arxiv.org/abs/2308.00692}.

\bibitem[Li et~al.(2025)Li, Meng, Lin, Luo, Tian, Ma, Huang, and Chua]{ScreenSpotPro}
[37] Li, K., Meng, Z., Lin, H., Luo, Z., Tian, Y., Ma, J., Huang, Z., and Chua, T.-S.
\newblock Screenspot-pro: Gui grounding for professional high-resolution computer use.
\newblock In \emph{Proceedings of the 33rd ACM International Conference on Multimedia}, pp.\  8778--8786, 2025.

\bibitem[Li et~al.(2026)Li, Hou, Zhu, Zhang, Cheng, Wang, Leng, Li, Lin, Yao, Zeng, Liu, Dang, Guo, Huang, Zhao, Ping, Zhao, Zhao, Wang, Lu, Xue, Tang, Wang, Wang, Gao, Lu, Liu, Yang, Chen, and Zhao]{RynnBrain112B}
[38] Li, K., Hou, B., Zhu, M., Zhang, T., Cheng, Z., Wang, Z., Leng, S., Li, X., Lin, X., Yao, B., Zeng, M., Liu, J., Dang, R., Guo, J., Huang, S., Zhao, H., Ping, H., Zhao, Y., Zhao, T., Wang, K., Lu, T., Xue, S., Tang, J., Wang, Y., Wang, Z., Gao, J., Lu, S., Liu, C., Yang, J., Chen, M., and Zhao, D.
\newblock Rynnbrain 1.1: Towards more capable and generalizable embodied foundation model, 2026.
\newblock URL \url{https://arxiv.org/abs/2607.17977}.

\bibitem[Li et~al.(2022)Li, Zhang, Zhang, Yang, Li, Zhong, Wang, Yuan, Zhang, Hwang, et~al.]{glip}
[39] Li, L.~H., Zhang, P., Zhang, H., Yang, J., Li, C., Zhong, Y., Wang, L., Yuan, L., Zhang, L., Hwang, J.-N., et~al.
\newblock Grounded language-image pre-training.
\newblock In \emph{2022 IEEE/CVF Conference on Computer Vision and Pattern Recognition (CVPR)}, pp.\  10955--10965. IEEE, 2022.

\bibitem[Liang et~al.(2023)Liang, Huang, Xia, Xu, Hausman, Ichter, Florence, and Zeng]{CodeAsPolicies}
[40] Liang, J., Huang, W., Xia, F., Xu, P., Hausman, K., Ichter, B., Florence, P., and Zeng, A.
\newblock Code as policies: Language model programs for embodied control.
\newblock In \emph{2023 IEEE International conference on robotics and automation (ICRA)}, pp.\  9493--9500. IEEE, 2023.

\bibitem[Lin et~al.(2024)Lin, Li, Gao, Yang, Wu, Bai, Lei, Wang, and Shou]{lin2024showui}
[41] Lin, K.~Q., Li, L., Gao, D., Yang, Z., Wu, S., Bai, Z., Lei, W., Wang, L., and Shou, M.~Z.
\newblock Showui: One vision-language-action model for gui visual agent, 2024.
\newblock URL \url{https://arxiv.org/abs/2411.17465}.

\bibitem[Lin et~al.(2014)Lin, Maire, Belongie, Hays, Perona, Ramanan, Doll{\'a}r, and Zitnick]{COCO}
[42] Lin, T.-Y., Maire, M., Belongie, S., Hays, J., Perona, P., Ramanan, D., Doll{\'a}r, P., and Zitnick, C.~L.
\newblock Microsoft coco: Common objects in context.
\newblock In \emph{European conference on computer vision}, pp.\  740--755. Springer, 2014.

\bibitem[Lipman et~al.(2023)Lipman, Chen, Ben-Hamu, Nickel, and Le]{FlowMatching}
[43] Lipman, Y., Chen, R. T.~Q., Ben-Hamu, H., Nickel, M., and Le, M.
\newblock Flow matching for generative modeling, 2023.
\newblock URL \url{https://arxiv.org/abs/2210.02747}.

\bibitem[Liu et~al.(2024)Liu, Zeng, Ren, Li, Zhang, Yang, Jiang, Li, Yang, Su, et~al.]{groundingdino}
[44] Liu, S., Zeng, Z., Ren, T., Li, F., Zhang, H., Yang, J., Jiang, Q., Li, C., Yang, J., Su, H., et~al.
\newblock Grounding dino: Marrying dino with grounded pre-training for open-set object detection.
\newblock In \emph{European conference on computer vision}, pp.\  38--55. Springer, 2024.

\bibitem[Liu et~al.(2025)Liu, Xie, Ding, Li, Yang, Wu, Wang, Sun, Liu, Wang, Ye, Li, Dong, Yu, Lu, Mo, Yan, Tian, Zhang, Huang, Liu, Su, Luo, Yue, Qi, Chen, Zhou, Qiao, Chen, and Wang]{liu2025scalecua}
[45] Liu, Z., Xie, J., Ding, Z., Li, Z., Yang, B., Wu, Z., Wang, X., Sun, Q., Liu, S., Wang, W., Ye, S., Li, Q., Dong, X., Yu, Y., Lu, C., Mo, Y., Yan, Y., Tian, Z., Zhang, X., Huang, Y., Liu, Y., Su, W., Luo, G., Yue, X., Qi, B., Chen, K., Zhou, B., Qiao, Y., Chen, Q., and Wang, W.
\newblock Scalecua: Scaling open-source computer use agents with cross-platform data, 2025.
\newblock URL \url{https://arxiv.org/abs/2509.15221}.

\bibitem[Long et~al.(2022)Long, Qin, Panteleev, Bissacco, Fujii, and Raptis]{HierText}
[46] Long, S., Qin, S., Panteleev, D., Bissacco, A., Fujii, Y., and Raptis, M.
\newblock Towards end-to-end unified scene text detection and layout analysis.
\newblock In \emph{2022 IEEE/CVF Conference on Computer Vision and Pattern Recognition (CVPR)}, pp.\  1039--1049. IEEE, 2022.

\bibitem[Loshchilov \& Hutter(2019)Loshchilov and Hutter]{AdamW}
[47] Loshchilov, I. and Hutter, F.
\newblock Decoupled weight decay regularization, 2019.
\newblock URL \url{https://arxiv.org/abs/1711.05101}.

\bibitem[Lu et~al.(2026)Lu, Chai, Guo, Yin, Liu, Wang, Xiao, Ren, Zhao, Liu, et~al.]{UIR13B}
[48] Lu, Z., Chai, Y., Guo, Y., Yin, X., Liu, L., Wang, H., Xiao, H., Ren, S., Zhao, P., Liu, G., et~al.
\newblock Ui-r1: Enhancing efficient action prediction of gui agents by reinforcement learning.
\newblock In \emph{Proceedings of the AAAI Conference on Artificial Intelligence}, volume~40, pp.\  17608--17616, 2026.

\bibitem[Mao et~al.(2016)Mao, Huang, Toshev, Camburu, Yuille, and Murphy]{RefCOCOg}
[49] Mao, J., Huang, J., Toshev, A., Camburu, O., Yuille, A.~L., and Murphy, K.
\newblock Generation and comprehension of unambiguous object descriptions.
\newblock In \emph{Proceedings of the IEEE conference on computer vision and pattern recognition}, pp.\  11--20, 2016.

\bibitem[Meng et~al.(2024)Meng, Yang, Tian, Dai, Wu, Gao, and Jiang]{DeepStack}
[50] Meng, L., Yang, J., Tian, R., Dai, X., Wu, Z., Gao, J., and Jiang, Y.-G.
\newblock Deepstack: Deeply stacking visual tokens is surprisingly simple and effective for lmms.
\newblock \emph{Advances in Neural Information Processing Systems}, 37:\penalty0 23464--23487, 2024.

\bibitem[Nagaraja et~al.(2016)Nagaraja, Morariu, and Davis]{RefCOCOgUMD}
[51] Nagaraja, V.~K., Morariu, V.~I., and Davis, L.~S.
\newblock Modeling context between objects for referring expression understanding.
\newblock In \emph{European conference on computer vision}, pp.\  792--807. Springer, 2016.

\bibitem[Nasiriany et~al.(2024)Nasiriany, Maddukuri, Zhang, Parikh, Lo, Joshi, Mandlekar, and Zhu]{nasiriany2024robocasa}
[52] Nasiriany, S., Maddukuri, A., Zhang, L., Parikh, A., Lo, A., Joshi, A., Mandlekar, A., and Zhu, Y.
\newblock {RoboCasa}: Large-scale simulation of everyday tasks for generalist robots.
\newblock In \emph{Robotics: Science and Systems}, 2024.

\bibitem[{NVIDIA}(2026)]{nvidia2026cosmos3}
[53] {NVIDIA}.
\newblock Cosmos 3: Omnimodal world models for physical ai.
\newblock \emph{arXiv preprint arXiv:2606.02800}, 2026.

\bibitem[{NVIDIA} et~al.(2025){NVIDIA}, Bjorck, Casta{\~n}eda, Cherniadev, et~al.]{nvidia2025gr00tn1}
[54] {NVIDIA}, Bjorck, J., Casta{\~n}eda, F., Cherniadev, N., et~al.
\newblock {GR00T N1}: An open foundation model for generalist humanoid robots.
\newblock \emph{arXiv preprint arXiv:2503.14734}, 2025.

\bibitem[Peebles \& Xie(2023)Peebles and Xie]{ActionDiT}
[55] Peebles, W. and Xie, S.
\newblock Scalable diffusion models with transformers.
\newblock In \emph{2023 IEEE/CVF International Conference on Computer Vision (ICCV)}, pp.\  4172--4182. IEEE, 2023.

\bibitem[Pfitzmann et~al.(2022)Pfitzmann, Auer, Dolfi, Nassar, and Staar]{DocLayNet}
[56] Pfitzmann, B., Auer, C., Dolfi, M., Nassar, A.~S., and Staar, P.
\newblock Doclaynet: A large human-annotated dataset for document-layout segmentation.
\newblock In \emph{Proceedings of the 28th ACM SIGKDD conference on knowledge discovery and data mining}, pp.\  3743--3751, 2022.

\bibitem[Ping et~al.(2026)Ping, Chen, Hui, Yu, Li, Yan, and Chang]{ping2026longactharnessingintrinsicactivation}
[57] Ping, B., Chen, Z., Hui, T., Yu, Q., Li, C., Yan, J., and Chang, B.
\newblock {LongAct}: Harnessing intrinsic activation patterns for long-context reinforcement learning, 2026.
\newblock URL \url{https://arxiv.org/abs/2604.14922}.

\bibitem[Qin et~al.(2025)Qin, Ye, Fang, Wang, Liang, Tian, Zhang, Li, Li, Huang, Zhong, Li, Yang, Miao, Lin, Liu, Jiang, Ma, Li, Xiao, Cai, Li, Zheng, Jin, Li, Zhou, Wang, Chen, Li, Yang, Liu, Lin, Peng, Liu, and Shi]{UITARS2B}
[58] Qin, Y., Ye, Y., Fang, J., Wang, H., Liang, S., Tian, S., Zhang, J., Li, J., Li, Y., Huang, S., Zhong, W., Li, K., Yang, J., Miao, Y., Lin, W., Liu, L., Jiang, X., Ma, Q., Li, J., Xiao, X., Cai, K., Li, C., Zheng, Y., Jin, C., Li, C., Zhou, X., Wang, M., Chen, H., Li, Z., Yang, H., Liu, H., Lin, F., Peng, T., Liu, X., and Shi, G.
\newblock Ui-tars: Pioneering automated gui interaction with native agents, 2025.
\newblock URL \url{https://arxiv.org/abs/2501.12326}.

\bibitem[Qiu et~al.(2026)Qiu, Wang, Zheng, Huang, Wen, Yang, Men, Yu, Huang, Huang, et~al.]{GatedAttention}
[59] Qiu, Z., Wang, Z., Zheng, B., Huang, Z., Wen, K., Yang, S., Men, R., Yu, L., Huang, F., Huang, S., et~al.
\newblock Gated attention for large language models: Non-linearity, sparsity, and attention-sink-free.
\newblock \emph{Advances in Neural Information Processing Systems}, 38:\penalty0 100092--100118, 2026.

\bibitem[{Qwen Team}(2026{\natexlab{a}})]{qwen35}
[60] {Qwen Team}.
\newblock {Qwen3.5}: Towards native multimodal agents, February 2026{\natexlab{a}}.
\newblock URL \url{https://qwen.ai/blog?id=qwen3.5}.

\bibitem[{Qwen Team}(2026{\natexlab{b}})]{qwen3627b}
[61] {Qwen Team}.
\newblock {Qwen3.6-27B}: Flagship-level coding in a {27B} dense model, April 2026{\natexlab{b}}.
\newblock URL \url{https://qwen.ai/blog?id=qwen3.6-27b}.

\bibitem[{Qwen Team}(2026{\natexlab{c}})]{qwen37}
[62] {Qwen Team}.
\newblock {Qwen3.7}: The agent frontier, May 2026{\natexlab{c}}.
\newblock URL \url{https://qwen.ai/blog?id=qwen3.7}.

\bibitem[{Qwen Team}(2026{\natexlab{d}})]{qwen38}
[63] {Qwen Team}.
\newblock {Qwen3.8-Max}: A new bar for coding and cowork, August 2026{\natexlab{d}}.
\newblock URL \url{https://qwen.ai/blog?id=qwen3.8}.

\bibitem[Rajbhandari et~al.(2020)Rajbhandari, Rasley, Ruwase, and He]{ZeRO2}
[64] Rajbhandari, S., Rasley, J., Ruwase, O., and He, Y.
\newblock Zero: Memory optimizations toward training trillion parameter models.
\newblock In \emph{SC20: international conference for high performance computing, networking, storage and analysis}, pp.\  1--16. IEEE, 2020.

\bibitem[Ranjan et~al.(2021)Ranjan, Sharma, Nguyen, and Hoai]{FSC147}
[65] Ranjan, V., Sharma, U., Nguyen, T., and Hoai, M.
\newblock Learning to count everything.
\newblock In \emph{2021 IEEE/CVF Conference on Computer Vision and Pattern Recognition (CVPR)}, pp.\  3393--3402. IEEE, 2021.

\bibitem[Redmon et~al.(2016)Redmon, Divvala, Girshick, and Farhadi]{yolo}
[66] Redmon, J., Divvala, S., Girshick, R., and Farhadi, A.
\newblock You only look once: Unified, real-time object detection.
\newblock In \emph{Proceedings of the IEEE conference on computer vision and pattern recognition}, pp.\  779--788, 2016.

\bibitem[Ren et~al.(2024)Ren, Huang, Wei, Zhao, Fu, Feng, and Jin]{ren2024pixellm}
[67] Ren, Z., Huang, Z., Wei, Y., Zhao, Y., Fu, D., Feng, J., and Jin, X.
\newblock Pixellm: Pixel reasoning with large multimodal model, 2024.
\newblock URL \url{https://arxiv.org/abs/2312.02228}.

\bibitem[Shao et~al.(2024)Shao, Wang, Zhu, Xu, Song, Bi, Zhang, Zhang, Li, Wu, and Guo]{GRPO}
[68] Shao, Z., Wang, P., Zhu, Q., Xu, R., Song, J., Bi, X., Zhang, H., Zhang, M., Li, Y.~K., Wu, Y., and Guo, D.
\newblock Deepseekmath: Pushing the limits of mathematical reasoning in open language models, 2024.
\newblock URL \url{https://arxiv.org/abs/2402.03300}.

\bibitem[Shen et~al.(2026)Shen, Liang, Lu, Jiang, Wang, Wei, Liu, Yang, Yu, You, Hao, He, Xie, and Wu]{shen2026ld4wamlearninglatentdynamics}
[69] Shen, Z., Liang, J., Lu, J., Jiang, F., Wang, Y., Wei, C., Liu, J., Yang, J., Yu, Q., You, J., Hao, C., He, G., Xie, C., and Wu, R.
\newblock {LD4WAM}: Learning latent dynamics from human videos for world action models, 2026.
\newblock URL \url{https://arxiv.org/abs/2608.22403}.

\bibitem[Song et~al.(2025)Song, Blukis, Tremblay, Tyree, Su, and Birchfield]{RoboSpatial}
[70] Song, C.~H., Blukis, V., Tremblay, J., Tyree, S., Su, Y., and Birchfield, S.
\newblock Robospatial: Teaching spatial understanding to 2d and 3d vision-language models for robotics.
\newblock In \emph{2025 IEEE/CVF Conference on Computer Vision and Pattern Recognition (CVPR)}, pp.\  15768--15780. IEEE, 2025.

\bibitem[Song et~al.(2026)Song, Zhou, Zhao, Chen, Ding, Yan, Huang, Tang, Wang, and Li]{reconvla}
[71] Song, W., Zhou, Z., Zhao, H., Chen, J., Ding, P., Yan, H., Huang, Y., Tang, F., Wang, D., and Li, H.
\newblock Reconvla: Reconstructive vision-language-action model as effective robot perceiver.
\newblock In \emph{Proceedings of the AAAI Conference on Artificial Intelligence}, volume~40, pp.\  18549--18557, 2026.

\bibitem[Team et~al.(2026)Team, Bai, Bai, Bao, Cai, Cai, Cao, Cao, Chai, Charles, et~al.]{KimiK3}
[72] Team, K., Bai, T., Bai, Y., Bao, Y., Cai, J., Cai, X., Cao, P., Cao, Y., Chai, Z., Charles, Y., et~al.
\newblock Kimi k3: Open frontier intelligence.
\newblock \emph{arXiv preprint arXiv:2607.24653}, 2026.

\bibitem[Tu et~al.(2026)Tu, Shukla, Yoo, Li, Li, Xie, Su, and Tu]{sgvla}
[73] Tu, R., Shukla, A., Yoo, S., Li, X., Li, J., Xie, J., Su, H., and Tu, Z.
\newblock Sg-vla: Learning spatially-grounded vision-language-action models for mobile manipulation.
\newblock \emph{arXiv preprint arXiv:2603.22760}, 2026.

\bibitem[Wan et~al.(2025)Wan, Wang, Ai, Wen, Mao, Xie, Chen, Yu, Zhao, Yang, Zeng, Wang, Zhang, Zhou, Wang, Chen, Zhu, Zhao, Yan, Huang, Feng, Zhang, Li, Wu, Chu, Feng, Zhang, Sun, Fang, Wang, Gui, Weng, Shen, Lin, Wang, Wang, Zhou, Wang, Shen, Yu, Shi, Huang, Xu, Kou, Lv, Li, Liu, Wang, Zhang, Huang, Li, Wu, Liu, Pan, Zheng, Hong, Shi, Feng, Jiang, Han, Wu, and Liu]{WAN}
[74] Wan, T., Wang, A., Ai, B., Wen, B., Mao, C., Xie, C.-W., Chen, D., Yu, F., Zhao, H., Yang, J., Zeng, J., Wang, J., Zhang, J., Zhou, J., Wang, J., Chen, J., Zhu, K., Zhao, K., Yan, K., Huang, L., Feng, M., Zhang, N., Li, P., Wu, P., Chu, R., Feng, R., Zhang, S., Sun, S., Fang, T., Wang, T., Gui, T., Weng, T., Shen, T., Lin, W., Wang, W., Wang, W., Zhou, W., Wang, W., Shen, W., Yu, W., Shi, X., Huang, X., Xu, X., Kou, Y., Lv, Y., Li, Y., Liu, Y., Wang, Y., Zhang, Y., Huang, Y., Li, Y., Wu, Y., Liu, Y., Pan, Y., Zheng, Y., Hong, Y., Shi, Y., Feng, Y., Jiang, Z., Han, Z., Wu, Z.-F., and Liu, Z.
\newblock Wan: Open and advanced large-scale video generative models.
\newblock \emph{arXiv preprint arXiv:2503.20314}, 2025.

\bibitem[Wang et~al.(2026{\natexlab{a}})Wang, Liu, Kuang, Wei, Liu, Li, Man, Chen, Tao, Liu, et~al.]{locateanything}
[75] Wang, S., Liu, S., Kuang, Y., Wei, X., Liu, Y., Li, Z., Man, Y., Chen, G., Tao, A., Liu, G., et~al.
\newblock Locateanything: Fast and high-quality vision-language grounding with parallel box decoding.
\newblock In \emph{European Conference on Computer Vision}, pp.\  336--357. Springer, 2026{\natexlab{a}}.

\bibitem[Wang et~al.(2026{\natexlab{b}})Wang, Huang, Li, Zhang, Liang, Jin, Chen, Chi, Zhou, Yu, et~al.]{openwam}
[76] Wang, Y., Huang, S., Li, M., Zhang, C., Liang, J., Jin, W., Chen, Y., Chi, X., Zhou, D., Yu, Q., et~al.
\newblock Openwam: An open, modular exploration towards systematic world-action model pretraining.
\newblock \emph{arXiv preprint arXiv:2609.07398}, 2026{\natexlab{b}}.

\bibitem[Wu et~al.(2026)Wu, Kong, Chen, Yu, Ye, Li, Wang, and Dong]{wu2026sugarscalablehumanvideodrivengeneralizable}
[77] Wu, T., Kong, X., Chen, Y., Yu, Q., Ye, H., Li, J., Wang, Y., and Dong, H.
\newblock {SUGAR}: A scalable human-video-driven generalizable humanoid loco-manipulation learning framework, 2026.
\newblock URL \url{https://arxiv.org/abs/2605.20373}.

\bibitem[Wu et~al.(2024)Wu, Chen, Pan, Liu, Liu, Dai, Gao, Ma, Wu, Wang, et~al.]{Deepseekvl2}
[78] Wu, Z., Chen, X., Pan, Z., Liu, X., Liu, W., Dai, D., Gao, H., Ma, Y., Wu, C., Wang, B., et~al.
\newblock Deepseek-vl2: Mixture-of-experts vision-language models for advanced multimodal understanding.
\newblock \emph{arXiv preprint arXiv:2412.10302}, 2024.

\bibitem[Wu et~al.(2025)Wu, Wu, Xu, Wang, Sun, Jia, Cheng, Ding, Chen, Liang, et~al.]{ScreenSpotV2}
[79] Wu, Z., Wu, Z., Xu, F., Wang, Y., Sun, Q., Jia, C., Cheng, K., Ding, Z., Chen, L., Liang, P.~P., et~al.
\newblock Os-atlas: Foundation action model for generalist gui agents.
\newblock In \emph{International Conference on Learning Representations}, volume 2025, pp.\  5090--5108, 2025.

\bibitem[Xiao et~al.(2024)Xiao, Wu, Xu, Dai, Hu, Lu, Zeng, Liu, and Yuan]{xiao2024florence}
[80] Xiao, B., Wu, H., Xu, W., Dai, X., Hu, H., Lu, Y., Zeng, M., Liu, C., and Yuan, L.
\newblock Florence-2: Advancing a unified representation for a variety of vision tasks.
\newblock In \emph{2024 IEEE/CVF Conference on Computer Vision and Pattern Recognition (CVPR)}, pp.\  4818--4829. IEEE, 2024.

\bibitem[Xie et~al.(2026)Xie, Deng, Li, Yang, Wu, Chen, Hu, Wang, Xu, Wang, et~al.]{JEDI3B}
[81] Xie, T., Deng, J., Li, X., Yang, J., Wu, H., Chen, J., Hu, W., Wang, X., Xu, Y., Wang, Z., et~al.
\newblock Scaling computer-use grounding via user interface decomposition and synthesis.
\newblock \emph{Advances in Neural Information Processing Systems}, 38, 2026.

\bibitem[Yang et~al.(2025{\natexlab{a}})Yang, Li, Yang, Zhang, Hui, Zheng, Yu, Gao, Huang, Lv, et~al.]{Qwen34BLLM}
[82] Yang, A., Li, A., Yang, B., Zhang, B., Hui, B., Zheng, B., Yu, B., Gao, C., Huang, C., Lv, C., et~al.
\newblock Qwen3 technical report.
\newblock \emph{arXiv preprint arXiv:2505.09388}, 2025{\natexlab{a}}.

\bibitem[Yang et~al.(2025{\natexlab{b}})Yang, Kautz, and Hatamizadeh]{GatedDeltaNet}
[83] Yang, S., Kautz, J., and Hatamizadeh, A.
\newblock Gated delta networks: Improving mamba2 with delta rule.
\newblock In \emph{International Conference on Learning Representations}, volume 2025, pp.\  29687--29707, 2025{\natexlab{b}}.

\bibitem[Ye et~al.(2025)Ye, Zhang, Xu, Liu, Wang, Zhu, Zheng, Gao, Cao, Lu, Liao, Zheng, Huang, Zhou, and Yan]{GUIOwl32B}
[84] Ye, J., Zhang, X., Xu, H., Liu, H., Wang, J., Zhu, Z., Zheng, Z., Gao, F., Cao, J., Lu, Z., Liao, J., Zheng, Q., Huang, F., Zhou, J., and Yan, M.
\newblock Mobile-agent-v3: Fundamental agents for gui automation, 2025.
\newblock URL \url{https://arxiv.org/abs/2508.15144}.

\bibitem[Ye et~al.(2026)Ye, Ge, Zheng, Gao, Yu, Kurian, Indupuru, Tan, Zhu, Xiang, et~al.]{dreamzero}
[85] Ye, S., Ge, Y., Zheng, K., Gao, S., Yu, S., Kurian, G., Indupuru, S., Tan, Y.~L., Zhu, C., Xiang, J., et~al.
\newblock World action models are zero-shot policies.
\newblock \emph{arXiv preprint arXiv:2602.15922}, 2026.

\bibitem[You et~al.(2026)You, Yu, Chen, Cai, Zhong, Wang, Ping, Liang, Shen, Yan, Li, Wu, Qi, and Chen]{you2026affordancewamaffordanceawarejointworldaction}
[86] You, J., Yu, Q., Chen, Y., Cai, M., Zhong, Z., Wang, Y., Ping, B., Liang, J., Shen, Z., Yan, H., Li, Y., Wu, R., Qi, X., and Chen, Y.
\newblock {AffordanceWAM}: Affordance-aware joint world-action modeling for robot manipulation, 2026.
\newblock URL \url{https://arxiv.org/abs/2609.22332}.

\bibitem[Yu et~al.(2025)Yu, Lin, Zhao, Yin, Wei, Peng, Wei, Sun, Han, Ge, Zhang, Jiang, Wang, and Tao]{yu2025perceptionr1}
[87] Yu, E., Lin, K., Zhao, L., Yin, J., Wei, Y., Peng, Y., Wei, H., Sun, J., Han, C., Ge, Z., Zhang, X., Jiang, D., Wang, J., and Tao, W.
\newblock Perception-r1: Pioneering perception policy with reinforcement learning, 2025.
\newblock URL \url{https://arxiv.org/abs/2504.07954}.

\bibitem[Yu et~al.(2016)Yu, Poirson, Yang, Berg, and Berg]{RefCOCO}
[88] Yu, L., Poirson, P., Yang, S., Berg, A.~C., and Berg, T.~L.
\newblock Modeling context in referring expressions.
\newblock In \emph{European conference on computer vision}, pp.\  69--85. Springer, 2016.

\bibitem[Yu et~al.(2026)Yu, You, Wang, Liang, Ping, Tian, Chen, Cai, Gong, Wu, et~al.]{affordancevla}
[89] Yu, Q., You, J., Wang, Y., Liang, J., Ping, B., Tian, Y., Chen, Y., Cai, M., Gong, Z., Wu, R., et~al.
\newblock {AffordanceVLA}: A vision-language-action model empowering action generation through affordance-aware understanding.
\newblock \emph{arXiv preprint arXiv:2606.06155}, 2026.

\bibitem[Yuan et~al.(2024)Yuan, Duan, Blukis, Pumacay, Krishna, Murali, Mousavian, and Fox]{RoboPoint13B}
[90] Yuan, W., Duan, J., Blukis, V., Pumacay, W., Krishna, R., Murali, A., Mousavian, A., and Fox, D.
\newblock Robopoint: A vision-language model for spatial affordance prediction for robotics, 2024.
\newblock URL \url{https://arxiv.org/abs/2406.10721}.

\bibitem[Yue et~al.(2025)Yue, Lin, Song, Wang, Ren, Gu, Li, Li, Zhao, Li, et~al.]{MimoVL}
[91] Yue, Z., Lin, Z., Song, Y., Wang, W., Ren, S., Gu, S., Li, S., Li, P., Zhao, L., Li, L., et~al.
\newblock Mimo-vl technical report.
\newblock \emph{arXiv preprint arXiv:2506.03569}, 2025.

\bibitem[Zhang et~al.(2022)Zhang, Li, Liu, Zhang, Su, Zhu, Ni, and Shum]{DINOR50}
[92] Zhang, H., Li, F., Liu, S., Zhang, L., Su, H., Zhu, J., Ni, L.~M., and Shum, H.-Y.
\newblock Dino: Detr with improved denoising anchor boxes for end-to-end object detection, 2022.
\newblock URL \url{https://arxiv.org/abs/2203.03605}.

\bibitem[Zhang et~al.(2023)Zhang, Li, Li, Ren, Zou, Liu, Huang, Gao, Zhang, Li, and Yang]{zhang2024llavagrounding}
[93] Zhang, H., Li, H., Li, F., Ren, T., Zou, X., Liu, S., Huang, S., Gao, J., Zhang, L., Li, C., and Yang, J.
\newblock Llava-grounding: Grounded visual chat with large multimodal models, 2023.
\newblock URL \url{https://arxiv.org/abs/2312.02949}.

\bibitem[Zhao et~al.(2024)Zhao, Kang, Wang, and He]{DocLayoutYOLO}
[94] Zhao, Z., Kang, H., Wang, B., and He, C.
\newblock Doclayout-yolo: Enhancing document layout analysis through diverse synthetic data and global-to-local adaptive perception, 2024.
\newblock URL \url{https://arxiv.org/abs/2410.12628}.

\bibitem[Zhou et~al.(2026)Zhou, An, Chi, Han, Rong, Zhang, Wang, Wang, Huang, Sheng, et~al.]{RoboRefer2B}
[95] Zhou, E., An, J., Chi, C., Han, Y., Rong, S., Zhang, C., Wang, P., Wang, Z., Huang, T., Sheng, L., et~al.
\newblock Roborefer: Towards spatial referring with reasoning in vision-language models for robotics.
\newblock \emph{Advances in Neural Information Processing Systems}, 38:\penalty0 28404--28481, 2026.

\bibitem[Zhu et~al.(2018)Zhu, Wen, Bian, Ling, and Hu]{VisDrone}
[96] Zhu, P., Wen, L., Bian, X., Ling, H., and Hu, Q.
\newblock Vision meets drones: A challenge, 2018.
\newblock URL \url{https://arxiv.org/abs/1804.07437}.

\end{thebibliography}
\bibliographystyle{arxiv-numbered}

\clearpage
\etocdepthtag.toc{appendix}
\beginappendix
\suppressfloats[t]
\setlength{\parskip}{4pt plus 0.5pt minus 0.5pt}
\pretocmd{\section}{\FinishPlacedArxivWrap}{}{}
\pretocmd{\subsection}{\FinishPlacedArxivWrap}{}{}
\pretocmd{\subsubsection}{\FinishPlacedArxivWrap}{}{}
\let\OriginalSuppMid\mid
\renewcommand{\mid}{\allowbreak\OriginalSuppMid}
\label{app:appendix}

\ifdefined\BenchmarkTableFont\else
\newcommand{\BenchmarkTableFont}{\normalfont}
\newcommand{\BenchHead}[1]{\begin{tabular}{@{}c@{}}#1\end{tabular}}
\colorlet{benchmarktype}{black!8}
\definecolor{benchmarkpurple}{HTML}{F5F1FA}
\definecolor{benchmarkgreen}{HTML}{F0F3E8}
\fi

\ifdefined\GPIBoxHelpersLoaded\else
\def\GPIBoxHelpersLoaded{1}
\ifdefined\tcolorbox
  \providecommand{\finding}[2]{%
    \begin{tcolorbox}[colback=blue!4,colframe=blue!55!black,
      boxrule=0.6pt,arc=2pt,left=5pt,right=5pt,top=3pt,bottom=3pt,
      before skip=5pt,after skip=5pt]\small
      \textbf{\textit{Finding #1:}} #2
    \end{tcolorbox}}
  \providecommand{\takeaway}[2]{%
    \begin{tcolorbox}[colback=blue!4,colframe=blue!55!black,
      boxrule=0.6pt,arc=2pt,left=5pt,right=5pt,top=3pt,bottom=3pt,
      before skip=5pt,after skip=5pt]\small
      \textbf{\textit{Takeaway #1:}} #2
    \end{tcolorbox}}
\else
  \providecommand{\finding}[2]{\begin{gpiquestions}\textbf{\textit{Finding #1:}} #2\end{gpiquestions}}
  \providecommand{\takeaway}[2]{\begin{gpiquestions}\textbf{\textit{Takeaway #1:}} #2\end{gpiquestions}}
\fi
\fi

\section{Grounding Model Details}
\label{app:groundingpi_model_details}

\providecommand{\gpitok}[1]{\texttt{\detokenize{#1}}}

\subsection{Input/Output Protocol}
\label{app:groundingpi_protocol}

GroundingPI uses the same language-conditioned interface for box and point prediction. A query specifies the task and its semantic targets; the response associates each label, referring expression, or OCR transcription with a geometric payload. The examples below illustrate the protocol rather than training records. Displayed line breaks are for readability and are omitted in the serialized response.

\paragraph{Vocabulary and entry structure.}
\Cref{tab:groundingpi_protocol_tokens} summarizes the token roles. Coordinates are atomic vocabulary entries; labels and punctuation use ordinary language tokens. Every entry follows the template
\begin{quote}
\small
\gpitok{<|object_ref_start|>}\textit{label}\gpitok{<|object_ref_end|>}\\
\gpitok{<|box_start|>}\textit{payload}\gpitok{<|box_end|>}
\end{quote}
A box uses four consecutive coordinate tokens in $xyxy$ order; a point uses two in $(x,y)$ order. Multiple instances with the same label share one wrapper, with tuples separated by commas without spaces. Distinct entries are separated by a comma and a space. Explicitly queried but absent categories retain their label and use the ordinary text payload \gpitok{None}.

\begin{table}[hb]
\centering\small
\setlength{\tabcolsep}{4pt}
\renewcommand{\arraystretch}{1.1}
\caption{Tokens and delimiters in GroundingPI's input/output protocol.}
\label{tab:groundingpi_protocol_tokens}
\begin{tabular}{@{}p{0.44\linewidth}p{0.51\linewidth}@{}}
\toprule
Token or delimiter & Role \\
\midrule
\gpitok{<0>}--\gpitok{<999>} & One atomic token per quantized coordinate \\
\gpitok{</c>} & Separator between categories in the input query \\
\gpitok{<|object_ref_start|>}, \gpitok{<|object_ref_end|>} & Delimit a semantic label or transcription \\
\gpitok{<|box_start|>}, \gpitok{<|box_end|>} & Delimit a box, point, or ordered-point payload \\
\gpitok{None} & Ordinary text indicating an absent queried target \\
\gpitok{<|im_end|>} & End of the assistant response \\
\bottomrule
\end{tabular}
\end{table}

\paragraph{Task prompts.}
\Cref{tab:groundingpi_prompt_templates} lists the canonical user prompts. Category names are joined by \gpitok{</c>} without additional spaces or commas; requested category strings are preserved in grounding and layout responses. Referring prompts provide the target description, and OCR responses use recognized text as their labels. Dense and ordinary grounding share a prompt. Visual prompts encode example boxes with the same coordinate vocabulary as the output. The image and user prompt are supplied through the native multimodal chat template.

\begin{table}[hb]
\centering\small
\setlength{\tabcolsep}{4pt}
\renewcommand{\arraystretch}{1.18}
\caption{Canonical prompts and output geometry. Italic fields are replaced by query-specific text or coordinates.}
\label{tab:groundingpi_prompt_templates}
\begin{tabular}{@{}>{\raggedright\arraybackslash}p{0.17\linewidth}>{\raggedright\arraybackslash}p{0.66\linewidth}>{\raggedright\arraybackslash}p{0.11\linewidth}@{}}
\toprule
Task & User prompt & Output \\
\midrule
Grounding / dense grounding & Locate all the instances that match the following categories: \textit{cat1}\gpitok{</c>}\textit{cat2}\gpitok{</c>}\ldots. & Boxes \\
Referring & Locate the target referred to by the following description: \textit{phrase}. & Boxes \\
Object pointing & Point to: \textit{cat1}\gpitok{</c>}\textit{cat2}\gpitok{</c>}\ldots. & Points \\
Referring pointing & Point to the target referred to by the following description: \textit{phrase}. & Points \\
GUI grounding & Point to the UI element to click for the following instruction: \textit{instruction}. & Point \\
OCR & OCR task detect all the text in box format. & Text + boxes \\
Layout grounding & Detect all document layout elements that match the following categories: \textit{cat1}\gpitok{</c>}\textit{cat2}\gpitok{</c>}\ldots. & Boxes \\
Visual prompting & Given reference boxes \gpitok{<|box_start|>}\textit{reference boxes}\gpitok{<|box_end|>} indicating one or more objects, find all similar objects in the image and output their bounding boxes. & Boxes \\
\bottomrule
\end{tabular}
\end{table}

\paragraph{Illustrative responses.}
The following example contains two cups and an absent car. The comma after the first entry is followed by a space in the serialized response.
\begin{center}
\fcolorbox{black!20}{black!2}{\begin{minipage}{0.95\linewidth}
\small\raggedright
\gpitok{<|object_ref_start|>cup<|object_ref_end|>}\\
\gpitok{<|box_start|><10><20><30><40>,<50><60><70><80><|box_end|>,}\\
\gpitok{<|object_ref_start|>car<|object_ref_end|>}\\
\gpitok{<|box_start|>None<|box_end|><|im_end|>}
\end{minipage}}
\end{center}
Pointing uses the same wrapper with two coordinates per instance:
\begin{center}
\fcolorbox{black!20}{black!2}{\begin{minipage}{0.95\linewidth}
\small\raggedright
\gpitok{<|object_ref_start|>cup<|object_ref_end|>}\\
\gpitok{<|box_start|><20><30>,<60><70><|box_end|><|im_end|>}
\end{minipage}}
\end{center}
OCR binds the transcription directly to its text region:
\begin{center}
\fcolorbox{black!20}{black!2}{\begin{minipage}{0.95\linewidth}
\small\raggedright
\gpitok{<|object_ref_start|>OPEN<|object_ref_end|>}\\
\gpitok{<|box_start|><100><200><400><300><|box_end|><|im_end|>}
\end{minipage}}
\end{center}

\paragraph{Coordinate conventions.}
Image coordinates are normalized relative to image width and height. An integer coordinate $v$ on the intermediate $[0,1000]$ grid is mapped to $q(v)=\lfloor(999v+500)/1000\rfloor$ and encoded by the corresponding atomic coordinate token. Image-space coordinates are recovered by multiplying $q/999$ by the corresponding image dimension. Source $xywh$ boxes are converted to $xyxy$ before normalization, and polygon annotations are converted to their axis-aligned enclosing boxes. Point targets follow their task definition; box centers are used only where the annotation convention specifies them. Non-finite, out-of-range, inverted, or degenerate geometry is rejected rather than silently repaired.

\paragraph{Ordering and response boundaries.}
Within a label, box instances are stably sorted by $x_1$ and unordered points by $x$; ordered trajectories retain temporal order, including repeated locations. Trajectories use a separate metric-coordinate convention described in \Cref{app:driving_setup}. The marker \gpitok{<|box_end|>} closes one payload, whereas \gpitok{<|im_end|>} ends the response. Structural and coordinate tokens identify the entries and their geometry.

\subsection{Architecture, Tokenizer, and Visual Processing}
\label{app:groundingpi_architecture}

GroundingPI combines a MoonViT-V2 (Kimi K3) visual encoder, a learnable multimodal projector, and a Qwen3-4B language backbone. Visual embeddings are inserted into the language sequence, which uses one-dimensional rotary position indices. All 36 language layers use full attention. \Cref{tab:groundingpi_architecture_details} summarizes the architecture.

\begin{table}[htbp]
\centering\small
\setlength{\tabcolsep}{4pt}
\renewcommand{\arraystretch}{1.12}
\caption{GroundingPI architecture and visual processing. Positional capacity is distinct from the training sequence limit.}
\label{tab:groundingpi_architecture_details}
\begin{tabular}{@{}p{0.25\linewidth}p{0.70\linewidth}@{}}
\toprule
Component & Configuration \\
\midrule
Vision encoder & 27 layers; hidden width 1024; FFN width 4096; 12 attention heads; QKV hidden width 1536; patch size 14 \\
Spatial aggregation & $2\times2$ neighboring patches; four 1024-dimensional features form one 4096-dimensional feature \\
Projector & LayerNorm(1024), spatial aggregation, two bias-free linear layers ($4096\to4096\to2560$) with GELU, and RMSNorm(2560); normalization $\epsilon=10^{-5}$ \\
Language backbone & 36 full-attention layers; hidden width 2560; FFN width 9728; 32 query heads and 8 KV heads; head dimension 128 \\
Language numerics & SiLU; attention dropout 0; RMSNorm $\epsilon=10^{-6}$; RoPE base $5{,}000{,}000$; no sliding window \\
Positional capacity & 262,144 positions \\
Visual processing & Dynamic patch budget of at most 4096 patches, producing at most 1024 projected visual tokens; channel mean/std $(0.5,0.5,0.5)$ \\
\bottomrule
\end{tabular}
\end{table}

\sbox{\ArxivNaturalTable}{%
\small
\setlength{\tabcolsep}{5pt}
\begin{tabular}{@{}lr@{}}
\toprule
Module & Parameters \\
\midrule
Vision encoder & 401,214,464 \\
Projector & 27,267,584 \\
Decoder excluding vocabulary matrices & 3,633,511,936 \\
Input embedding & 390,835,200 \\
Output head & 390,835,200 \\
\midrule
Total & 4,843,664,384 \\
\bottomrule
\end{tabular}%
}
\FinishArxivWrap
\begin{wraptable}{R}{\wd\ArxivNaturalTable}
\centering\small
\setlength{\tabcolsep}{5pt}
\caption{GroundingPI parameter counts, including untied vocabulary matrices.}
\label{tab:groundingpi_parameter_counts}
\usebox{\ArxivNaturalTable}
\end{wraptable}

\paragraph{Vocabulary and parameterization.}
The tokenizer extends the 151,669-entry base vocabulary with 1,000 coordinate tokens and \gpitok{</c>}, giving 152,670 entries. Coordinate IDs are 151669--152668, and the category separator has ID 152669. EOS and padding are distinct (151645 and 151643). The input embedding and output head are untied; all semantic, structural, and coordinate tokens are predicted by the same vocabulary head. The complete multimodal model contains approximately 4.844B parameters, including the visual encoder and vocabulary matrices (\Cref{tab:groundingpi_parameter_counts}); Qwen3-4B denotes the language-backbone family.

\subsection{Data Engine}
\label{app:groundingpi_data_engine}

\paragraph{Candidate generation and field fusion.}
The engine combines complementary predictions for object localization, segmentation, text recognition, GUI elements, and document regions. For category-level grounding, category discovery precedes category-conditioned localization. Candidate instances are mapped to the original image and aligned within a common query scope and annotation granularity. Fusion operates separately on semantic and geometric fields: verified information is retained, while disagreements trigger additional evidence for the affected fields. Object--part containment and word--line relations are preserved rather than merged as duplicates.

\paragraph{Task-dependent validation.}
Each field is marked as accepted, rejected, or unresolved. Validation checks geometry, syntax, semantic correspondence, and consistency with available evidence. Required fields are determined by the query, independently of which fields a teacher produces; missing required fields remain unresolved. A supervision item is accepted only when all required fields are accepted. Queries requiring all instances additionally require a coverage check over the relevant region. For an absent-target answer, absence is itself a required fact to verify; an empty prediction does not establish it. Consequently, an unresolved instance can block an all-target or counting query while leaving independently verified local supervision usable. Multi-teacher agreement supports validation but does not guarantee annotation correctness.

\paragraph{Targeted observations and expert iteration.}
Accepted annotations train a unified grounding expert, whose subsequent predictions pass through the same fusion and validation procedure. Complementary teachers and local observations are invoked when semantic identity, geometry, or coverage remains unresolved. Crop predictions are mapped back to the original coordinates; supervision requiring detail unavailable in the original training input retains the necessary local view. Additional evidence can both add annotations and revise existing labels. Revisions propagate to dependent tasks, and invalidated labels are withdrawn until their dependencies are resolved.

\paragraph{Deriving supervision.}
A verified instance record can support several tasks: category--box pairs provide grounding, attributes and relations support referring, valid instance regions support pointing, exemplar correspondence supports visual prompting, and text regions paired with transcriptions support OCR. Each derived query retains its own required fields and coverage conditions. This reuse preserves task-specific semantics while sharing the underlying visual evidence.

\subsection{Base VLM Training (Pretrain 1)}
\label{app:groundingpi_vlm_training}

\paragraph{Training organization.}
GroundingPI training comprises three successive phases: (i) Base VLM training (Pretrain~1), (ii) coordinate alignment (Pretrain~2), also termed supervised fine-tuning (SFT), and (iii) reinforcement learning (RL) with GRPO. Pretrain~1 contains Stages~1--3 below, while Pretrain~2 corresponds to Stage~4. Thus, Stage~1--4 numbering describes the supervised training recipe within the first two phases. Each stage starts from the preceding checkpoint, and RL follows the Stage~4 SFT checkpoint. The aggregate training exposure is $500.41\mathrm{B}+221.17\mathrm{B}$ tokens.

\paragraph{Stage 1: Vision--language projector alignment.}
We freeze the visual encoder and language model and update only the projector using image-caption supervision. Causal cross-entropy on the assistant response aligns visual features with the language input space, without an additional feature-distance or contrastive objective. The context and packing lengths are both 8192 tokens.

\paragraph{Stage 2: Joint multimodal pretraining.}
We unfreeze the visual encoder, projector, and language model and jointly train on text-only and general image--text data. Text examples provide full-token causal supervision, while multimodal examples supervise assistant responses. The language model uses a peak learning rate of $10^{-5}$, and the visual encoder and projector each use $10^{-6}$. The context and packing lengths remain 8192 tokens.

\paragraph{Stage 3: General visual and video understanding.}
We continue updating all modules on general visual question answering and instruction-following data, image captions, and videos. The context and packing lengths increase to 32768 tokens to accommodate longer multimodal sequences. This stage does not separately mix in the spatial-specialization data or an independent text-only quota; image-caption supervision remains part of the general visual training mixture.

\subsection{Coordinate Alignment (Pretrain 2 / SFT)}
\label{app:groundingpi_coordinate_alignment}

\paragraph{Stage 4: Spatial perception specialization.}
Starting from the Stage~3 checkpoint, coordinate alignment updates all modules using eight spatial task groups: detection, GUI grounding, referring-expression grounding, referring-expression pointing, OCR, document layout, dense pointing and counting, and visual prompting. This phase is the SFT phase described in the main text. It uses an 8192-token context and packing length and applies autoregressive supervision to semantic labels, box and point coordinates, and protocol markers. Quantized coordinate tokens share the same cross-entropy objective as other valid response tokens; no additional IoU regression loss is introduced.

\paragraph{Supervision and loss normalization.}
Stages~1--4 use causal next-token prediction. Let $\mathcal S_d$ contain valid sample--position pairs $(b,t)$ in domain $d\in\{\mathrm{text},\mathrm{vlm}\}$. The domain loss is $\mathcal L_d=-\bigl(\max(|\mathcal S_d|,1)\bigr)^{-1}\sum_{(b,t)\in\mathcal S_d}\log p_\theta(z_{b,t}\mid V_b,z_{b,<t})$, with $V_b=\varnothing$ for text-only examples. Text supervision covers valid causal targets; multimodal supervision covers assistant responses, excluding prompt, visual, padding, and empty thinking-prefix positions. Stage~2 minimizes $\mathcal L_{\mathrm{text}}+\mathcal L_{\mathrm{vlm}}$, with each domain normalized over its valid targets across data-parallel workers. A missing domain contributes zero. Stages~1, 3, and~4 use $\mathcal L_{\mathrm{vlm}}$ without a separate text-only domain. In Stage~4, this assistant-only loss includes the semantic, coordinate, and protocol tokens of structured spatial responses. Packing preserves each sample's causal boundaries and supervision mask.

\paragraph{Optimization.}
\Cref{tab:groundingpi_vlm_sft} summarizes the configuration for Pretrain~1 and Pretrain~2. The supervised training recipe uses 128 GPUs, BF16 precision, ZeRO-1, FlashAttention, and activation recomputation. Only the projector is trainable in Stage~1; Stages~2--4 update all modules. Language parameters include the input embedding and untied output head. All four stages use one gradient-accumulation step, AdamW with betas $(0.9,0.95)$ and $\epsilon=10^{-8}$, and gradient clipping at 1.0. Weight decay is zero in Stage~1 and 0.1 thereafter. Cosine schedules use 3\% warmup and decay to 2\% of the peak learning rate in Stage~1 and 10\% in Stages~2--4. The configured budget is one epoch per stage. Global batches count packed sequences; context length is the complete multimodal sequence budget, distinct from the visual budget of 1024 projected tokens. Seed and data seed are both 42.

\begin{table}[htbp]
\centering\small
\setlength{\tabcolsep}{3pt}
\renewcommand{\arraystretch}{1.13}
\caption{Training configuration for Base VLM training (Pretrain~1, Stages~1--3) and coordinate alignment (Pretrain~2 / SFT, Stage~4). Each column uses 128 GPUs; language-side updates include both vocabulary matrices.}
\label{tab:groundingpi_vlm_sft}
\begin{tabular}{@{}>{\raggedright\arraybackslash}p{0.26\linewidth}>{\raggedright\arraybackslash}p{0.16\linewidth}>{\raggedright\arraybackslash}p{0.18\linewidth}>{\raggedright\arraybackslash}p{0.16\linewidth}>{\raggedright\arraybackslash}p{0.16\linewidth}@{}}
\toprule
& \multicolumn{3}{c}{Base VLM training (Pretrain~1)} & Pretrain~2 / SFT \\
\cmidrule(lr){2-4}\cmidrule(l){5-5}
Setting & Stage~1: Projector alignment & Stage~2: Joint multimodal pretraining & Stage~3: General visual/video understanding & Stage~4: Coordinate alignment \\
\midrule
Trainable modules & Projector only & All & All & All \\
Per-device batch & 6 & 2 & 2 & 2 \\
Global batch & 768 & 256 & 256 & 256 \\
Context / packing length & 8192 / 8192 & 8192 / 8192 & 32768 / 32768 & 8192 / 8192 \\
Language peak LR & Frozen & $10^{-5}$ & $10^{-5}$ & $10^{-5}$ \\
Projector peak LR & $5\times10^{-5}$ & $10^{-6}$ & $10^{-6}$ & $10^{-6}$ \\
Vision peak LR & Frozen & $10^{-6}$ & $10^{-6}$ & $10^{-6}$ \\
Weight decay & 0 & 0.1 & 0.1 & 0.1 \\
Minimum LR / peak LR & 2\% & 10\% & 10\% & 10\% \\
Warmup / epoch budget & 3\% / 1 & 3\% / 1 & 3\% / 1 & 3\% / 1 \\
\bottomrule
\end{tabular}
\end{table}

\subsection{Reinforcement Learning (RL)}
\label{app:groundingpi_vlm_rl}

RL is the third overall training phase. We initialize the policy from the coordinate-aligned Pretrain~2 / SFT checkpoint and optimize complete generated outputs with task-specific GRPO rewards.

\paragraph{Policy update.}
We sample image--query pairs $q=(I,P)$ from the training distribution $\mathcal D$ and draw $G=8$ responses per pair from $\pi_{\mathrm{old}}(\cdot\mid q)$. Let $\mathcal B=\{(q_i,Y_i)\}_{i=1}^{N}$ denote the complete response batch, with each sampled prompt repeated for its $G$ responses. Grounding uses two active reward components with weights 0.7 and 0.3; OCR uses one composite reward with weight 1. Inactive task components are excluded. For active component $h$, $Z_h(R_{h,i})=(R_{h,i}-\mu_{h,q_i})/(s_{h,q_i}+\delta)$, where $\mu_{h,q_i}$ and $s_{h,q_i}$ are the mean and sample standard deviation over responses to the same prompt, and $\delta=10^{-8}$. Thus $\widetilde A_i=0.7Z_{\mathrm{set}}(R_{\mathrm{set},i})+0.3Z_{\mathrm{strict}}(R_{\mathrm{strict},i})$ for grounding and $\widetilde A_i=Z_{\mathrm{OCR}}(R_{\mathrm{OCR},i})$ for OCR. These values are standardized across all $N$ responses: $A_i=(\widetilde A_i-\mu_{\widetilde A})/(s_{\widetilde A}+\delta)$. A constant component within a prompt contributes zero before this final normalization.

The same response-level advantage is assigned to each output token. Define the token context $c_{i,t}=(q_i,y_{i,<t})$ and policy ratio $r_{i,t}=\pi_\theta(y_{i,t}\mid c_{i,t})/\pi_{\mathrm{old}}(y_{i,t}\mid c_{i,t})$. We maximize the response-averaged objective
\begin{equation}
\mathcal J(\theta)=\mathbb E_{\mathcal B}\!\left[\frac{1}{N}\sum_{i=1}^{N}\frac{1}{|Y_i|}\sum_{t=1}^{|Y_i|}\left\{\min\!\left(r_{i,t}A_i,\operatorname{clip}(r_{i,t},1-\epsilon,1+\epsilon)A_i\right)-\beta d_{i,t}\right\}\right].
\label{eq:groundingpi-grpo}
\end{equation}
Following GRPO~\citep{GRPO}, the sampled KL penalty is $d_{i,t}=\rho_{i,t}-\log\rho_{i,t}-1\ge0$, where $\rho_{i,t}=\pi_{\mathrm{ref}}(y_{i,t}\mid c_{i,t})/\pi_\theta(y_{i,t}\mid c_{i,t})$. Its expectation equals $D_{\mathrm{KL}}(\pi_\theta\Vert\pi_{\mathrm{ref}})$ when the token is sampled from $\pi_\theta$; evaluated on old-policy rollouts, it is a sampled regularizer rather than an unbiased current-policy KL estimate. We use a frozen copy of the Pretrain~2 / SFT checkpoint as the reference, $\epsilon=0.2$, and $\beta=0.02$. Advantages and the old/reference policies are held fixed during the policy update. Only nonpadding response positions contribute to the objective. The vision encoder and projector remain frozen throughout reinforcement post-training.

\begin{table}[htbp]
\centering\small
\setlength{\tabcolsep}{4pt}
\renewcommand{\arraystretch}{1.1}
\caption{GroundingPI reinforcement post-training configuration.}
\label{tab:groundingpi_vlm_rl}
\begin{tabular}{@{}p{0.43\linewidth}p{0.52\linewidth}@{}}
\toprule
Setting & Value \\
\midrule
GPUs / precision / sharding & 256 / BF16 / ZeRO-2 \\
Responses per prompt & 8 \\
Per-device response batch / accumulation & 1 / 1 \\
Distinct-prompt / response batch & 32 / 256 \\
Updates per rollout & 1 \\
Trainable modules & Language parameters, including embedding and output head \\
Learning rate / schedule & $5\times10^{-7}$; constant; no warmup \\
Optimizer & Fused AdamW; betas $(0.9,0.999)$; $\epsilon=10^{-8}$ \\
Weight decay / gradient clipping & 0.01 / 1.0 \\
Activation recomputation & Language model \\
Seed / data seed & 20260812 / 20260812 \\
\bottomrule
\end{tabular}
\end{table}

\subsubsection{Grounding Rewards}
\label{app:groundingpi_grounding_reward}

\paragraph{Set completeness.}
For a nonempty reference set $\{(g_j,c_j)\}_{j=1}^{n}$, with $n\ge1$, let $\{(b_k,\hat c_k)\}_{k=1}^{m}$ denote the predictions. If $m=0$, set $R_{\mathrm{set}}=0$ without performing a match. Otherwise, select $k_j^\star=\arg\max_{1\le k\le m}\operatorname{IoU}(g_j,b_k)$ for each reference, then validate its class: $s_j=\operatorname{IoU}(g_j,b_{k_j^\star})\mathbb I[c_j=\hat c_{k_j^\star}]$. With $S=\sum_{j=1}^{n}s_j$, soft recall and precision are $R_s=S/n$ and $P_s=S/m$, and $R_{\mathrm{set}}=2P_sR_s/(P_s+R_s+10^{-8})$. Matching is independent for each reference and can reuse a prediction; this F1-style coverage surrogate can therefore exceed one. Empty reference sets lie outside this definition.

\paragraph{Strict localization and output quality.}
The complementary reward is a weighted sum of the components in \Cref{tab:groundingpi_strict_reward}, clipped to $[0,1]$. Its detection term is $\overline F=(F_{0.50}+F_{0.75}+F_{0.95})/3$, where $F_\tau$ is detection F1 at IoU threshold $\tau$; the matched-box IoU term provides continuous localization feedback. Format, count, and ordering scores assess the structured response, while nonnegative duplicate and oversized-box penalties discourage redundant or imprecise predictions. These auxiliary scores are distinct from the OCR count and format terms below. Both grounding components are standardized separately before their weighted combination, as described in \Cref{app:groundingpi_vlm_rl}.

\begin{table}[htbp]
\centering\small
\setlength{\tabcolsep}{5pt}
\caption{Components of the strict grounding reward; the weighted sum is clipped to $[0,1]$.}
\label{tab:groundingpi_strict_reward}
\begin{tabular}{@{}p{0.76\linewidth}r@{}}
\toprule
Component & Weight \\
\midrule
Structured-format validity & 0.10 \\
Agreement between predicted and reference counts & 0.10 \\
Mean detection F1 at IoU $0.50$, $0.75$, and $0.95$ & 0.50 \\
Matched-box IoU & 0.25 \\
Compliance with the output ordering convention & 0.05 \\
Oversized-box penalty & $-0.07$ \\
Duplicate-box penalty & $-0.03$ \\
\bottomrule
\end{tabular}
\end{table}

\subsubsection{OCR Rewards}
\label{app:groundingpi_ocr_reward}

\paragraph{Joint text--geometry matching.}
Let $P=\{(b_i,s_i)\}_{i=1}^{n}$ and $T=\{(\hat b_j,\hat s_j)\}_{j=1}^{m}$ contain predicted and reference text instances, with $n\ge0$, $m\ge1$, and valid geometry and transcriptions. Define $I_{ij}=\operatorname{IoU}(b_i,\hat b_j)$. Text normalization $\mathcal N$ applies Unicode NFKC, case folding, and alphanumeric filtering; symbol-only strings retain distinct Unicode-based keys. For nonempty normalized transcriptions, edit similarity is $E_{ij}=1-\operatorname{Lev}(\mathcal N(s_i),\mathcal N(\hat s_j))/\max(|\mathcal N(s_i)|,|\mathcal N(\hat s_j)|)$. Blank transcriptions and empty reference sets lie outside these definitions. All affinities use Hungarian maximum-weight one-to-one assignment. If $M(A)$ is the assigned affinity sum, define $F(A)=2M(A)/(n+m)$; with no predictions, $M(A)=F(A)=0$.

The hard affinity is $A^{\mathrm{hard},\tau}_{ij}=\mathbb I[\mathcal N(s_i)=\mathcal N(\hat s_j)]\mathbb I[I_{ij}\ge\tau]$, giving $H_\tau=F(A^{\mathrm{hard},\tau})$ and
\[\overline H=\tfrac{1}{10}\sum_{\tau\in\{0.50,0.55,\ldots,0.95\}}H_\tau.\]
The soft affinity is $A^{\mathrm{soft}}_{ij}=\sqrt{I_{ij}}\,E_{ij}\mathbb I[I_{ij}\ge0.10]\mathbb I[E_{ij}\ge0.20]$, with $S_{\mathrm{soft}}=F(A^{\mathrm{soft}})$. These terms provide strict text--region agreement and continuous feedback for partial matches. Count agreement is $C=\min(n,m)/\max(n,m)$, with $C=0$ when $n=0$. The reference-view score is $V(P,T)=\operatorname{clip}_{[0,1]}(0.45\overline H+0.35S_{\mathrm{soft}}+0.10H_{0.50}+0.10C)$.

\paragraph{Word/line granularity.}
Complete, nonempty word and line references $T_w$ and $T_l$ receive scores $V_w=V(P,T_w)$ and $V_l=V(P,T_l)$. Let $V_{\mathrm{hi}}=\max(V_w,V_l)$ and $V_{\mathrm{lo}}=\min(V_w,V_l)$. For a local granularity group $g$, predictions $P_g$ are scored against each complete local representation: $G_g=\max\{V(P_g,T_w^g),V(P_g,T_l^g)\}$. This selects between whole word and line views rather than combining isolated matches from incompatible views.

For singleton reference $j$, let $B_j$ and $S_j$ contain its valid box and text alternatives. Define
\[ I_{ij}^\star=\max_{b\in B_j}\operatorname{IoU}(b_i,b),\qquad E_{ij}^\star=\max_{s\in S_j}E(s_i,s), \]
with the maxima taken independently. The corresponding hard and soft affinities use these starred quantities, with exact text agreement given by $E_{ij}^\star=1$, and retain Hungarian one-to-one assignment. The singleton consensus score is $G_{\mathrm{cons}}=0.60\overline H^\star+0.40S^\star$. Let $\Gamma$ denote the annotation-dependent aggregate of the granularity and singleton scores; conflicting singleton blocks receive reduced weight. The global-view score $D$ and content score $Q_g$ use the coefficients in \Cref{tab:groundingpi_ocr_aggregation}.

\begin{table}[htbp]
\centering\small
\setlength{\tabcolsep}{4pt}
\renewcommand{\arraystretch}{1.13}
\caption{Aggregation of complete word/line views and local granularity groups.}
\label{tab:groundingpi_ocr_aggregation}
\begin{tabular}{@{}p{0.27\linewidth}p{0.34\linewidth}p{0.34\linewidth}@{}}
\toprule
Annotation structure & Global views $D$ & Content score $Q_g$ \\
\midrule
With granularity conflict & $0.90V_{\mathrm{hi}}+0.10V_{\mathrm{lo}}$ & $0.35D+0.55\Gamma$ \\
Without granularity conflict & $0.75V_{\mathrm{hi}}+0.25V_{\mathrm{lo}}$ & $0.50D+0.40\Gamma$ \\
\bottomrule
\end{tabular}
\end{table}

Set $m_g=\max(|T_w|,|T_l|,1)$ and $C_g=\min(n,m_g)/m_g$. Format score $f_g$ is 1 for complete valid output, 0.6 for incomplete structure with reliably parseable instances, 0.3 for partially valid instances, and 0 for unparseable output. The final reward is $R_g=\operatorname{clip}_{[0,1]}((Q_g+0.10f_gC_g)f_g^2-\Pi)$, where $\Pi=0.05P_{\mathrm{dup}}+0.10P_{\mathrm{over}}+0.05P_{\mathrm{invalid}}$ penalizes duplicate, excessive, and invalid predictions.

\paragraph{Complementary references for complex text.}
For curved, rotated, or complex text arrangements, consider three nonempty reference views: two primary views $T_a,T_b$ and a supplementary view $T_c$. Each receives $W=0.85V+0.15U$, where $U$ is text--geometry F1 at IoU 0.50 using exact NFKC/case-folded surface strings with punctuation retained. Primary scores are fused as $W_p=0.70\max(W_a,W_b)+0.30\min(W_a,W_b)$, then $W_t=0.82W_p+0.18W_c$. When a nonempty auxiliary geometry view is available, $Q_c=0.95W_t+0.05\Gamma_{\mathrm{aux}}$, with $\Gamma_{\mathrm{aux}}=0.60\overline H_{\mathrm{geo}}+0.30S_{\mathrm{geo}}+0.10C_{\mathrm{geo}}$; otherwise $Q_c=W_t$. The auxiliary term evaluates geometry only.

Set $m_c=\operatorname{median}(|T_a|,|T_b|,|T_c|)\ge1$ and $C_c=\min(n,m_c)/m_c$. Format scores $f_c$ are 1, 0.45, 0.35, 0.15, and 0 for strictly valid output, complete output with extraneous content, incomplete but parseable structure, recoverable instances without a valid overall structure, and unparseable output, respectively. The reward is $R_c=\operatorname{clip}_{[0,1]}((0.90Q_c+0.10f_cC_c)f_c^2-\Pi-0.08P_{\mathrm{large}})$. The last term penalizes boxes covering over 80\% of the image without supporting reference geometry.

Each OCR sample selects the branch appropriate to its annotations. All content, format, coverage, and penalty terms are combined into one scalar before within-prompt standardization. OCR subterms are not independently standardized; the subsequent response-batch normalization follows \Cref{app:groundingpi_vlm_rl}. Training rewards provide optimization feedback and are distinct from the benchmark metrics.

\section{Training Details and Evaluation Setup}
\label{app:training_setup}

This appendix describes the downstream interfaces and task-specific protocols for the physical-intelligence study in \Cref{sec:physical_intelligence_performance}. Grounding-model optimization is provided separately in \Cref{app:groundingpi_vlm_training,app:groundingpi_coordinate_alignment,app:groundingpi_vlm_rl}.

\subsection{Autonomous Driving}
\label{app:driving_setup}

\paragraph{Observation and trajectory interface.}
The driving task uses the current front-camera image and up to seven ego-state records at 0.5-second intervals, including the current state and covering at most three seconds of history. Historical positions are expressed relative to the current ego pose, together with available velocity, acceleration, and steering information. Short histories retain their observed length; missing states are not replaced by fabricated zeros. The output is six cumulative future positions over three seconds, $Y=((x_1,y_1),\ldots,(x_6,y_6))$, with $x$ forward and $y$ left, measured in meters. Current visual observations and historical states condition the prediction; future waypoints are the supervised targets.

\paragraph{Metric coordinates and serialization.}
The trajectory interface uses fixed ranges $x\in[-60,60]$ and $y\in[-20,20]$ meters. For axis bounds $[a,b]$ with $a<b$ and an in-range coordinate $z$, define the integer code $q_z=\operatorname{round}_{\mathrm{even}}(999(z-a)/(b-a))\in\{0,\ldots,999\}$ and reconstruct $\hat z=a+(b-a)q_z/999$. Rounding uses ties to even. These are metric-coordinate bins, distinct from image normalization in \Cref{app:groundingpi_protocol}. GroundingPI's existing atomic coordinate vocabulary represents the trajectory through one \gpitok{trajectory} entry with six ordered coordinate pairs. Temporal order and repeated points are retained; the output describes cumulative locations, so no additional cumulative sum is applied. The adaptation objective is causal cross-entropy on the trajectory response.

\paragraph{Open-loop metric.}
Predictions are decoded to meters and compared with unquantized reference waypoints. Let $\bar d_k$ be the mean Euclidean error at future step $k$. The cumulative-horizon metric is $\mathrm{L2}@h=(2h)^{-1}\sum_{k=1}^{2h}\bar d_k$ for $h\in\{1,2,3\}$ seconds, and the reported average is $\mathrm{L2}_{\mathrm{Avg}}=\tfrac13\sum_{h=1}^{3}\mathrm{L2}@h$. This averages errors up to each horizon, rather than only the endpoint errors. Open-loop trajectory error measures agreement with recorded trajectories; it does not establish closed-loop control performance.

\subsection{Robot Manipulation}
\label{app:manipulation_setup}

The manipulation study compares the foundations used by vision-language-action (VLA) and world-action model (WAM) systems without allowing their native action heads, conditioning paths, or optimization budgets to become additional variables. We replace the model-specific control modules with the same backbone-to-action interface and the same fixed-capacity action expert. The evaluated variable is therefore the pretrained foundation that supplies the representation for action learning.

The compared foundations are Qwen3-VL-4B~\citep{qwen3vl} and PaliGemma-3B~\citep{Paligemma}; Wan2.2-TI2V-5B~\citep{WAN} and Cosmos-Predict2.5-2B~\citep{CosmosPredict25}; LocateAnything-3B~\citep{locateanything} and Rex-Omni-3B~\citep{rexomni}; and RynnBrain-2B~\citep{dang2026rynnbrain} and RynnBrain1.1-2B~\citep{RynnBrain112B}, alongside GroundingPI. These groups characterize the pretrained foundations rather than different downstream action heads.

\subsubsection{Unified Backbone-to-Action Interface}

\paragraph{Layer-wise backbone features.}
For each observation and language instruction, the foundation backbone is executed once. Given a backbone with $N_B$ transformer blocks, we sample eight hidden states at normalized depths $\ell_j=\operatorname{round}(j(N_B-1)/7)$ for $j\in\{0,\ldots,7\}$.
This rule preserves comparable relative depths across foundations with different numbers of layers and avoids selecting backbone-specific layers. Each sampled state $H_{\ell_j}$ is normalized, mapped from the backbone width $D_B$ to the common action width by a backbone-specific linear projector, augmented with a learned depth embedding, and compressed by a shared learned-query resampler: $Z_j=\operatorname{LN}(H_{\ell_j})W_B+e_j$ and $C_j=\operatorname{Resampler}_{64}(Z_j)$.
The projector $W_B$ is shared across the eight depths of a backbone, and the same resampler is shared across both depths and models. Every cross-attention block consequently receives the same number and width of condition tokens, independent of the backbone's native width or token count.

\paragraph{Controlling the VLA--WAM comparison.}
For VLA foundations, the eight states are taken from the language-conditioned vision--language stack. For WAM foundations, they are taken from the video-generation backbone under a fixed feature-extraction state, including the observed-frame construction, diffusion timestep, condition mask, noise realization, input resolution, and number of frames. In both cases, a single backbone forward pass produces all eight conditions, which are cached and reused throughout action denoising. We remove any model-specific raw-text or proprioceptive cross-attention bypass: language, perception, and dynamics information can reach the action expert only through the evaluated foundation representations. Thus, the two paradigms differ in the pretrained foundation that constructs the physical representation, while sharing the same downstream control architecture and compute schedule.

\subsubsection{Fixed Layer-wise Action DiT}

The common action expert is a fixed-depth, fixed-width $\pi$-style~\citep{black2024pi0} Action DiT~\citep{ActionDiT} trained with flow matching~\citep{FlowMatching}. It contains 16 \emph{atomic} transformer blocks, alternating between layer-wise cross-attention and action self-attention. Blocks $0,2,\ldots,14$ attend to $C_0,C_1,\ldots,C_7$, respectively, and blocks $1,3,\ldots,15$ perform self-attention over the action-side sequence. The architecture therefore contains eight cross-attention blocks, eight self-attention blocks, and 16 feed-forward networks; it is not a stack of 16 paired self- and cross-attention blocks.

The action-side input concatenates the encoded proprioceptive state, learned planning tokens, and a noisy action chunk with its flow timestep. Timestep-conditioned adaptive layer normalization is applied before attention, whereas the feed-forward path uses standard layer normalization. Only the action-token positions are decoded. Action and state input/output projections are embodiment-specific, but the transformer capacity and conditioning interface are identical across all foundations. The complete shared configuration is summarized in \Cref{tab:unified_action_config}.

\begin{table}[htbp]
    \centering
    \small
    \caption{Unified action architecture used for every manipulation foundation.}
    \label{tab:unified_action_config}
    \begin{tabular}{@{}p{0.34\linewidth}p{0.58\linewidth}@{}}
        \toprule
        \textbf{Component} & \textbf{Shared configuration} \\
        \midrule
        Backbone conditions & 8 normalized-depth hidden states \\
        Condition interface & 64 tokens per depth, width 1024 \\
        Action topology & 16 atomic blocks: 8 cross-attention and 8 self-attention blocks, interleaved from cross-attention \\
        Residual / attention width & 1024; 16 heads; head dimension 64 \\
        Feed-forward width & 4096 \\
        Planning tokens & 32 \\
        Context bypass & None \\
        Action horizon & 16 \\
        Time sampling & Beta base distribution $(\alpha,\beta)=(1.5,1.0)$, $s=0.999$, 1,000 timestep buckets \\
        Inference & 4 Euler steps; one cached backbone forward per observation \\
        \bottomrule
    \end{tabular}
\end{table}

Given a normalized action chunk $a$, Gaussian noise $\epsilon$, and flow timestep $t$, training constructs $a_t=(1-t)\epsilon+ta$ with target velocity $v^\star=a-\epsilon$ and minimizes $\lVert \hat v_{\theta}(a_t,t)-v^{\star}\rVert_2^2$ on the action positions. Every model uses the same noise distribution, action normalization, and solver settings listed in \Cref{tab:unified_action_config}.

\subsubsection{Training Budget and Evaluation Protocol}

Within each benchmark, every foundation is trained on the same action demonstrations with the same sampling and augmentation pipeline. RoboTwin 2.0 Full uses the Clean and Randomized protocols, whereas Clean2Random uses only Clean demonstrations for training and holds Randomized scenes out for evaluation. RoboCasa-GR1 uses the same task suite and demonstration pool for every foundation. The reduced-data experiments change only the available fraction of the action dataset. All other optimization settings are matched across foundations, as summarized in \Cref{tab:manipulation_training_budget}.

\begin{table}[htbp]
    \centering
    \small
    \caption{Matched downstream training budget for the manipulation comparison.}
    \label{tab:manipulation_training_budget}
    \begin{tabular}{@{}p{0.34\linewidth}p{0.58\linewidth}@{}}
        \toprule
        \textbf{Item} & \textbf{Configuration} \\
        \midrule
        Optimization length & 40,000 updates \\
        Hardware / precision & 40 accelerators; bfloat16; DeepSpeed ZeRO-2~\citep{ZeRO2} \\
        Batch size & 32 per device; 1280 global \\
        Optimizer & AdamW~\citep{AdamW}, $\beta_1=0.95$, $\beta_2=0.999$, $\epsilon=10^{-8}$ \\
        Weight decay / clipping & $10^{-5}$; gradient norm 1.0 \\
        Learning rates & $10^{-4}$ for the action expert and interface; $10^{-5}$ for adapted backbone parameters \\
        Schedule & Cosine decay; 2,000 warmup steps; minimum LR $5\times10^{-7}$ \\
        Flow samples & 2 noise samples per training observation \\
        RoboTwin 2.0 data & Full: Clean and Randomized; Clean2Random: Clean training, Randomized evaluation \\
        RoboCasa-GR1 data & Identical demonstration pool across foundations at each data fraction \\
        \bottomrule
    \end{tabular}
\end{table}

This protocol fixes the quantities most likely to confound a backbone comparison: action-head depth and width, condition-token budget, state and action paths, action data, batch size, number of updates, optimizer and schedule, flow objective, and inference computation. Backbone-specific width projectors are the only interface parameters whose size varies with the native backbone width; they are shared across depth and remain small relative to the common action expert. The comparison therefore measures how effectively the representations inherited from VLA- and WAM-style foundations support effective and generalizable action learning under matched downstream capacity and supervision.

\section{Additional Transfer and Ablation Analyses}
\label{app:transfer_analysis}

\subsection{Data Efficiency and the Burden on Action Demonstrations}
\label{app:data_efficiency_analysis}

\Cref{fig:data-efficiency} varies the fraction of RoboCasa-GR1 demonstrations while retaining the action architecture and remaining training choices. GroundingPI leads every reduced-data setting. Its 28.75\% SR with half the demonstrations exceeds the strongest baseline trained on three quarters, RynnBrain at 27.75\%. At full data, RynnBrain reaches 39.00\%, compared with GroundingPI's 37.75\%. Thus, the main advantage in this experiment is effective adaptation with limited demonstrations, rather than the highest full-data ID score.

The distinction between learning \emph{where to interact} and \emph{how to act} interprets this result in terms of the paper's motivation. A reusable perceptual foundation may let downstream supervision focus more on control, but the experiment does not separately measure perceptual and motor-learning sample complexity. It also does not vary physical prompts. The evidence concerns downstream demonstration efficiency in manipulation; it does not establish lower total pretraining cost, a universal data-scaling law, or driving-data efficiency.

\subsection{Grounding and Action Scaling}
\label{app:scaling_analysis}

\Cref{tab:gpi_scaling_results} records the performance trajectory in \Cref{fig:pretraining-scaling}, ordered by increasing grounding-training exposure. The initial backbone, downstream action architecture, action demonstrations, and action-training recipe are held fixed.

\begin{ArxivInlineBlock}
\sbox{\ArxivNaturalTable}{%
\small

\begin{tabular}{@{}lrrr@{}}
\toprule
Exposure order & Grounding Avg & RoboCasa-GR1 ID SR & RoboCasa-GR1 OOD Avg SR \\
\midrule
1 & 61.69 & 33.92 & 26.59 \\
2 & 69.71 & 35.92 & 30.00 \\
3 & 72.22 & 38.75 & 32.06 \\
4 (full) & 73.68 & 37.75 & 33.38 \\
\bottomrule
\end{tabular}%
}

\begin{wraptable}{r}{\wd\ArxivNaturalTable}
\centering\small
\caption{Performance at successive grounding-training exposures. The last row is the full configuration.}
\label{tab:gpi_scaling_results}
\usebox{\ArxivNaturalTable}
\end{wraptable}

The endpoint gains are 11.99 pp in grounding, 3.83 pp in ID SR, and 6.78 pp in OOD Avg SR, using the recorded values before rounding. Grounding and OOD performance improve at every step, whereas ID SR falls by 1.00 pp at the final step. The evidence supports joint improvement in perception and transfer, especially generalization, but not a monotonic law connecting grounding score to every action metric. Four settings without repeated-seed uncertainty estimates are insufficient to establish a scaling law or determine whether the final ID decrease is systematic.
\end{ArxivInlineBlock}

RoboCasa-GR1 OOD Avg weights Container, Appearance, and Type by their evaluation-suite sizes:
\begin{equation}
\mathrm{SR}_{\mathrm{OOD}}=
\frac{14\,\mathrm{SR}_{\mathrm{Container}}+18\,\mathrm{SR}_{\mathrm{Appearance}}+32\,\mathrm{SR}_{\mathrm{Type}}}{64}.
\label{eq:gpi_ood_average}
\end{equation}
These are evaluation weights, not training-mixture proportions. The aggregate is computed before rounding the displayed suite scores.

\subsection{Which Perceptual Capabilities Transfer?}
\label{app:composition_analysis}

\sbox{\ArxivNaturalTable}{%
\small

\begin{tabular}{@{}lrrr@{}}
\toprule
Included groups & RoboCasa-GR1 ID & RoboCasa-GR1 OOD Avg & nuScenes L2 Avg \\
\midrule
A+D+F & 30.92 & 29.28 & 0.310 \\
A+B+C+F & 34.92 & 31.28 & 0.299 \\
A+B+C+D+F & 35.83 & 31.84 & 0.296 \\
A+B+D+E+F & 35.67 & 32.28 & 0.298 \\
A+C+D+E+F & 33.08 & 29.91 & 0.311 \\
A+B+C+D+E & 32.42 & 29.75 & 0.308 \\
F & 24.42 & 7.28 & 0.391 \\
All six & 37.75 & 33.38 & 0.296 \\
\bottomrule
\end{tabular}%
}
\FinishArxivWrap
\begin{wraptable}{R}{\wd\ArxivNaturalTable}
\centering\small
\caption{Task-group ablation. Manipulation uses SR (\%, higher is better); driving uses average open-loop L2 error (m, lower is better).}
\label{tab:gpi_composition_results}
\usebox{\ArxivNaturalTable}
\end{wraptable}

The six groups retain the naming in \Cref{fig:grounding-data-mix}: A Basic Grounding; B Dense Grounding; C Referring; D Basic Pointing; E Robo Pointing; and F Else (OCR, Layout, GUI). Group membership specifies task inclusion only. \Cref{tab:gpi_composition_results} reports the downstream results; the action-learning setup is unchanged across configurations.

\paragraph{Basic grounding provides an important foundation.}
Adding A and D to F raises ID SR from 24.42\% to 30.92\%, OOD Avg SR from 7.28\% to 29.28\%, and reduces driving error from 0.391 to 0.310\,m. This supports the importance of basic object localization together with pointing, particularly for generalization. Since A and D change jointly and no A-only leave-one-out row is available, their individual contributions cannot be identified. The table does not establish that either group is independently necessary or sufficient.

\paragraph{Dense grounding has broad marginal value.}
Removing B reduces ID/OOD SR by 4.67/3.47 pp and increases L2 error by 0.015\,m, larger changes than removing C or E. Dense and tiny-object perception share the need to preserve small spatial distinctions and separate nearby instances. Such demands plausibly recur in cluttered manipulation and distant road objects, making this supervision relevant across domains. The intervention removes the dense group as a whole; it does not disentangle object size, crowding, or annotation coverage.

\paragraph{Why tiny targets demand precise localization.}
For equal axis-aligned boxes of width $w>0$ and height $h>0$ displaced only horizontally by $\delta$, with $|\delta|<w$, the intersection and union areas are $(w-|\delta|)h$ and $(w+|\delta|)h$. Thus, for $0<\tau<1$,
\begin{equation}
\operatorname{IoU}(\delta)=\frac{w-|\delta|}{w+|\delta|},
\qquad
\operatorname{IoU}(\delta)\geq\tau
\;\Longleftrightarrow\;
|\delta|\leq w\frac{1-\tau}{1+\tau}.
\label{eq:gpi_tiny_iou}
\end{equation}
The tolerated absolute displacement shrinks linearly with target width. This illustrates the precision demanded by tiny targets and complements the need to separate nearby instances in dense scenes.

\paragraph{Embodied labels are not the only route to action.}
Removing E reduces ID/OOD SR by 1.92/1.53 pp and leaves driving error unchanged at the displayed precision. This is a smaller marginal effect than removing B, not evidence that embodied supervision is useless: its information may overlap with other groups, and the tasks may not emphasize every affordance it teaches. Removing C causes 2.08/1.09 pp losses and a 0.002\,m error increase. Existing language understanding may reduce its marginal benefit, but that explanation would require a controlled change to the language foundation.

\paragraph{Complementarity, not independent effect sizes.}
Removing F produces the largest manipulation decreases, 5.33/3.63 pp, yet F alone is the weakest configuration. These observations establish complementarity at the group level. The full mixture is best on both manipulation metrics and ties the best displayed driving error; it is not uniquely best on every metric. Leave-one-out changes depend on the retained groups and should not be added as independent contributions. No interaction effect or OCR-only causal effect is identified by this incomplete factorial design.

\subsection{OCR as a Perceptual Catalyst: A Hypothesis}
\label{app:ocr_catalyst}

\begin{ArxivInlineBlock}
\sbox{\ArxivNaturalTable}{%
\small
\setlength{\tabcolsep}{6pt}
\begin{tabular}{@{}lccc@{}}
\toprule
& \multicolumn{2}{c}{RoboCasa-GR1} & nuScenes \\
\cmidrule(lr){2-3}
OCR schedule
& ID SR (\%$\uparrow$)
& OOD Avg. SR (\%$\uparrow$)
& L2 Avg. (m$\downarrow$) \\
\midrule
Cold-start mixing
& 35.17 & 32.38 & 0.296 \\
Batch mixing
& 34.92 & 32.53 & 0.296 \\
Cold-start + batch mixing
& \textbf{37.75} & \textbf{33.38} & 0.296 \\
\bottomrule
\end{tabular}%
}

\begin{wraptable}{r}{\wd\ArxivNaturalTable}
\centering
\small
\setlength{\tabcolsep}{6pt}
\caption{\textbf{OCR mixing schedules.} We vary when OCR supervision is incorporated to probe its catalytic role in downstream transfer. Cold-start + batch mixing corresponds to the full six-group configuration.}
\label{tab:gpi_ocr_mixing}
\usebox{\ArxivNaturalTable}
\end{wraptable}

\paragraph{OCR mixing schedules.}
To probe OCR's catalytic role, we compare cold-start mixing, batch mixing, and their combination (\Cref{tab:gpi_ocr_mixing}). The combined schedule achieves the highest manipulation success on both ID and OOD splits, while nuScenes L2 remains unchanged at the reported precision. This pattern is consistent with complementary benefits from early OCR exposure and continued joint training for manipulation transfer, motivating the following analysis of transcription beyond geometric supervision.

OCR may complement grounding by coupling fine visual discrimination with region--text alignment, providing a localized captioning proxy for perceptual learning (\Cref{sec:grounding_to_action_transfer}). This motivation is consistent with the benefits of local visual semantics~\citep{LocalityAlignment}. The hypothesis concerns the contribution of transcription beyond shared geometric supervision.
\end{ArxivInlineBlock}

\paragraph{Transcription beyond geometry.}
Partition supervised OCR response positions into box, transcription, and protocol roles, $\mathcal T_b$, $\mathcal T_t$, and $\mathcal T_p$, with nonempty union $\mathcal T$. The SFT loss decomposes as
\begin{equation}
\mathcal L_{\mathrm{OCR}}=\mathcal L_b+\mathcal L_t+\mathcal L_p,
\qquad
\mathcal L_r=-\frac{1}{|\mathcal T|}\sum_{i\in\mathcal T_r}
\log p_\theta(y_i\mid I,Q,y_{<i}),\quad r\in\{b,t,p\}.
\label{eq:gpi_ocr_components}
\end{equation}
Because transcription precedes coordinates, teacher-forced box prediction already conditions on the reference text. A \emph{box-only} control therefore removes $\mathcal L_t$ while retaining protocol supervision, all transcription tokens, and the original denominator $|\mathcal T|$. Comparing it with the full loss isolates the additional transcription gradient under identical conditioning and matched training budgets. Deleting transcription tokens changes the box-prediction task; renormalizing over unmasked positions changes the weight of the retained losses.

\paragraph{Local transfer condition.}
For trainable visual parameters $\theta_v$, let $g_t=\nabla_{\theta_v}\mathcal L_t$ and $g_g=\nabla_{\theta_v}\mathcal L_g$, where $\mathcal L_g$ is a non-text grounding loss. With other parameters fixed and Hessian norm bounded by $H$ along the update,
\begin{equation}
\mathcal L_g(\theta_v-\eta g_t)-\mathcal L_g(\theta_v)
\leq-\eta\langle g_g,g_t\rangle
+\frac{H\eta^2}{2}\lVert g_t\rVert_2^2.
\label{eq:gpi_ocr_alignment}
\end{equation}
Positive alignment permits a local decrease for sufficiently small $\eta>0$; total OCR-gradient alignment could instead arise from box supervision alone. For AdamW, the corresponding first-order diagnostic is $\langle g_g,\Delta\theta_v\rangle$ using the actual update. The catalyst hypothesis predicts that adding transcription loss under the fixed-conditioning control improves non-text dense/tiny grounding and its subsequent transfer to action.

\subsection{Output Representation and Generation Cost}
\label{app:efficiency_analysis}

The coordinate ablation favors quantization by 2.47 pp in grounding Avg. The reported textual speed is 0.25$\times$ the quantized configuration's speed; the figure does not specify a closed-loop action-latency metric. The visual-encoder gaps are 0.51 pp for MoonViT and 2.03 pp for Qwen3-ViT relative to MoonViT-V2, while reinforcement learning contributes 0.82 pp under the reported comparison. These are configuration-level results; the encoder comparison also changes pretrained visual representations and does not isolate cross-layer fusion.

In \Cref{tab:grounding-token-efficiency}, SEED1.5-VL statistics are external values from \citep{rexomni}, rather than a new reproduction. Dividing the reported tokens per box gives approximately 19.6$\times$ and 14.6$\times$ fewer tokens for GroundingPI on COCO and Dense200. These ratios quantify serialization, not measured speedups: model computation, sampling, and the distribution of predictions also matter. Four atomic coordinates specify a box, with additional tokens for labels, separators, and wrappers. Sharing one label wrapper across multiple instances amortizes overhead, explaining why tokens per box can fall in dense images even as total output length rises. \Cref{fig:efficiency-report} is consistent with increasing generation cost as more boxes are emitted; it neither measures action execution nor establishes real-time performance.

\section{Embodied-Foundation Design: Mechanisms and System Roles}
\label{app:base_model_discussion}

We analyze how pretrained visual features reach the manipulation action interface under a new objective. These local, conditional analyses characterize feature access and adaptation; the cross-model comparisons do not isolate architectural causes.

\subsection{Architectural Facts and Scope of Comparison}
\label{app:base_architecture_facts}

RynnBrain-2B uses Qwen3-VL with full-attention language layers~\citep{dang2026rynnbrain,qwen3vl}. RynnBrain1.1-2B uses Qwen3.5 with 18 linear-attention and six full-attention layers, plus attention-output gating~\citep{RynnBrain112B,qwen35}. Both employ DeepStack; here this denotes intermediate ViT-feature injection into early language layers~\citep{RynnBrain112B,qwen3vl}. Qwen3-VL injects projected features into its first three language layers. GroundingPI combines MoonViT-V2, a single visual interface, and a full-attention Qwen3-4B decoder.

RynnBrain1.1 trails RynnBrain in manipulation but improves driving, motivating analysis of task-dependent feature access. The comparison fixes downstream capacity and optimization while varying complete pretrained foundations. RynnBrain and Qwen3-VL share DeepStack, whereas GroundingPI also differs in encoder, connector, pretraining, positional treatment, and language-model scale. The Qwen2.5-VL foundations of Rex-Omni and LocateAnything leave general base-model quality as another potential influence alongside grounding specialization~\citep{rexomni,locateanything}.

\subsection{Gated Memory and Attention: Conditional Transfer Sensitivities}
\label{app:gated_transfer}

For the eight-depth manipulation interface in \Cref{app:manipulation_setup}, write
\begin{equation}
C_s=\mathcal R\!\left(\operatorname{LN}(H_{\ell_s})W_B+e_s\right),\quad s=0,\ldots,7,
\qquad
\hat y=\mathcal A_\phi(\xi;C_0,\ldots,C_7).
\label{eq:gpi_action_readout}
\end{equation}
Here $\mathcal R$ is the resampler, $W_B$ the shared projector, $e_s$ the depth embedding, and $\mathcal A_\phi$ the Action DiT predicting flow velocity. Local derivatives hold parameters and action-side inputs $\xi$ fixed.

\paragraph{Recurrent transmission.}
For one GDN head~\citep{GatedDeltaNet}, let $k_t,q_t\in\mathbb R^{d_k}$, $v_t\in\mathbb R^{d_v}$, and $S_t\in\mathbb R^{d_v\times d_k}$, with
\begin{equation}
S_t=S_{t-1}A_t+\beta_t v_tk_t^\top,
\qquad A_t=\alpha_t(I-\beta_t k_tk_t^\top),
\qquad o_t=S_tq_t.
\label{eq:gpi_gdn_update}
\end{equation}
Assume $0\leq\alpha_t,\beta_t\leq1$, $\lVert k_t\rVert_2\leq1$, and uniformly $\lVert q_t\rVert_2\leq Q$, consistent with normalized keys and queries. Fix all keys, queries, gates, other values, and the preceding state, and perturb only $v_j$. Expanding the recurrence gives
\begin{align}
\Delta o_t&=\kappa_{jt}\Delta v_j,
\qquad \kappa_{jt}=\beta_j k_j^\top A_{j+1}\cdots A_tq_t,\quad t\geq j,
\label{eq:gpi_gdn_path}\\
|\kappa_{jt}|&\leq\beta_j Q\prod_{r=j+1}^{t}\alpha_r.
\label{eq:gpi_gdn_bound}
\end{align}
The ordered product is $I$ for $t=j$; indices denote sequence positions, not physical distance. The bound follows from $\lVert A_t\rVert_2\leq\alpha_t$ and decays as $\bar\alpha^{t-j}$ if $\alpha_r\leq\bar\alpha\in(0,1)$ uniformly along the path.

\paragraph{Access through action conditions.}
Treat $o_1,\ldots,o_T$ as independent inputs to the downstream network. Let $R_t=\partial\hat y/\partial o_t$ include subsequent backbone layers, all eight feature readouts, projection, resampling, and the Action DiT. The conditional value-path Jacobian is
\begin{equation}
J_j^{\mathrm{rec}}=\sum_{t=j}^{T}\kappa_{jt}R_t,
\qquad
\nabla_{v_j}^{\mathrm{rec}}\mathcal L_{\mathrm{act}}
=(J_j^{\mathrm{rec}})^\top\nabla_{\hat y}\mathcal L_{\mathrm{act}}.
\label{eq:gpi_gdn_action_access}
\end{equation}
The $t=j$ term retains access through the token's own hidden state. Action sensitivity thus depends jointly on $\kappa_{jt}$ and $R_t$, including downstream amplification or cancellation; image perturbations can additionally follow residual and full-attention paths.

\paragraph{Output gating.}
For gated softmax attention~\citep{GatedAttention}, write $b=W_o(g\odot u)$, where $u$ is the attention output and $g=\sigma(z)$ is an elementwise sigmoid gate. With $\lambda=\nabla_b\mathcal L_{\mathrm{act}}$,
\begin{equation}
\nabla_u\mathcal L_{\mathrm{act}}=g\odot W_o^\top\lambda,
\qquad
\nabla_z\mathcal L_{\mathrm{act}}
=g\odot(1-g)\odot u\odot W_o^\top\lambda.
\label{eq:gpi_gate_gradients}
\end{equation}
Small gates attenuate branch-feature gradients, while saturation attenuates gate-logit gradients; residual paths and optimizer rescaling remain. Spatial readout from the eight resampled conditions is therefore the relevant diagnostic of action-feature accessibility, beyond retention or output-gate values alone.

\subsection{Multi-level Visual Injection and Action Readout}
\label{app:deepstack_transfer}

DeepStack enriches visual evidence through intermediate feature injection~\citep{DeepStack,qwen3vl}. Under limited action adaptation, its utility may depend on compatibility with the shared projector and resampler in \Cref{eq:gpi_action_readout}. GroundingPI uses a single ViT-to-language interface, but both designs provide eight feature levels to the Action DiT. The hypothesis in \Cref{sec:effects_base_model} therefore concerns adaptation of a shared cross-depth readout.

\paragraph{Readout sensitivity.}
For vectorized states, write $h_{\ell+1}=F_\ell(h_\ell)+U_\ell f_\ell$, where $U_\ell$ inserts projected visual features $f_\ell$ at visual-token positions. Let $\mathcal I$ index injection layers, $\mathcal J_\ell=\partial F_\ell/\partial h_\ell$, $h_{m_s}=\operatorname{vec}(H_{\ell_s})$, and $c_s=\operatorname{vec}(C_s)$. With initial state, parameters, and action-side inputs fixed,
\begin{align}
B_\ell&=\sum_{s:m_s>\ell}
\frac{\partial\hat y}{\partial c_s}
\frac{\partial c_s}{\partial h_{m_s}}
\mathcal J_{m_s-1}\cdots\mathcal J_{\ell+1}U_\ell,\nonumber\\
\delta\hat y&=\sum_{\ell\in\mathcal I}B_\ell\,\delta f_\ell
+O(\lVert\delta f\rVert_2^2).
\label{eq:gpi_deepstack_jacobian}
\end{align}
Empty products are identities; $\delta f$ concatenates feature perturbations, and the remainder assumes locally bounded second derivatives. Each $B_\ell$ includes all affected action readouts. This quantifies sensitivity to feature perturbations, which are not themselves prediction errors.

\paragraph{Task alignment under finite adaptation.}
For two injection settings with other parameters and inputs unchanged, let $e_0=\hat y_0-y^\star$ denote error relative to the flow-matching target and $d=\hat y_1-\hat y_0$ the prediction change. Then
\begin{equation}
\mathbb E\lVert e_0+d\rVert_2^2-\mathbb E\lVert e_0\rVert_2^2
=2\mathbb E[e_0^\top d]+\mathbb E\lVert d\rVert_2^2.
\label{eq:gpi_deepstack_task_error}
\end{equation}
Additional evidence helps when correction of existing error outweighs the quadratic term; for small changes, $d\approx\sum_\ell B_\ell\delta f_\ell$. If extra branches can be zeroed with all other components unchanged, the multi-injection model retains the single-interface model as a special case, so path count alone cannot raise optimal task loss. Any practical difficulty instead concerns learning a compatible readout within the available data and optimization budget. Shared projection and resampling couple adaptation across feature levels, while depth embeddings and separate action cross-attention blocks can compensate. A matched-capacity comparison of shared and depth-specific projectors across action-data budgets would directly test this compatibility hypothesis.

\subsection{Designing Perceptual Foundations for System~1}
\label{app:sys1_design}

\paragraph{Different roles may favor different foundations.}
A useful design hypothesis separates knowledge-intensive deliberation from rapid perception--action execution. A planning system may benefit from extensive world knowledge, coding, and long-horizon reasoning. An execution system must translate the current goal and observations into reliable interaction while incorporating timely feedback. Existing dual-system architectures provide concrete precedents: RoboDual couples a generalist with a specialist policy, and GR00T N1 couples vision--language interpretation with a diffusion action module~\citep{RoboDual,nvidia2025gr00tn1}. These systems motivate the distinction; they do not establish that one universal division of modules is optimal.

Here, System~1 denotes the functional perception--action execution loop. Its module boundaries need not match the naming in another architecture: GR00T N1 calls its VLM System~2 and its diffusion action module System~1. Our proposal concerns the perceptual foundation supplying an action learner, rather than identifying an unadapted grounding VLM with a complete controller. The benefit sought is precise, task-conditioned information available when an action must be selected or corrected.

\paragraph{Execution quality can limit the value of planning.}
The coding-agent analogy highlights the need for a dependable execution layer: increasingly sophisticated generated programs have limited practical value if they repeatedly fail to run or if their feedback cannot guide correction. In robotics, stronger plans similarly depend on correctly identifying interaction targets and executing feasible actions. Code as Policies offers a concrete connection between these roles by generating programs that combine perception outputs with control APIs~\citep{CodeAsPolicies}. Our analogy motivates evaluating the complete feedback loop; it is not an empirical comparison between coding and robot-learning systems.

\paragraph{What perception-native pretraining may contribute.}
Dense and tiny-object grounding requires distinctions that can matter directly for interaction: which instance is intended, where it is, and which local region is relevant. Basic grounding supplies reusable localization, while OCR may complement it through local visual precision and semantic alignment. These observations suggest pretraining priorities for a foundation supporting execution. Broad semantic knowledge remains useful, but may be insufficient when the limiting factor is spatial discrimination. The controlled backbone comparison supports this concern for the evaluated Qwen- and PaliGemma-based foundations; it does not establish inferiority of every adaptation recipe or every member of the $\pi$ family. Grounding also leaves dynamics, contact, and feedback control to the downstream policy.

\paragraph{Interpreting the Astra comparison.}
Astra is a frontier reference for grounding quality in this study. Comparing a compact grounding model with it assesses how closely specialized perception can approach a strong general-purpose reference across the evaluated capabilities. It does not define a theoretical performance ceiling or evaluate Astra as a VLA backbone. Assigning a frontier model to higher-level planning and a perception-native foundation to execution is a prospective design, compatible with the observed capability differences but not tested as an integrated system here.

\paragraph{Tests of the proposed division.}
A direct evaluation would hold the planner and action expert fixed while changing the perceptual foundation, then measure task success, robustness under distribution shift, and observation-to-action latency. Perception failures should be distinguished from planning and control failures, especially for small, crowded, or ambiguous targets. Conversely, varying planner capability under a fixed execution layer would test when better reasoning helps and when interaction remains the limiting factor. The present results motivate these experiments through grounding quality, controlled action transfer, and manipulation-data efficiency; they do not yet demonstrate a real-time dual-system controller.

\section{Limitations and Future Work}
\label{sec:limitations}

GroundingPI's aggregate strength is not uniform dominance: tiny-box localization, high-IoU boundaries, TotalText OCR, FSC147 visual prompting, and frontier-level GUI grounding retain clear gaps. Prompt and parser compatibility also constrain several baseline measurements; unsupported evaluations are not evidence of absent intrinsic capability. The manipulation comparison controls the downstream recipe, but pretrained foundations differ in scale, architecture, and supervision. Driving is evaluated open loop, and action-data efficiency is tested only in manipulation. OCR-specific causality and the proposed architectural mechanisms remain hypotheses requiring matched interventions. Finally, physical prompting and complementary System-1 execution alongside frontier reasoning are directions suggested by the interface, rather than capabilities independently established by the current experiments.

\section{Comprehensive Grounding Benchmark Results}
\label{app:grounding_results}

This appendix reports the complete recorded baseline comparisons underlying the selected main-text tables, including local evaluation results and externally reported baselines. Published reference scores replace local entries only for the explicitly marked substitutions. Missing metrics are not reconstructed from other scores.

\paragraph{Reporting conventions.}
The notation defined in \Cref{sec:benchmark_reporting} applies throughout. R and P denote recall and precision. For box grounding, the nine columns report R/P/F1 at IoU 0.50 and 0.95, followed by their mIoU aggregates. Object pointing uses point-in-mask R/P/F1. OCR uses the recorded loose-match F1 metrics; parse error is the percentage of outputs that cannot be parsed. Named dataset-family averages are unweighted arithmetic means and are reported only when every component is available. Threshold-specific and mean-overlap metrics are retained as recorded; missing recall or precision is not inferred from F1. The headline Avg is the mean of 11 independently computed capability scores. Benchmark-level results below are reported separately and do not define this average through a direct mean of their table entries. Size groups use the language backbone configuration (the 7B understanding branch for MoT models and total language-model parameters for MoE models). Qwen3.7-Max, Kimi-K2.6, Kimi-K3, and GPT-6 Astra are grouped separately in the >1T category.

\paragraph{Benchmark suite and metrics.}
Common and long-tailed detection use COCO~\citep{COCO} and LVIS~\citep{LVIS}; dense and tiny-object detection use Dense200~\citep{rexomni} and VisDrone~\citep{VisDrone}. Referring grounding covers HumanRef~\citep{HumanRef}, RefCOCO and RefCOCO+~\citep{RefCOCO}, and RefCOCOg~\citep{RefCOCOg,RefCOCOgUMD}. Spatial pointing uses RefSpatial~\citep{RoboRefer2B} and RoboSpatial~\citep{RoboSpatial}. GUI grounding uses ScreenSpot-Pro~\citep{ScreenSpotPro}, ScreenSpot-V2~\citep{ScreenSpotV2}, and OSWorld-G~\citep{JEDI3B}. OCR covers HierText~\citep{HierText}, ICDAR2015~\citep{ICDAR2015}, TotalText~\citep{TotalText}, and SROIE~\citep{SROIE}; layout grounding uses DocLayNet~\citep{DocLayNet} and M6Doc~\citep{M6Doc}. Visual prompting uses FSC147~\citep{FSC147} and exemplar-conditioned detection. RefCOCO avg is the arithmetic mean of the three family-level F1mIoU scores, distinct from the separately reported RefCOCOg validation/test columns. RefSpatial avg averages Location and Placement. ScreenSpot reports action accuracy, and OSWorld-G reports exact accuracy. Scores marked as external retain their source protocol and are not presented as locally reproduced measurements.

\subsection{Common and Long-tailed Object Detection}
\label{app:common_longtailed_detection}

\paragraph{COCO.}
GroundingPI achieves 62.98 F1mIoU, exceeding Rex-Omni (56.28), LocateAnything Hybrid (59.12), and Astra (62.75). Its 81.16 F1 at IoU 0.50 reflects strong object coverage, but its 22.84 at IoU 0.95 trails several baselines. The aggregate gain therefore should not be read as uniformly superior boundary precision: extremely strict localization remains a limitation even on common objects.

\begin{table}[htbp]
  \centering
  \BenchmarkTableFont
  \caption{\textbf{COCO.} Complete box-grounding metrics.  Model-name stars denote external scores from Table~2 of \citet{rexomni}.}
  \label{tab:app-coco}
  \begingroup
  \fontsize{8}{9.7}\selectfont
  \setlength{\tabcolsep}{3pt}
  \renewcommand{\arraystretch}{1.15}
  \resizebox{\linewidth}{!}{%
  \begin{NiceTabular}{@{}>{\raggedright\arraybackslash}p{177pt}cccccccccc@{}}
  \CodeBefore
    \rowcolor{benchmarktype}{3}
    \rowcolor{benchmarktype}{7}
    \rowcolor{benchmarktype}{9}
    \rowcolor{benchmarkpurple}{26}
    \rowcolor{benchmarktype}{27}
    \rowcolor{benchmarktype}{35}
  \Body
  \toprule
  \textbf{Model} & \BenchHead{Zero-shot} & \multicolumn{3}{c}{\textbf{IoU 0.50}} & \multicolumn{3}{c}{\textbf{IoU 0.95}} & \multicolumn{3}{c}{\textbf{mIoU}} \\
  \cmidrule(lr){3-5}\cmidrule(lr){6-8}\cmidrule(lr){9-11}
   &  & \BenchHead{R} & \BenchHead{P} & \BenchHead{F1} & \BenchHead{R} & \BenchHead{P} & \BenchHead{F1} & \BenchHead{R} & \BenchHead{P} & \BenchHead{F1} \\
  \midrule
  \multicolumn{11}{@{}l}{\hspace{0.4em}\strut\textbf{Closed-set Specialized Detectors}} \\
  DINO-R50\textsuperscript{*}\hspace{0.35em}\citep{DINOR50} & NO & 62.60 & 76.50 & 68.80 & 17.80 & 25.80 & 21.10 & 50.00 & 62.40 & 55.60 \\
  DETR-R50\textsuperscript{*}\hspace{0.35em}\citep{DETRR50} & NO & 59.60 & 73.90 & 65.90 & 10.60 & 19.00 & 13.60 & 42.90 & 55.30 & 48.30 \\
  DyHead-R50\textsuperscript{*}\hspace{0.35em}\citep{DyHeadR50} & NO & 58.10 & 76.60 & 66.10 & 11.90 & 20.60 & 15.00 & 44.80 & 60.10 & 51.30 \\
  \addlinespace[2pt]
  \multicolumn{11}{@{}l}{\hspace{0.4em}\strut\textbf{Open-set Specialized Detectors}} \\
  GroundingDINO\hspace{0.35em}\citep{groundingdino} & YES & 79.60 & 83.23 & \textbf{81.37} & 24.66 & 28.07 & 26.25 & 61.13 & 63.65 & 60.56 \\
  \addlinespace[2pt]
  \multicolumn{11}{@{}l}{\hspace{0.4em}\strut\textbf{Vision-Language Models (<10B)}} \\
  Rex-Omni\hspace{0.35em}\citep{rexomni} & YES & 69.90 & 80.99 & 75.04 & 17.93 & 21.73 & 19.65 & 53.92 & 59.36 & 56.28 \\
  LocateAnything Fast\hspace{0.35em}\citep{locateanything} & YES & 66.15 & 73.34 & 69.56 & 24.73 & 26.90 & 25.77 & 50.89 & 59.26 & 53.06 \\
  LocateAnything Hybrid\hspace{0.35em}\citep{locateanything} & YES & 74.67 & 80.80 & 77.61 & 26.64 & 28.66 & 27.61 & 53.43 & 66.19 & 59.12 \\
  LocateAnything Slow NTP\hspace{0.35em}\citep{locateanything} & YES & 75.02 & 81.87 & 78.30 & 24.97 & 27.19 & 26.03 & 54.70 & 65.09 & 59.37 \\
  Qwen2.5-VL-7B\hspace{0.35em}\citep{qwen25vl} & UNK & 57.18 & 62.87 & 59.89 & 12.16 & 13.23 & 12.67 & 44.67 & 49.05 & 46.27 \\
  Qwen3-VL-2B\hspace{0.35em}\citep{qwen3vl} & UNK & 63.50 & 67.14 & 65.27 & 21.77 & 22.57 & 22.16 & 42.64 & 44.86 & 42.48 \\
  Qwen3-VL-4B\hspace{0.35em}\citep{qwen3vl} & UNK & 66.05 & 70.64 & 68.27 & 23.69 & 24.87 & 24.27 & 45.13 & 48.02 & 46.53 \\
  Qwen3-VL-8B\hspace{0.35em}\citep{qwen3vl} & UNK & 65.70 & 69.48 & 67.54 & 23.91 & 25.11 & 24.49 & 45.09 & 47.58 & 46.30 \\
  Qwen3.5-4B\hspace{0.35em}\citep{qwen35} & UNK & 65.86 & 67.75 & 66.79 & 22.94 & 23.67 & 23.30 & 48.97 & 46.28 & 47.58 \\
  Qwen3.5-9B\hspace{0.35em}\citep{qwen35} & UNK & 73.17 & 68.13 & 70.56 & 23.83 & 24.48 & 24.15 & 58.80 & 46.60 & 51.99 \\
  RynnBrain1.1\hspace{0.35em}\citep{RynnBrain112B} & UNK & 36.49 & 60.51 & 45.53 & 11.56 & 16.53 & 13.60 & 28.72 & 45.31 & 35.15 \\
  SenseNova-Vision\hspace{0.35em}\citep{SenseNovaVision7BMoT} & UNK & 79.07 & 80.54 & 79.80 & \textbf{27.19} & \textbf{30.19} & \textbf{28.61} & 63.13 & 55.37 & 57.49 \\
  MiMo-VL-7B-SFT\hspace{0.35em}\citep{MimoVL} & UNK & 65.48 & 72.18 & 68.67 & 10.02 & 11.11 & 10.54 & 45.72 & 50.47 & 47.98 \\
  MiMo-VL-7B-RL\hspace{0.35em}\citep{MimoVL} & UNK & 65.64 & 74.52 & 69.79 & 9.17 & 10.29 & 9.69 & 44.48 & 50.24 & 47.18 \\
  BAGEL\hspace{0.35em}\citep{BAGEL7BMoT} & UNK & 65.57 & 75.20 & 70.06 & 11.04 & 13.01 & 11.94 & 46.30 & 47.15 & 45.98 \\
  RynnBrain\hspace{0.35em}\citep{dang2026rynnbrain} & UNK & 21.27 & 49.89 & 29.82 & 4.89 & 10.87 & 6.74 & 15.46 & 35.37 & 21.51 \\
  GroundingPI & YES & 81.41 & 80.90 & 81.16 & 22.90 & 22.77 & 22.84 & 63.17 & 62.78 & \textbf{62.98} \\
  \addlinespace[2pt]
  \multicolumn{11}{@{}l}{\hspace{0.4em}\strut\textbf{Vision-Language Models (10B--1T)}} \\
  Qwen3.6-27B\hspace{0.35em}\citep{qwen3627b} & UNK & 78.19 & 79.00 & 78.59 & 25.02 & 25.68 & 25.35 & 61.26 & 62.19 & 61.72 \\
  Qwen3-VL-32B\hspace{0.35em}\citep{qwen3vl} & UNK & 76.47 & 77.79 & 77.12 & 23.05 & 23.84 & 23.44 & 59.50 & 60.84 & 60.16 \\
  Qwen3.8-27B\hspace{0.35em}\citep{qwen38} & UNK & 77.48 & 79.51 & 78.48 & 23.60 & 24.44 & 24.01 & 59.97 & 61.72 & 60.84 \\
  Qwen3.5-35B-A3B\hspace{0.35em}\citep{qwen35} & UNK & 78.37 & 79.65 & 79.00 & 25.20 & 25.92 & 25.56 & 61.45 & 62.71 & 62.07 \\
  DeepSeek-VL2-Small-16B\hspace{0.35em}\citep{Deepseekvl2} & UNK & 20.86 & 45.22 & 28.55 & 0.85 & 1.73 & 1.14 & 10.12 & 21.19 & 13.70 \\
  DeepSeek-VL2-27B\hspace{0.35em}\citep{Deepseekvl2} & UNK & 34.58 & \textbf{84.65} & 49.10 & 12.89 & 28.12 & 17.67 & 28.60 & \textbf{67.97} & 40.25 \\
  SEED1.5-VL\textsuperscript{*}\hspace{0.35em}\citep{SEED15VL} & YES & 65.30 & 78.60 & 71.30 & 12.70 & 16.40 & 14.30 & 46.80 & 56.90 & 51.40 \\
  \addlinespace[2pt]
  \multicolumn{11}{@{}l}{\hspace{0.4em}\strut\textbf{Vision-Language Models (>1T)}} \\
  Qwen3.7-Max\hspace{0.35em}\citep{qwen37} & UNK & 78.98 & 82.62 & 80.76 & 24.19 & 25.15 & 24.66 & 60.97 & 64.72 & 62.79 \\
  Kimi-K2.6 & UNK & 72.27 & 80.10 & 75.99 & 25.04 & 27.19 & 26.07 & 58.28 & 64.34 & 61.16 \\
  Kimi-K3\hspace{0.35em}\citep{KimiK3} & UNK & 72.69 & 84.33 & 78.08 & 23.40 & 26.13 & 24.69 & 56.99 & 65.37 & 60.89 \\
  GPT-6 Astra & UNK & \textbf{82.65} & 77.22 & 79.81 & 26.50 & 25.74 & 26.11 & \textbf{64.61} & 61.03 & 62.75 \\
  \addlinespace[2pt]
  \bottomrule
  
  \end{NiceTabular}%
  }
  \endgroup
\end{table}

\paragraph{LVIS.}
GroundingPI reaches 56.02 F1mIoU and 75.22 F1 at IoU 0.50, improving over Astra's 54.97 and 72.58. SenseNova-Vision remains slightly ahead on F1mIoU (56.12) and more clearly ahead at IoU 0.95 (28.88 versus 22.42). Together with COCO, these results support broad category coverage while identifying high-IoU localization as a separate challenge.

\begin{table}[htbp]
  \centering
  \BenchmarkTableFont
  \caption{\textbf{LVIS.} Complete box-grounding metrics.  The starred SEED1.5-VL scores are from Table~3 of \citet{rexomni}.}
  \label{tab:app-lvis}
  \begingroup
  \fontsize{8}{9.7}\selectfont
  \setlength{\tabcolsep}{3pt}
  \renewcommand{\arraystretch}{1.15}
  \resizebox{\linewidth}{!}{%
  \begin{NiceTabular}{@{}>{\raggedright\arraybackslash}p{177pt}cccccccccc@{}}
  \CodeBefore
    \rowcolor{benchmarktype}{3}
    \rowcolor{benchmarktype}{5}
    \rowcolor{benchmarkpurple}{22}
    \rowcolor{benchmarktype}{23}
    \rowcolor{benchmarktype}{31}
  \Body
  \toprule
  \textbf{Model} & \BenchHead{Zero-shot} & \multicolumn{3}{c}{\textbf{IoU 0.50}} & \multicolumn{3}{c}{\textbf{IoU 0.95}} & \multicolumn{3}{c}{\textbf{mIoU}} \\
  \cmidrule(lr){3-5}\cmidrule(lr){6-8}\cmidrule(lr){9-11}
   &  & \BenchHead{R} & \BenchHead{P} & \BenchHead{F1} & \BenchHead{R} & \BenchHead{P} & \BenchHead{F1} & \BenchHead{R} & \BenchHead{P} & \BenchHead{F1} \\
  \midrule
  \multicolumn{11}{@{}l}{\hspace{0.4em}\strut\textbf{Open-set Specialized Detectors}} \\
  GroundingDINO\hspace{0.35em}\citep{groundingdino} & YES & 52.64 & \textbf{82.61} & 64.30 & 23.34 & 26.55 & 24.84 & 46.06 & \textbf{61.34} & 52.61 \\
  \addlinespace[2pt]
  \multicolumn{11}{@{}l}{\hspace{0.4em}\strut\textbf{Vision-Language Models (<10B)}} \\
  Rex-Omni\hspace{0.35em}\citep{rexomni} & YES & 58.50 & 73.11 & 64.99 & 17.76 & 20.04 & 18.83 & 45.50 & 48.06 & 46.74 \\
  LocateAnything Fast\hspace{0.35em}\citep{locateanything} & YES & 48.06 & 61.58 & 53.98 & 20.37 & 25.12 & 22.50 & 38.43 & 48.55 & 42.90 \\
  LocateAnything Hybrid\hspace{0.35em}\citep{locateanything} & YES & 55.98 & 71.98 & 62.98 & 22.43 & 27.72 & 24.79 & 44.30 & 56.25 & 49.56 \\
  LocateAnything Slow NTP\hspace{0.35em}\citep{locateanything} & YES & 57.98 & 75.03 & 65.41 & 20.96 & 26.03 & 23.22 & 44.95 & 57.46 & 50.44 \\
  Qwen2.5-VL-7B\hspace{0.35em}\citep{qwen25vl} & UNK & 49.52 & 66.52 & 56.78 & 8.16 & 9.87 & 8.93 & 33.37 & 43.60 & 37.80 \\
  Qwen3-VL-2B\hspace{0.35em}\citep{qwen3vl} & UNK & 55.20 & 68.93 & 61.30 & 16.50 & 19.12 & 17.71 & 40.66 & 49.71 & 44.73 \\
  Qwen3-VL-4B\hspace{0.35em}\citep{qwen3vl} & UNK & 59.60 & 76.16 & 66.87 & 18.71 & 22.25 & 20.32 & 44.84 & 56.15 & 49.86 \\
  Qwen3-VL-8B\hspace{0.35em}\citep{qwen3vl} & UNK & 59.47 & 74.39 & 66.10 & 18.00 & 21.19 & 19.46 & 44.25 & 54.49 & 48.84 \\
  Qwen3.5-4B\hspace{0.35em}\citep{qwen35} & UNK & 58.19 & 70.41 & 63.72 & 17.87 & 20.52 & 19.10 & 42.65 & 50.95 & 46.43 \\
  Qwen3.5-9B\hspace{0.35em}\citep{qwen35} & UNK & 60.86 & 72.03 & 65.98 & 19.33 & 22.00 & 20.58 & 45.28 & 53.07 & 48.87 \\
  RynnBrain1.1\hspace{0.35em}\citep{RynnBrain112B} & UNK & 26.86 & 52.27 & 35.48 & 8.76 & 13.56 & 10.64 & 20.18 & 36.64 & 26.01 \\
  SenseNova-Vision\hspace{0.35em}\citep{SenseNovaVision7BMoT} & UNK & 64.50 & 75.97 & 69.76 & \textbf{27.09} & \textbf{30.94} & \textbf{28.88} & 52.08 & 60.84 & \textbf{56.12} \\
  MiMo-VL-7B-SFT\hspace{0.35em}\citep{MimoVL} & UNK & 42.88 & 60.78 & 50.28 & 6.26 & 8.50 & 7.21 & 28.39 & 39.89 & 33.17 \\
  MiMo-VL-7B-RL\hspace{0.35em}\citep{MimoVL} & UNK & 43.50 & 61.87 & 51.09 & 5.55 & 7.52 & 6.38 & 27.53 & 38.74 & 32.18 \\
  BAGEL\hspace{0.35em}\citep{BAGEL7BMoT} & UNK & 34.68 & 51.47 & 41.44 & 8.74 & 11.58 & 9.96 & 24.88 & 35.76 & 29.34 \\
  RynnBrain\hspace{0.35em}\citep{dang2026rynnbrain} & UNK & 15.06 & 43.29 & 22.35 & 3.24 & 8.90 & 4.75 & 10.34 & 28.98 & 15.25 \\
  GroundingPI & YES & \textbf{72.91} & 77.67 & \textbf{75.22} & 22.00 & 22.86 & 22.42 & \textbf{54.49} & 57.63 & 56.02 \\
  \addlinespace[2pt]
  \multicolumn{11}{@{}l}{\hspace{0.4em}\strut\textbf{Vision-Language Models (10B--1T)}} \\
  Qwen3.6-27B\hspace{0.35em}\citep{qwen3627b} & UNK & 62.89 & 75.25 & 68.52 & 19.87 & 22.54 & 21.12 & 47.15 & 55.67 & 51.05 \\
  Qwen3-VL-32B\hspace{0.35em}\citep{qwen3vl} & UNK & 61.21 & 72.44 & 66.36 & 18.01 & 20.44 & 19.15 & 45.41 & 53.20 & 48.99 \\
  Qwen3.8-27B\hspace{0.35em}\citep{qwen38} & UNK & 62.86 & 74.17 & 68.05 & 18.79 & 21.14 & 19.90 & 46.43 & 54.15 & 49.99 \\
  Qwen3.5-35B-A3B\hspace{0.35em}\citep{qwen35} & UNK & 62.87 & 74.82 & 68.33 & 20.04 & 22.75 & 21.31 & 47.14 & 55.41 & 50.94 \\
  DeepSeek-VL2-Small-16B\hspace{0.35em}\citep{Deepseekvl2} & UNK & 12.80 & 28.41 & 17.65 & 0.50 & 1.20 & 0.71 & 6.10 & 13.05 & 8.31 \\
  DeepSeek-VL2-27B\hspace{0.35em}\citep{Deepseekvl2} & UNK & 26.13 & 70.96 & 38.19 & 10.40 & 24.05 & 14.52 & 21.06 & 54.62 & 30.39 \\
  SEED1.5-VL\textsuperscript{*}\hspace{0.35em}\citep{SEED15VL} & YES & 54.70 & 82.00 & 65.60 & 15.00 & 28.10 & 19.50 & 38.50 & 59.30 & 46.70 \\
  \addlinespace[2pt]
  \multicolumn{11}{@{}l}{\hspace{0.4em}\strut\textbf{Vision-Language Models (>1T)}} \\
  Qwen3.7-Max\hspace{0.35em}\citep{qwen37} & UNK & 63.86 & 79.06 & 70.65 & 20.27 & 23.49 & 21.76 & 48.54 & 59.04 & 53.28 \\
  Kimi-K2.6 & UNK & 57.79 & 76.44 & 65.82 & 21.58 & 26.91 & 23.95 & 45.23 & 58.95 & 51.19 \\
  Kimi-K3\hspace{0.35em}\citep{KimiK3} & UNK & 57.17 & 76.66 & 65.50 & 17.46 & 21.54 & 19.29 & 42.08 & 55.14 & 47.73 \\
  GPT-6 Astra & UNK & 71.35 & 73.85 & 72.58 & 23.47 & 24.60 & 24.02 & 54.23 & 55.73 & 54.97 \\
  \addlinespace[2pt]
  \bottomrule
  
  \end{NiceTabular}%
  }
  \endgroup
\end{table}

\subsection{Dense and Tiny Object Detection}
\label{app:dense_tiny_detection}

\paragraph{Dense200.}
GroundingPI attains 74.53 F1mIoU, exceeding SenseNova-Vision by 6.40 pp and Astra by 9.49 pp. At IoU 0.50, recall and precision are both high (89.88 and 92.80), indicating that the result balances finding instances and avoiding excess predictions. Its 27.85 F1 at IoU 0.95 also exceeds Astra's 12.64. Unlike the common-object comparison, the dense-scene improvement extends to strict localization.

\begin{table}[htbp]
  \centering
  \BenchmarkTableFont
  \caption{\textbf{Dense200.} Complete box-grounding metrics. Local BAGEL and DeepSeek runs did not reproduce the reference results, so only available external scores are reported for these models. The starred BAGEL scores are taken from Table~1 of \citet{SenseNovaVision7BMoT}. No matching external result is available for full DeepSeek-VL2; Small and Tiny checkpoint results are not substituted. The starred DeepSeek-VL2-Small and SEED1.5-VL scores are from Table~4 of \citet{rexomni}, also reported in Table~2 of \citet{locateanything}.}
  \label{tab:app-dense200}
  \begingroup
  \fontsize{8}{9.7}\selectfont
  \setlength{\tabcolsep}{3pt}
  \renewcommand{\arraystretch}{1.15}
  \resizebox{\linewidth}{!}{%
  \begin{NiceTabular}{@{}>{\raggedright\arraybackslash}p{177pt}ccccccccc@{}}
  \CodeBefore
    \rowcolor{benchmarktype}{3}
    \rowcolor{benchmarktype}{5}
    \rowcolor{benchmarkpurple}{22}
    \rowcolor{benchmarktype}{23}
    \rowcolor{benchmarktype}{31}
  \Body
  \toprule
  \textbf{Model} & \multicolumn{3}{c}{\textbf{IoU 0.50}} & \multicolumn{3}{c}{\textbf{IoU 0.95}} & \multicolumn{3}{c}{\textbf{mIoU}} \\
  \cmidrule(lr){2-4}\cmidrule(lr){5-7}\cmidrule(lr){8-10}
   & \BenchHead{R} & \BenchHead{P} & \BenchHead{F1} & \BenchHead{R} & \BenchHead{P} & \BenchHead{F1} & \BenchHead{R} & \BenchHead{P} & \BenchHead{F1} \\
  \midrule
  \multicolumn{10}{@{}l}{\hspace{0.4em}\strut\textbf{Open-set Specialized Detectors}} \\
  GroundingDINO\hspace{0.35em}\citep{groundingdino} & 22.43 & 36.30 & 27.73 & 10.92 & 18.38 & 13.70 & 20.16 & 32.62 & 24.92 \\
  \addlinespace[2pt]
  \multicolumn{10}{@{}l}{\hspace{0.4em}\strut\textbf{Vision-Language Models (<10B)}} \\
  Rex-Omni\hspace{0.35em}\citep{rexomni} & 70.61 & 75.13 & 72.80 & 8.66 & 9.17 & 8.91 & 51.79 & 54.88 & 53.29 \\
  LocateAnything Fast\hspace{0.35em}\citep{locateanything} & 22.45 & 26.12 & 24.15 & 9.16 & 10.50 & 9.78 & 19.30 & 22.31 & 20.70 \\
  LocateAnything Hybrid\hspace{0.35em}\citep{locateanything} & 58.06 & 63.32 & 60.57 & 21.52 & 22.90 & 22.19 & 48.13 & 52.18 & 50.07 \\
  LocateAnything Slow NTP\hspace{0.35em}\citep{locateanything} & 74.89 & 81.46 & 78.03 & 20.64 & 22.11 & 21.35 & 59.65 & 64.62 & 62.04 \\
  Qwen2.5-VL-7B\hspace{0.35em}\citep{qwen25vl} & 0.50 & 0.57 & 0.53 & 0.00 & 0.00 & 0.00 & 0.22 & 0.24 & 0.23 \\
  Qwen3-VL-2B\hspace{0.35em}\citep{qwen3vl} & 9.93 & 13.28 & 11.36 & 1.27 & 1.93 & 1.53 & 7.18 & 9.65 & 8.23 \\
  Qwen3-VL-4B\hspace{0.35em}\citep{qwen3vl} & 17.61 & 22.94 & 19.92 & 2.68 & 3.76 & 3.13 & 12.33 & 16.23 & 14.02 \\
  Qwen3-VL-8B\hspace{0.35em}\citep{qwen3vl} & 15.42 & 16.40 & 15.90 & 2.26 & 2.58 & 2.41 & 11.13 & 11.77 & 11.44 \\
  Qwen3.5-4B\hspace{0.35em}\citep{qwen35} & 43.44 & 48.47 & 45.82 & 8.20 & 9.01 & 8.58 & 32.20 & 36.12 & 34.04 \\
  Qwen3.5-9B\hspace{0.35em}\citep{qwen35} & 39.02 & 42.00 & 40.46 & 7.37 & 8.19 & 7.76 & 28.82 & 31.33 & 30.02 \\
  RynnBrain1.1\hspace{0.35em}\citep{RynnBrain112B} & 0.10 & 4.00 & 0.19 & 0.02 & 0.50 & 0.04 & 0.06 & 2.40 & 0.12 \\
  SenseNova-Vision\hspace{0.35em}\citep{SenseNovaVision7BMoT} & 79.18 & 86.84 & 82.83 & 24.52 & 25.72 & 25.10 & 65.31 & 71.20 & 68.13 \\
  MiMo-VL-7B-SFT\hspace{0.35em}\citep{MimoVL} & 11.75 & 12.38 & 12.06 & 0.14 & 0.14 & 0.14 & 5.62 & 5.95 & 5.78 \\
  MiMo-VL-7B-RL\hspace{0.35em}\citep{MimoVL} & 13.10 & 14.14 & 13.60 & 0.08 & 0.07 & 0.07 & 6.31 & 6.88 & 6.58 \\
  BAGEL\textsuperscript{*}\hspace{0.35em}\citep{BAGEL7BMoT} & -- & -- & -- & -- & -- & -- & -- & -- & 42.40 \\
  RynnBrain\hspace{0.35em}\citep{dang2026rynnbrain} & 0.03 & 1.50 & 0.06 & 0.00 & 0.00 & 0.00 & 0.02 & 0.80 & 0.03 \\
  GroundingPI & \textbf{89.88} & \textbf{92.80} & \textbf{91.31} & \textbf{27.52} & \textbf{28.19} & \textbf{27.85} & \textbf{73.45} & \textbf{75.63} & \textbf{74.53} \\
  \addlinespace[2pt]
  \multicolumn{10}{@{}l}{\hspace{0.4em}\strut\textbf{Vision-Language Models (10B--1T)}} \\
  Qwen3.6-27B\hspace{0.35em}\citep{qwen3627b} & 51.74 & 56.09 & 53.83 & 9.91 & 10.76 & 10.32 & 39.12 & 42.56 & 40.77 \\
  Qwen3-VL-32B\hspace{0.35em}\citep{qwen3vl} & 24.97 & 28.35 & 26.56 & 2.89 & 4.04 & 3.37 & 17.78 & 20.69 & 19.12 \\
  Qwen3.8-27B\hspace{0.35em}\citep{qwen38} & 44.31 & 47.44 & 45.82 & 8.92 & 9.56 & 9.23 & 33.28 & 35.40 & 34.30 \\
  Qwen3.5-35B-A3B\hspace{0.35em}\citep{qwen35} & 46.43 & 49.58 & 47.95 & 9.22 & 10.00 & 9.60 & 35.18 & 37.36 & 36.24 \\
  DeepSeek-VL2-Small-16B\textsuperscript{*}\hspace{0.35em}\citep{Deepseekvl2} & -- & -- & 16.00 & -- & -- & 3.90 & -- & -- & 12.70 \\
  DeepSeek-VL2-27B\textsuperscript{*}\hspace{0.35em}\citep{Deepseekvl2} & -- & -- & -- & -- & -- & -- & -- & -- & -- \\
  SEED1.5-VL\textsuperscript{*}\hspace{0.35em}\citep{SEED15VL} & -- & -- & 76.90 & -- & -- & 5.30 & -- & -- & 53.20 \\
  \addlinespace[2pt]
  \multicolumn{10}{@{}l}{\hspace{0.4em}\strut\textbf{Vision-Language Models (>1T)}} \\
  Qwen3.7-Max\hspace{0.35em}\citep{qwen37} & 37.56 & 43.69 & 40.40 & 7.20 & 8.65 & 7.86 & 28.94 & 33.84 & 31.20 \\
  Kimi-K2.6 & 55.64 & 59.00 & 57.27 & 9.21 & 9.91 & 9.55 & 41.38 & 44.04 & 42.67 \\
  Kimi-K3\hspace{0.35em}\citep{KimiK3} & 65.95 & 78.99 & 71.88 & 10.40 & 11.88 & 11.09 & 47.61 & 56.40 & 51.64 \\
  GPT-6 Astra & 89.01 & 84.63 & 86.76 & 12.90 & 12.39 & 12.64 & 66.54 & 63.61 & 65.04 \\
  \addlinespace[2pt]
  \bottomrule
  
  \end{NiceTabular}%
  }
  \endgroup
\end{table}

\paragraph{VisDrone.}
GroundingPI's 40.44 F1mIoU exceeds Astra (37.12), Rex-Omni (27.19), and LocateAnything Hybrid (28.57), but trails SenseNova-Vision (42.35) and Qwen3.7-Max (41.59). Its 3.31 F1 at IoU 0.95 remains low, as do the other results. Small absolute coordinate errors can substantially change tiny-box overlap; improved dense grounding has therefore not eliminated the precision bottleneck for tiny objects.

\begin{table}[htbp]
  \centering
  \BenchmarkTableFont
  \caption{\textbf{VisDrone.} Complete box-grounding metrics. Local BAGEL and DeepSeek runs did not reproduce the reference results, so only available external scores are reported for these models. The starred BAGEL scores are taken from Table~1 of \citet{SenseNovaVision7BMoT}. No matching external result is available for full DeepSeek-VL2; Small and Tiny checkpoint results are not substituted. The starred DeepSeek-VL2-Small and SEED1.5-VL scores are from Table~4 of \citet{rexomni}, also reported in Table~2 of \citet{locateanything}.}
  \label{tab:app-visdrone}
  \begingroup
  \fontsize{8}{9.7}\selectfont
  \setlength{\tabcolsep}{3pt}
  \renewcommand{\arraystretch}{1.15}
  \resizebox{\linewidth}{!}{%
  \begin{NiceTabular}{@{}>{\raggedright\arraybackslash}p{177pt}ccccccccc@{}}
  \CodeBefore
    \rowcolor{benchmarktype}{3}
    \rowcolor{benchmarktype}{5}
    \rowcolor{benchmarkpurple}{22}
    \rowcolor{benchmarktype}{23}
    \rowcolor{benchmarktype}{31}
  \Body
  \toprule
  \textbf{Model} & \multicolumn{3}{c}{\textbf{IoU 0.50}} & \multicolumn{3}{c}{\textbf{IoU 0.95}} & \multicolumn{3}{c}{\textbf{mIoU}} \\
  \cmidrule(lr){2-4}\cmidrule(lr){5-7}\cmidrule(lr){8-10}
   & \BenchHead{R} & \BenchHead{P} & \BenchHead{F1} & \BenchHead{R} & \BenchHead{P} & \BenchHead{F1} & \BenchHead{R} & \BenchHead{P} & \BenchHead{F1} \\
  \midrule
  \multicolumn{10}{@{}l}{\hspace{0.4em}\strut\textbf{Open-set Specialized Detectors}} \\
  GroundingDINO\hspace{0.35em}\citep{groundingdino} & 32.38 & \textbf{87.09} & 47.21 & 2.81 & \textbf{7.42} & \textbf{4.08} & 23.65 & \textbf{63.54} & 34.47 \\
  \addlinespace[2pt]
  \multicolumn{10}{@{}l}{\hspace{0.4em}\strut\textbf{Vision-Language Models (<10B)}} \\
  Rex-Omni\hspace{0.35em}\citep{rexomni} & 42.47 & 52.33 & 46.89 & 1.07 & 1.27 & 1.16 & 24.78 & 30.12 & 27.19 \\
  LocateAnything Fast\hspace{0.35em}\citep{locateanything} & 13.58 & 15.44 & 14.45 & 0.85 & 0.97 & 0.90 & 9.27 & 10.44 & 9.82 \\
  LocateAnything Hybrid\hspace{0.35em}\citep{locateanything} & 40.66 & 46.99 & 43.59 & 2.34 & 2.64 & 2.48 & 26.78 & 30.62 & 28.57 \\
  LocateAnything Slow NTP\hspace{0.35em}\citep{locateanything} & 59.28 & 70.06 & 64.22 & 2.73 & 3.13 & 2.91 & 37.32 & 43.81 & 40.31 \\
  Qwen2.5-VL-7B\hspace{0.35em}\citep{qwen25vl} & 27.63 & 42.83 & 33.59 & 0.44 & 0.63 & 0.52 & 15.48 & 23.55 & 18.67 \\
  Qwen3-VL-2B\hspace{0.35em}\citep{qwen3vl} & 37.16 & 46.15 & 41.17 & 1.16 & 1.46 & 1.29 & 23.01 & 28.23 & 25.36 \\
  Qwen3-VL-4B\hspace{0.35em}\citep{qwen3vl} & 47.85 & 52.36 & 50.00 & 1.78 & 1.86 & 1.82 & 29.92 & 32.47 & 31.14 \\
  Qwen3-VL-8B\hspace{0.35em}\citep{qwen3vl} & 46.61 & 44.84 & 45.71 & 1.74 & 1.70 & 1.72 & 28.80 & 27.81 & 28.29 \\
  Qwen3.5-4B\hspace{0.35em}\citep{qwen35} & 54.93 & 50.78 & 52.77 & 2.60 & 2.53 & 2.57 & 34.91 & 32.71 & 33.77 \\
  Qwen3.5-9B\hspace{0.35em}\citep{qwen35} & 57.52 & 45.04 & 50.52 & 2.87 & 2.48 & 2.66 & 36.75 & 29.69 & 32.84 \\
  RynnBrain1.1\hspace{0.35em}\citep{RynnBrain112B} & 7.42 & 46.41 & 12.80 & 0.38 & 2.05 & 0.64 & 4.83 & 29.23 & 8.29 \\
  SenseNova-Vision\hspace{0.35em}\citep{SenseNovaVision7BMoT} & \textbf{65.10} & 69.99 & \textbf{67.45} & 2.86 & 3.07 & 2.96 & \textbf{40.91} & 43.88 & \textbf{42.35} \\
  MiMo-VL-7B-SFT\hspace{0.35em}\citep{MimoVL} & 26.42 & 28.73 & 27.53 & 0.14 & 0.15 & 0.15 & 12.60 & 13.88 & 13.21 \\
  MiMo-VL-7B-RL\hspace{0.35em}\citep{MimoVL} & 27.94 & 32.86 & 30.20 & 0.28 & 0.34 & 0.31 & 13.79 & 16.62 & 15.07 \\
  BAGEL\textsuperscript{*}\hspace{0.35em}\citep{BAGEL7BMoT} & -- & -- & -- & -- & -- & -- & -- & -- & 23.00 \\
  RynnBrain\hspace{0.35em}\citep{dang2026rynnbrain} & 8.78 & 32.85 & 13.86 & 0.20 & 0.75 & 0.32 & 4.97 & 18.52 & 7.84 \\
  GroundingPI & 57.57 & 68.12 & 62.41 & \textbf{3.08} & 3.59 & 3.31 & 37.46 & 43.95 & 40.44 \\
  \addlinespace[2pt]
  \multicolumn{10}{@{}l}{\hspace{0.4em}\strut\textbf{Vision-Language Models (10B--1T)}} \\
  Qwen3.6-27B\hspace{0.35em}\citep{qwen3627b} & 57.36 & 56.32 & 56.83 & 2.61 & 2.73 & 2.67 & 36.44 & 36.43 & 36.44 \\
  Qwen3-VL-32B\hspace{0.35em}\citep{qwen3vl} & 48.94 & 46.14 & 47.50 & 1.48 & 1.46 & 1.47 & 29.51 & 28.24 & 28.86 \\
  Qwen3.8-27B\hspace{0.35em}\citep{qwen38} & 56.44 & 49.01 & 52.46 & 2.76 & 2.62 & 2.69 & 35.48 & 31.42 & 33.33 \\
  Qwen3.5-35B-A3B\hspace{0.35em}\citep{qwen35} & 59.78 & 50.81 & 54.93 & 2.92 & 2.76 & 2.84 & 37.97 & 32.86 & 35.23 \\
  DeepSeek-VL2-Small-16B\textsuperscript{*}\hspace{0.35em}\citep{Deepseekvl2} & -- & -- & 35.80 & -- & -- & 1.70 & -- & -- & 23.30 \\
  DeepSeek-VL2-27B\textsuperscript{*}\hspace{0.35em}\citep{Deepseekvl2} & -- & -- & -- & -- & -- & -- & -- & -- & -- \\
  SEED1.5-VL\textsuperscript{*}\hspace{0.35em}\citep{SEED15VL} & -- & -- & 55.90 & -- & -- & 0.60 & -- & -- & 27.40 \\
  \addlinespace[2pt]
  \multicolumn{10}{@{}l}{\hspace{0.4em}\strut\textbf{Vision-Language Models (>1T)}} \\
  Qwen3.7-Max\hspace{0.35em}\citep{qwen37} & 60.27 & 71.40 & 65.36 & 2.84 & 3.45 & 3.11 & 38.28 & 45.52 & 41.59 \\
  Kimi-K2.6 & 41.36 & 65.79 & 50.79 & 1.20 & 1.73 & 1.42 & 24.42 & 38.73 & 29.95 \\
  Kimi-K3\hspace{0.35em}\citep{KimiK3} & 41.51 & 76.49 & 53.81 & 0.98 & 1.59 & 1.21 & 23.57 & 42.47 & 30.31 \\
  GPT-6 Astra & 64.60 & 65.83 & 65.18 & 2.69 & 2.80 & 2.74 & 36.70 & 37.58 & 37.12 \\
  \addlinespace[2pt]
  \bottomrule
  
  \end{NiceTabular}%
  }
  \endgroup
\end{table}

\subsection{Referring Object Detection}
\label{app:referring_detection}

\paragraph{HumanRef.}
GroundingPI reaches 88.56 F1mIoU and 76.47 F1 at IoU 0.95, compared with Astra's 83.01 and 71.19. Recall and precision at IoU 0.50 are closely balanced (93.10 and 93.93). The gain thus includes accurate localization of the referred target, rather than only a change in the number of returned predictions.

\begin{table}[htbp]
  \centering
  \BenchmarkTableFont
  \caption{\textbf{HumanRef.} Complete box-grounding metrics. The starred BAGEL scores are taken from Table~1 of \citet{SenseNovaVision7BMoT}, because local runs did not reproduce the reported performance. The starred SEED1.5-VL scores are from Table~5 of \citet{rexomni}.}
  \label{tab:app-humanref}
  \begingroup
  \fontsize{8}{9.7}\selectfont
  \setlength{\tabcolsep}{3pt}
  \renewcommand{\arraystretch}{1.15}
  \resizebox{\linewidth}{!}{%
  \begin{NiceTabular}{@{}>{\raggedright\arraybackslash}p{177pt}ccccccccc@{}}
  \CodeBefore
    \rowcolor{benchmarktype}{3}
    \rowcolor{benchmarktype}{5}
    \rowcolor{benchmarkpurple}{22}
    \rowcolor{benchmarktype}{23}
    \rowcolor{benchmarktype}{31}
  \Body
  \toprule
  \textbf{Model} & \multicolumn{3}{c}{\textbf{IoU 0.50}} & \multicolumn{3}{c}{\textbf{IoU 0.95}} & \multicolumn{3}{c}{\textbf{mIoU}} \\
  \cmidrule(lr){2-4}\cmidrule(lr){5-7}\cmidrule(lr){8-10}
   & \BenchHead{R} & \BenchHead{P} & \BenchHead{F1} & \BenchHead{R} & \BenchHead{P} & \BenchHead{F1} & \BenchHead{R} & \BenchHead{P} & \BenchHead{F1} \\
  \midrule
  \multicolumn{10}{@{}l}{\hspace{0.4em}\strut\textbf{Open-set Specialized Detectors}} \\
  GroundingDINO\hspace{0.35em}\citep{groundingdino} & 54.80 & 46.81 & 50.49 & 37.38 & 31.26 & 34.05 & 50.32 & 42.59 & 46.13 \\
  \addlinespace[2pt]
  \multicolumn{10}{@{}l}{\hspace{0.4em}\strut\textbf{Vision-Language Models (<10B)}} \\
  Rex-Omni\hspace{0.35em}\citep{rexomni} & 85.91 & 84.95 & 85.43 & 65.72 & 65.09 & 65.40 & 80.33 & 79.42 & 79.87 \\
  LocateAnything Fast\hspace{0.35em}\citep{locateanything} & 68.01 & 71.34 & 69.64 & 53.25 & 55.15 & 54.19 & 62.15 & 64.72 & 63.41 \\
  LocateAnything Hybrid\hspace{0.35em}\citep{locateanything} & 83.01 & 83.07 & 83.04 & 68.61 & 68.63 & 68.62 & 78.51 & 78.53 & 78.52 \\
  LocateAnything Slow NTP\hspace{0.35em}\citep{locateanything} & 82.74 & 84.01 & 83.37 & 68.21 & 69.43 & 68.82 & 78.31 & 79.55 & 78.93 \\
  Qwen2.5-VL-7B\hspace{0.35em}\citep{qwen25vl} & 46.52 & 53.67 & 49.84 & 19.66 & 22.18 & 20.84 & 39.16 & 44.94 & 41.85 \\
  Qwen3-VL-2B\hspace{0.35em}\citep{qwen3vl} & 61.41 & 81.93 & 70.20 & 42.96 & 56.06 & 48.64 & 55.87 & 74.16 & 63.73 \\
  Qwen3-VL-4B\hspace{0.35em}\citep{qwen3vl} & 68.32 & 86.41 & 76.31 & 49.90 & 62.00 & 55.30 & 62.85 & 79.13 & 70.06 \\
  Qwen3-VL-8B\hspace{0.35em}\citep{qwen3vl} & 69.44 & 85.61 & 76.68 & 51.08 & 61.73 & 55.90 & 63.86 & 78.21 & 70.31 \\
  Qwen3.5-4B\hspace{0.35em}\citep{qwen35} & 68.23 & 88.91 & 77.21 & 52.68 & 67.07 & 59.01 & 63.26 & 81.84 & 71.36 \\
  Qwen3.5-9B\hspace{0.35em}\citep{qwen35} & 69.20 & 90.43 & 78.40 & 55.32 & 71.92 & 62.54 & 64.89 & 84.78 & 73.51 \\
  RynnBrain1.1\hspace{0.35em}\citep{RynnBrain112B} & 53.79 & 72.00 & 61.58 & 31.15 & 38.70 & 34.52 & 46.09 & 59.84 & 52.07 \\
  SenseNova-Vision\hspace{0.35em}\citep{SenseNovaVision7BMoT} & 86.59 & 81.80 & 84.12 & 69.92 & 66.09 & 67.95 & 81.64 & 77.17 & 79.34 \\
  MiMo-VL-7B-SFT\hspace{0.35em}\citep{MimoVL} & 72.62 & 79.17 & 75.76 & 26.25 & 28.45 & 27.31 & 59.07 & 64.10 & 61.48 \\
  MiMo-VL-7B-RL\hspace{0.35em}\citep{MimoVL} & 67.28 & 79.57 & 72.91 & 21.95 & 25.90 & 23.76 & 54.40 & 63.99 & 58.81 \\
  BAGEL\textsuperscript{*}\hspace{0.35em}\citep{BAGEL7BMoT} & -- & -- & -- & -- & -- & -- & -- & -- & 74.60 \\
  RynnBrain\hspace{0.35em}\citep{dang2026rynnbrain} & 46.21 & 59.43 & 51.99 & 23.77 & 28.00 & 25.71 & 38.46 & 47.76 & 42.60 \\
  GroundingPI & \textbf{93.10} & \textbf{93.93} & \textbf{93.51} & \textbf{76.07} & \textbf{76.87} & \textbf{76.47} & \textbf{88.17} & \textbf{88.97} & \textbf{88.56} \\
  \addlinespace[2pt]
  \multicolumn{10}{@{}l}{\hspace{0.4em}\strut\textbf{Vision-Language Models (10B--1T)}} \\
  Qwen3.6-27B\hspace{0.35em}\citep{qwen3627b} & 84.36 & 90.96 & 87.54 & 65.50 & 70.02 & 67.69 & 78.65 & 84.53 & 81.48 \\
  Qwen3-VL-32B\hspace{0.35em}\citep{qwen3vl} & 64.38 & 76.91 & 70.09 & 47.88 & 55.88 & 51.57 & 59.48 & 70.26 & 64.42 \\
  Qwen3.8-27B\hspace{0.35em}\citep{qwen38} & 80.35 & 87.49 & 83.77 & 60.01 & 64.83 & 62.33 & 74.22 & 80.57 & 77.26 \\
  Qwen3.5-35B-A3B\hspace{0.35em}\citep{qwen35} & 79.42 & 90.26 & 84.50 & 63.08 & 71.18 & 66.89 & 74.43 & 84.42 & 79.11 \\
  DeepSeek-VL2-Small-16B\hspace{0.35em}\citep{Deepseekvl2} & 39.95 & 50.27 & 44.52 & 1.10 & 1.08 & 1.09 & 20.34 & 24.49 & 22.21 \\
  DeepSeek-VL2-27B\hspace{0.35em}\citep{Deepseekvl2} & 61.08 & 75.73 & 67.62 & 32.32 & 38.40 & 35.10 & 52.27 & 64.15 & 57.60 \\
  SEED1.5-VL\textsuperscript{*}\hspace{0.35em}\citep{SEED15VL} & -- & -- & 88.20 & -- & -- & 60.00 & -- & -- & 81.60 \\
  \addlinespace[2pt]
  \multicolumn{10}{@{}l}{\hspace{0.4em}\strut\textbf{Vision-Language Models (>1T)}} \\
  Qwen3.7-Max\hspace{0.35em}\citep{qwen37} & 65.55 & 85.23 & 74.10 & 43.58 & 55.18 & 48.70 & 59.43 & 76.46 & 66.88 \\
  Kimi-K2.6 & 81.86 & 82.07 & 81.97 & 56.48 & 56.41 & 56.44 & 73.75 & 73.77 & 73.76 \\
  Kimi-K3\hspace{0.35em}\citep{KimiK3} & 87.30 & 89.14 & 88.21 & 61.10 & 61.48 & 61.29 & 79.38 & 80.40 & 79.89 \\
  GPT-6 Astra & 90.75 & 88.58 & 89.64 & 71.80 & 70.60 & 71.19 & 83.99 & 82.08 & 83.01 \\
  \addlinespace[2pt]
  \bottomrule
  
  \end{NiceTabular}%
  }
  \endgroup
\end{table}

\paragraph{RefCOCOg validation and test.}
GroundingPI obtains 85.62/84.36 F1mIoU on validation/test, versus 74.98/78.91 for Astra and 76.43/77.67 for LocateAnything Hybrid. Its gains over these baselines also hold at IoU 0.95. DeepSeek-VL2-27B is stronger at that strict threshold and nearly matches the test aggregate (84.07), showing that aggregate and strict-boundary rankings need not coincide.

\begin{table}[htbp]
  \centering
  \BenchmarkTableFont
  \caption{\textbf{RefCOCOg validation and test.} Complete recorded F1 metrics at IoU 0.50, 0.95, and mIoU. The starred BAGEL scores are taken from Table~1 of \citet{SenseNovaVision7BMoT}, because local runs did not reproduce the reported performance. The starred SEED1.5-VL scores are from Table~5 of \citet{rexomni}.}
  \label{tab:app-refcocog-splits}
  \begingroup
  \fontsize{8}{9.7}\selectfont
  \setlength{\tabcolsep}{3pt}
  \renewcommand{\arraystretch}{1.15}
  \resizebox{\linewidth}{!}{%
  \begin{NiceTabular}{@{}>{\raggedright\arraybackslash}p{177pt}cccccc@{}}
  \CodeBefore
    \rowcolor{benchmarktype}{3}
    \rowcolor{benchmarktype}{5}
    \rowcolor{benchmarkpurple}{22}
    \rowcolor{benchmarktype}{23}
    \rowcolor{benchmarktype}{31}
  \Body
  \toprule
  \textbf{Model} & \multicolumn{3}{c}{\textbf{RefCOCOg val}} & \multicolumn{3}{c}{\textbf{RefCOCOg test}} \\
  \cmidrule(lr){2-4}\cmidrule(lr){5-7}
   & \BenchHead{F1@.50} & \BenchHead{F1@.95} & \BenchHead{F1mIoU} & \BenchHead{F1@.50} & \BenchHead{F1@.95} & \BenchHead{F1mIoU} \\
  \midrule
  \multicolumn{7}{@{}l}{\hspace{0.4em}\strut\textbf{Open-set Specialized Detectors}} \\
  GroundingDINO\hspace{0.35em}\citep{groundingdino} & 58.37 & 22.47 & 49.77 & 58.52 & 24.38 & 50.43 \\
  \addlinespace[2pt]
  \multicolumn{7}{@{}l}{\hspace{0.4em}\strut\textbf{Vision-Language Models (<10B)}} \\
  Rex-Omni\hspace{0.35em}\citep{rexomni} & 87.01 & 35.23 & 73.90 & 87.36 & 36.51 & 74.76 \\
  LocateAnything Fast\hspace{0.35em}\citep{locateanything} & 87.99 & 39.39 & 75.30 & 88.34 & 42.08 & 76.50 \\
  LocateAnything Hybrid\hspace{0.35em}\citep{locateanything} & 88.50 & 40.40 & 76.43 & 88.91 & 42.63 & 77.67 \\
  LocateAnything Slow NTP\hspace{0.35em}\citep{locateanything} & 88.18 & 34.81 & 74.93 & 88.64 & 36.99 & 76.46 \\
  Qwen2.5-VL-7B\hspace{0.35em}\citep{qwen25vl} & 78.19 & 15.10 & 61.86 & 78.49 & 16.31 & 62.93 \\
  Qwen3-VL-2B\hspace{0.35em}\citep{qwen3vl} & 85.25 & 32.13 & 71.80 & 85.87 & 33.88 & 72.64 \\
  Qwen3-VL-4B\hspace{0.35em}\citep{qwen3vl} & 88.53 & 36.61 & 75.27 & 88.69 & 36.61 & 75.88 \\
  Qwen3-VL-8B\hspace{0.35em}\citep{qwen3vl} & 88.92 & 35.61 & 75.79 & 89.26 & 37.03 & 76.26 \\
  Qwen3.5-4B\hspace{0.35em}\citep{qwen35} & 89.24 & 34.89 & 75.03 & 89.00 & 35.14 & 75.54 \\
  Qwen3.5-9B\hspace{0.35em}\citep{qwen35} & 89.89 & 36.63 & 76.20 & 89.42 & 36.89 & 76.28 \\
  RynnBrain1.1\hspace{0.35em}\citep{RynnBrain112B} & 83.21 & 24.36 & 67.66 & 83.75 & 25.14 & 68.24 \\
  SenseNova-Vision\hspace{0.35em}\citep{SenseNovaVision7BMoT} & 89.94 & 43.58 & 78.69 & 89.85 & 44.91 & 79.48 \\
  MiMo-VL-7B-SFT\hspace{0.35em}\citep{MimoVL} & 86.51 & 14.72 & 66.44 & 86.59 & 15.26 & 66.78 \\
  MiMo-VL-7B-RL\hspace{0.35em}\citep{MimoVL} & 88.28 & 12.56 & 64.89 & 87.69 & 12.92 & 64.19 \\
  BAGEL\textsuperscript{*}\hspace{0.35em}\citep{BAGEL7BMoT} & -- & -- & 76.40 & -- & -- & 77.80 \\
  RynnBrain\hspace{0.35em}\citep{dang2026rynnbrain} & 73.72 & 17.32 & 57.99 & 73.77 & 18.29 & 58.38 \\
  GroundingPI & \textbf{94.99} & 54.04 & \textbf{85.62} & \textbf{94.51} & 49.07 & \textbf{84.36} \\
  \addlinespace[2pt]
  \multicolumn{7}{@{}l}{\hspace{0.4em}\strut\textbf{Vision-Language Models (10B--1T)}} \\
  Qwen3.6-27B\hspace{0.35em}\citep{qwen3627b} & 91.66 & 38.19 & 77.76 & 91.36 & 38.80 & 78.43 \\
  Qwen3-VL-32B\hspace{0.35em}\citep{qwen3vl} & 86.27 & 34.17 & 73.79 & 86.03 & 34.83 & 74.02 \\
  Qwen3.8-27B\hspace{0.35em}\citep{qwen38} & 90.14 & 35.69 & 76.03 & 90.86 & 36.98 & 77.37 \\
  Qwen3.5-35B-A3B\hspace{0.35em}\citep{qwen35} & 90.92 & 38.46 & 77.32 & 90.88 & 39.09 & 78.38 \\
  DeepSeek-VL2-Small-16B\hspace{0.35em}\citep{Deepseekvl2} & 59.86 & 0.90 & 26.18 & 61.06 & 0.89 & 27.49 \\
  DeepSeek-VL2-27B\hspace{0.35em}\citep{Deepseekvl2} & 92.46 & \textbf{55.36} & 82.73 & 92.19 & \textbf{58.33} & 84.07 \\
  SEED1.5-VL\textsuperscript{*}\hspace{0.35em}\citep{SEED15VL} & 84.70 & 30.90 & 71.90 & 85.20 & 32.10 & 73.20 \\
  \addlinespace[2pt]
  \multicolumn{7}{@{}l}{\hspace{0.4em}\strut\textbf{Vision-Language Models (>1T)}} \\
  Qwen3.7-Max\hspace{0.35em}\citep{qwen37} & 91.56 & 48.45 & 80.54 & 92.36 & 48.75 & 81.71 \\
  Kimi-K2.6 & 80.15 & 41.17 & 70.33 & 81.19 & 42.36 & 71.87 \\
  Kimi-K3\hspace{0.35em}\citep{KimiK3} & 88.02 & 34.39 & 73.39 & 87.92 & 35.06 & 73.99 \\
  GPT-6 Astra & 92.13 & 35.89 & 74.98 & 89.93 & 40.25 & 78.91 \\
  \addlinespace[2pt]
  \bottomrule
  
  \end{NiceTabular}%
  }
  \endgroup
\end{table}

\paragraph{RefCOCO family.}
GroundingPI obtains 85.79, 81.96, and 84.01 F1mIoU on RefCOCO, RefCOCOg, and RefCOCO+, respectively, giving the reported family mean of 83.92. The especially large advantage over Qwen3-VL-4B on RefCOCO+ (84.01 versus 73.36) supports discrimination from descriptive language. DeepSeek-VL2-27B leads these three complete-table aggregates, so GroundingPI's main-table advantage does not imply a universal referring-grounding lead.

\begin{table}[htbp]
  \centering
  \BenchmarkTableFont
  \caption{\textbf{RefCOCO family.} The three datasets are reported separately; their F1mIoU arithmetic mean is RefCOCO avg in the main text.}
  \label{tab:app-refcoco-family}
  \begingroup
  \fontsize{8}{9.7}\selectfont
  \setlength{\tabcolsep}{3pt}
  \renewcommand{\arraystretch}{1.15}
  \resizebox{\linewidth}{!}{%
  \begin{NiceTabular}{@{}>{\raggedright\arraybackslash}p{177pt}ccccccccc@{}}
  \CodeBefore
    \rowcolor{benchmarktype}{3}
    \rowcolor{benchmarktype}{5}
    \rowcolor{benchmarkpurple}{22}
    \rowcolor{benchmarktype}{23}
    \rowcolor{benchmarktype}{30}
  \Body
  \toprule
  \textbf{Model} & \multicolumn{3}{c}{\textbf{RefCOCO}} & \multicolumn{3}{c}{\textbf{RefCOCOg}} & \multicolumn{3}{c}{\textbf{RefCOCOplus}} \\
  \cmidrule(lr){2-4}\cmidrule(lr){5-7}\cmidrule(lr){8-10}
   & \BenchHead{F1@.50} & \BenchHead{F1@.95} & \BenchHead{F1mIoU} & \BenchHead{F1@.50} & \BenchHead{F1@.95} & \BenchHead{F1mIoU} & \BenchHead{F1@.50} & \BenchHead{F1@.95} & \BenchHead{F1mIoU} \\
  \midrule
  \multicolumn{10}{@{}l}{\hspace{0.4em}\strut\textbf{Open-set Specialized Detectors}} \\
  GroundingDINO\hspace{0.35em}\citep{groundingdino} & 51.19 & 23.28 & 44.33 & 57.99 & 23.17 & 49.69 & 49.20 & 20.89 & 41.42 \\
  \addlinespace[2pt]
  \multicolumn{10}{@{}l}{\hspace{0.4em}\strut\textbf{Vision-Language Models (<10B)}} \\
  Rex-Omni\hspace{0.35em}\citep{rexomni} & 84.37 & 32.85 & 71.43 & 84.74 & 34.74 & 72.34 & 76.92 & 29.53 & 64.10 \\
  LocateAnything Fast\hspace{0.35em}\citep{locateanything} & 92.48 & 43.95 & 80.61 & 88.47 & 40.69 & 76.17 & 84.43 & 39.53 & 73.08 \\
  LocateAnything Hybrid\hspace{0.35em}\citep{locateanything} & 92.73 & 44.32 & 81.37 & 89.42 & 41.46 & 77.73 & 85.80 & 40.57 & 75.08 \\
  LocateAnything Slow NTP\hspace{0.35em}\citep{locateanything} & 92.43 & 38.43 & 80.21 & 88.65 & 35.63 & 76.08 & 85.22 & 35.69 & 73.90 \\
  Qwen2.5-VL-7B\hspace{0.35em}\citep{qwen25vl} & 82.42 & 17.71 & 66.93 & 75.28 & 15.56 & 60.28 & 73.04 & 16.39 & 59.27 \\
  Qwen3-VL-2B\hspace{0.35em}\citep{qwen3vl} & 88.33 & 26.65 & 72.39 & 85.89 & 25.73 & 69.81 & 81.21 & 24.34 & 66.47 \\
  Qwen3-VL-4B\hspace{0.35em}\citep{qwen3vl} & 92.09 & 32.75 & 77.63 & 88.79 & 32.25 & 74.73 & 86.76 & 30.94 & 73.36 \\
  Qwen3-VL-8B\hspace{0.35em}\citep{qwen3vl} & 91.11 & 29.87 & 75.83 & 88.74 & 28.82 & 73.33 & 86.43 & 29.02 & 71.98 \\
  Qwen3.5-4B\hspace{0.35em}\citep{qwen35} & 90.11 & 29.07 & 74.46 & 88.42 & 26.98 & 71.61 & 84.13 & 27.79 & 69.48 \\
  Qwen3.5-9B\hspace{0.35em}\citep{qwen35} & 91.97 & 35.09 & 78.31 & 89.53 & 33.47 & 75.27 & 87.45 & 33.90 & 74.64 \\
  RynnBrain1.1\hspace{0.35em}\citep{RynnBrain112B} & 82.61 & 20.88 & 65.80 & 84.28 & 21.48 & 67.57 & 72.45 & 19.18 & 57.52 \\
  SenseNova-Vision\hspace{0.35em}\citep{SenseNovaVision7BMoT} & 90.12 & 45.07 & 79.66 & 89.18 & 44.73 & 78.69 & 84.40 & 41.68 & 74.28 \\
  MiMo-VL-7B-SFT\hspace{0.35em}\citep{MimoVL} & 86.05 & 14.45 & 66.26 & 84.87 & 14.33 & 64.76 & 76.79 & 13.13 & 58.86 \\
  MiMo-VL-7B-RL\hspace{0.35em}\citep{MimoVL} & 90.44 & 14.15 & 68.45 & 87.69 & 13.11 & 64.80 & 84.83 & 13.55 & 64.42 \\
  BAGEL\hspace{0.35em}\citep{BAGEL7BMoT} & 79.39 & 25.09 & 65.23 & 77.69 & 22.20 & 62.30 & 68.34 & 20.45 & 54.45 \\
  RynnBrain\hspace{0.35em}\citep{dang2026rynnbrain} & 73.99 & 13.60 & 55.59 & 76.49 & 14.29 & 57.91 & 65.51 & 11.72 & 48.88 \\
  GroundingPI & \textbf{95.85} & 48.32 & 85.79 & 93.17 & 43.48 & 81.96 & \textbf{93.62} & 48.41 & 84.01 \\
  \addlinespace[2pt]
  \multicolumn{10}{@{}l}{\hspace{0.4em}\strut\textbf{Vision-Language Models (10B--1T)}} \\
  Qwen3.6-27B\hspace{0.35em}\citep{qwen3627b} & 93.46 & 39.83 & 80.78 & 91.33 & 38.69 & 78.37 & 89.57 & 38.36 & 77.44 \\
  Qwen3-VL-32B\hspace{0.35em}\citep{qwen3vl} & 90.33 & 34.50 & 77.21 & 84.49 & 31.54 & 71.75 & 85.34 & 33.08 & 73.17 \\
  Qwen3.8-27B\hspace{0.35em}\citep{qwen38} & 92.93 & 38.44 & 79.83 & 90.37 & 37.15 & 76.91 & 89.07 & 37.76 & 76.91 \\
  Qwen3.5-35B-A3B\hspace{0.35em}\citep{qwen35} & 92.57 & 36.44 & 79.30 & 90.62 & 34.77 & 76.49 & 87.61 & 34.38 & 75.06 \\
  DeepSeek-VL2-Small-16B\hspace{0.35em}\citep{Deepseekvl2} & 66.95 & 1.10 & 30.95 & 64.47 & 1.29 & 29.60 & 64.93 & 1.25 & 30.58 \\
  DeepSeek-VL2-27B\hspace{0.35em}\citep{Deepseekvl2} & 94.73 & \textbf{74.35} & \textbf{90.43} & \textbf{94.02} & \textbf{65.74} & \textbf{87.32} & 91.33 & \textbf{71.62} & \textbf{87.19} \\
  \addlinespace[2pt]
  \multicolumn{10}{@{}l}{\hspace{0.4em}\strut\textbf{Vision-Language Models (>1T)}} \\
  Qwen3.7-Max\hspace{0.35em}\citep{qwen37} & 95.11 & 45.55 & 82.99 & 92.78 & 44.87 & 80.38 & 91.78 & 32.41 & 77.23 \\
  Kimi-K2.6 & 82.55 & 39.91 & 72.27 & 82.06 & 40.18 & 71.83 & 76.00 & 33.87 & 64.89 \\
  Kimi-K3\hspace{0.35em}\citep{KimiK3} & 88.46 & 35.78 & 75.12 & 88.57 & 34.51 & 74.14 & 81.86 & 33.77 & 69.31 \\
  GPT-6 Astra & 93.61 & 37.14 & 79.45 & 87.21 & 44.27 & 76.15 & 90.45 & 36.30 & 77.84 \\
  \addlinespace[2pt]
  \bottomrule
  
  \end{NiceTabular}%
  }
  \endgroup
\end{table}

\subsection{Object Pointing}
\label{app:object_pointing}

Following Rex-Omni~\citep{rexomni}, SAM~\citep{SAM} converts ground-truth boxes into object masks. A point is correct when it lies inside the corresponding mask. Detection-style recall, precision, and F1 are then computed; we denote the latter F1@Point.

\paragraph{Referring object pointing.}
GroundingPI reaches 88.79 F1@Point on HumanRef and 90.26/90.06 on RefCOCOg validation/test, exceeding Astra and the selected grounding specialists. HumanRef precision is 93.24, while recall is 84.76: target selection is reliable, but missed instances remain. The box-to-mask conversion fixes the scoring region; these results measure point correctness rather than box-boundary quality.

\begin{table}[htbp]
  \centering
  \BenchmarkTableFont
  \caption{\textbf{Referring object pointing.} Starred BAGEL scores are retained despite unreliable support for the unified pointing protocol. Kimi-K3, both MiMo variants, and both DeepSeek variants are N/A under that protocol. These outcomes do not establish intrinsic pointing capability. The starred Molmo and SEED1.5-VL scores are from Table~7 of \citet{rexomni}; its Molmo checkpoint is Molmo-7B-D.}
  \label{tab:app-point-referring}
  \begingroup
  \fontsize{8}{9.7}\selectfont
  \setlength{\tabcolsep}{3pt}
  \renewcommand{\arraystretch}{1.15}
  \resizebox{\linewidth}{!}{%
  \begin{NiceTabular}{@{}>{\raggedright\arraybackslash}p{177pt}ccccc@{}}
  \CodeBefore
    \rowcolor{benchmarktype}{3}
    \rowcolor{benchmarktype}{5}
    \rowcolor{benchmarkpurple}{23}
    \rowcolor{benchmarktype}{24}
    \rowcolor{benchmarktype}{32}
  \Body
  \toprule
  \textbf{Model} & \multicolumn{3}{c}{\textbf{HumanRef}} & \multicolumn{1}{c}{\textbf{RefCOCOg val}} & \multicolumn{1}{c}{\textbf{RefCOCOg test}} \\
  \cmidrule(lr){2-4}\cmidrule(lr){5-5}\cmidrule(lr){6-6}
   & \BenchHead{R@Point} & \BenchHead{P@Point} & \BenchHead{F1@Point} & \BenchHead{F1@Point} & \BenchHead{F1@Point} \\
  \midrule
  \multicolumn{6}{@{}l}{\hspace{0.4em}\strut\textbf{Open-set Specialized Detectors}} \\
  GroundingDINO\hspace{0.35em}\citep{groundingdino} & 50.22 & 43.90 & 46.85 & 49.34 & 49.97 \\
  \addlinespace[2pt]
  \multicolumn{6}{@{}l}{\hspace{0.4em}\strut\textbf{Vision-Language Models (<10B)}} \\
  Rex-Omni\hspace{0.35em}\citep{rexomni} & 84.05 & 82.76 & 83.40 & 84.96 & 85.32 \\
  LocateAnything Fast\hspace{0.35em}\citep{locateanything} & 66.10 & 68.28 & 67.18 & 73.84 & 74.94 \\
  LocateAnything Hybrid\hspace{0.35em}\citep{locateanything} & 71.55 & 71.33 & 71.44 & 75.89 & 76.65 \\
  LocateAnything Slow NTP\hspace{0.35em}\citep{locateanything} & 71.40 & 71.52 & 71.46 & 77.17 & 77.59 \\
  Qwen2.5-VL-7B\hspace{0.35em}\citep{qwen25vl} & 47.50 & 54.76 & 50.87 & 81.65 & 82.48 \\
  Qwen3-VL-2B\hspace{0.35em}\citep{qwen3vl} & 58.74 & 70.48 & 64.08 & 76.17 & 76.01 \\
  Qwen3-VL-4B\hspace{0.35em}\citep{qwen3vl} & 57.64 & 79.67 & 66.89 & 76.43 & 77.64 \\
  Qwen3-VL-8B\hspace{0.35em}\citep{qwen3vl} & 71.13 & 81.15 & 75.81 & 81.97 & 82.09 \\
  Qwen3.5-4B\hspace{0.35em}\citep{qwen35} & 74.96 & 82.10 & 78.37 & 79.35 & 79.31 \\
  Qwen3.5-9B\hspace{0.35em}\citep{qwen35} & 74.86 & 81.88 & 78.21 & 77.59 & 77.85 \\
  RynnBrain1.1\hspace{0.35em}\citep{RynnBrain112B} & 54.15 & 73.98 & 62.53 & 74.42 & 74.17 \\
  SenseNova-Vision\hspace{0.35em}\citep{SenseNovaVision7BMoT} & 74.70 & 73.39 & 74.04 & 74.63 & 75.42 \\
  MiMo-VL-7B-SFT\hspace{0.35em}\citep{MimoVL} & N/A\textsuperscript{*} & N/A\textsuperscript{*} & N/A\textsuperscript{*} & N/A\textsuperscript{*} & N/A\textsuperscript{*} \\
  MiMo-VL-7B-RL\hspace{0.35em}\citep{MimoVL} & N/A\textsuperscript{*} & N/A\textsuperscript{*} & N/A\textsuperscript{*} & N/A\textsuperscript{*} & N/A\textsuperscript{*} \\
  BAGEL\hspace{0.35em}\citep{BAGEL7BMoT} & 42.77\textsuperscript{*} & 48.66\textsuperscript{*} & 45.52\textsuperscript{*} & 55.92\textsuperscript{*} & 54.57\textsuperscript{*} \\
  RynnBrain\hspace{0.35em}\citep{dang2026rynnbrain} & 51.00 & 72.06 & 59.73 & 73.34 & 73.67 \\
  Molmo-7B\textsuperscript{*}\hspace{0.35em}\citep{Molmo} & -- & -- & 70.00 & 83.70 & 83.60 \\
  GroundingPI & \textbf{84.76} & \textbf{93.24} & \textbf{88.79} & \textbf{90.26} & \textbf{90.06} \\
  \addlinespace[2pt]
  \multicolumn{6}{@{}l}{\hspace{0.4em}\strut\textbf{Vision-Language Models (10B--1T)}} \\
  Qwen3.6-27B\hspace{0.35em}\citep{qwen3627b} & 82.01 & 84.28 & 83.13 & 82.73 & 82.88 \\
  Qwen3-VL-32B\hspace{0.35em}\citep{qwen3vl} & 64.78 & 72.21 & 68.30 & 76.52 & 76.23 \\
  Qwen3.8-27B\hspace{0.35em}\citep{qwen38} & 73.17 & 75.18 & 74.16 & 75.84 & 75.86 \\
  Qwen3.5-35B-A3B\hspace{0.35em}\citep{qwen35} & 77.76 & 82.37 & 80.00 & 75.35 & 76.09 \\
  DeepSeek-VL2-Small-16B\hspace{0.35em}\citep{Deepseekvl2} & N/A\textsuperscript{*} & N/A\textsuperscript{*} & N/A\textsuperscript{*} & N/A\textsuperscript{*} & N/A\textsuperscript{*} \\
  DeepSeek-VL2-27B\hspace{0.35em}\citep{Deepseekvl2} & N/A\textsuperscript{*} & N/A\textsuperscript{*} & N/A\textsuperscript{*} & N/A\textsuperscript{*} & N/A\textsuperscript{*} \\
  SEED1.5-VL\textsuperscript{*}\hspace{0.35em}\citep{SEED15VL} & -- & -- & 83.10 & 83.60 & 84.20 \\
  \addlinespace[2pt]
  \multicolumn{6}{@{}l}{\hspace{0.4em}\strut\textbf{Vision-Language Models (>1T)}} \\
  Qwen3.7-Max\hspace{0.35em}\citep{qwen37} & 77.39 & 85.61 & 81.30 & 71.21 & 72.40 \\
  Kimi-K2.6 & 55.79\textsuperscript{\textdagger} & 56.15\textsuperscript{\textdagger} & 55.97\textsuperscript{\textdagger} & 39.59\textsuperscript{\textdagger} & 39.60\textsuperscript{\textdagger} \\
  Kimi-K3\hspace{0.35em}\citep{KimiK3} & N/A\textsuperscript{*} & N/A\textsuperscript{*} & N/A\textsuperscript{*} & N/A\textsuperscript{*} & N/A\textsuperscript{*} \\
  GPT-6 Astra & 82.85 & 85.08 & 83.83 & 87.80 & 84.90 \\
  \addlinespace[2pt]
  \bottomrule
  
  \end{NiceTabular}%
  }
  \endgroup
\end{table}

\paragraph{Common and long-tailed object pointing.}
GroundingPI reaches 84.79 F1@Point on COCO and 79.51 on LVIS, versus Astra's 82.17 and 77.14. On COCO, Astra has higher recall (86.56 versus 84.20), while GroundingPI has higher precision (85.38 versus 78.14). Its F1 advantage therefore reflects a better balance of coverage and false positives, not uniformly higher recall.

\begin{table}[htbp]
  \centering
  \BenchmarkTableFont
  \caption{\textbf{Object pointing on COCO and LVIS.} Starred BAGEL scores are retained despite unreliable support for the unified pointing protocol. Kimi-K3, both MiMo variants, and both DeepSeek variants are N/A under that protocol. These outcomes do not establish intrinsic pointing capability. The starred Molmo and SEED1.5-VL scores are from Table~7 of \citet{rexomni}; its Molmo checkpoint is Molmo-7B-D.}
  \label{tab:app-point-common}
  \begingroup
  \fontsize{8}{9.7}\selectfont
  \setlength{\tabcolsep}{3pt}
  \renewcommand{\arraystretch}{1.15}
  \resizebox{\linewidth}{!}{%
  \begin{NiceTabular}{@{}>{\raggedright\arraybackslash}p{177pt}cccccc@{}}
  \CodeBefore
    \rowcolor{benchmarktype}{3}
    \rowcolor{benchmarktype}{5}
    \rowcolor{benchmarkpurple}{23}
    \rowcolor{benchmarktype}{24}
    \rowcolor{benchmarktype}{32}
  \Body
  \toprule
  \textbf{Model} & \multicolumn{3}{c}{\textbf{COCO}} & \multicolumn{3}{c}{\textbf{LVIS}} \\
  \cmidrule(lr){2-4}\cmidrule(lr){5-7}
   & \BenchHead{R@Point} & \BenchHead{P@Point} & \BenchHead{F1@Point} & \BenchHead{R@Point} & \BenchHead{P@Point} & \BenchHead{F1@Point} \\
  \midrule
  \multicolumn{7}{@{}l}{\hspace{0.4em}\strut\textbf{Open-set Specialized Detectors}} \\
  GroundingDINO\hspace{0.35em}\citep{groundingdino} & 68.92 & 71.97 & 70.41 & 44.91 & 71.15 & 55.07 \\
  \addlinespace[2pt]
  \multicolumn{7}{@{}l}{\hspace{0.4em}\strut\textbf{Vision-Language Models (<10B)}} \\
  Rex-Omni\hspace{0.35em}\citep{rexomni} & 77.81 & 81.77 & 79.74 & 63.46 & 78.14 & 70.04 \\
  LocateAnything Fast\hspace{0.35em}\citep{locateanything} & 68.99 & 74.26 & 71.53 & 55.81 & 70.05 & 62.13 \\
  LocateAnything Hybrid\hspace{0.35em}\citep{locateanything} & 73.80 & 73.77 & 73.78 & 60.96 & 69.36 & 64.89 \\
  LocateAnything Slow NTP\hspace{0.35em}\citep{locateanything} & 74.68 & 75.05 & 74.86 & 62.36 & 72.42 & 67.01 \\
  Qwen2.5-VL-7B\hspace{0.35em}\citep{qwen25vl} & 61.24 & 65.76 & 63.42 & 46.46 & 56.82 & 51.12 \\
  Qwen3-VL-2B\hspace{0.35em}\citep{qwen3vl} & 55.65 & 54.98 & 55.31 & 43.70 & 50.11 & 46.69 \\
  Qwen3-VL-4B\hspace{0.35em}\citep{qwen3vl} & 63.13 & 67.69 & 65.33 & 49.45 & 62.17 & 55.08 \\
  Qwen3-VL-8B\hspace{0.35em}\citep{qwen3vl} & 64.92 & 66.76 & 65.83 & 52.25 & 61.41 & 56.46 \\
  Qwen3.5-4B\hspace{0.35em}\citep{qwen35} & 70.15 & 68.85 & 69.50 & 55.61 & 64.72 & 59.82 \\
  Qwen3.5-9B\hspace{0.35em}\citep{qwen35} & 71.87 & 72.56 & 72.21 & 58.90 & 70.07 & 64.00 \\
  RynnBrain1.1\hspace{0.35em}\citep{RynnBrain112B} & 21.06 & 32.96 & 25.70 & 13.35 & 25.63 & 17.56 \\
  SenseNova-Vision\hspace{0.35em}\citep{SenseNovaVision7BMoT} & 70.85 & 75.21 & 72.96 & 57.21 & 69.26 & 62.66 \\
  MiMo-VL-7B-SFT\hspace{0.35em}\citep{MimoVL} & N/A\textsuperscript{*} & N/A\textsuperscript{*} & N/A\textsuperscript{*} & N/A\textsuperscript{*} & N/A\textsuperscript{*} & N/A\textsuperscript{*} \\
  MiMo-VL-7B-RL\hspace{0.35em}\citep{MimoVL} & N/A\textsuperscript{*} & N/A\textsuperscript{*} & N/A\textsuperscript{*} & N/A\textsuperscript{*} & N/A\textsuperscript{*} & N/A\textsuperscript{*} \\
  BAGEL\hspace{0.35em}\citep{BAGEL7BMoT} & 33.57\textsuperscript{*} & 39.16\textsuperscript{*} & 36.15\textsuperscript{*} & 22.73\textsuperscript{*} & 31.65\textsuperscript{*} & 26.46\textsuperscript{*} \\
  RynnBrain\hspace{0.35em}\citep{dang2026rynnbrain} & 9.46 & 17.73 & 12.33 & 5.38 & 13.73 & 7.73 \\
  Molmo-7B\textsuperscript{*}\hspace{0.35em}\citep{Molmo} & -- & -- & 77.30 & -- & -- & 40.30 \\
  GroundingPI & 84.20 & \textbf{85.38} & \textbf{84.79} & \textbf{76.55} & \textbf{82.71} & \textbf{79.51} \\
  \addlinespace[2pt]
  \multicolumn{7}{@{}l}{\hspace{0.4em}\strut\textbf{Vision-Language Models (10B--1T)}} \\
  Qwen3.6-27B\hspace{0.35em}\citep{qwen3627b} & 74.01 & 71.98 & 72.98 & 62.98 & 72.61 & 67.45 \\
  Qwen3-VL-32B\hspace{0.35em}\citep{qwen3vl} & 72.70 & 70.24 & 71.45 & 60.44 & 66.42 & 63.29 \\
  Qwen3.8-27B\hspace{0.35em}\citep{qwen38} & 72.69 & 75.38 & 74.01 & 61.98 & 74.47 & 67.65 \\
  Qwen3.5-35B-A3B\hspace{0.35em}\citep{qwen35} & 72.21 & 71.42 & 71.81 & 61.05 & 70.99 & 65.65 \\
  DeepSeek-VL2-Small-16B\hspace{0.35em}\citep{Deepseekvl2} & N/A\textsuperscript{*} & N/A\textsuperscript{*} & N/A\textsuperscript{*} & N/A\textsuperscript{*} & N/A\textsuperscript{*} & N/A\textsuperscript{*} \\
  DeepSeek-VL2-27B\hspace{0.35em}\citep{Deepseekvl2} & N/A\textsuperscript{*} & N/A\textsuperscript{*} & N/A\textsuperscript{*} & N/A\textsuperscript{*} & N/A\textsuperscript{*} & N/A\textsuperscript{*} \\
  SEED1.5-VL\textsuperscript{*}\hspace{0.35em}\citep{SEED15VL} & -- & -- & 78.20 & -- & -- & 70.70 \\
  \addlinespace[2pt]
  \multicolumn{7}{@{}l}{\hspace{0.4em}\strut\textbf{Vision-Language Models (>1T)}} \\
  Qwen3.7-Max\hspace{0.35em}\citep{qwen37} & 71.61 & 72.65 & 72.13 & 60.97 & 74.22 & 66.94 \\
  Kimi-K2.6 & 29.83\textsuperscript{\textdagger} & 31.52\textsuperscript{\textdagger} & 30.65\textsuperscript{\textdagger} & 23.20\textsuperscript{\textdagger} & 28.32\textsuperscript{\textdagger} & 25.51\textsuperscript{\textdagger} \\
  Kimi-K3\hspace{0.35em}\citep{KimiK3} & N/A\textsuperscript{*} & N/A\textsuperscript{*} & N/A\textsuperscript{*} & N/A\textsuperscript{*} & N/A\textsuperscript{*} & N/A\textsuperscript{*} \\
  GPT-6 Astra & \textbf{86.56} & 78.14 & 82.17 & 74.92 & 79.50 & 77.14 \\
  \addlinespace[2pt]
  \bottomrule
  
  \end{NiceTabular}%
  }
  \endgroup
\end{table}

\paragraph{Dense and tiny-object pointing.}
GroundingPI reaches 81.27 on Dense200 and 67.47 on VisDrone. Astra leads Dense200 at 86.57 through substantially higher recall, whereas GroundingPI has higher precision (85.91 versus 83.74). On VisDrone, GroundingPI's 74.17 precision supports a higher F1 than Astra's 65.62. This contrast reinforces the need to evaluate both point selection and full-box localization.

\begin{table}[htbp]
  \centering
  \BenchmarkTableFont
  \caption{\textbf{Object pointing on Dense200 and VisDrone.} Starred BAGEL scores are retained despite unreliable support for the unified pointing protocol. Kimi-K3, both MiMo variants, and both DeepSeek variants are N/A under that protocol. These outcomes do not establish intrinsic pointing capability. The starred Molmo and SEED1.5-VL scores are from Table~7 of \citet{rexomni}; its Molmo checkpoint is Molmo-7B-D.}
  \label{tab:app-point-dense}
  \begingroup
  \fontsize{8}{9.7}\selectfont
  \setlength{\tabcolsep}{3pt}
  \renewcommand{\arraystretch}{1.15}
  \resizebox{\linewidth}{!}{%
  \begin{NiceTabular}{@{}>{\raggedright\arraybackslash}p{177pt}cccccc@{}}
  \CodeBefore
    \rowcolor{benchmarktype}{3}
    \rowcolor{benchmarktype}{5}
    \rowcolor{benchmarkpurple}{23}
    \rowcolor{benchmarktype}{24}
    \rowcolor{benchmarktype}{32}
  \Body
  \toprule
  \textbf{Model} & \multicolumn{3}{c}{\textbf{Dense200}} & \multicolumn{3}{c}{\textbf{VisDrone}} \\
  \cmidrule(lr){2-4}\cmidrule(lr){5-7}
   & \BenchHead{R@Point} & \BenchHead{P@Point} & \BenchHead{F1@Point} & \BenchHead{R@Point} & \BenchHead{P@Point} & \BenchHead{F1@Point} \\
  \midrule
  \multicolumn{7}{@{}l}{\hspace{0.4em}\strut\textbf{Open-set Specialized Detectors}} \\
  GroundingDINO\hspace{0.35em}\citep{groundingdino} & 22.00 & 65.45 & 32.93 & 27.20 & 71.76 & 39.45 \\
  \addlinespace[2pt]
  \multicolumn{7}{@{}l}{\hspace{0.4em}\strut\textbf{Vision-Language Models (<10B)}} \\
  Rex-Omni\hspace{0.35em}\citep{rexomni} & 75.59 & 77.76 & 76.66 & 47.81 & 56.92 & 51.97 \\
  LocateAnything Fast\hspace{0.35em}\citep{locateanything} & 64.63 & 66.65 & 65.63 & 55.22 & 57.55 & 56.36 \\
  LocateAnything Hybrid\hspace{0.35em}\citep{locateanything} & 77.43 & 78.73 & 78.07 & 59.18 & 55.54 & 57.30 \\
  LocateAnything Slow NTP\hspace{0.35em}\citep{locateanything} & 78.87 & 81.39 & 80.11 & 61.17 & 60.77 & 60.97 \\
  Qwen2.5-VL-7B\hspace{0.35em}\citep{qwen25vl} & 12.12 & 36.42 & 18.19 & 12.81 & 18.21 & 15.04 \\
  Qwen3-VL-2B\hspace{0.35em}\citep{qwen3vl} & 14.06 & 15.95 & 14.95 & 10.64 & 11.98 & 11.27 \\
  Qwen3-VL-4B\hspace{0.35em}\citep{qwen3vl} & 14.13 & 46.93 & 21.72 & 21.27 & 26.25 & 23.50 \\
  Qwen3-VL-8B\hspace{0.35em}\citep{qwen3vl} & 20.61 & 32.96 & 25.36 & 17.68 & 18.54 & 18.10 \\
  Qwen3.5-4B\hspace{0.35em}\citep{qwen35} & 56.82 & 60.30 & 58.51 & 32.72 & 31.57 & 32.13 \\
  Qwen3.5-9B\hspace{0.35em}\citep{qwen35} & 61.02 & 70.33 & 65.35 & 44.42 & 45.04 & 44.73 \\
  RynnBrain1.1\hspace{0.35em}\citep{RynnBrain112B} & 2.10 & 34.50 & 3.95 & 7.68 & 46.55 & 13.18 \\
  SenseNova-Vision\hspace{0.35em}\citep{SenseNovaVision7BMoT} & 74.86 & 82.98 & 78.71 & 59.36 & 64.48 & 61.81 \\
  MiMo-VL-7B-SFT\hspace{0.35em}\citep{MimoVL} & N/A\textsuperscript{*} & N/A\textsuperscript{*} & N/A\textsuperscript{*} & N/A\textsuperscript{*} & N/A\textsuperscript{*} & N/A\textsuperscript{*} \\
  MiMo-VL-7B-RL\hspace{0.35em}\citep{MimoVL} & N/A\textsuperscript{*} & N/A\textsuperscript{*} & N/A\textsuperscript{*} & N/A\textsuperscript{*} & N/A\textsuperscript{*} & N/A\textsuperscript{*} \\
  BAGEL\hspace{0.35em}\citep{BAGEL7BMoT} & 8.48\textsuperscript{*} & 16.84\textsuperscript{*} & 11.28\textsuperscript{*} & 5.09\textsuperscript{*} & 7.33\textsuperscript{*} & 6.01\textsuperscript{*} \\
  RynnBrain\hspace{0.35em}\citep{dang2026rynnbrain} & 0.51 & 1.00 & 0.68 & 5.64 & 34.46 & 9.69 \\
  Molmo-7B\textsuperscript{*}\hspace{0.35em}\citep{Molmo} & -- & -- & 33.10 & -- & -- & 29.20 \\
  GroundingPI & 77.10 & \textbf{85.91} & 81.27 & 61.88 & \textbf{74.17} & \textbf{67.47} \\
  \addlinespace[2pt]
  \multicolumn{7}{@{}l}{\hspace{0.4em}\strut\textbf{Vision-Language Models (10B--1T)}} \\
  Qwen3.6-27B\hspace{0.35em}\citep{qwen3627b} & 70.52 & 75.69 & 73.01 & 46.68 & 43.94 & 45.27 \\
  Qwen3-VL-32B\hspace{0.35em}\citep{qwen3vl} & 44.04 & 46.11 & 45.05 & 27.88 & 25.71 & 26.75 \\
  Qwen3.8-27B\hspace{0.35em}\citep{qwen38} & 73.97 & 75.15 & 74.55 & 51.07 & 51.25 & 51.16 \\
  Qwen3.5-35B-A3B\hspace{0.35em}\citep{qwen35} & 68.54 & 72.89 & 70.65 & 45.64 & 43.92 & 44.76 \\
  DeepSeek-VL2-Small-16B\hspace{0.35em}\citep{Deepseekvl2} & N/A\textsuperscript{*} & N/A\textsuperscript{*} & N/A\textsuperscript{*} & N/A\textsuperscript{*} & N/A\textsuperscript{*} & N/A\textsuperscript{*} \\
  DeepSeek-VL2-27B\hspace{0.35em}\citep{Deepseekvl2} & N/A\textsuperscript{*} & N/A\textsuperscript{*} & N/A\textsuperscript{*} & N/A\textsuperscript{*} & N/A\textsuperscript{*} & N/A\textsuperscript{*} \\
  SEED1.5-VL\textsuperscript{*}\hspace{0.35em}\citep{SEED15VL} & -- & -- & 72.10 & -- & -- & 56.70 \\
  \addlinespace[2pt]
  \multicolumn{7}{@{}l}{\hspace{0.4em}\strut\textbf{Vision-Language Models (>1T)}} \\
  Qwen3.7-Max\hspace{0.35em}\citep{qwen37} & 57.56 & 78.12 & 66.28 & 55.04 & 63.65 & 59.04 \\
  Kimi-K2.6 & 36.14\textsuperscript{\textdagger} & 34.27\textsuperscript{\textdagger} & 35.18\textsuperscript{\textdagger} & 11.22\textsuperscript{\textdagger} & 17.37\textsuperscript{\textdagger} & 13.63\textsuperscript{\textdagger} \\
  Kimi-K3\hspace{0.35em}\citep{KimiK3} & N/A\textsuperscript{*} & N/A\textsuperscript{*} & N/A\textsuperscript{*} & N/A\textsuperscript{*} & N/A\textsuperscript{*} & N/A\textsuperscript{*} \\
  GPT-6 Astra & \textbf{89.52} & 83.74 & \textbf{86.57} & \textbf{64.39} & 66.99 & 65.62 \\
  \addlinespace[2pt]
  \bottomrule
  
  \end{NiceTabular}%
  }
  \endgroup
\end{table}

\subsection{Robot and Spatial Pointing}
\label{app:robot_spatial_pointing}

\paragraph{Robot and spatial pointing.}
GroundingPI obtains 76.00/75.00 on RefSpatial Location/Placement and 75.32 on Unseen, with no marked drop between the two familiar splits and the unseen split. Astra remains higher on all three, but GroundingPI leads their RoboSpatial Context comparison (73.77 versus 65.69). These results suggest complementary strengths in instruction-conditioned placement and contextual localization; they do not directly measure closed-loop manipulation.

\begin{table}[p]
  \centering
  \BenchmarkTableFont
  \caption{\textbf{Robot and spatial pointing.} Point-in-mask accuracy is reported for each dataset. The starred RefSpatial baselines follow Table~11 of \citet{rexomni}; RoboRefer uses the setting without a depth prior. Values retain the precision recorded in our evaluation tables.}
  \label{tab:app-robot-spatial}
  \begingroup
  \fontsize{8}{9.7}\selectfont
  \setlength{\tabcolsep}{4pt}
  \renewcommand{\arraystretch}{1.15}
  \resizebox{\linewidth}{!}{%
  \begin{NiceTabular}{@{}>{\raggedright\arraybackslash}p{177pt}cccc@{}}
  \CodeBefore
    \rowcolor{benchmarktype}{3}
    \rowcolor{benchmarktype}{5}
    \rowcolor{benchmarkpurple}{24}
    \rowcolor{benchmarktype}{25}
    \rowcolor{benchmarktype}{36}
  \Body
  \toprule
  \textbf{Model} & \multicolumn{4}{c}{\textbf{Point-in-mask accuracy}} \\
  \cmidrule(lr){2-5}
   & \BenchHead{RefSpatial\\Location} & \BenchHead{RefSpatial\\Placement} & \BenchHead{RefSpatial\\Unseen} & \BenchHead{RoboSpatial\\Context} \\
  \midrule
  \multicolumn{5}{@{}l}{\hspace{0.4em}\strut\textbf{Open-set Specialized Detectors}} \\
  GroundingDINO\hspace{0.35em}\citep{groundingdino} & 26.50 & 2.00 & 4.33 & 4.92 \\
  \addlinespace[2pt]
  \multicolumn{5}{@{}l}{\hspace{0.4em}\strut\textbf{Vision-Language Models (<10B)}} \\
  Rex-Omni\hspace{0.35em}\citep{rexomni} & 51.00 & 52.50 & 37.01 & 59.02 \\
  LocateAnything Fast\hspace{0.35em}\citep{locateanything} & 53.00 & 19.00 & 16.88 & 15.57 \\
  LocateAnything Hybrid\hspace{0.35em}\citep{locateanything} & 55.00 & 17.33 & 20.78 & 14.75 \\
  LocateAnything Slow NTP\hspace{0.35em}\citep{locateanything} & 54.00 & 27.20 & 16.88 & 16.39 \\
  Qwen2.5-VL-7B\hspace{0.35em}\citep{qwen25vl} & 43.00 & 15.50 & 18.18 & 26.23 \\
  Qwen3-VL-2B\hspace{0.35em}\citep{qwen3vl} & 42.00 & 24.00 & 11.69 & 32.79 \\
  Qwen3-VL-4B\hspace{0.35em}\citep{qwen3vl} & 48.00 & 50.00 & 27.27 & 64.75 \\
  Qwen3-VL-8B\hspace{0.35em}\citep{qwen3vl} & 51.00 & 45.00 & 28.57 & 59.02 \\
  Qwen3.5-4B\hspace{0.35em}\citep{qwen35} & 65.00 & 38.00 & 38.96 & 50.82 \\
  Qwen3.5-9B\hspace{0.35em}\citep{qwen35} & 65.00 & 46.83 & 37.01 & 60.66 \\
  RynnBrain1.1\hspace{0.35em}\citep{RynnBrain112B} & 49.70 & 51.50 & 36.90 & 54.10 \\
  SenseNova-Vision\hspace{0.35em}\citep{SenseNovaVision7BMoT} & 34.51 & 6.25 & 8.54 & 0.82 \\
  MiMo-VL-7B-SFT\hspace{0.35em}\citep{MimoVL} & 1.00\textsuperscript{\textdagger} & 8.03\textsuperscript{\textdagger} & 4.64\textsuperscript{\textdagger} & 4.10\textsuperscript{\textdagger} \\
  MiMo-VL-7B-RL\hspace{0.35em}\citep{MimoVL} & 1.00\textsuperscript{\textdagger} & 2.20\textsuperscript{\textdagger} & 2.61\textsuperscript{\textdagger} & 4.92\textsuperscript{\textdagger} \\
  BAGEL\hspace{0.35em}\citep{BAGEL7BMoT} & 51.79 & 12.01 & 23.38 & 13.93 \\
  RynnBrain\hspace{0.35em}\citep{dang2026rynnbrain} & 42.00 & 37.00 & 22.08 & 28.69 \\
  Molmo-7B\textsuperscript{*}\hspace{0.35em}\citep{Molmo} & 21.90 & 12.80 & 12.20 & -- \\
  RoboRefer\textsuperscript{*}\hspace{0.35em}\citep{RoboRefer2B} & 51.00 & 49.00 & 39.00 & -- \\
  GroundingPI & 76.00 & 75.00 & 75.32 & \textbf{73.77} \\
  \addlinespace[2pt]
  \multicolumn{5}{@{}l}{\hspace{0.4em}\strut\textbf{Vision-Language Models (10B--1T)}} \\
  Qwen3.6-27B\hspace{0.35em}\citep{qwen3627b} & 72.00 & 66.00 & 61.04 & 63.93 \\
  Qwen3-VL-32B\hspace{0.35em}\citep{qwen3vl} & 62.00 & 52.00 & 41.56 & 63.93 \\
  Qwen3.8-27B\hspace{0.35em}\citep{qwen38} & 64.00 & 56.00 & 46.75 & 63.93 \\
  Qwen3.5-35B-A3B\hspace{0.35em}\citep{qwen35} & 70.00 & 58.00 & 53.25 & 71.31 \\
  DeepSeek-VL2-Small-16B\hspace{0.35em}\citep{Deepseekvl2} & 2.11\textsuperscript{\textdagger} & 1.33\textsuperscript{\textdagger} & 0.32\textsuperscript{\textdagger} & 5.74\textsuperscript{\textdagger} \\
  DeepSeek-VL2-27B\hspace{0.35em}\citep{Deepseekvl2} & 3.75\textsuperscript{\textdagger} & 0.66\textsuperscript{\textdagger} & 0.72\textsuperscript{\textdagger} & 5.74\textsuperscript{\textdagger} \\
  SpaceLLaVA\textsuperscript{*} & 5.80 & 4.30 & 4.00 & -- \\
  RoboPoint\textsuperscript{*}\hspace{0.35em}\citep{RoboPoint13B} & 22.90 & 9.30 & 8.40 & -- \\
  Molmo-72B\textsuperscript{*}\hspace{0.35em}\citep{Molmo} & 45.80 & 14.70 & 21.20 & -- \\
  Gemini-2.5-Pro\textsuperscript{*}\hspace{0.35em}\citep{Gemini2.5Pro} & 47.00 & 24.20 & 27.10 & -- \\
  \addlinespace[2pt]
  \multicolumn{5}{@{}l}{\hspace{0.4em}\strut\textbf{Vision-Language Models (>1T)}} \\
  Qwen3.7-Max\hspace{0.35em}\citep{qwen37} & 71.50 & 66.00 & 57.14 & 69.67 \\
  Kimi-K2.6 & 54.09 & 28.57 & 41.56 & 30.33 \\
  Kimi-K3\hspace{0.35em}\citep{KimiK3} & 67.85 & 50.00 & 54.98 & 54.92 \\
  GPT-6 Astra & \textbf{86.00} & \textbf{85.86} & \textbf{81.93} & 65.69 \\
  \addlinespace[2pt]
  \bottomrule
  
  \end{NiceTabular}%
  }
  \endgroup
\end{table}

\subsection{OCR}
\label{app:ocr}

\paragraph{HierText and ICDAR2015.}
GroundingPI obtains 41.70/55.68 F1mIoU, versus Astra's 39.58/48.87, with parse-error rates of 0.06\%/0.00\%. LocateAnything Slow NTP is stronger on HierText (42.94) than both GroundingPI and its own Hybrid mode (26.65), underscoring sensitivity to decoding mode. GroundingPI's ICDAR2015 advantage is larger at IoU 0.75 than at IoU 0.50, supporting improved joint transcription and region alignment under this metric.

\begin{table}[htbp]
  \centering
  \BenchmarkTableFont
  \caption{\textbf{OCR on HierText and ICDAR2015.} Each dataset reports four loose-match F1 measures and parse-error rate. GroundingDINO is N/A because OCR is unsupported. Kimi-K3 and both DeepSeek variants are N/A because their outputs do not satisfy the evaluation protocol; this does not establish a lack of OCR capability. SenseNova-Vision uses the HierText and ICDAR2015 F1mIoU scores of 31.20 and 49.50 reported in Table~1 of \citet{SenseNovaVision7BMoT}, because its local evaluation prompts could not be aligned. Other metrics for these datasets are unavailable; TotalText and SROIE use local results. The starred PaddleOCRv5 and SEED1.5-VL scores use the BBOX results in Table~10 of \citet{rexomni}.}
  \label{tab:app-ocr-hiertext-icdar}
  \begingroup
  \fontsize{8}{9.7}\selectfont
  \setlength{\tabcolsep}{2.5pt}
  \renewcommand{\arraystretch}{1.15}
  \resizebox{\linewidth}{!}{%
  \begin{NiceTabular}{@{}>{\raggedright\arraybackslash}p{177pt}cccccccccc@{}}
  \CodeBefore
    \rowcolor{benchmarktype}{3}
    \rowcolor{benchmarktype}{5}
    \rowcolor{benchmarktype}{7}
    \rowcolor{benchmarkpurple}{24}
    \rowcolor{benchmarktype}{25}
    \rowcolor{benchmarktype}{33}
  \Body
  \toprule
  \textbf{Model} & \multicolumn{5}{c}{\textbf{HierText}} & \multicolumn{5}{c}{\textbf{ICDAR2015}} \\
  \cmidrule(lr){2-6}\cmidrule(lr){7-11}
   & \BenchHead{F1@.50} & \BenchHead{F1@.75} & \BenchHead{F1@.95} & \BenchHead{F1mIoU} & \BenchHead{Parse\\err.} & \BenchHead{F1@.50} & \BenchHead{F1@.75} & \BenchHead{F1@.95} & \BenchHead{F1mIoU} & \BenchHead{Parse\\err.} \\
  \midrule
  \multicolumn{11}{@{}l}{\hspace{0.4em}\strut\textbf{Closed-set Specialized Detectors}} \\
  PaddleOCRv5\textsuperscript{*}\hspace{0.35em}\citep{PaddleOCRv5} & 45.20 & -- & 3.40 & 30.50 & -- & 38.20 & -- & 1.20 & 25.60 & -- \\
  \addlinespace[2pt]
  \multicolumn{11}{@{}l}{\hspace{0.4em}\strut\textbf{Open-set Specialized Detectors}} \\
  GroundingDINO\hspace{0.35em}\citep{groundingdino} & N/A\textsuperscript{*} & N/A\textsuperscript{*} & N/A\textsuperscript{*} & N/A\textsuperscript{*} & N/A\textsuperscript{*} & N/A\textsuperscript{*} & N/A\textsuperscript{*} & N/A\textsuperscript{*} & N/A\textsuperscript{*} & N/A\textsuperscript{*} \\
  \addlinespace[2pt]
  \multicolumn{11}{@{}l}{\hspace{0.4em}\strut\textbf{Vision-Language Models (<10B)}} \\
  Rex-Omni\hspace{0.35em}\citep{rexomni} & 54.16 & 35.67 & 2.10 & 34.46 & 1.92 & 73.39 & 50.07 & 0.96 & 45.65 & \textbf{0.00} \\
  LocateAnything Fast\hspace{0.35em}\citep{locateanything} & 29.91 & 25.83 & 3.13 & 22.59 & 31.46 & 44.51 & 28.32 & 0.40 & 26.73 & 3.43 \\
  LocateAnything Hybrid\hspace{0.35em}\citep{locateanything} & 35.50 & 30.43 & 3.61 & 26.65 & 19.04 & 45.61 & 29.41 & 0.40 & 27.48 & 1.21 \\
  LocateAnything Slow NTP\hspace{0.35em}\citep{locateanything} & 58.34 & \textbf{48.65} & \textbf{5.28} & 42.94 & 1.10 & 53.07 & 33.29 & 0.65 & 31.81 & \textbf{0.00} \\
  Qwen2.5-VL-7B\hspace{0.35em}\citep{qwen25vl} & 29.61 & 14.28 & 0.51 & 15.49 & \textbf{0.00} & 55.50 & 23.37 & 1.19 & 27.72 & \textbf{0.00} \\
  Qwen3-VL-2B\hspace{0.35em}\citep{qwen3vl} & 25.08 & 11.38 & 0.33 & 12.65 & 1.74 & 49.47 & 24.52 & 1.93 & 27.43 & 0.40 \\
  Qwen3-VL-4B\hspace{0.35em}\citep{qwen3vl} & 41.06 & 23.42 & 0.97 & 23.48 & 0.93 & 51.52 & 26.37 & 1.67 & 28.41 & 0.20 \\
  Qwen3-VL-8B\hspace{0.35em}\citep{qwen3vl} & 42.32 & 23.47 & 0.99 & 23.89 & 0.29 & 54.20 & 29.71 & 1.40 & 30.82 & \textbf{0.00} \\
  Qwen3.5-4B\hspace{0.35em}\citep{qwen35} & 30.47 & 16.50 & 0.76 & 17.00 & 12.65 & 41.16 & 16.20 & 0.37 & 19.61 & \textbf{0.00} \\
  Qwen3.5-9B\hspace{0.35em}\citep{qwen35} & 48.60 & 30.10 & 1.90 & 29.63 & 13.93 & 56.02 & 26.65 & 0.87 & 29.90 & \textbf{0.00} \\
  RynnBrain1.1\hspace{0.35em}\citep{RynnBrain112B} & 3.55 & 2.04 & 0.11 & 2.07 & 30.24 & 36.78 & 15.63 & 0.20 & 17.81 & 6.05 \\
  SenseNova-Vision\textsuperscript{*}\hspace{0.35em}\citep{SenseNovaVision7BMoT} & -- & -- & -- & 31.20 & -- & -- & -- & -- & 49.50 & -- \\
  MiMo-VL-7B-SFT\hspace{0.35em}\citep{MimoVL} & 24.77 & 12.82 & 0.36 & 13.44 & 1.33 & 64.31 & 30.58 & 0.46 & 34.33 & 1.01 \\
  MiMo-VL-7B-RL\hspace{0.35em}\citep{MimoVL} & 26.13 & 13.01 & 0.32 & 13.88 & 0.29 & 56.84 & 24.32 & 0.37 & 28.79 & 7.86 \\
  BAGEL\hspace{0.35em}\citep{BAGEL7BMoT} & 11.91 & 3.30 & 0.05 & 4.87 & 0.75 & 36.29 & 10.41 & 0.17 & 15.48 & 0.20 \\
  RynnBrain\hspace{0.35em}\citep{dang2026rynnbrain} & 4.08 & 1.92 & 0.06 & 2.13 & 0.64 & 36.05 & 15.15 & 0.47 & 17.81 & 0.60 \\
  GroundingPI & 60.02 & 46.16 & 4.11 & 41.70 & 0.06 & 76.79 & \textbf{65.86} & \textbf{3.59} & \textbf{55.68} & \textbf{0.00} \\
  \addlinespace[2pt]
  \multicolumn{11}{@{}l}{\hspace{0.4em}\strut\textbf{Vision-Language Models (10B--1T)}} \\
  Qwen3.6-27B\hspace{0.35em}\citep{qwen3627b} & 36.27 & 25.16 & 1.94 & 23.49 & 4.12 & 53.83 & 32.09 & 1.75 & 31.64 & 0.20 \\
  Qwen3-VL-32B\hspace{0.35em}\citep{qwen3vl} & 15.55 & 9.08 & 0.49 & 9.00 & 54.32 & 42.26 & 26.11 & 1.69 & 25.42 & 25.40 \\
  Qwen3.8-27B\hspace{0.35em}\citep{qwen38} & 51.50 & 35.06 & 2.46 & 32.95 & 0.06 & 68.67 & 39.98 & 0.95 & 39.72 & \textbf{0.00} \\
  Qwen3.5-35B-A3B\hspace{0.35em}\citep{qwen35} & 49.83 & 33.22 & 2.16 & 31.38 & 3.71 & 51.85 & 25.41 & 1.22 & 28.04 & 3.02 \\
  DeepSeek-VL2-Small-16B\hspace{0.35em}\citep{Deepseekvl2} & N/A\textsuperscript{*} & N/A\textsuperscript{*} & N/A\textsuperscript{*} & N/A\textsuperscript{*} & N/A\textsuperscript{*} & N/A\textsuperscript{*} & N/A\textsuperscript{*} & N/A\textsuperscript{*} & N/A\textsuperscript{*} & N/A\textsuperscript{*} \\
  DeepSeek-VL2-27B\hspace{0.35em}\citep{Deepseekvl2} & N/A\textsuperscript{*} & N/A\textsuperscript{*} & N/A\textsuperscript{*} & N/A\textsuperscript{*} & N/A\textsuperscript{*} & N/A\textsuperscript{*} & N/A\textsuperscript{*} & N/A\textsuperscript{*} & N/A\textsuperscript{*} & N/A\textsuperscript{*} \\
  SEED1.5-VL\textsuperscript{*}\hspace{0.35em}\citep{SEED15VL} & 27.10 & -- & 0.20 & 12.00 & -- & 38.60 & -- & 0.00 & 18.70 & -- \\
  \addlinespace[2pt]
  \multicolumn{11}{@{}l}{\hspace{0.4em}\strut\textbf{Vision-Language Models (>1T)}} \\
  Qwen3.7-Max\hspace{0.35em}\citep{qwen37} & \textbf{62.98} & 47.45 & 3.93 & \textbf{42.95} & \textbf{0.00} & 69.09 & 45.15 & 1.97 & 43.07 & \textbf{0.00} \\
  Kimi-K2.6 & 45.89\textsuperscript{\textdagger} & 24.25\textsuperscript{\textdagger} & 1.25\textsuperscript{\textdagger} & 25.26\textsuperscript{\textdagger} & 0.12\textsuperscript{\textdagger} & 56.81\textsuperscript{\textdagger} & 31.14\textsuperscript{\textdagger} & 1.95\textsuperscript{\textdagger} & 32.25\textsuperscript{\textdagger} & \textbf{0.00}\textsuperscript{\textdagger} \\
  Kimi-K3\hspace{0.35em}\citep{KimiK3} & N/A\textsuperscript{*} & N/A\textsuperscript{*} & N/A\textsuperscript{*} & N/A\textsuperscript{*} & N/A\textsuperscript{*} & N/A\textsuperscript{*} & N/A\textsuperscript{*} & N/A\textsuperscript{*} & N/A\textsuperscript{*} & N/A\textsuperscript{*} \\
  GPT-6 Astra & 61.35 & 41.89 & 3.38 & 39.58 & 0.23 & \textbf{79.02} & 55.72 & 0.87 & 48.87 & \textbf{0.00} \\
  \addlinespace[2pt]
  \bottomrule
  
  \end{NiceTabular}%
  }
  \endgroup
\end{table}

\paragraph{TotalText and SROIE.}
GroundingPI is strongest in the SROIE comparison at 72.47 F1mIoU, versus 65.49 for LocateAnything Slow NTP and 53.57 for Astra. TotalText is less favorable: GroundingPI's 49.32 trails Astra (53.55) and Rex-Omni (52.35), despite zero parse error for all three. The remaining gap therefore concerns valid text--region predictions rather than output syntax alone; its precise source requires instance-level error analysis.

\begin{table}[htbp]
  \centering
  \BenchmarkTableFont
  \caption{\textbf{OCR on TotalText and SROIE.} Each dataset reports four loose-match F1 measures and parse-error rate. GroundingDINO is N/A because OCR is unsupported. Kimi-K3 and both DeepSeek variants are N/A because their outputs do not satisfy the evaluation protocol; this does not establish a lack of OCR capability. The starred PaddleOCRv5 and SEED1.5-VL scores use the BBOX results in Table~10 of \citet{rexomni}.}
  \label{tab:app-ocr-totaltext-sroie}
  \begingroup
  \fontsize{8}{9.7}\selectfont
  \setlength{\tabcolsep}{2.5pt}
  \renewcommand{\arraystretch}{1.15}
  \resizebox{\linewidth}{!}{%
  \begin{NiceTabular}{@{}>{\raggedright\arraybackslash}p{177pt}cccccccccc@{}}
  \CodeBefore
    \rowcolor{benchmarktype}{3}
    \rowcolor{benchmarktype}{5}
    \rowcolor{benchmarktype}{7}
    \rowcolor{benchmarkpurple}{24}
    \rowcolor{benchmarktype}{25}
    \rowcolor{benchmarktype}{33}
  \Body
  \toprule
  \textbf{Model} & \multicolumn{5}{c}{\textbf{TotalText}} & \multicolumn{5}{c}{\textbf{SROIE}} \\
  \cmidrule(lr){2-6}\cmidrule(lr){7-11}
   & \BenchHead{F1@.50} & \BenchHead{F1@.75} & \BenchHead{F1@.95} & \BenchHead{F1mIoU} & \BenchHead{Parse\\err.} & \BenchHead{F1@.50} & \BenchHead{F1@.75} & \BenchHead{F1@.95} & \BenchHead{F1mIoU} & \BenchHead{Parse\\err.} \\
  \midrule
  \multicolumn{11}{@{}l}{\hspace{0.4em}\strut\textbf{Closed-set Specialized Detectors}} \\
  PaddleOCRv5\textsuperscript{*}\hspace{0.35em}\citep{PaddleOCRv5} & 40.20 & -- & 0.70 & 25.70 & -- & 77.70 & -- & 5.60 & 58.60 & -- \\
  \addlinespace[2pt]
  \multicolumn{11}{@{}l}{\hspace{0.4em}\strut\textbf{Open-set Specialized Detectors}} \\
  GroundingDINO\hspace{0.35em}\citep{groundingdino} & N/A\textsuperscript{*} & N/A\textsuperscript{*} & N/A\textsuperscript{*} & N/A\textsuperscript{*} & N/A\textsuperscript{*} & N/A\textsuperscript{*} & N/A\textsuperscript{*} & N/A\textsuperscript{*} & N/A\textsuperscript{*} & N/A\textsuperscript{*} \\
  \addlinespace[2pt]
  \multicolumn{11}{@{}l}{\hspace{0.4em}\strut\textbf{Vision-Language Models (<10B)}} \\
  Rex-Omni\hspace{0.35em}\citep{rexomni} & 74.04 & 59.21 & 4.47 & 52.35 & \textbf{0.00} & 72.28 & 47.65 & 2.49 & 48.35 & 3.61 \\
  LocateAnything Fast\hspace{0.35em}\citep{locateanything} & 60.96 & 49.64 & 5.48 & 44.40 & 3.00 & 33.18 & 29.73 & 1.86 & 24.89 & 45.83 \\
  LocateAnything Hybrid\hspace{0.35em}\citep{locateanything} & 62.40 & 51.07 & 5.66 & 45.49 & 1.33 & 40.53 & 35.84 & 2.12 & 30.05 & 34.17 \\
  LocateAnything Slow NTP\hspace{0.35em}\citep{locateanything} & 70.79 & 53.40 & 4.88 & 49.13 & 0.33 & 88.23 & 77.85 & 4.49 & 65.49 & 1.11 \\
  Qwen2.5-VL-7B\hspace{0.35em}\citep{qwen25vl} & 56.62 & 28.47 & 2.54 & 31.13 & \textbf{0.00} & 30.76 & 14.04 & 0.35 & 15.58 & \textbf{0.00} \\
  Qwen3-VL-2B\hspace{0.35em}\citep{qwen3vl} & 60.76 & 29.25 & 6.45 & 33.55 & \textbf{0.00} & 34.86 & 12.51 & 0.15 & 15.93 & \textbf{0.00} \\
  Qwen3-VL-4B\hspace{0.35em}\citep{qwen3vl} & 65.02 & 36.49 & 7.75 & 38.35 & \textbf{0.00} & 68.73 & 41.96 & 0.98 & 40.41 & \textbf{0.00} \\
  Qwen3-VL-8B\hspace{0.35em}\citep{qwen3vl} & 61.40 & 35.44 & 7.40 & 36.70 & \textbf{0.00} & 49.41 & 24.33 & 0.46 & 25.88 & \textbf{0.00} \\
  Qwen3.5-4B\hspace{0.35em}\citep{qwen35} & 54.40 & 28.83 & 4.75 & 31.56 & \textbf{0.00} & 45.96 & 20.70 & 0.30 & 23.21 & 1.11 \\
  Qwen3.5-9B\hspace{0.35em}\citep{qwen35} & 63.78 & 35.22 & 4.51 & 37.26 & 1.00 & 51.52 & 20.59 & 0.81 & 26.74 & 1.94 \\
  RynnBrain1.1\hspace{0.35em}\citep{RynnBrain112B} & 30.38 & 13.04 & 4.29 & 16.90 & 10.67 & 7.23 & 3.24 & 0.06 & 3.59 & 0.83 \\
  SenseNova-Vision\hspace{0.35em}\citep{SenseNovaVision7BMoT} & 18.30 & 11.21 & 1.02 & 11.40 & \textbf{0.00} & 46.97 & 41.90 & 4.23 & 36.26 & \textbf{0.00} \\
  MiMo-VL-7B-SFT\hspace{0.35em}\citep{MimoVL} & 64.24 & 42.31 & 1.79 & 39.76 & \textbf{0.00} & 30.13 & 13.02 & 0.36 & 15.01 & 5.28 \\
  MiMo-VL-7B-RL\hspace{0.35em}\citep{MimoVL} & 65.24 & 40.22 & 2.19 & 39.55 & 0.67 & 39.77 & 15.71 & 0.47 & 19.00 & 0.28 \\
  BAGEL\hspace{0.35em}\citep{BAGEL7BMoT} & 51.19 & 19.02 & 1.02 & 24.25 & \textbf{0.00} & 18.63 & 4.94 & 0.07 & 7.74 & 1.11 \\
  RynnBrain\hspace{0.35em}\citep{dang2026rynnbrain} & 35.58 & 17.41 & 1.57 & 18.90 & \textbf{0.00} & 7.08 & 2.37 & 0.02 & 3.19 & 4.72 \\
  GroundingPI & 72.92 & 53.53 & 5.42 & 49.32 & \textbf{0.00} & \textbf{92.82} & \textbf{84.72} & \textbf{7.83} & \textbf{72.47} & \textbf{0.00} \\
  \addlinespace[2pt]
  \multicolumn{11}{@{}l}{\hspace{0.4em}\strut\textbf{Vision-Language Models (10B--1T)}} \\
  Qwen3.6-27B\hspace{0.35em}\citep{qwen3627b} & 63.52 & 42.49 & 5.38 & 41.21 & 2.33 & 52.05 & 29.37 & 0.63 & 29.37 & \textbf{0.00} \\
  Qwen3-VL-32B\hspace{0.35em}\citep{qwen3vl} & 44.51 & 28.93 & \textbf{8.27} & 28.84 & 18.67 & 10.12 & 6.31 & 0.20 & 6.02 & 73.89 \\
  Qwen3.8-27B\hspace{0.35em}\citep{qwen38} & 65.73 & 42.57 & 3.99 & 41.57 & \textbf{0.00} & 61.27 & 38.97 & 1.70 & 37.24 & 0.28 \\
  Qwen3.5-35B-A3B\hspace{0.35em}\citep{qwen35} & 59.86 & 34.72 & 5.31 & 35.95 & 1.67 & 65.83 & 36.31 & 0.92 & 36.64 & 0.28 \\
  DeepSeek-VL2-Small-16B\hspace{0.35em}\citep{Deepseekvl2} & N/A\textsuperscript{*} & N/A\textsuperscript{*} & N/A\textsuperscript{*} & N/A\textsuperscript{*} & N/A\textsuperscript{*} & N/A\textsuperscript{*} & N/A\textsuperscript{*} & N/A\textsuperscript{*} & N/A\textsuperscript{*} & N/A\textsuperscript{*} \\
  DeepSeek-VL2-27B\hspace{0.35em}\citep{Deepseekvl2} & N/A\textsuperscript{*} & N/A\textsuperscript{*} & N/A\textsuperscript{*} & N/A\textsuperscript{*} & N/A\textsuperscript{*} & N/A\textsuperscript{*} & N/A\textsuperscript{*} & N/A\textsuperscript{*} & N/A\textsuperscript{*} & N/A\textsuperscript{*} \\
  SEED1.5-VL\textsuperscript{*}\hspace{0.35em}\citep{SEED15VL} & 35.00 & -- & 0.30 & 19.50 & -- & 51.90 & -- & 0.80 & 28.10 & -- \\
  \addlinespace[2pt]
  \multicolumn{11}{@{}l}{\hspace{0.4em}\strut\textbf{Vision-Language Models (>1T)}} \\
  Qwen3.7-Max\hspace{0.35em}\citep{qwen37} & 72.58 & 52.18 & 7.92 & 48.64 & \textbf{0.00} & 57.72 & 44.25 & 1.50 & 38.54 & \textbf{0.00} \\
  Kimi-K2.6 & 67.66\textsuperscript{\textdagger} & 40.90\textsuperscript{\textdagger} & 5.52\textsuperscript{\textdagger} & 40.99\textsuperscript{\textdagger} & \textbf{0.00}\textsuperscript{\textdagger} & 76.94\textsuperscript{\textdagger} & 49.43\textsuperscript{\textdagger} & 1.92\textsuperscript{\textdagger} & 46.83\textsuperscript{\textdagger} & \textbf{0.00}\textsuperscript{\textdagger} \\
  Kimi-K3\hspace{0.35em}\citep{KimiK3} & N/A\textsuperscript{*} & N/A\textsuperscript{*} & N/A\textsuperscript{*} & N/A\textsuperscript{*} & N/A\textsuperscript{*} & N/A\textsuperscript{*} & N/A\textsuperscript{*} & N/A\textsuperscript{*} & N/A\textsuperscript{*} & N/A\textsuperscript{*} \\
  GPT-6 Astra & \textbf{74.61} & \textbf{62.13} & 4.96 & \textbf{53.55} & \textbf{0.00} & 70.81 & 62.54 & 5.27 & 53.57 & \textbf{0.00} \\
  \addlinespace[2pt]
  \bottomrule
  
  \end{NiceTabular}%
  }
  \endgroup
\end{table}

\subsection{GUI Grounding}
\label{app:gui_grounding}

\paragraph{ScreenSpot-Pro.}
GroundingPI attains 65.78 overall accuracy with zero parse error, improving over Qwen3-VL-4B (56.74), LocateAnything Hybrid (57.05), and the reported GUI-Owl reference (58.00). Its CAD icon accuracy rises to 56.25 from Qwen3-VL-4B's 26.56, while their CAD text scores tie at 57.87. Astra's 93.17 overall remains substantially higher. Broad gains thus coexist with a sizable frontier-model gap on professional interfaces.

\begin{table}[htbp]
  \centering
  \BenchmarkTableFont
  \caption{\textbf{ScreenSpot-Pro.} Action accuracy is broken down by domain and target type, followed by overall action accuracy and parse-error rate. GroundingDINO is N/A because GUI grounding is unsupported. BAGEL is N/A because its outputs do not satisfy the unified protocol. The starred JEDI, UI-R1, and UI-TARS scores are from Table~8 of \citet{rexomni}; GUI-Owl-32B scores are from Table~3 of \citet{locateanything}.}
  \label{tab:app-gui-pro}
  \begingroup
  \fontsize{7}{8.7}\selectfont
  \setlength{\tabcolsep}{2pt}
  \renewcommand{\arraystretch}{1.15}
  \resizebox{\linewidth}{!}{%
  \begin{NiceTabular}{@{}>{\raggedright\arraybackslash}p{161pt}cccccccccccccc@{}}
  \CodeBefore
    \rowcolor{benchmarktype}{3}
    \rowcolor{benchmarktype}{5}
    \rowcolor{benchmarkpurple}{25}
    \rowcolor{benchmarktype}{26}
    \rowcolor{benchmarktype}{34}
  \Body
  \toprule
  \textbf{Model} & \multicolumn{2}{c}{\textbf{Dev}} & \multicolumn{2}{c}{\textbf{Creative}} & \multicolumn{2}{c}{\textbf{CAD}} & \multicolumn{2}{c}{\textbf{Sci}} & \multicolumn{2}{c}{\textbf{Office}} & \multicolumn{2}{c}{\textbf{OS}} & \multicolumn{2}{c}{\textbf{Overall}} \\
  \cmidrule(lr){2-3}\cmidrule(lr){4-5}\cmidrule(lr){6-7}\cmidrule(lr){8-9}\cmidrule(lr){10-11}\cmidrule(lr){12-13}\cmidrule(lr){14-15}
   & \BenchHead{Text} & \BenchHead{Icon} & \BenchHead{Text} & \BenchHead{Icon} & \BenchHead{Text} & \BenchHead{Icon} & \BenchHead{Text} & \BenchHead{Icon} & \BenchHead{Text} & \BenchHead{Icon} & \BenchHead{Text} & \BenchHead{Icon} & \BenchHead{Action\\acc.} & \BenchHead{Parse\\err.} \\
  \midrule
  \multicolumn{15}{@{}l}{\hspace{0.4em}\strut\textbf{Open-set Specialized Detectors}} \\
  GroundingDINO\hspace{0.35em}\citep{groundingdino} & N/A\textsuperscript{*} & N/A\textsuperscript{*} & N/A\textsuperscript{*} & N/A\textsuperscript{*} & N/A\textsuperscript{*} & N/A\textsuperscript{*} & N/A\textsuperscript{*} & N/A\textsuperscript{*} & N/A\textsuperscript{*} & N/A\textsuperscript{*} & N/A\textsuperscript{*} & N/A\textsuperscript{*} & N/A\textsuperscript{*} & N/A\textsuperscript{*} \\
  \addlinespace[2pt]
  \multicolumn{15}{@{}l}{\hspace{0.4em}\strut\textbf{Vision-Language Models (<10B)}} \\
  Rex-Omni\hspace{0.35em}\citep{rexomni} & 61.04 & 9.66 & 53.03 & 12.59 & 23.35 & 9.38 & 57.64 & 26.36 & 65.54 & 24.53 & 42.06 & 13.48 & 36.75 & -- \\
  LocateAnything Fast\hspace{0.35em}\citep{locateanything} & 68.18 & 44.83 & 59.60 & 32.87 & 58.38 & 35.94 & 71.53 & 53.64 & 75.14 & 56.60 & 51.40 & 37.08 & 56.04 & -- \\
  LocateAnything Hybrid\hspace{0.35em}\citep{locateanything} & 70.13 & 44.83 & 59.60 & 36.36 & 58.88 & 37.50 & 71.53 & 51.82 & 72.88 & 58.49 & 58.88 & 40.45 & 57.05 & -- \\
  LocateAnything Slow NTP\hspace{0.35em}\citep{locateanything} & 70.78 & 48.28 & 61.11 & 39.86 & 60.41 & 39.06 & 75.69 & 51.82 & 74.58 & 54.72 & 57.94 & 40.45 & 58.57 & -- \\
  Qwen2.5-VL-7B\hspace{0.35em}\citep{qwen25vl} & 40.91 & 3.45 & 36.36 & 9.09 & 18.27 & 3.12 & 47.22 & 6.36 & 56.50 & 13.21 & 34.58 & 11.24 & 26.57 & -- \\
  Qwen3-VL-2B\hspace{0.35em}\citep{qwen3vl} & 47.40 & 7.59 & 29.29 & 8.39 & 23.86 & 7.81 & 38.89 & 18.18 & 49.15 & 22.64 & 37.38 & 20.22 & 27.77 & -- \\
  Qwen3-VL-4B\hspace{0.35em}\citep{qwen3vl} & 72.73 & 31.72 & 66.67 & 25.17 & 57.87 & 26.56 & 77.08 & 35.45 & 84.75 & 47.17 & 78.50 & 34.83 & 56.74 & -- \\
  Qwen3-VL-8B\hspace{0.35em}\citep{qwen3vl} & 75.32 & 30.34 & 71.21 & 20.98 & 60.91 & 26.56 & 76.39 & 40.00 & 83.62 & 37.74 & 73.83 & 33.71 & 56.86 & -- \\
  Qwen3.5-4B\hspace{0.35em}\citep{qwen35} & 79.22 & 42.07 & 67.68 & 25.17 & 65.99 & 35.94 & 74.31 & 34.55 & 82.49 & 50.94 & 69.16 & 43.82 & 59.27 & -- \\
  Qwen3.5-9B\hspace{0.35em}\citep{qwen35} & 70.78 & 37.93 & 64.14 & 30.77 & 40.10 & 28.12 & 72.92 & 33.64 & 74.58 & 49.06 & 69.16 & 38.20 & 53.13 & -- \\
  RynnBrain1.1\hspace{0.35em}\citep{RynnBrain112B} & 52.60 & 15.86 & 45.96 & 12.59 & 21.32 & 10.94 & 52.08 & 21.82 & 60.45 & 20.75 & 47.66 & 20.22 & 34.66 & 0.38 \\
  SenseNova-Vision\hspace{0.35em}\citep{SenseNovaVision7BMoT} & N/A\textsuperscript{\textdagger} & N/A\textsuperscript{\textdagger} & N/A\textsuperscript{\textdagger} & N/A\textsuperscript{\textdagger} & N/A\textsuperscript{\textdagger} & N/A\textsuperscript{\textdagger} & N/A\textsuperscript{\textdagger} & N/A\textsuperscript{\textdagger} & N/A\textsuperscript{\textdagger} & N/A\textsuperscript{\textdagger} & N/A\textsuperscript{\textdagger} & N/A\textsuperscript{\textdagger} & N/A\textsuperscript{\textdagger} & N/A\textsuperscript{\textdagger} \\
  MiMo-VL-7B-SFT\hspace{0.35em}\citep{MimoVL} & 28.57 & 4.83 & 31.82 & 4.20 & 19.29 & 7.81 & 52.78 & 12.73 & 40.11 & 20.75 & 24.30 & 6.74 & 23.21 & 6.83 \\
  MiMo-VL-7B-RL\hspace{0.35em}\citep{MimoVL} & 27.92 & 2.07 & 38.38 & 4.20 & 27.41 & 9.38 & 57.64 & 17.27 & 53.11 & 24.53 & 29.91 & 7.87 & 27.58 & 9.30 \\
  BAGEL\hspace{0.35em}\citep{BAGEL7BMoT} & N/A\textsuperscript{*} & N/A\textsuperscript{*} & N/A\textsuperscript{*} & N/A\textsuperscript{*} & N/A\textsuperscript{*} & N/A\textsuperscript{*} & N/A\textsuperscript{*} & N/A\textsuperscript{*} & N/A\textsuperscript{*} & N/A\textsuperscript{*} & N/A\textsuperscript{*} & N/A\textsuperscript{*} & N/A\textsuperscript{*} & N/A\textsuperscript{*} \\
  RynnBrain\hspace{0.35em}\citep{dang2026rynnbrain} & 44.81 & 7.59 & 35.35 & 5.59 & 15.23 & 7.81 & 40.28 & 13.64 & 54.80 & 18.87 & 42.06 & 11.24 & 27.07 & 2.47 \\
  JEDI\textsuperscript{*}\hspace{0.35em}\citep{JEDI3B} & 61.00 & 13.80 & 53.50 & 8.40 & 27.40 & 9.40 & 54.20 & 18.20 & 64.40 & 32.10 & 38.30 & 9.00 & 36.10 & -- \\
  UI-R1\textsuperscript{*}\hspace{0.35em}\citep{UIR13B} & 22.70 & 4.10 & 27.30 & 3.50 & 11.20 & 6.30 & 42.40 & 11.80 & 32.20 & 11.30 & 13.10 & 4.50 & 17.80 & -- \\
  UI-TARS\textsuperscript{*}\hspace{0.35em}\citep{UITARS2B} & 47.40 & 4.10 & 42.90 & 6.30 & 17.80 & 4.70 & 56.90 & 17.30 & 50.30 & 17.00 & 21.50 & 5.60 & 27.70 & -- \\
  GroundingPI & 74.03 & 58.62 & 70.20 & 56.64 & 57.87 & 56.25 & 85.42 & 64.55 & 77.40 & 62.26 & 56.07 & 52.81 & 65.78 & \textbf{0.00} \\
  \addlinespace[2pt]
  \multicolumn{15}{@{}l}{\hspace{0.4em}\strut\textbf{Vision-Language Models (10B--1T)}} \\
  Qwen3.6-27B\hspace{0.35em}\citep{qwen3627b} & 87.01 & 53.79 & 76.77 & 44.76 & 75.13 & 46.88 & 81.25 & 48.18 & 87.57 & 64.15 & 79.44 & 57.30 & 69.64 & -- \\
  Qwen3-VL-32B\hspace{0.35em}\citep{qwen3vl} & 73.38 & 20.00 & 76.26 & 24.48 & 59.39 & 34.38 & 86.11 & 30.91 & 83.62 & 41.51 & 60.75 & 22.47 & 55.66 & 0.13 \\
  Qwen3.8-27B\hspace{0.35em}\citep{qwen38} & 85.71 & 64.83 & 45.45 & 41.96 & 50.76 & 14.06 & 67.36 & 36.36 & 87.57 & 56.60 & 74.77 & 47.19 & 58.76 & 0.76 \\
  Qwen3.5-35B-A3B\hspace{0.35em}\citep{qwen35} & 82.47 & 44.83 & 56.57 & 27.97 & 54.82 & 14.06 & 53.47 & 29.09 & 61.58 & 41.51 & 53.27 & 20.22 & 49.08 & 0.38 \\
  DeepSeek-VL2-Small-16B\hspace{0.35em}\citep{Deepseekvl2} & 0.00\textsuperscript{\textdagger} & 0.00\textsuperscript{\textdagger} & 0.00\textsuperscript{\textdagger} & 0.00\textsuperscript{\textdagger} & 0.00\textsuperscript{\textdagger} & 0.00\textsuperscript{\textdagger} & 0.00\textsuperscript{\textdagger} & 0.00\textsuperscript{\textdagger} & 0.00\textsuperscript{\textdagger} & 1.89\textsuperscript{\textdagger} & 0.93\textsuperscript{\textdagger} & 0.00\textsuperscript{\textdagger} & 0.13\textsuperscript{\textdagger} & 25.62\textsuperscript{\textdagger} \\
  DeepSeek-VL2-27B\hspace{0.35em}\citep{Deepseekvl2} & 0.00\textsuperscript{\textdagger} & 0.00\textsuperscript{\textdagger} & 0.00\textsuperscript{\textdagger} & 0.00\textsuperscript{\textdagger} & 0.51\textsuperscript{\textdagger} & 0.00\textsuperscript{\textdagger} & 0.00\textsuperscript{\textdagger} & 0.00\textsuperscript{\textdagger} & 0.00\textsuperscript{\textdagger} & 0.00\textsuperscript{\textdagger} & 0.00\textsuperscript{\textdagger} & 0.00\textsuperscript{\textdagger} & 0.06\textsuperscript{\textdagger} & 34.28\textsuperscript{\textdagger} \\
  GUI-Owl\textsuperscript{*}\hspace{0.35em}\citep{GUIOwl32B} & 84.40 & 39.30 & 65.20 & 18.20 & 62.40 & 28.10 & 82.60 & 39.10 & 81.40 & 39.60 & 70.10 & 36.00 & 58.00 & -- \\
  \addlinespace[2pt]
  \multicolumn{15}{@{}l}{\hspace{0.4em}\strut\textbf{Vision-Language Models (>1T)}} \\
  Qwen3.7-Max\hspace{0.35em}\citep{qwen37} & 79.22 & 37.50 & 62.63 & 34.27 & 46.19 & 31.25 & 77.08 & 31.82 & 78.53 & 47.17 & 61.68 & 38.20 & 55.06 & 10.57 \\
  Kimi-K2.6 & -- & -- & -- & -- & -- & -- & -- & -- & -- & -- & -- & -- & 6.07\textsuperscript{\textdagger} & 8.22\textsuperscript{\textdagger} \\
  Kimi-K3\hspace{0.35em}\citep{KimiK3} & 29.87\textsuperscript{\textdagger} & 20.00\textsuperscript{\textdagger} & 43.94\textsuperscript{\textdagger} & 25.17\textsuperscript{\textdagger} & 25.38\textsuperscript{\textdagger} & 3.12\textsuperscript{\textdagger} & 29.86\textsuperscript{\textdagger} & 12.73\textsuperscript{\textdagger} & 27.12\textsuperscript{\textdagger} & 13.21\textsuperscript{\textdagger} & 28.97\textsuperscript{\textdagger} & 8.99\textsuperscript{\textdagger} & 25.36\textsuperscript{\textdagger} & 3.73\textsuperscript{\textdagger} \\
  GPT-6 Astra & \textbf{96.10} & \textbf{84.83} & \textbf{95.45} & \textbf{89.51} & \textbf{95.43} & \textbf{85.94} & \textbf{94.44} & \textbf{87.27} & \textbf{98.87} & \textbf{96.23} & \textbf{97.20} & \textbf{89.89} & \textbf{93.17} & \textbf{0.00} \\
  \addlinespace[2pt]
  \bottomrule
  
  \end{NiceTabular}%
  }
  \endgroup
\end{table}

\paragraph{ScreenSpot-V2 and OSWorld-G.}
GroundingPI obtains 96.15 and 74.82, compared with Qwen3-VL-4B's 92.30 and 56.91. On ScreenSpot-V2, desktop and web icon accuracy reaches 96.43 and 97.54, versus 87.14 and 86.21 for Qwen3-VL-4B. Astra remains stronger overall on both benchmarks (97.88/86.70). Near-ceiling ScreenSpot-V2 performance therefore does not imply that broader GUI grounding is solved.

\begin{table}[htbp]
  \centering
  \BenchmarkTableFont
  \caption{\textbf{ScreenSpot-V2 and OSWorld-G.} ScreenSpot-V2 includes text/icon results for mobile, desktop, and web environments, overall action accuracy, and parse-error rate; OSWorld-G reports exact accuracy and parse-error rate. GroundingDINO is N/A because GUI grounding is unsupported. BAGEL is N/A because its outputs do not satisfy the unified protocol. The starred ScreenSpot-V2 scores are from Table~8 of \citet{rexomni}.}
  \label{tab:app-gui-v2-osworld}
  \begingroup
  \fontsize{8}{9.7}\selectfont
  \setlength{\tabcolsep}{2.5pt}
  \renewcommand{\arraystretch}{1.15}
  \resizebox{\linewidth}{!}{%
  \begin{NiceTabular}{@{}>{\raggedright\arraybackslash}p{177pt}cccccccccc@{}}
  \CodeBefore
    \rowcolor{benchmarktype}{3}
    \rowcolor{benchmarktype}{5}
    \rowcolor{benchmarkpurple}{25}
    \rowcolor{benchmarktype}{26}
    \rowcolor{benchmarktype}{33}
  \Body
  \toprule
  \textbf{Model} & \multicolumn{8}{c}{\textbf{ScreenSpot-V2}} & \multicolumn{2}{c}{\textbf{OSWorld-G}} \\
  \cmidrule(lr){2-9}\cmidrule(lr){10-11}
   & \BenchHead{Mobile\\text} & \BenchHead{Mobile\\icon} & \BenchHead{Desktop\\text} & \BenchHead{Desktop\\icon} & \BenchHead{Web\\text} & \BenchHead{Web\\icon} & \BenchHead{Action\\acc.} & \BenchHead{Parse\\err.} & \BenchHead{Exact\\acc.} & \BenchHead{Parse\\err.} \\
  \midrule
  \multicolumn{11}{@{}l}{\hspace{0.4em}\strut\textbf{Open-set Specialized Detectors}} \\
  GroundingDINO\hspace{0.35em}\citep{groundingdino} & N/A\textsuperscript{*} & N/A\textsuperscript{*} & N/A\textsuperscript{*} & N/A\textsuperscript{*} & N/A\textsuperscript{*} & N/A\textsuperscript{*} & N/A\textsuperscript{*} & N/A\textsuperscript{*} & N/A\textsuperscript{*} & N/A\textsuperscript{*} \\
  \addlinespace[2pt]
  \multicolumn{11}{@{}l}{\hspace{0.4em}\strut\textbf{Vision-Language Models (<10B)}} \\
  Rex-Omni\hspace{0.35em}\citep{rexomni} & 96.90 & 82.46 & 97.94 & 80.71 & 89.74 & 76.35 & 88.29 & -- & 46.10 & -- \\
  LocateAnything Fast\hspace{0.35em}\citep{locateanything} & 94.48 & 81.04 & 93.81 & 86.43 & 88.89 & 83.25 & 88.44 & -- & 59.93 & -- \\
  LocateAnything Hybrid\hspace{0.35em}\citep{locateanything} & 96.21 & 83.89 & 92.78 & 89.29 & 88.89 & 86.21 & 89.94 & -- & 60.46 & -- \\
  LocateAnything Slow NTP\hspace{0.35em}\citep{locateanything} & 95.86 & 83.41 & 93.30 & 89.29 & 90.60 & 85.22 & 90.02 & -- & 61.17 & -- \\
  Qwen2.5-VL-7B\hspace{0.35em}\citep{qwen25vl} & 98.28 & 82.94 & 91.75 & 67.86 & 92.74 & 79.31 & 87.34 & -- & 34.22 & -- \\
  Qwen3-VL-2B\hspace{0.35em}\citep{qwen3vl} & 93.10 & 70.62 & 79.90 & 61.43 & 79.91 & 62.07 & 76.49 & -- & 33.51 & -- \\
  Qwen3-VL-4B\hspace{0.35em}\citep{qwen3vl} & 97.93 & 87.20 & 96.91 & 87.14 & 94.44 & 86.21 & 92.30 & -- & 56.91 & -- \\
  Qwen3-VL-8B\hspace{0.35em}\citep{qwen3vl} & 98.62 & 89.57 & 98.45 & 87.86 & 94.87 & 89.16 & 93.71 & -- & 56.91 & -- \\
  Qwen3.5-4B\hspace{0.35em}\citep{qwen35} & \textbf{98.97} & 88.15 & 97.94 & 89.29 & 95.73 & 90.15 & 93.95 & -- & 54.61 & -- \\
  Qwen3.5-9B\hspace{0.35em}\citep{qwen35} & 96.55 & 88.15 & 97.94 & 93.57 & 87.61 & 78.82 & 90.57 & -- & 60.99 & -- \\
  RynnBrain1.1\hspace{0.35em}\citep{RynnBrain112B} & 82.07 & 71.09 & 75.26 & 52.86 & 73.08 & 57.64 & 70.44 & 4.25 & 33.33 & 4.96 \\
  SenseNova-Vision\hspace{0.35em}\citep{SenseNovaVision7BMoT} & N/A\textsuperscript{\textdagger} & N/A\textsuperscript{\textdagger} & N/A\textsuperscript{\textdagger} & N/A\textsuperscript{\textdagger} & N/A\textsuperscript{\textdagger} & N/A\textsuperscript{\textdagger} & N/A\textsuperscript{\textdagger} & N/A\textsuperscript{\textdagger} & N/A\textsuperscript{\textdagger} & N/A\textsuperscript{\textdagger} \\
  MiMo-VL-7B-SFT\hspace{0.35em}\citep{MimoVL} & 91.03 & 68.25 & 86.60 & 50.00 & 63.68 & 56.65 & 71.54 & 8.18 & 34.04 & 6.38 \\
  MiMo-VL-7B-RL\hspace{0.35em}\citep{MimoVL} & 90.69 & 67.30 & 85.57 & 49.29 & 78.21 & 62.56 & 74.69 & 5.66 & 37.06 & 2.30 \\
  BAGEL\hspace{0.35em}\citep{BAGEL7BMoT} & N/A\textsuperscript{*} & N/A\textsuperscript{*} & N/A\textsuperscript{*} & N/A\textsuperscript{*} & N/A\textsuperscript{*} & N/A\textsuperscript{*} & N/A\textsuperscript{*} & N/A\textsuperscript{*} & N/A\textsuperscript{*} & N/A\textsuperscript{*} \\
  RynnBrain\hspace{0.35em}\citep{dang2026rynnbrain} & 89.31 & 69.19 & 80.93 & 56.43 & 79.91 & 60.10 & 74.69 & 5.27 & 35.99 & 1.77 \\
  JEDI\textsuperscript{*}\hspace{0.35em}\citep{JEDI3B} & 96.60 & 81.50 & 96.90 & 78.60 & 88.50 & 83.70 & 88.60 & -- & -- & -- \\
  UI-R1\textsuperscript{*}\hspace{0.35em}\citep{UIR13B} & 84.30 & \textbf{96.20} & 75.40 & 89.20 & 63.60 & 92.30 & 85.40 & -- & -- & -- \\
  UI-TARS\textsuperscript{*}\hspace{0.35em}\citep{UITARS2B} & 95.20 & 79.10 & 90.70 & 68.60 & 87.20 & 78.30 & 84.70 & -- & -- & -- \\
  GroundingPI & 97.93 & 91.00 & 97.42 & 96.43 & 96.15 & \textbf{97.54} & 96.15 & \textbf{0.00} & 74.82 & \textbf{0.00} \\
  \addlinespace[2pt]
  \multicolumn{11}{@{}l}{\hspace{0.4em}\strut\textbf{Vision-Language Models (10B--1T)}} \\
  Qwen3.6-27B\hspace{0.35em}\citep{qwen3627b} & \textbf{98.97} & 92.42 & 98.45 & 92.86 & 96.58 & 89.16 & 95.13 & -- & 68.44 & -- \\
  Qwen3-VL-32B\hspace{0.35em}\citep{qwen3vl} & 98.62 & 90.05 & 98.45 & 86.43 & 96.58 & 91.63 & 94.34 & 0.08 & 60.28 & 0.71 \\
  Qwen3.8-27B\hspace{0.35em}\citep{qwen38} & 97.93 & 90.52 & 98.45 & 94.29 & 94.87 & 89.66 & 94.50 & 0.24 & 63.48 & \textbf{0.00} \\
  Qwen3.5-35B-A3B\hspace{0.35em}\citep{qwen35} & 97.59 & 89.10 & 96.39 & 92.14 & 94.02 & 86.70 & 93.00 & 1.42 & 62.23 & 0.18 \\
  DeepSeek-VL2-Small-16B\hspace{0.35em}\citep{Deepseekvl2} & 2.07\textsuperscript{\textdagger} & 0.00\textsuperscript{\textdagger} & 4.64\textsuperscript{\textdagger} & 1.43\textsuperscript{\textdagger} & 0.85\textsuperscript{\textdagger} & 0.49\textsuperscript{\textdagger} & 1.57\textsuperscript{\textdagger} & 41.82\textsuperscript{\textdagger} & 0.89\textsuperscript{\textdagger} & 11.52\textsuperscript{\textdagger} \\
  DeepSeek-VL2-27B\hspace{0.35em}\citep{Deepseekvl2} & 6.90\textsuperscript{\textdagger} & 1.42\textsuperscript{\textdagger} & 2.58\textsuperscript{\textdagger} & 0.00\textsuperscript{\textdagger} & 2.56\textsuperscript{\textdagger} & 1.48\textsuperscript{\textdagger} & 2.91\textsuperscript{\textdagger} & 25.63\textsuperscript{\textdagger} & 0.00\textsuperscript{\textdagger} & 32.62\textsuperscript{\textdagger} \\
  \addlinespace[2pt]
  \multicolumn{11}{@{}l}{\hspace{0.4em}\strut\textbf{Vision-Language Models (>1T)}} \\
  Qwen3.7-Max\hspace{0.35em}\citep{qwen37} & 89.31 & 82.94 & 71.13 & 76.43 & 81.62 & 79.80 & 81.13 & 13.21 & 49.29 & 24.47 \\
  Kimi-K2.6 & -- & -- & -- & -- & -- & -- & 52.36\textsuperscript{\textdagger} & 7.15\textsuperscript{\textdagger} & 10.11\textsuperscript{\textdagger} & 7.98\textsuperscript{\textdagger} \\
  Kimi-K3\hspace{0.35em}\citep{KimiK3} & 95.86\textsuperscript{\textdagger} & 88.63\textsuperscript{\textdagger} & 97.42\textsuperscript{\textdagger} & 89.29\textsuperscript{\textdagger} & 63.68\textsuperscript{\textdagger} & 61.08\textsuperscript{\textdagger} & 82.70\textsuperscript{\textdagger} & 1.57\textsuperscript{\textdagger} & 68.26\textsuperscript{\textdagger} & 4.96\textsuperscript{\textdagger} \\
  GPT-6 Astra & \textbf{98.97} & 95.26 & \textbf{98.97} & \textbf{98.57} & \textbf{98.29} & 97.04 & \textbf{97.88} & 0.24 & \textbf{86.70} & 0.18 \\
  \addlinespace[2pt]
  \bottomrule
  
  \end{NiceTabular}%
  }
  \endgroup
\end{table}

\subsection{Layout Grounding}
\label{app:layout_grounding}

Document layout analysis is treated as detection of labeled document regions, using the grounding evaluation convention of \citep{rexomni}.

\paragraph{DocLayNet.}
GroundingPI reaches 85.08 F1mIoU, improving over DocLayout-YOLO's reported 81.10 and Astra's 77.54, while SenseNova-Vision leads slightly at 85.53. At IoU 0.95, GroundingPI achieves 43.47, versus 34.93 for Astra and 49.33 for SenseNova-Vision. The result supports broad document-region grounding, with the strongest specialist comparison still exposing room for tighter boundaries.

\begin{table}[htbp]
  \centering
  \BenchmarkTableFont
  \caption{\textbf{DocLayNet.} Complete box-grounding metrics. GroundingDINO lacks a compatible document-region interface; it, Kimi-K3, and both DeepSeek variants are N/A. Starred MiMo and BAGEL scores retain observations under suspected output-protocol incompatibility and are not formal capability measurements. The starred DocLayout-YOLO and SEED1.5-VL scores are from Table~9 of \citet{rexomni}.}
  \label{tab:app-doclaynet}
  \begingroup
  \fontsize{8}{9.7}\selectfont
  \setlength{\tabcolsep}{3pt}
  \renewcommand{\arraystretch}{1.15}
  \resizebox{\linewidth}{!}{%
  \begin{NiceTabular}{@{}>{\raggedright\arraybackslash}p{177pt}ccccccccc@{}}
  \CodeBefore
    \rowcolor{benchmarktype}{3}
    \rowcolor{benchmarktype}{5}
    \rowcolor{benchmarktype}{7}
    \rowcolor{benchmarkpurple}{24}
    \rowcolor{benchmarktype}{25}
    \rowcolor{benchmarktype}{33}
  \Body
  \toprule
  \textbf{Model} & \multicolumn{3}{c}{\textbf{IoU 0.50}} & \multicolumn{3}{c}{\textbf{IoU 0.95}} & \multicolumn{3}{c}{\textbf{mIoU}} \\
  \cmidrule(lr){2-4}\cmidrule(lr){5-7}\cmidrule(lr){8-10}
   & \BenchHead{R} & \BenchHead{P} & \BenchHead{F1} & \BenchHead{R} & \BenchHead{P} & \BenchHead{F1} & \BenchHead{R} & \BenchHead{P} & \BenchHead{F1} \\
  \midrule
  \multicolumn{10}{@{}l}{\hspace{0.4em}\strut\textbf{Closed-set Specialized Detectors}} \\
  DocLayout-YOLO\textsuperscript{*}\hspace{0.35em}\citep{DocLayoutYOLO} & -- & -- & 91.20 & -- & -- & \textbf{52.10} & -- & -- & 81.10 \\
  \addlinespace[2pt]
  \multicolumn{10}{@{}l}{\hspace{0.4em}\strut\textbf{Open-set Specialized Detectors}} \\
  GroundingDINO\hspace{0.35em}\citep{groundingdino} & N/A\textsuperscript{*} & N/A\textsuperscript{*} & N/A\textsuperscript{*} & N/A\textsuperscript{*} & N/A\textsuperscript{*} & N/A\textsuperscript{*} & N/A\textsuperscript{*} & N/A\textsuperscript{*} & N/A\textsuperscript{*} \\
  \addlinespace[2pt]
  \multicolumn{10}{@{}l}{\hspace{0.4em}\strut\textbf{Vision-Language Models (<10B)}} \\
  Rex-Omni\hspace{0.35em}\citep{rexomni} & 83.48 & 88.85 & 86.08 & 27.13 & 27.95 & 27.53 & 66.25 & 69.98 & 68.06 \\
  LocateAnything Fast\hspace{0.35em}\citep{locateanything} & 55.10 & 59.26 & 57.11 & 25.03 & 26.55 & 25.77 & 47.73 & 51.12 & 49.37 \\
  LocateAnything Hybrid\hspace{0.35em}\citep{locateanything} & 87.95 & 90.90 & 89.40 & 38.72 & 39.82 & 39.26 & 76.14 & 78.57 & 77.34 \\
  LocateAnything Slow NTP\hspace{0.35em}\citep{locateanything} & 91.55 & 94.58 & 93.04 & 39.56 & 40.64 & 40.09 & 78.98 & 81.45 & 80.19 \\
  Qwen2.5-VL-7B\hspace{0.35em}\citep{qwen25vl} & 30.23 & 29.07 & 29.64 & 2.52 & 2.92 & 2.70 & 16.41 & 16.72 & 16.55 \\
  Qwen3-VL-2B\hspace{0.35em}\citep{qwen3vl} & 40.38 & 35.70 & 37.89 & 2.80 & 2.83 & 2.81 & 21.10 & 19.19 & 20.09 \\
  Qwen3-VL-4B\hspace{0.35em}\citep{qwen3vl} & 64.86 & 69.50 & 67.10 & 8.40 & 8.68 & 8.54 & 39.90 & 41.76 & 40.81 \\
  Qwen3-VL-8B\hspace{0.35em}\citep{qwen3vl} & 60.09 & 57.80 & 58.92 & 7.13 & 7.18 & 7.15 & 37.76 & 36.58 & 37.16 \\
  Qwen3.5-4B\hspace{0.35em}\citep{qwen35} & 59.37 & 47.02 & 52.48 & 5.04 & 4.47 & 4.74 & 35.03 & 28.51 & 31.43 \\
  Qwen3.5-9B\hspace{0.35em}\citep{qwen35} & 63.44 & 49.12 & 55.37 & 6.96 & 6.09 & 6.50 & 39.12 & 31.10 & 34.65 \\
  RynnBrain1.1\hspace{0.35em}\citep{RynnBrain112B} & 5.50 & 24.01 & 8.96 & 1.52 & 6.68 & 2.48 & 3.70 & 16.21 & 6.02 \\
  SenseNova-Vision\hspace{0.35em}\citep{SenseNovaVision7BMoT} & 95.45 & 95.78 & 95.61 & \textbf{49.23} & \textbf{49.43} & 49.33 & \textbf{85.38} & \textbf{85.67} & \textbf{85.53} \\
  MiMo-VL-7B-SFT\hspace{0.35em}\citep{MimoVL} & 3.37\textsuperscript{*} & 5.01\textsuperscript{*} & 4.03\textsuperscript{*} & 0.17\textsuperscript{*} & 0.24\textsuperscript{*} & 0.20\textsuperscript{*} & 1.21\textsuperscript{*} & 1.99\textsuperscript{*} & 1.50\textsuperscript{*} \\
  MiMo-VL-7B-RL\hspace{0.35em}\citep{MimoVL} & 5.43\textsuperscript{*} & 8.01\textsuperscript{*} & 6.47\textsuperscript{*} & 0.11\textsuperscript{*} & 0.19\textsuperscript{*} & 0.14\textsuperscript{*} & 1.98\textsuperscript{*} & 3.15\textsuperscript{*} & 2.43\textsuperscript{*} \\
  BAGEL\hspace{0.35em}\citep{BAGEL7BMoT} & 10.83\textsuperscript{*} & 10.93\textsuperscript{*} & 10.88\textsuperscript{*} & 0.60\textsuperscript{*} & 0.69\textsuperscript{*} & 0.64\textsuperscript{*} & 4.34\textsuperscript{*} & 4.64\textsuperscript{*} & 4.48\textsuperscript{*} \\
  RynnBrain\hspace{0.35em}\citep{dang2026rynnbrain} & 7.16 & 17.74 & 10.20 & 0.98 & 1.88 & 1.29 & 4.01 & 9.40 & 5.62 \\
  GroundingPI & 96.07 & \textbf{96.04} & 96.05 & 43.47 & 43.47 & 43.47 & 85.10 & 85.06 & 85.08 \\
  \addlinespace[2pt]
  \multicolumn{10}{@{}l}{\hspace{0.4em}\strut\textbf{Vision-Language Models (10B--1T)}} \\
  Qwen3.6-27B\hspace{0.35em}\citep{qwen3627b} & 73.85 & 72.14 & 72.98 & 10.44 & 10.10 & 10.27 & 47.15 & 46.10 & 46.62 \\
  Qwen3-VL-32B\hspace{0.35em}\citep{qwen3vl} & 40.07 & 36.22 & 38.05 & 4.08 & 4.22 & 4.15 & 21.49 & 20.37 & 20.91 \\
  Qwen3.8-27B\hspace{0.35em}\citep{qwen38} & 69.25 & 70.21 & 69.73 & 8.62 & 8.74 & 8.68 & 41.97 & 42.46 & 42.21 \\
  Qwen3.5-35B-A3B\hspace{0.35em}\citep{qwen35} & 63.83 & 56.73 & 60.07 & 7.40 & 6.99 & 7.19 & 40.02 & 36.03 & 37.92 \\
  DeepSeek-VL2-Small-16B\hspace{0.35em}\citep{Deepseekvl2} & N/A\textsuperscript{*} & N/A\textsuperscript{*} & N/A\textsuperscript{*} & N/A\textsuperscript{*} & N/A\textsuperscript{*} & N/A\textsuperscript{*} & N/A\textsuperscript{*} & N/A\textsuperscript{*} & N/A\textsuperscript{*} \\
  DeepSeek-VL2-27B\hspace{0.35em}\citep{Deepseekvl2} & N/A\textsuperscript{*} & N/A\textsuperscript{*} & N/A\textsuperscript{*} & N/A\textsuperscript{*} & N/A\textsuperscript{*} & N/A\textsuperscript{*} & N/A\textsuperscript{*} & N/A\textsuperscript{*} & N/A\textsuperscript{*} \\
  SEED1.5-VL\textsuperscript{*}\hspace{0.35em}\citep{SEED15VL} & -- & -- & 54.90 & -- & -- & 4.30 & -- & -- & 28.70 \\
  \addlinespace[2pt]
  \multicolumn{10}{@{}l}{\hspace{0.4em}\strut\textbf{Vision-Language Models (>1T)}} \\
  Qwen3.7-Max\hspace{0.35em}\citep{qwen37} & 75.85 & 65.60 & 70.35 & 12.76 & 12.08 & 12.41 & 49.90 & 44.31 & 46.93 \\
  Kimi-K2.6 & 29.23\textsuperscript{\textdagger} & 28.21\textsuperscript{\textdagger} & 28.71\textsuperscript{\textdagger} & 2.29\textsuperscript{\textdagger} & 2.43\textsuperscript{\textdagger} & 2.36\textsuperscript{\textdagger} & 15.64\textsuperscript{\textdagger} & 15.55\textsuperscript{\textdagger} & 15.59\textsuperscript{\textdagger} \\
  Kimi-K3\hspace{0.35em}\citep{KimiK3} & N/A\textsuperscript{*} & N/A\textsuperscript{*} & N/A\textsuperscript{*} & N/A\textsuperscript{*} & N/A\textsuperscript{*} & N/A\textsuperscript{*} & N/A\textsuperscript{*} & N/A\textsuperscript{*} & N/A\textsuperscript{*} \\
  GPT-6 Astra & \textbf{97.45} & 95.72 & \textbf{96.58} & 35.04 & 34.83 & 34.93 & 78.24 & 76.86 & 77.54 \\
  \addlinespace[2pt]
  \bottomrule
  
  \end{NiceTabular}%
  }
  \endgroup
\end{table}

\paragraph{M6Doc.}
GroundingPI reaches 74.82 F1mIoU, exceeding LocateAnything Hybrid (65.94), Slow NTP (68.35), and Astra (60.59). Its 92.31/31.83 F1 at IoU 0.50/0.95 also exceeds Astra's 80.84/20.58. Gains across both loose and strict overlap criteria indicate that the advantage includes region coverage and localization precision, rather than only easier matching.

\begin{table}[htbp]
  \centering
  \BenchmarkTableFont
  \caption{\textbf{M6Doc.} Complete box-grounding metrics. GroundingDINO lacks a compatible document-region interface; it, Kimi-K3, and both DeepSeek variants are N/A. Starred MiMo and BAGEL scores retain observations under suspected output-protocol incompatibility and are not formal capability measurements. The starred SEED1.5-VL scores are from Table~9 of \citet{rexomni}.}
  \label{tab:app-m6doc}
  \begingroup
  \fontsize{8}{9.7}\selectfont
  \setlength{\tabcolsep}{3pt}
  \renewcommand{\arraystretch}{1.15}
  \resizebox{\linewidth}{!}{%
  \begin{NiceTabular}{@{}>{\raggedright\arraybackslash}p{177pt}ccccccccc@{}}
  \CodeBefore
    \rowcolor{benchmarktype}{3}
    \rowcolor{benchmarktype}{5}
    \rowcolor{benchmarkpurple}{22}
    \rowcolor{benchmarktype}{23}
    \rowcolor{benchmarktype}{31}
  \Body
  \toprule
  \textbf{Model} & \multicolumn{3}{c}{\textbf{IoU 0.50}} & \multicolumn{3}{c}{\textbf{IoU 0.95}} & \multicolumn{3}{c}{\textbf{mIoU}} \\
  \cmidrule(lr){2-4}\cmidrule(lr){5-7}\cmidrule(lr){8-10}
   & \BenchHead{R} & \BenchHead{P} & \BenchHead{F1} & \BenchHead{R} & \BenchHead{P} & \BenchHead{F1} & \BenchHead{R} & \BenchHead{P} & \BenchHead{F1} \\
  \midrule
  \multicolumn{10}{@{}l}{\hspace{0.4em}\strut\textbf{Open-set Specialized Detectors}} \\
  GroundingDINO\hspace{0.35em}\citep{groundingdino} & N/A\textsuperscript{*} & N/A\textsuperscript{*} & N/A\textsuperscript{*} & N/A\textsuperscript{*} & N/A\textsuperscript{*} & N/A\textsuperscript{*} & N/A\textsuperscript{*} & N/A\textsuperscript{*} & N/A\textsuperscript{*} \\
  \addlinespace[2pt]
  \multicolumn{10}{@{}l}{\hspace{0.4em}\strut\textbf{Vision-Language Models (<10B)}} \\
  Rex-Omni\hspace{0.35em}\citep{rexomni} & 73.09 & 78.27 & 75.59 & 18.16 & 18.78 & 18.46 & 53.36 & 56.64 & 54.95 \\
  LocateAnything Fast\hspace{0.35em}\citep{locateanything} & 59.82 & 61.54 & 60.67 & 18.06 & 18.50 & 18.28 & 46.39 & 47.69 & 47.03 \\
  LocateAnything Hybrid\hspace{0.35em}\citep{locateanything} & 83.54 & 84.94 & 84.23 & 26.18 & 26.59 & 26.39 & 65.38 & 66.50 & 65.94 \\
  LocateAnything Slow NTP\hspace{0.35em}\citep{locateanything} & 88.16 & 89.64 & 88.89 & 25.62 & 26.08 & 25.85 & 67.78 & 68.94 & 68.35 \\
  Qwen2.5-VL-7B\hspace{0.35em}\citep{qwen25vl} & 19.20 & 22.91 & 20.89 & 2.24 & 2.65 & 2.43 & 11.86 & 14.26 & 12.95 \\
  Qwen3-VL-2B\hspace{0.35em}\citep{qwen3vl} & 22.31 & 24.68 & 23.44 & 2.24 & 2.58 & 2.40 & 12.61 & 14.08 & 13.31 \\
  Qwen3-VL-4B\hspace{0.35em}\citep{qwen3vl} & 37.94 & 44.13 & 40.80 & 4.88 & 5.62 & 5.22 & 23.03 & 26.69 & 24.73 \\
  Qwen3-VL-8B\hspace{0.35em}\citep{qwen3vl} & 35.84 & 36.64 & 36.23 & 4.73 & 5.01 & 4.87 & 21.56 & 22.36 & 21.95 \\
  Qwen3.5-4B\hspace{0.35em}\citep{qwen35} & 17.32 & 15.67 & 16.45 & 1.47 & 1.54 & 1.50 & 8.44 & 8.13 & 8.28 \\
  Qwen3.5-9B\hspace{0.35em}\citep{qwen35} & 30.92 & 28.08 & 29.43 & 4.08 & 4.17 & 4.12 & 17.80 & 16.94 & 17.35 \\
  RynnBrain1.1\hspace{0.35em}\citep{RynnBrain112B} & 4.11 & 26.77 & 7.13 & 0.89 & 6.20 & 1.56 & 2.60 & 17.51 & 4.53 \\
  SenseNova-Vision\hspace{0.35em}\citep{SenseNovaVision7BMoT} & 50.01 & 63.15 & 55.82 & 8.87 & 10.60 & 9.66 & 32.12 & 39.99 & 35.62 \\
  MiMo-VL-7B-SFT\hspace{0.35em}\citep{MimoVL} & 2.88\textsuperscript{*} & 5.77\textsuperscript{*} & 3.84\textsuperscript{*} & 0.03\textsuperscript{*} & 0.10\textsuperscript{*} & 0.05\textsuperscript{*} & 0.91\textsuperscript{*} & 2.14\textsuperscript{*} & 1.28\textsuperscript{*} \\
  MiMo-VL-7B-RL\hspace{0.35em}\citep{MimoVL} & 4.63\textsuperscript{*} & 9.36\textsuperscript{*} & 6.19\textsuperscript{*} & 0.07\textsuperscript{*} & 0.12\textsuperscript{*} & 0.09\textsuperscript{*} & 1.53\textsuperscript{*} & 3.24\textsuperscript{*} & 2.07\textsuperscript{*} \\
  BAGEL\hspace{0.35em}\citep{BAGEL7BMoT} & 10.62\textsuperscript{*} & 15.75\textsuperscript{*} & 12.68\textsuperscript{*} & 0.40\textsuperscript{*} & 0.62\textsuperscript{*} & 0.48\textsuperscript{*} & 4.94\textsuperscript{*} & 7.61\textsuperscript{*} & 5.99\textsuperscript{*} \\
  RynnBrain\hspace{0.35em}\citep{dang2026rynnbrain} & 3.98 & 21.54 & 6.72 & 0.32 & 2.06 & 0.55 & 2.00 & 10.88 & 3.38 \\
  GroundingPI & \textbf{91.75} & \textbf{92.88} & \textbf{92.31} & \textbf{31.65} & \textbf{32.00} & \textbf{31.83} & \textbf{74.38} & \textbf{75.27} & \textbf{74.82} \\
  \addlinespace[2pt]
  \multicolumn{10}{@{}l}{\hspace{0.4em}\strut\textbf{Vision-Language Models (10B--1T)}} \\
  Qwen3.6-27B\hspace{0.35em}\citep{qwen3627b} & 45.19 & 47.32 & 46.23 & 6.69 & 6.91 & 6.80 & 27.92 & 29.17 & 28.53 \\
  Qwen3-VL-32B\hspace{0.35em}\citep{qwen3vl} & 46.89 & 48.85 & 47.85 & 7.02 & 7.37 & 7.19 & 29.66 & 31.02 & 30.32 \\
  Qwen3.8-27B\hspace{0.35em}\citep{qwen38} & 34.42 & 35.88 & 35.13 & 4.19 & 4.47 & 4.33 & 19.43 & 20.43 & 19.92 \\
  Qwen3.5-35B-A3B\hspace{0.35em}\citep{qwen35} & 44.90 & 46.66 & 45.76 & 6.63 & 6.99 & 6.81 & 28.07 & 29.32 & 28.68 \\
  DeepSeek-VL2-Small-16B\hspace{0.35em}\citep{Deepseekvl2} & N/A\textsuperscript{*} & N/A\textsuperscript{*} & N/A\textsuperscript{*} & N/A\textsuperscript{*} & N/A\textsuperscript{*} & N/A\textsuperscript{*} & N/A\textsuperscript{*} & N/A\textsuperscript{*} & N/A\textsuperscript{*} \\
  DeepSeek-VL2-27B\hspace{0.35em}\citep{Deepseekvl2} & N/A\textsuperscript{*} & N/A\textsuperscript{*} & N/A\textsuperscript{*} & N/A\textsuperscript{*} & N/A\textsuperscript{*} & N/A\textsuperscript{*} & N/A\textsuperscript{*} & N/A\textsuperscript{*} & N/A\textsuperscript{*} \\
  SEED1.5-VL\textsuperscript{*}\hspace{0.35em}\citep{SEED15VL} & -- & -- & 48.00 & -- & -- & 3.40 & -- & -- & 28.00 \\
  \addlinespace[2pt]
  \multicolumn{10}{@{}l}{\hspace{0.4em}\strut\textbf{Vision-Language Models (>1T)}} \\
  Qwen3.7-Max\hspace{0.35em}\citep{qwen37} & 48.23 & 50.84 & 49.50 & 9.36 & 9.87 & 9.61 & 31.50 & 33.16 & 32.31 \\
  Kimi-K2.6 & 21.96\textsuperscript{\textdagger} & 23.87\textsuperscript{\textdagger} & 22.87\textsuperscript{\textdagger} & 2.80\textsuperscript{\textdagger} & 3.16\textsuperscript{\textdagger} & 2.97\textsuperscript{\textdagger} & 12.89\textsuperscript{\textdagger} & 14.16\textsuperscript{\textdagger} & 13.50\textsuperscript{\textdagger} \\
  Kimi-K3\hspace{0.35em}\citep{KimiK3} & N/A\textsuperscript{*} & N/A\textsuperscript{*} & N/A\textsuperscript{*} & N/A\textsuperscript{*} & N/A\textsuperscript{*} & N/A\textsuperscript{*} & N/A\textsuperscript{*} & N/A\textsuperscript{*} & N/A\textsuperscript{*} \\
  GPT-6 Astra & 78.30 & 83.51 & 80.84 & 19.68 & 21.54 & 20.58 & 58.50 & 62.78 & 60.59 \\
  \addlinespace[2pt]
  \bottomrule
  
  \end{NiceTabular}%
  }
  \endgroup
\end{table}

\subsection{Visual Prompting}
\label{app:visual_prompting}

\paragraph{FSC147.}
GroundingPI reaches 87.04 F1 at IoU 0.50 but only 3.49 at IoU 0.95, yielding 57.99 F1mIoU. This slightly exceeds Rex-Omni (57.15) but trails SenseNova-Vision (62.51) and Astra (61.04). Exemplar correspondence is therefore effective at coarse overlap, while precise exemplar-conditioned box boundaries remain a clear limitation.

\begin{table}[htbp]
  \centering
  \BenchmarkTableFont
  \caption{\textbf{FSC147 visual prompting.} Complete box-grounding metrics. LocateAnything variants lack a supported visual-prompt interface; both DeepSeek variants have incompatible output protocols. Their entries are N/A. Starred GroundingDINO scores come from an unsupported visual-prompt task interface; starred MiMo and BAGEL scores are retained observations under suspected output-protocol incompatibility.}
  \label{tab:app-fsc147}
  \begingroup
  \fontsize{8}{9.7}\selectfont
  \setlength{\tabcolsep}{3pt}
  \renewcommand{\arraystretch}{1.15}
  \resizebox{\linewidth}{!}{%
  \begin{NiceTabular}{@{}>{\raggedright\arraybackslash}p{177pt}ccccccccc@{}}
  \CodeBefore
    \rowcolor{benchmarktype}{3}
    \rowcolor{benchmarktype}{5}
    \rowcolor{benchmarkpurple}{22}
    \rowcolor{benchmarktype}{23}
    \rowcolor{benchmarktype}{30}
  \Body
  \toprule
  \textbf{Model} & \multicolumn{3}{c}{\textbf{IoU 0.50}} & \multicolumn{3}{c}{\textbf{IoU 0.95}} & \multicolumn{3}{c}{\textbf{mIoU}} \\
  \cmidrule(lr){2-4}\cmidrule(lr){5-7}\cmidrule(lr){8-10}
   & \BenchHead{R} & \BenchHead{P} & \BenchHead{F1} & \BenchHead{R} & \BenchHead{P} & \BenchHead{F1} & \BenchHead{R} & \BenchHead{P} & \BenchHead{F1} \\
  \midrule
  \multicolumn{10}{@{}l}{\hspace{0.4em}\strut\textbf{Open-set Specialized Detectors}} \\
  GroundingDINO\hspace{0.35em}\citep{groundingdino} & 7.94\textsuperscript{*} & 20.16\textsuperscript{*} & 11.39\textsuperscript{*} & 1.56\textsuperscript{*} & 4.94\textsuperscript{*} & 2.37\textsuperscript{*} & 6.34\textsuperscript{*} & 16.01\textsuperscript{*} & 9.08\textsuperscript{*} \\
  \addlinespace[2pt]
  \multicolumn{10}{@{}l}{\hspace{0.4em}\strut\textbf{Vision-Language Models (<10B)}} \\
  Rex-Omni\hspace{0.35em}\citep{rexomni} & 78.97 & 77.92 & 78.44 & 9.13 & 9.01 & 9.07 & 57.56 & 56.75 & 57.15 \\
  LocateAnything Fast\hspace{0.35em}\citep{locateanything} & N/A\textsuperscript{*} & N/A\textsuperscript{*} & N/A\textsuperscript{*} & N/A\textsuperscript{*} & N/A\textsuperscript{*} & N/A\textsuperscript{*} & N/A\textsuperscript{*} & N/A\textsuperscript{*} & N/A\textsuperscript{*} \\
  LocateAnything Hybrid\hspace{0.35em}\citep{locateanything} & N/A\textsuperscript{*} & N/A\textsuperscript{*} & N/A\textsuperscript{*} & N/A\textsuperscript{*} & N/A\textsuperscript{*} & N/A\textsuperscript{*} & N/A\textsuperscript{*} & N/A\textsuperscript{*} & N/A\textsuperscript{*} \\
  LocateAnything Slow NTP\hspace{0.35em}\citep{locateanything} & N/A\textsuperscript{*} & N/A\textsuperscript{*} & N/A\textsuperscript{*} & N/A\textsuperscript{*} & N/A\textsuperscript{*} & N/A\textsuperscript{*} & N/A\textsuperscript{*} & N/A\textsuperscript{*} & N/A\textsuperscript{*} \\
  Qwen2.5-VL-7B\hspace{0.35em}\citep{qwen25vl} & 14.70 & 34.29 & 20.57 & 0.09 & 1.20 & 0.16 & 7.54 & 17.39 & 10.49 \\
  Qwen3-VL-2B\hspace{0.35em}\citep{qwen3vl} & 22.55 & 39.85 & 28.80 & 0.30 & 0.73 & 0.42 & 11.21 & 19.07 & 14.11 \\
  Qwen3-VL-4B\hspace{0.35em}\citep{qwen3vl} & 18.74 & 83.98 & 30.65 & 0.59 & 1.60 & 0.86 & 11.60 & 51.16 & 18.90 \\
  Qwen3-VL-8B\hspace{0.35em}\citep{qwen3vl} & 19.80 & 74.93 & 31.33 & 0.39 & 0.85 & 0.53 & 11.09 & 40.67 & 17.41 \\
  Qwen3.5-4B\hspace{0.35em}\citep{qwen35} & 27.06 & 41.98 & 32.91 & 0.14 & 0.21 & 0.17 & 11.70 & 18.62 & 14.37 \\
  Qwen3.5-9B\hspace{0.35em}\citep{qwen35} & 29.37 & 60.29 & 39.49 & 0.64 & 1.21 & 0.84 & 15.31 & 31.30 & 20.56 \\
  RynnBrain1.1\hspace{0.35em}\citep{RynnBrain112B} & 5.25 & 47.70 & 9.47 & 0.02 & 0.14 & 0.03 & 2.68 & 22.33 & 4.78 \\
  SenseNova-Vision\hspace{0.35em}\citep{SenseNovaVision7BMoT} & 78.56 & 88.88 & 83.40 & \textbf{10.25} & \textbf{10.78} & \textbf{10.51} & 59.46 & \textbf{65.89} & \textbf{62.51} \\
  MiMo-VL-7B-SFT\hspace{0.35em}\citep{MimoVL} & 13.59\textsuperscript{*} & 16.68\textsuperscript{*} & 14.98\textsuperscript{*} & 0.04\textsuperscript{*} & 0.04\textsuperscript{*} & 0.04\textsuperscript{*} & 4.65\textsuperscript{*} & 5.79\textsuperscript{*} & 5.15\textsuperscript{*} \\
  MiMo-VL-7B-RL\hspace{0.35em}\citep{MimoVL} & 13.58\textsuperscript{*} & 21.88\textsuperscript{*} & 16.76\textsuperscript{*} & 0.01\textsuperscript{*} & 0.02\textsuperscript{*} & 0.01\textsuperscript{*} & 4.51\textsuperscript{*} & 7.23\textsuperscript{*} & 5.56\textsuperscript{*} \\
  BAGEL\hspace{0.35em}\citep{BAGEL7BMoT} & 10.94\textsuperscript{*} & 52.39\textsuperscript{*} & 18.09\textsuperscript{*} & 0.31\textsuperscript{*} & 1.06\textsuperscript{*} & 0.48\textsuperscript{*} & 6.62\textsuperscript{*} & 31.67\textsuperscript{*} & 10.95\textsuperscript{*} \\
  RynnBrain\hspace{0.35em}\citep{dang2026rynnbrain} & 2.04 & 37.23 & 3.87 & 0.02 & 0.17 & 0.03 & 1.00 & 17.12 & 1.88 \\
  GroundingPI & 86.33 & 87.76 & 87.04 & 3.49 & 3.49 & 3.49 & 57.69 & 58.29 & 57.99 \\
  \addlinespace[2pt]
  \multicolumn{10}{@{}l}{\hspace{0.4em}\strut\textbf{Vision-Language Models (10B--1T)}} \\
  Qwen3.6-27B\hspace{0.35em}\citep{qwen3627b} & 28.56 & 53.78 & 37.31 & 1.60 & 3.45 & 2.19 & 19.05 & 35.48 & 24.78 \\
  Qwen3-VL-32B\hspace{0.35em}\citep{qwen3vl} & 29.33 & 79.24 & 42.82 & 1.25 & 1.90 & 1.51 & 18.24 & 47.65 & 26.36 \\
  Qwen3.8-27B\hspace{0.35em}\citep{qwen38} & 14.80 & 86.51 & 25.27 & 0.60 & 2.42 & 0.97 & 9.84 & 57.57 & 16.80 \\
  Qwen3.5-35B-A3B\hspace{0.35em}\citep{qwen35} & 37.48 & 80.34 & 51.12 & 1.44 & 2.31 & 1.77 & 23.36 & 49.13 & 31.65 \\
  DeepSeek-VL2-Small-16B\hspace{0.35em}\citep{Deepseekvl2} & N/A\textsuperscript{*} & N/A\textsuperscript{*} & N/A\textsuperscript{*} & N/A\textsuperscript{*} & N/A\textsuperscript{*} & N/A\textsuperscript{*} & N/A\textsuperscript{*} & N/A\textsuperscript{*} & N/A\textsuperscript{*} \\
  DeepSeek-VL2-27B\hspace{0.35em}\citep{Deepseekvl2} & N/A\textsuperscript{*} & N/A\textsuperscript{*} & N/A\textsuperscript{*} & N/A\textsuperscript{*} & N/A\textsuperscript{*} & N/A\textsuperscript{*} & N/A\textsuperscript{*} & N/A\textsuperscript{*} & N/A\textsuperscript{*} \\
  \addlinespace[2pt]
  \multicolumn{10}{@{}l}{\hspace{0.4em}\strut\textbf{Vision-Language Models (>1T)}} \\
  Qwen3.7-Max\hspace{0.35em}\citep{qwen37} & 19.91 & \textbf{90.30} & 32.62 & 0.82 & 1.92 & 1.15 & 13.31 & 57.98 & 21.62 \\
  Kimi-K2.6 & 52.17 & 55.60 & 53.83 & 2.75 & 3.06 & 2.89 & 31.33 & 33.69 & 32.47 \\
  Kimi-K3\hspace{0.35em}\citep{KimiK3} & 67.29 & 76.03 & 71.39 & 3.98 & 4.16 & 4.07 & 40.56 & 44.83 & 42.59 \\
  GPT-6 Astra & \textbf{89.83} & 87.12 & \textbf{88.48} & 8.34 & 8.03 & 8.18 & \textbf{61.83} & 60.25 & 61.04 \\
  \addlinespace[2pt]
  \bottomrule
  
  \end{NiceTabular}%
  }
  \endgroup
\end{table}

\paragraph{Dense200 visual prompting.}
GroundingPI obtains 75.43 F1mIoU, compared with Astra's 68.94 and Rex-Omni's 55.50. Astra has slightly higher F1 at IoU 0.50 (92.04 versus 91.88), whereas GroundingPI is substantially higher at IoU 0.95 (28.42 versus 12.29). The mean-score advantage thus reflects tighter localization, not just recognizing more exemplar-matched instances.

\begin{table}[htbp]
  \centering
  \BenchmarkTableFont
  \caption{\textbf{Dense200 visual prompting.} Complete box-grounding metrics. LocateAnything variants lack a supported visual-prompt interface; both DeepSeek variants have incompatible output protocols. Their entries are N/A. Kimi-K3 is N/A because the required prompt format is unsupported. Starred GroundingDINO scores come from an unsupported visual-prompt task interface; starred MiMo and BAGEL scores are retained observations under suspected output-protocol incompatibility.}
  \label{tab:app-visual-dense200}
  \begingroup
  \fontsize{8}{9.7}\selectfont
  \setlength{\tabcolsep}{3pt}
  \renewcommand{\arraystretch}{1.15}
  \resizebox{\linewidth}{!}{%
  \begin{NiceTabular}{@{}>{\raggedright\arraybackslash}p{177pt}ccccccccc@{}}
  \CodeBefore
    \rowcolor{benchmarktype}{3}
    \rowcolor{benchmarktype}{5}
    \rowcolor{benchmarkpurple}{22}
    \rowcolor{benchmarktype}{23}
    \rowcolor{benchmarktype}{30}
  \Body
  \toprule
  \textbf{Model} & \multicolumn{3}{c}{\textbf{IoU 0.50}} & \multicolumn{3}{c}{\textbf{IoU 0.95}} & \multicolumn{3}{c}{\textbf{mIoU}} \\
  \cmidrule(lr){2-4}\cmidrule(lr){5-7}\cmidrule(lr){8-10}
   & \BenchHead{R} & \BenchHead{P} & \BenchHead{F1} & \BenchHead{R} & \BenchHead{P} & \BenchHead{F1} & \BenchHead{R} & \BenchHead{P} & \BenchHead{F1} \\
  \midrule
  \multicolumn{10}{@{}l}{\hspace{0.4em}\strut\textbf{Open-set Specialized Detectors}} \\
  GroundingDINO\hspace{0.35em}\citep{groundingdino} & 2.03\textsuperscript{*} & 9.98\textsuperscript{*} & 3.38\textsuperscript{*} & 1.35\textsuperscript{*} & 5.64\textsuperscript{*} & 2.18\textsuperscript{*} & 1.90\textsuperscript{*} & 9.00\textsuperscript{*} & 3.14\textsuperscript{*} \\
  \addlinespace[2pt]
  \multicolumn{10}{@{}l}{\hspace{0.4em}\strut\textbf{Vision-Language Models (<10B)}} \\
  Rex-Omni\hspace{0.35em}\citep{rexomni} & 73.05 & 74.81 & 73.92 & 10.89 & 11.12 & 11.00 & 54.93 & 56.08 & 55.50 \\
  LocateAnything Fast\hspace{0.35em}\citep{locateanything} & N/A\textsuperscript{*} & N/A\textsuperscript{*} & N/A\textsuperscript{*} & N/A\textsuperscript{*} & N/A\textsuperscript{*} & N/A\textsuperscript{*} & N/A\textsuperscript{*} & N/A\textsuperscript{*} & N/A\textsuperscript{*} \\
  LocateAnything Hybrid\hspace{0.35em}\citep{locateanything} & N/A\textsuperscript{*} & N/A\textsuperscript{*} & N/A\textsuperscript{*} & N/A\textsuperscript{*} & N/A\textsuperscript{*} & N/A\textsuperscript{*} & N/A\textsuperscript{*} & N/A\textsuperscript{*} & N/A\textsuperscript{*} \\
  LocateAnything Slow NTP\hspace{0.35em}\citep{locateanything} & N/A\textsuperscript{*} & N/A\textsuperscript{*} & N/A\textsuperscript{*} & N/A\textsuperscript{*} & N/A\textsuperscript{*} & N/A\textsuperscript{*} & N/A\textsuperscript{*} & N/A\textsuperscript{*} & N/A\textsuperscript{*} \\
  Qwen2.5-VL-7B\hspace{0.35em}\citep{qwen25vl} & 0.75 & 23.66 & 1.46 & 0.00 & 0.00 & 0.00 & 0.44 & 13.33 & 0.86 \\
  Qwen3-VL-2B\hspace{0.35em}\citep{qwen3vl} & 7.76 & 60.82 & 13.76 & 0.69 & 1.78 & 1.00 & 5.08 & 33.85 & 8.79 \\
  Qwen3-VL-4B\hspace{0.35em}\citep{qwen3vl} & 1.76 & 93.00 & 3.45 & 0.21 & 8.00 & 0.41 & 1.27 & 62.90 & 2.49 \\
  Qwen3-VL-8B\hspace{0.35em}\citep{qwen3vl} & 8.26 & 87.71 & 15.10 & 0.99 & 3.91 & 1.58 & 5.79 & 55.05 & 10.45 \\
  Qwen3.5-4B\hspace{0.35em}\citep{qwen35} & 2.82 & 78.69 & 5.44 & 0.34 & 7.71 & 0.65 & 2.01 & 54.25 & 3.87 \\
  Qwen3.5-9B\hspace{0.35em}\citep{qwen35} & 26.66 & 65.61 & 37.91 & 4.19 & 10.33 & 5.96 & 19.00 & 46.27 & 26.93 \\
  RynnBrain1.1\hspace{0.35em}\citep{RynnBrain112B} & 1.16 & 62.50 & 2.28 & 0.07 & 4.00 & 0.14 & 0.74 & 38.25 & 1.45 \\
  SenseNova-Vision\hspace{0.35em}\citep{SenseNovaVision7BMoT} & 69.56 & 84.68 & 76.38 & 19.15 & 22.06 & 20.50 & 57.21 & 69.70 & 62.84 \\
  MiMo-VL-7B-SFT\hspace{0.35em}\citep{MimoVL} & 2.87\textsuperscript{*} & 20.40\textsuperscript{*} & 5.03\textsuperscript{*} & 0.01\textsuperscript{*} & 0.00\textsuperscript{*} & 0.00\textsuperscript{*} & 1.08\textsuperscript{*} & 6.38\textsuperscript{*} & 1.83\textsuperscript{*} \\
  MiMo-VL-7B-RL\hspace{0.35em}\citep{MimoVL} & 2.58\textsuperscript{*} & 14.41\textsuperscript{*} & 4.38\textsuperscript{*} & 0.01\textsuperscript{*} & 0.01\textsuperscript{*} & 0.01\textsuperscript{*} & 0.89\textsuperscript{*} & 5.40\textsuperscript{*} & 1.51\textsuperscript{*} \\
  BAGEL\hspace{0.35em}\citep{BAGEL7BMoT} & 1.28\textsuperscript{*} & 56.29\textsuperscript{*} & 2.50\textsuperscript{*} & 0.02\textsuperscript{*} & 1.25\textsuperscript{*} & 0.05\textsuperscript{*} & 0.74\textsuperscript{*} & 29.87\textsuperscript{*} & 1.45\textsuperscript{*} \\
  RynnBrain\hspace{0.35em}\citep{dang2026rynnbrain} & 1.26 & 64.50 & 2.47 & 0.02 & 1.00 & 0.03 & 0.74 & 36.35 & 1.44 \\
  GroundingPI & 89.96 & \textbf{93.89} & 91.88 & \textbf{27.93} & \textbf{28.94} & \textbf{28.42} & \textbf{73.93} & \textbf{76.98} & \textbf{75.43} \\
  \addlinespace[2pt]
  \multicolumn{10}{@{}l}{\hspace{0.4em}\strut\textbf{Vision-Language Models (10B--1T)}} \\
  Qwen3.6-27B\hspace{0.35em}\citep{qwen3627b} & 31.92 & 59.87 & 41.64 & 5.79 & 7.69 & 6.61 & 23.81 & 45.89 & 31.33 \\
  Qwen3-VL-32B\hspace{0.35em}\citep{qwen3vl} & 18.96 & 55.73 & 28.30 & 2.51 & 6.05 & 3.55 & 13.84 & 41.11 & 20.69 \\
  Qwen3.8-27B\hspace{0.35em}\citep{qwen38} & 25.08 & 68.06 & 36.65 & 4.61 & 14.62 & 7.01 & 19.15 & 53.40 & 28.19 \\
  Qwen3.5-35B-A3B\hspace{0.35em}\citep{qwen35} & 41.09 & 71.35 & 52.15 & 5.85 & 7.14 & 6.43 & 29.95 & 51.04 & 37.73 \\
  DeepSeek-VL2-Small-16B\hspace{0.35em}\citep{Deepseekvl2} & N/A\textsuperscript{*} & N/A\textsuperscript{*} & N/A\textsuperscript{*} & N/A\textsuperscript{*} & N/A\textsuperscript{*} & N/A\textsuperscript{*} & N/A\textsuperscript{*} & N/A\textsuperscript{*} & N/A\textsuperscript{*} \\
  DeepSeek-VL2-27B\hspace{0.35em}\citep{Deepseekvl2} & N/A\textsuperscript{*} & N/A\textsuperscript{*} & N/A\textsuperscript{*} & N/A\textsuperscript{*} & N/A\textsuperscript{*} & N/A\textsuperscript{*} & N/A\textsuperscript{*} & N/A\textsuperscript{*} & N/A\textsuperscript{*} \\
  \addlinespace[2pt]
  \multicolumn{10}{@{}l}{\hspace{0.4em}\strut\textbf{Vision-Language Models (>1T)}} \\
  Qwen3.7-Max\hspace{0.35em}\citep{qwen37} & 28.05 & 72.46 & 40.44 & 5.40 & 9.51 & 6.89 & 22.37 & 55.16 & 31.81 \\
  Kimi-K2.6 & 43.14 & 51.19 & 46.82 & 5.48 & 6.84 & 6.09 & 29.90 & 35.39 & 32.41 \\
  Kimi-K3\hspace{0.35em}\citep{KimiK3} & N/A\textsuperscript{*} & N/A\textsuperscript{*} & N/A\textsuperscript{*} & N/A\textsuperscript{*} & N/A\textsuperscript{*} & N/A\textsuperscript{*} & N/A\textsuperscript{*} & N/A\textsuperscript{*} & N/A\textsuperscript{*} \\
  GPT-6 Astra & \textbf{95.44} & 88.71 & \textbf{92.04} & 12.84 & 11.75 & 12.29 & 71.48 & 66.44 & 68.94 \\
  \addlinespace[2pt]
  \bottomrule
  
  \end{NiceTabular}%
  }
  \endgroup
\end{table}

\paragraph{COCO visual prompting.}
GroundingPI attains 82.62 F1mIoU and 71.53 F1 at IoU 0.95, versus Astra's 68.43 and 40.54. Both mean recall and precision are high (84.03/81.25). These results support the shared coordinate vocabulary as an effective interface for visual as well as linguistic queries. Absolute scores should not be compared directly with category-prompted COCO, since the supplied target information differs.

\begin{table}[htbp]
  \centering
  \BenchmarkTableFont
  \caption{\textbf{COCO visual prompting.} Complete box-grounding metrics. LocateAnything variants lack a supported visual-prompt interface; both DeepSeek variants have incompatible output protocols. Their entries are N/A. Kimi-K3 is N/A because the required prompt format is unsupported. Starred GroundingDINO scores come from an unsupported visual-prompt task interface; starred MiMo and BAGEL scores are retained observations under suspected output-protocol incompatibility.}
  \label{tab:app-visual-coco}
  \begingroup
  \fontsize{8}{9.7}\selectfont
  \setlength{\tabcolsep}{3pt}
  \renewcommand{\arraystretch}{1.15}
  \resizebox{\linewidth}{!}{%
  \begin{NiceTabular}{@{}>{\raggedright\arraybackslash}p{177pt}ccccccccc@{}}
  \CodeBefore
    \rowcolor{benchmarktype}{3}
    \rowcolor{benchmarktype}{5}
    \rowcolor{benchmarkpurple}{22}
    \rowcolor{benchmarktype}{23}
    \rowcolor{benchmarktype}{30}
  \Body
  \toprule
  \textbf{Model} & \multicolumn{3}{c}{\textbf{IoU 0.50}} & \multicolumn{3}{c}{\textbf{IoU 0.95}} & \multicolumn{3}{c}{\textbf{mIoU}} \\
  \cmidrule(lr){2-4}\cmidrule(lr){5-7}\cmidrule(lr){8-10}
   & \BenchHead{R} & \BenchHead{P} & \BenchHead{F1} & \BenchHead{R} & \BenchHead{P} & \BenchHead{F1} & \BenchHead{R} & \BenchHead{P} & \BenchHead{F1} \\
  \midrule
  \multicolumn{10}{@{}l}{\hspace{0.4em}\strut\textbf{Open-set Specialized Detectors}} \\
  GroundingDINO\hspace{0.35em}\citep{groundingdino} & 26.84\textsuperscript{*} & 27.30\textsuperscript{*} & 27.07\textsuperscript{*} & 13.69\textsuperscript{*} & 13.52\textsuperscript{*} & 13.60\textsuperscript{*} & 23.62\textsuperscript{*} & 23.83\textsuperscript{*} & 23.73\textsuperscript{*} \\
  \addlinespace[2pt]
  \multicolumn{10}{@{}l}{\hspace{0.4em}\strut\textbf{Vision-Language Models (<10B)}} \\
  Rex-Omni\hspace{0.35em}\citep{rexomni} & 78.69 & 67.17 & 72.48 & 20.38 & 18.49 & 19.39 & 61.99 & 53.44 & 57.40 \\
  LocateAnything Fast\hspace{0.35em}\citep{locateanything} & N/A\textsuperscript{*} & N/A\textsuperscript{*} & N/A\textsuperscript{*} & N/A\textsuperscript{*} & N/A\textsuperscript{*} & N/A\textsuperscript{*} & N/A\textsuperscript{*} & N/A\textsuperscript{*} & N/A\textsuperscript{*} \\
  LocateAnything Hybrid\hspace{0.35em}\citep{locateanything} & N/A\textsuperscript{*} & N/A\textsuperscript{*} & N/A\textsuperscript{*} & N/A\textsuperscript{*} & N/A\textsuperscript{*} & N/A\textsuperscript{*} & N/A\textsuperscript{*} & N/A\textsuperscript{*} & N/A\textsuperscript{*} \\
  LocateAnything Slow NTP\hspace{0.35em}\citep{locateanything} & N/A\textsuperscript{*} & N/A\textsuperscript{*} & N/A\textsuperscript{*} & N/A\textsuperscript{*} & N/A\textsuperscript{*} & N/A\textsuperscript{*} & N/A\textsuperscript{*} & N/A\textsuperscript{*} & N/A\textsuperscript{*} \\
  Qwen2.5-VL-7B\hspace{0.35em}\citep{qwen25vl} & 49.63 & 66.36 & 56.79 & 7.34 & 8.69 & 7.96 & 35.70 & 46.44 & 40.36 \\
  Qwen3-VL-2B\hspace{0.35em}\citep{qwen3vl} & 60.57 & 73.14 & 66.26 & 14.91 & 16.36 & 15.61 & 45.16 & 53.21 & 48.85 \\
  Qwen3-VL-4B\hspace{0.35em}\citep{qwen3vl} & 65.02 & \textbf{88.20} & 74.85 & 22.72 & 27.10 & 24.71 & 52.88 & 69.62 & 60.09 \\
  Qwen3-VL-8B\hspace{0.35em}\citep{qwen3vl} & 65.86 & 82.25 & 73.15 & 20.83 & 23.53 & 22.10 & 51.86 & 63.15 & 56.94 \\
  Qwen3.5-4B\hspace{0.35em}\citep{qwen35} & 61.62 & 77.58 & 68.69 & 14.60 & 16.08 & 15.30 & 45.57 & 55.63 & 50.09 \\
  Qwen3.5-9B\hspace{0.35em}\citep{qwen35} & 71.95 & 74.02 & 72.97 & 19.76 & 20.51 & 20.13 & 55.46 & 57.49 & 56.46 \\
  RynnBrain1.1\hspace{0.35em}\citep{RynnBrain112B} & 58.34 & 79.03 & 67.13 & 13.07 & 15.37 & 14.12 & 43.49 & 56.88 & 49.28 \\
  SenseNova-Vision\hspace{0.35em}\citep{SenseNovaVision7BMoT} & 66.48 & 63.44 & 64.93 & 28.88 & 28.66 & 28.77 & 57.10 & 54.98 & 56.02 \\
  MiMo-VL-7B-SFT\hspace{0.35em}\citep{MimoVL} & 46.34\textsuperscript{*} & 44.23\textsuperscript{*} & 45.26\textsuperscript{*} & 4.11\textsuperscript{*} & 3.85\textsuperscript{*} & 3.98\textsuperscript{*} & 30.10\textsuperscript{*} & 28.59\textsuperscript{*} & 29.32\textsuperscript{*} \\
  MiMo-VL-7B-RL\hspace{0.35em}\citep{MimoVL} & 48.83\textsuperscript{*} & 49.32\textsuperscript{*} & 49.08\textsuperscript{*} & 4.23\textsuperscript{*} & 4.05\textsuperscript{*} & 4.14\textsuperscript{*} & 30.93\textsuperscript{*} & 30.85\textsuperscript{*} & 30.89\textsuperscript{*} \\
  BAGEL\hspace{0.35em}\citep{BAGEL7BMoT} & 59.41\textsuperscript{*} & 77.49\textsuperscript{*} & 67.26\textsuperscript{*} & 16.85\textsuperscript{*} & 19.41\textsuperscript{*} & 18.04\textsuperscript{*} & 45.64\textsuperscript{*} & 57.49\textsuperscript{*} & 50.87\textsuperscript{*} \\
  RynnBrain\hspace{0.35em}\citep{dang2026rynnbrain} & 57.72 & 78.39 & 66.49 & 10.88 & 12.72 & 11.73 & 41.65 & 54.31 & 47.13 \\
  GroundingPI & 89.68 & 86.48 & \textbf{88.05} & \textbf{72.60} & \textbf{70.50} & \textbf{71.53} & \textbf{84.03} & \textbf{81.25} & \textbf{82.62} \\
  \addlinespace[2pt]
  \multicolumn{10}{@{}l}{\hspace{0.4em}\strut\textbf{Vision-Language Models (10B--1T)}} \\
  Qwen3.6-27B\hspace{0.35em}\citep{qwen3627b} & 74.11 & 74.23 & 74.17 & 22.67 & 22.01 & 22.34 & 58.30 & 58.16 & 58.23 \\
  Qwen3-VL-32B\hspace{0.35em}\citep{qwen3vl} & 40.50 & 45.71 & 42.95 & 15.51 & 17.14 & 16.28 & 34.10 & 38.37 & 36.11 \\
  Qwen3.8-27B\hspace{0.35em}\citep{qwen38} & 67.57 & 70.43 & 68.97 & 17.29 & 17.61 & 17.45 & 51.30 & 53.16 & 52.21 \\
  Qwen3.5-35B-A3B\hspace{0.35em}\citep{qwen35} & 72.35 & 76.48 & 74.36 & 20.59 & 20.84 & 20.71 & 55.98 & 58.56 & 57.24 \\
  DeepSeek-VL2-Small-16B\hspace{0.35em}\citep{Deepseekvl2} & N/A\textsuperscript{*} & N/A\textsuperscript{*} & N/A\textsuperscript{*} & N/A\textsuperscript{*} & N/A\textsuperscript{*} & N/A\textsuperscript{*} & N/A\textsuperscript{*} & N/A\textsuperscript{*} & N/A\textsuperscript{*} \\
  DeepSeek-VL2-27B\hspace{0.35em}\citep{Deepseekvl2} & N/A\textsuperscript{*} & N/A\textsuperscript{*} & N/A\textsuperscript{*} & N/A\textsuperscript{*} & N/A\textsuperscript{*} & N/A\textsuperscript{*} & N/A\textsuperscript{*} & N/A\textsuperscript{*} & N/A\textsuperscript{*} \\
  \addlinespace[2pt]
  \multicolumn{10}{@{}l}{\hspace{0.4em}\strut\textbf{Vision-Language Models (>1T)}} \\
  Qwen3.7-Max\hspace{0.35em}\citep{qwen37} & 71.76 & 81.62 & 76.37 & 20.73 & 21.98 & 21.34 & 56.64 & 63.17 & 59.72 \\
  Kimi-K2.6 & 71.27 & 68.79 & 70.01 & 29.23 & 28.61 & 28.91 & 58.33 & 56.47 & 57.39 \\
  Kimi-K3\hspace{0.35em}\citep{KimiK3} & N/A\textsuperscript{*} & N/A\textsuperscript{*} & N/A\textsuperscript{*} & N/A\textsuperscript{*} & N/A\textsuperscript{*} & N/A\textsuperscript{*} & N/A\textsuperscript{*} & N/A\textsuperscript{*} & N/A\textsuperscript{*} \\
  GPT-6 Astra & \textbf{91.13} & 71.62 & 80.26 & 43.17 & 38.23 & 40.54 & 76.87 & 61.62 & 68.43 \\
  \addlinespace[2pt]
  \bottomrule
  
  \end{NiceTabular}%
  }
  \endgroup
\end{table}

\paragraph{LVIS visual prompting.}
GroundingPI reaches 78.32 F1mIoU, improving over Astra's 64.96 and Rex-Omni's 49.36. Mean recall and precision are balanced at 77.39/79.27, and F1 at IoU 0.95 reaches 65.86. Together with COCO, this suggests that exemplar conditioning can supply useful appearance information across category vocabularies. N/A entries for unsupported exemplar interfaces are not treated as zero-score capability measurements.

\begin{table}[htbp]
  \centering
  \BenchmarkTableFont
  \caption{\textbf{LVIS visual prompting.} Complete box-grounding metrics. LocateAnything variants lack a supported visual-prompt interface; both DeepSeek variants have incompatible output protocols. Their entries are N/A. Kimi-K3 is N/A because the required prompt format is unsupported. Starred GroundingDINO scores come from an unsupported visual-prompt task interface; starred MiMo and BAGEL scores are retained observations under suspected output-protocol incompatibility.}
  \label{tab:app-visual-lvis}
  \begingroup
  \fontsize{8}{9.7}\selectfont
  \setlength{\tabcolsep}{3pt}
  \renewcommand{\arraystretch}{1.15}
  \resizebox{\linewidth}{!}{%
  \begin{NiceTabular}{@{}>{\raggedright\arraybackslash}p{177pt}ccccccccc@{}}
  \CodeBefore
    \rowcolor{benchmarktype}{3}
    \rowcolor{benchmarktype}{5}
    \rowcolor{benchmarkpurple}{22}
    \rowcolor{benchmarktype}{23}
    \rowcolor{benchmarktype}{30}
  \Body
  \toprule
  \textbf{Model} & \multicolumn{3}{c}{\textbf{IoU 0.50}} & \multicolumn{3}{c}{\textbf{IoU 0.95}} & \multicolumn{3}{c}{\textbf{mIoU}} \\
  \cmidrule(lr){2-4}\cmidrule(lr){5-7}\cmidrule(lr){8-10}
   & \BenchHead{R} & \BenchHead{P} & \BenchHead{F1} & \BenchHead{R} & \BenchHead{P} & \BenchHead{F1} & \BenchHead{R} & \BenchHead{P} & \BenchHead{F1} \\
  \midrule
  \multicolumn{10}{@{}l}{\hspace{0.4em}\strut\textbf{Open-set Specialized Detectors}} \\
  GroundingDINO\hspace{0.35em}\citep{groundingdino} & 18.16\textsuperscript{*} & 18.90\textsuperscript{*} & 18.52\textsuperscript{*} & 9.85\textsuperscript{*} & 9.99\textsuperscript{*} & 9.92\textsuperscript{*} & 15.91\textsuperscript{*} & 16.38\textsuperscript{*} & 16.14\textsuperscript{*} \\
  \addlinespace[2pt]
  \multicolumn{10}{@{}l}{\hspace{0.4em}\strut\textbf{Vision-Language Models (<10B)}} \\
  Rex-Omni\hspace{0.35em}\citep{rexomni} & 69.31 & 61.02 & 64.90 & 17.42 & 15.42 & 16.36 & 52.74 & 46.39 & 49.36 \\
  LocateAnything Fast\hspace{0.35em}\citep{locateanything} & N/A\textsuperscript{*} & N/A\textsuperscript{*} & N/A\textsuperscript{*} & N/A\textsuperscript{*} & N/A\textsuperscript{*} & N/A\textsuperscript{*} & N/A\textsuperscript{*} & N/A\textsuperscript{*} & N/A\textsuperscript{*} \\
  LocateAnything Hybrid\hspace{0.35em}\citep{locateanything} & N/A\textsuperscript{*} & N/A\textsuperscript{*} & N/A\textsuperscript{*} & N/A\textsuperscript{*} & N/A\textsuperscript{*} & N/A\textsuperscript{*} & N/A\textsuperscript{*} & N/A\textsuperscript{*} & N/A\textsuperscript{*} \\
  LocateAnything Slow NTP\hspace{0.35em}\citep{locateanything} & N/A\textsuperscript{*} & N/A\textsuperscript{*} & N/A\textsuperscript{*} & N/A\textsuperscript{*} & N/A\textsuperscript{*} & N/A\textsuperscript{*} & N/A\textsuperscript{*} & N/A\textsuperscript{*} & N/A\textsuperscript{*} \\
  Qwen2.5-VL-7B\hspace{0.35em}\citep{qwen25vl} & 37.84 & 51.28 & 43.54 & 4.81 & 5.52 & 5.14 & 25.96 & 33.86 & 29.38 \\
  Qwen3-VL-2B\hspace{0.35em}\citep{qwen3vl} & 46.95 & 60.77 & 52.97 & 10.16 & 11.34 & 10.72 & 33.05 & 41.25 & 36.69 \\
  Qwen3-VL-4B\hspace{0.35em}\citep{qwen3vl} & 55.33 & 78.89 & 65.05 & 16.48 & 19.70 & 17.94 & 42.86 & 58.61 & 49.49 \\
  Qwen3-VL-8B\hspace{0.35em}\citep{qwen3vl} & 52.30 & 68.07 & 59.15 & 13.92 & 15.58 & 14.70 & 39.15 & 48.91 & 43.47 \\
  Qwen3.5-4B\hspace{0.35em}\citep{qwen35} & 50.61 & 68.83 & 58.33 & 10.27 & 11.63 & 10.91 & 35.31 & 46.12 & 39.98 \\
  Qwen3.5-9B\hspace{0.35em}\citep{qwen35} & 59.48 & 67.68 & 63.32 & 14.31 & 15.36 & 14.82 & 44.12 & 49.75 & 46.77 \\
  RynnBrain1.1\hspace{0.35em}\citep{RynnBrain112B} & 51.28 & 72.83 & 60.18 & 10.11 & 11.95 & 10.95 & 36.86 & 50.13 & 42.46 \\
  SenseNova-Vision\hspace{0.35em}\citep{SenseNovaVision7BMoT} & 52.53 & 51.26 & 51.89 & 24.74 & 24.52 & 24.63 & 45.21 & 44.23 & 44.71 \\
  MiMo-VL-7B-SFT\hspace{0.35em}\citep{MimoVL} & 30.01\textsuperscript{*} & 30.31\textsuperscript{*} & 30.16\textsuperscript{*} & 2.22\textsuperscript{*} & 2.22\textsuperscript{*} & 2.22\textsuperscript{*} & 18.12\textsuperscript{*} & 18.25\textsuperscript{*} & 18.18\textsuperscript{*} \\
  MiMo-VL-7B-RL\hspace{0.35em}\citep{MimoVL} & 32.31\textsuperscript{*} & 34.11\textsuperscript{*} & 33.18\textsuperscript{*} & 2.24\textsuperscript{*} & 2.26\textsuperscript{*} & 2.25\textsuperscript{*} & 19.00\textsuperscript{*} & 19.83\textsuperscript{*} & 19.40\textsuperscript{*} \\
  BAGEL\hspace{0.35em}\citep{BAGEL7BMoT} & 49.17\textsuperscript{*} & 66.61\textsuperscript{*} & 56.58\textsuperscript{*} & 11.79\textsuperscript{*} & 13.49\textsuperscript{*} & 12.58\textsuperscript{*} & 35.35\textsuperscript{*} & 45.66\textsuperscript{*} & 39.83\textsuperscript{*} \\
  RynnBrain\hspace{0.35em}\citep{dang2026rynnbrain} & 51.73 & 73.85 & 60.84 & 8.39 & 9.78 & 9.03 & 35.41 & 47.95 & 40.71 \\
  GroundingPI & \textbf{83.48} & \textbf{85.72} & \textbf{84.58} & \textbf{65.29} & \textbf{66.44} & \textbf{65.86} & \textbf{77.39} & \textbf{79.27} & \textbf{78.32} \\
  \addlinespace[2pt]
  \multicolumn{10}{@{}l}{\hspace{0.4em}\strut\textbf{Vision-Language Models (10B--1T)}} \\
  Qwen3.6-27B\hspace{0.35em}\citep{qwen3627b} & 61.25 & 66.40 & 63.72 & 15.65 & 15.73 & 15.69 & 46.15 & 49.28 & 47.66 \\
  Qwen3-VL-32B\hspace{0.35em}\citep{qwen3vl} & 31.80 & 36.42 & 33.96 & 11.01 & 11.84 & 11.41 & 25.85 & 29.17 & 27.41 \\
  Qwen3.8-27B\hspace{0.35em}\citep{qwen38} & 54.66 & 61.69 & 57.96 & 12.59 & 13.18 & 12.88 & 40.11 & 44.51 & 42.19 \\
  Qwen3.5-35B-A3B\hspace{0.35em}\citep{qwen35} & 58.42 & 68.94 & 63.24 & 14.40 & 15.16 & 14.77 & 43.42 & 49.89 & 46.42 \\
  DeepSeek-VL2-Small-16B\hspace{0.35em}\citep{Deepseekvl2} & N/A\textsuperscript{*} & N/A\textsuperscript{*} & N/A\textsuperscript{*} & N/A\textsuperscript{*} & N/A\textsuperscript{*} & N/A\textsuperscript{*} & N/A\textsuperscript{*} & N/A\textsuperscript{*} & N/A\textsuperscript{*} \\
  DeepSeek-VL2-27B\hspace{0.35em}\citep{Deepseekvl2} & N/A\textsuperscript{*} & N/A\textsuperscript{*} & N/A\textsuperscript{*} & N/A\textsuperscript{*} & N/A\textsuperscript{*} & N/A\textsuperscript{*} & N/A\textsuperscript{*} & N/A\textsuperscript{*} & N/A\textsuperscript{*} \\
  \addlinespace[2pt]
  \multicolumn{10}{@{}l}{\hspace{0.4em}\strut\textbf{Vision-Language Models (>1T)}} \\
  Qwen3.7-Max\hspace{0.35em}\citep{qwen37} & 59.47 & 72.75 & 65.44 & 15.59 & 16.77 & 16.16 & 45.54 & 54.11 & 49.44 \\
  Kimi-K2.6 & 59.81 & 59.89 & 59.85 & 22.77 & 22.63 & 22.70 & 47.16 & 47.08 & 47.12 \\
  Kimi-K3\hspace{0.35em}\citep{KimiK3} & N/A\textsuperscript{*} & N/A\textsuperscript{*} & N/A\textsuperscript{*} & N/A\textsuperscript{*} & N/A\textsuperscript{*} & N/A\textsuperscript{*} & N/A\textsuperscript{*} & N/A\textsuperscript{*} & N/A\textsuperscript{*} \\
  GPT-6 Astra & 82.06 & 72.77 & 77.13 & 35.17 & 33.49 & 34.31 & 69.08 & 61.33 & 64.96 \\
  \addlinespace[2pt]
  \bottomrule
  
  \end{NiceTabular}%
  }
  \endgroup
\end{table}

\section{Potential Applications}
\label{app:potential_applications}

\paragraph{Industrial inspection and flexible manufacturing.}
Language descriptions or visual exemplars could specify defects and component variants for localization, while OCR associates part markings and packaging text with their image regions. Queries can be revised as products change, and localized parts and markings provide inputs for downstream assembly checks.

\paragraph{Embodied and driving data annotation.}
On egocentric images and video keyframes, boxes and points can link descriptions such as ``the cup beside the plate'' or ``the drawer handle'' to objects, parts, and candidate interaction sites. The same query interface could support road-scene pre-annotation and retrieval of unusual obstacles, temporary signs, or construction equipment.

\paragraph{Dense object retrieval and counting.}
In aerial, retail, and agricultural imagery, instance localization could support ship, vehicle, product, or fruit counting. Referring expressions distinguish crowded targets by appearance or relative position, while visual examples specify unfamiliar objects that are difficult to name consistently.

\paragraph{Text and document information extraction.}
Joint text recognition and region localization can retain the positions of receipt fields, package labels, and equipment markings; layout predictions additionally identify titles, tables, and figures. These spatial associations support field verification, document retrieval, and matching text to the corresponding objects or document regions.

\paragraph{GUI interaction.}
Requests such as ``open the settings for this project'' can be mapped to candidate click locations in screenshots. Combining control text, icons, and spatial context helps specify which repeated button or menu entry an interface agent should act on.

\section{Qualitative Analysis}
\label{app:qualitative_analysis}

We examine selected visualizations across grounding tasks, emphasizing the spatial and semantic demands visible in each example. Source images and their displayed annotations are preserved. These examples provide qualitative context, not additional estimates of accuracy or recall. A panel marked N/A denotes an unavailable comparison.

\subsection{General Object Grounding}

\begin{figure}[htbp]
\centering
\includegraphics[width=\linewidth]{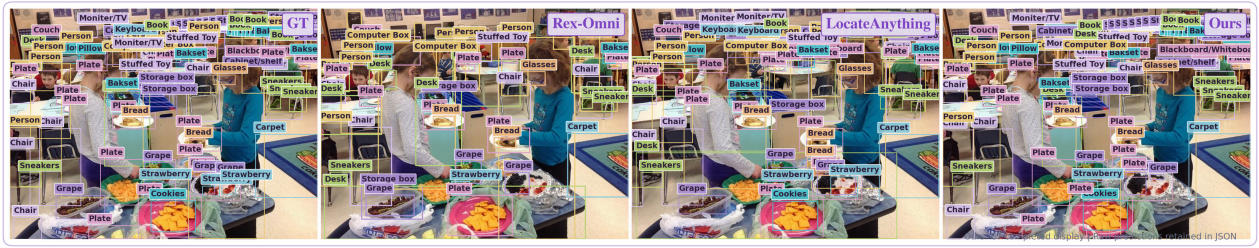}
\caption{Multi-scale grounding in a crowded classroom scene.}
\label{fig:qual-02}
\end{figure}

\paragraph{Grounding across foreground and background.} \Cref{fig:qual-02} spans children around a table, food and plates in the foreground, chairs, and densely arranged background objects. A useful structured description must separate overlapping people and furniture while retaining smaller items on the table and shelves. This scene illustrates why broad category recognition and precise instance localization must work together: recognizing the room does not determine the boundaries of each object.

\subsection{Dense Object Grounding}

\begin{figure}[htbp]
\centering
\includegraphics[width=\linewidth]{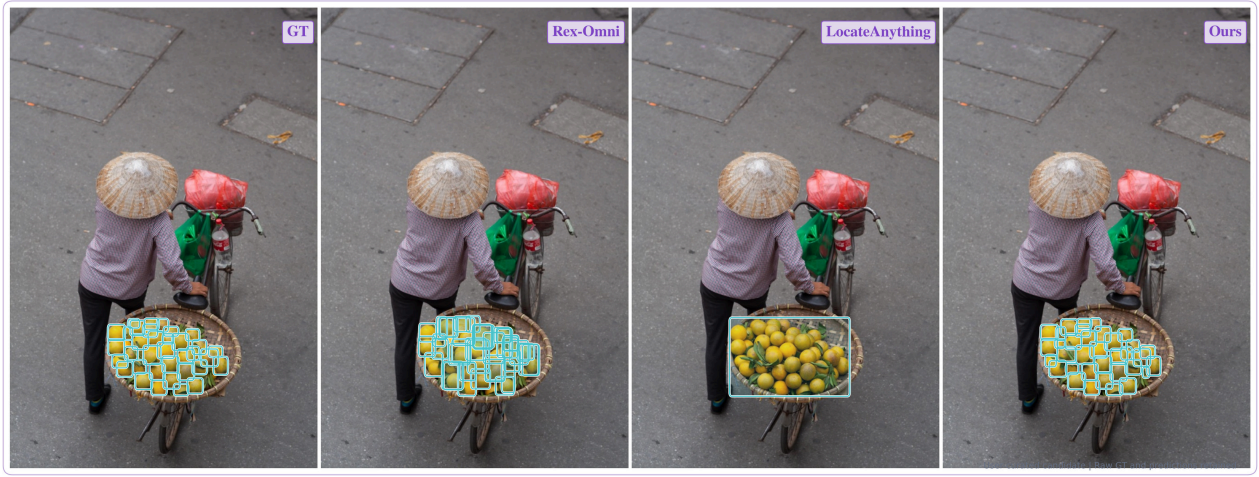}
\caption{Dense fruit grounding under partial occlusion.}
\label{fig:qual-05}
\end{figure}

\paragraph{Dense objects inside a container.} In \Cref{fig:qual-05}, a top-down view places many small fruits inside a basket below a cyclist. The basket is a salient enclosing region, but the desired instance granularity concerns its contents. The comparison contrasts individual fruit boxes, overlapping proposals, and a broad container-level region. Separating partially hidden fruits requires local appearance cues and an understanding of which object level the query requests. This selected display illustrates the distinction between container recognition and instance-complete grounding.

\subsection{Referring Grounding and Complex Visual Configurations}

\begin{figure}[htbp]
\centering
\includegraphics[width=\linewidth]{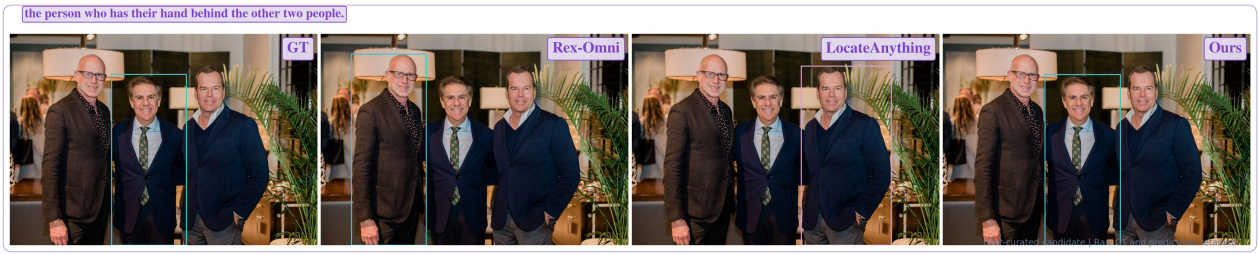}
\caption{Grounding a person through an occluded spatial relationship.}
\label{fig:qual-09}
\end{figure}

\paragraph{Relational referring with occluded body parts.} \Cref{fig:qual-09} asks for the person whose hand is behind the other two people. The displayed Ours region selects the central person, matching the illustrated reference, while the comparison panels select one of the flanking people. Resolving the expression requires connecting body parts and person-level regions despite overlapping torsos and largely hidden arms. The case illustrates compositional referring: the model must identify the entity satisfying a relationship rather than simply detect a visible hand or choose a salient face.

\subsection{Referring Point-in-Mask Grounding}

\begin{figure}[htbp]
\centering
\includegraphics[width=\linewidth]{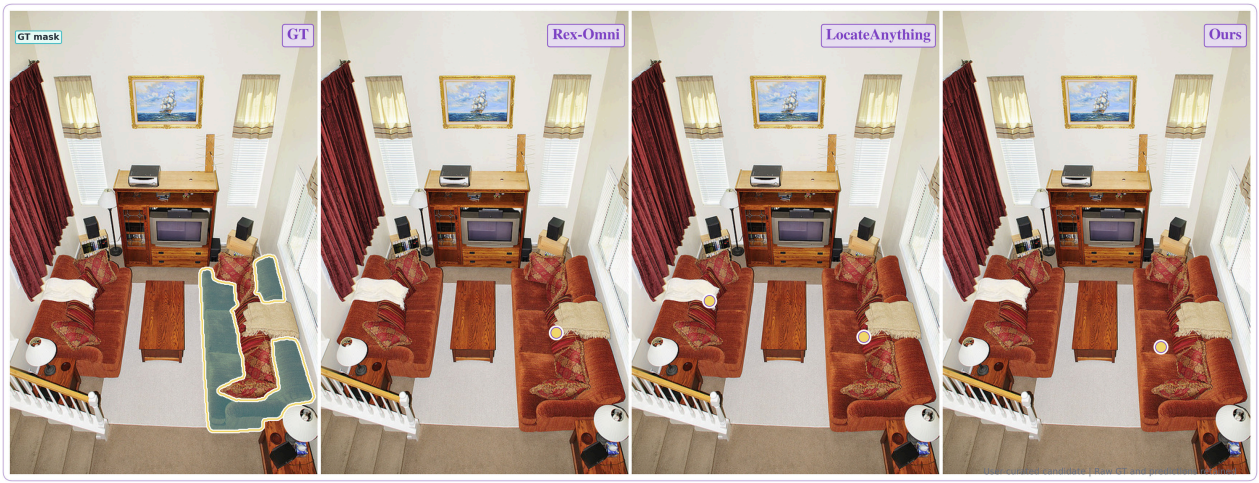}
\caption{Point grounding on the visible surface of a sofa.}
\label{fig:qual-11}
\end{figure}

\paragraph{Instance selection and interior-point placement.} The two similar sofas in \Cref{fig:qual-11} create an instance-selection challenge, while cushions and a blanket divide the visible surface of the target sofa. The reference mask highlights the right sofa's exposed regions. The displayed points illustrate different choices of instance and local surface, with the Ours point placed on an exposed seat region. For point-in-mask evaluation, predicting a nearby cushion or the other sofa can be incorrect even when a coarse bounding box would overlap the intended object.

\subsection{Dense Point Grounding}

\begin{figure}[htbp]
\centering
\includegraphics[width=\linewidth]{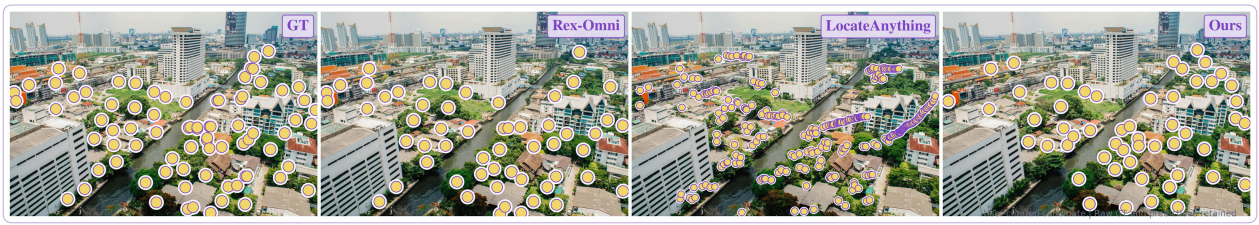}
\caption{Dense point grounding in an aerial urban scene.}
\label{fig:qual-13}
\end{figure}

\paragraph{Dense pointing across a wide field of view.} The aerial scene in \Cref{fig:qual-13} contains many small targets distributed across a wide field of view, with strong perspective variation and clutter from buildings, vegetation, and roads. The panels illustrate the difference between distributed instance-level points and dense runs of nearby points that can repeatedly sample one structure. The displayed Ours pattern follows the spatial distribution of the reference more closely in this selected rendering. Because the underlying query is not printed in the figure, we restrict the analysis to point distribution and avoid inferring an unshown target category.

\subsection{GUI Grounding}

\begin{figure}[htbp]
\centering
\includegraphics[width=\linewidth]{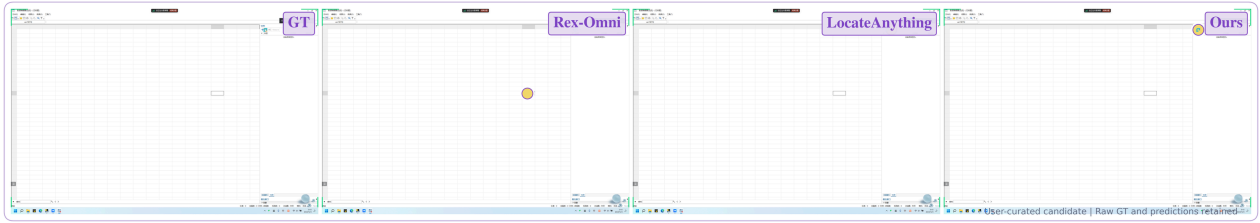}
\caption{GUI target grounding of a small side-panel control.}
\label{fig:qual-16}
\end{figure}

\paragraph{GUI grounding of a small peripheral control.} \Cref{fig:qual-16} contrasts a large spreadsheet-like canvas with a small target control in the upper part of the right-side panel. The displayed Ours point aligns with the marked control, while a comparison point is drawn toward a selected central cell. The example separates semantic interface grounding from visual salience: the active cell is prominent but does not identify the requested control. Fine localization is particularly important because neighboring toolbar icons occupy only a small portion of the screenshot.

\subsection{OCR and Artistic Text Understanding}

\begin{figure}[htbp]
\centering
\includegraphics[width=\linewidth]{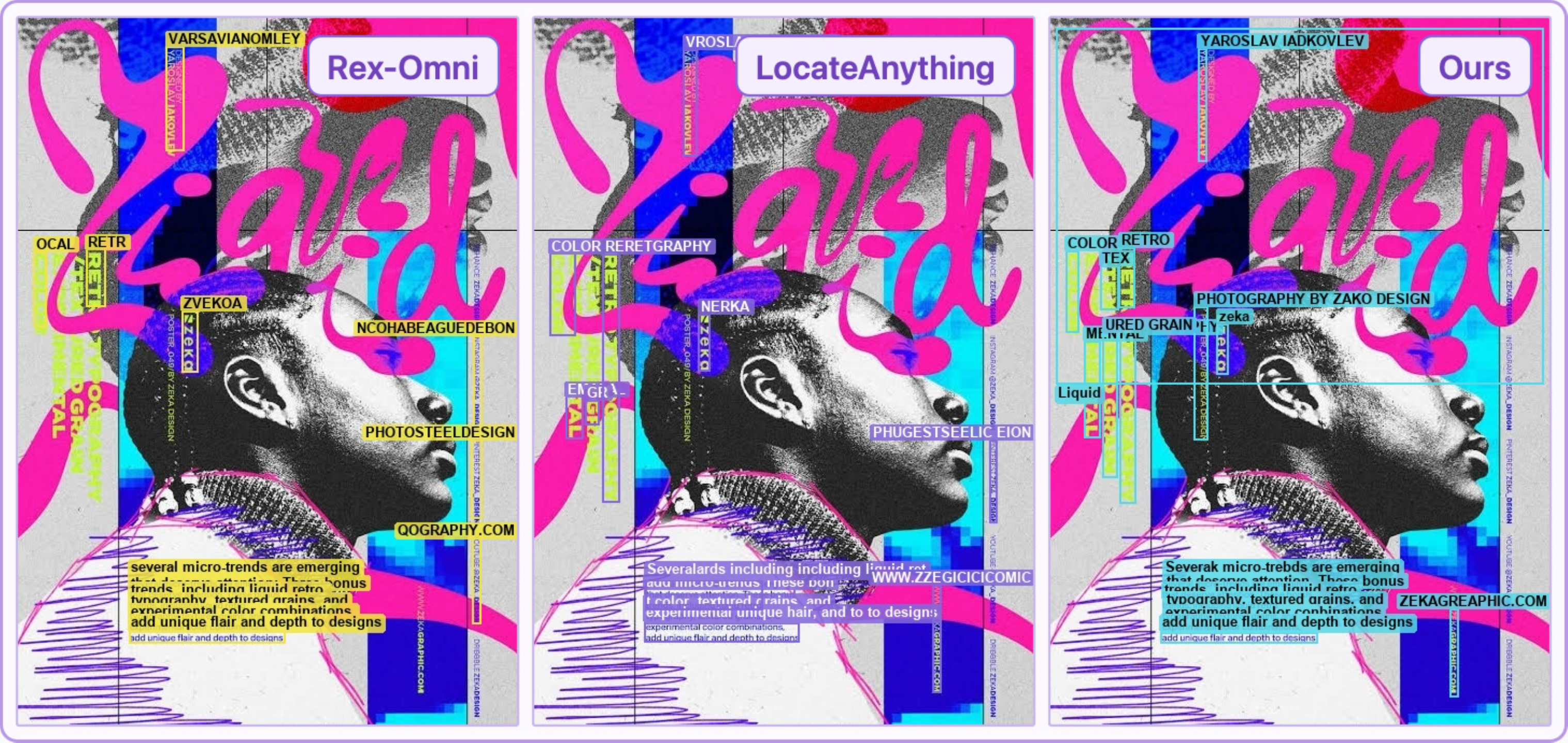}
\caption{OCR of layered artistic typography and small print.}
\label{fig:qual-19}
\end{figure}

\paragraph{Complex case: typography across scales and orientations.} \Cref{fig:qual-19} mixes a large pink script word, vertical labels, small horizontal print, and a photographic background. The Ours panel identifies ``Liquid'' despite the enlarged, curved, and overlapping letterforms, while also localizing several smaller text regions. This combination requires separating lettering from decorative strokes and reading across markedly different spatial scales. Word-level context is helpful, but the geometry of each text region remains necessary to associate a transcription with the correct visual evidence. The display also contains imperfect small-text transcriptions, so it should not be described as error-free OCR. The case instead illustrates both the value of a strong vision--language representation for artistic lettering and the remaining difficulty of dense, layered, low-resolution print.

\begin{figure}[htbp]
\centering
\includegraphics[width=\linewidth]{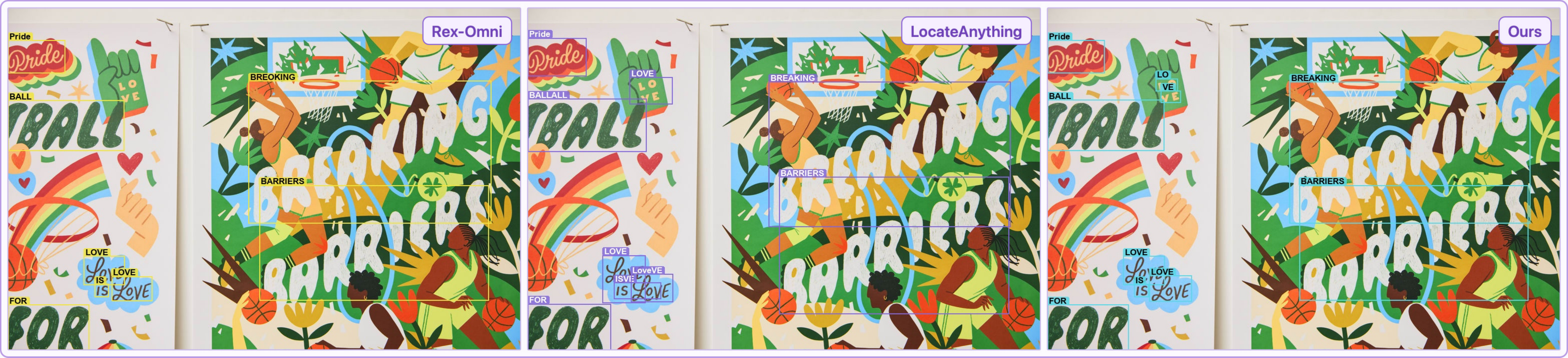}
\caption{OCR of slanted display text and mixed lettering styles.}
\label{fig:qual-20}
\end{figure}

\paragraph{Complex case: text integrated into graphic composition.} The wall graphics in \Cref{fig:qual-20} combine slanted display text, script, and short words distributed over illustrations. The Ours panel reads ``BREAKING'' and ``BARRIERS'' across the large diagonal composition and separates ``LO'' and ``VE'' on the foam-hand graphic. Recognizing these regions requires tolerance to irregular baselines, nonuniform glyph shapes, and competing decorative contours. The comparison also illustrates annotation granularity: ``LO'' and ``VE'' can be represented as two spatial text groups or combined as ``LOVE'', depending on the protocol. Strong visual--language understanding helps connect unusual letterforms to coherent text, while localization preserves where that evidence occurs. The example supports a qualitative capability discussion rather than a claim that artistic reading is exclusive to a particular model family.

\begin{figure}[htbp]
\centering
\includegraphics[width=\linewidth]{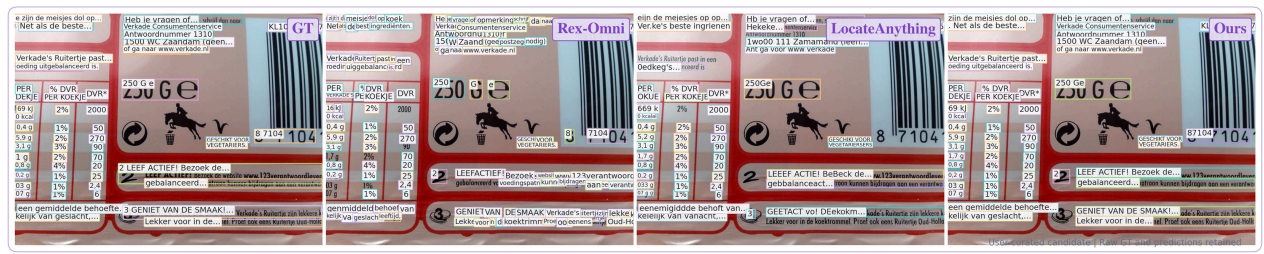}
\caption{OCR of multilingual packaging and numerical fields.}
\label{fig:qual-23}
\end{figure}

\paragraph{Multilingual text and numerical fields.} \Cref{fig:qual-23} combines Dutch prose, compact numerical columns, units, percentages, barcode digits, and graphic symbols. The displayed Ours regions retain line-level text and separate numerical entries, whereas comparison panels show alternative fragmentation and transcription errors. This example requires preserving punctuation and units while distinguishing text from nearby logos and illustrations. The challenge is not only recognizing a language but maintaining the spatial associations between heterogeneous textual elements in a crowded package layout.

\subsection{Visual-Prompt Grounding}

\begin{figure}[htbp]
\centering
\includegraphics[width=\linewidth]{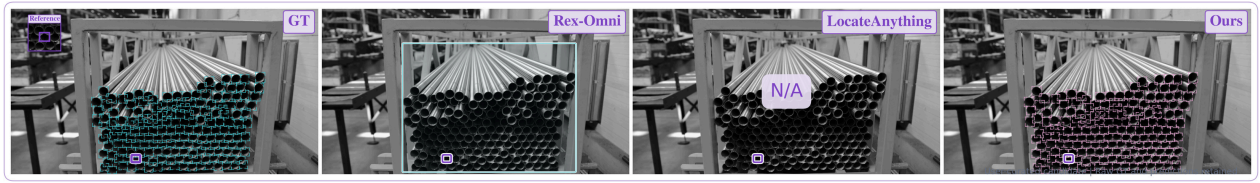}
\caption{Exemplar-guided grounding of densely stacked pipe openings.}
\label{fig:qual-26}
\end{figure}

\paragraph{Visual prompting and instance granularity.} \Cref{fig:qual-26} supplies a visual exemplar of a pipe opening in a tightly packed bundle. The desired unit is an individual opening, rather than the whole stack or the long metal tube extending behind it. The displayed Ours boxes retain this unit across the bundle, while the broad Rex-Omni box illustrates a group-level interpretation. Repeated dark interiors and small boundary gaps make instance separation difficult. The N/A panel is kept as an unavailable comparison and does not support a quantitative baseline claim.

\end{document}